\documentclass{ut-thesis}

\newif\ifpersianack
\persianacktrue

\usepackage{thesis-packages}
\usepackage{thesis-macros}

\author{Mohammad Mozaffari}
\title{Compression Trinity: Exploring Sparsity, Quantization, and Low-Rank Approximations for LLM Compression}
\degree{Doctor of Philosophy}
\department{Computer Science}
\gradyear{2026}

\begin{document}
  \frontmatter
  \maketitle
  \begin{abstract}
Prohibitive computational and environmental costs impede the scalable deployment of Large Language Models (LLMs). Traditional compression techniques (sparsity, quantization, low-rank approximations) are typically applied in isolation, limiting effectiveness as each hits an accuracy-efficiency wall. Addressing this is critical for sustainable LLM deployment.

This thesis proposes the ``Compression Trinity,'' a unified framework applying these pillars jointly. By leveraging sparsity to reduce computation, quantization to minimize memory bandwidth, and low-rank approximations to recover accuracy, the framework unlocks new efficiency frontiers. We demonstrate these orthogonal methods are complementary, overcoming the limitations of isolated strategies.

To accelerate pretraining, we apply the Trinity to the optimizer and model architecture. \textbf{MKOR} approximates curvature via block-diagonal sparsity and low-rank inversion, maintaining numerical stability for quantized states. It reduces curvature update complexity from $\mathcal{O}(d^3)$ to $\mathcal{O}(d^2)$, accelerating convergence by up to $1.85\times$ over KFAC. \textbf{\slope} accelerates training by up to $1.25\times$ via a double-pruned backward pass for N:M sparsity, using low-rank ``lazy'' adapters in the final 1\% of training to recover accuracy.

For post-training compression, we progressively explore the Trinity to improve deployment efficiency. \textbf{\optima} establishes sparsity limits under strict resource constraints, stabilizing static masks in a zero-training regime by formulating weight reconstruction as globally optimal column-wise quadratic programs, improving zero-shot accuracy by up to 3.97\%. Given a fine-tuning budget, \textbf{\patch} breaks the ceiling of static masks by learning a dynamic hybrid sparsity ratio between 0\% and 50\%, yielding up to $1.38\times$ speedups. Finally, to overcome the structural limits of sparsity, \textbf{\slim} realizes the full Compression Trinity in one shot using mathematically derived low-rank adapters to recover information lost to quantization and sparsity, improving accuracy by up to 5.66\% over state-of-the-art methods and outperforming uncompressed dense models at equal parameter budgets by 0.6\%.

Together, these contributions demonstrate that the joint application of the Compression Trinity is essential for unlocking the next generation of efficient, scalable, high-performance LLMs.
\end{abstract}

  \begin{acknowledgements}
\addcontentsline{toc}{chapter}{Acknowledgements}

First and foremost, I would like to express my deepest gratitude to my family. To my father, Gholamhossein, and my mother, Pouran, whose unwavering love, sacrifices, and encouragement have been the foundation of everything I have achieved. To my sister, Maryam, for her constant support and for always believing in me. This thesis is as much yours as it is mine.

I would like to thank my supervisor, Professor Maryam Mehri Dehnavi, for providing the environment and resources that made this research possible.

I am grateful to my thesis committee members, Professor Angela Demke Brown, Professor Dan Roy, Professor Nandita Vijaykumar, and Professor Murat Erdogdu, for their valuable feedback and thoughtful questions that strengthened this work. I would also like to extend my thanks to Professor Dan Alistarh for serving as the external appraiser and for taking the time to carefully evaluate this thesis.

I owe a special debt of gratitude to Amir Yazdanbakhsh, who has been a mentor, collaborator, and source of inspiration throughout much of my research journey. His guidance and generosity with his time have profoundly shaped the way I approach problems. I am also grateful to Zhao Zhang for being a wonderful collaborator and for the many productive discussions we shared.

I would like to acknowledge my undergraduate supervisors, Professor Maryam Sabbaghiyan and Professor Amir Masoud Rabiei, who first sparked my passion for research and set me on this path. Their early mentorship laid the groundwork for everything that followed.

I have been fortunate to share the lab with exceptional friends and colleagues who made the journey both intellectually stimulating and enjoyable. I would like to thank Behrooz Zarebavani, Lucas Wilkinson, Younes Hourri, Kazem Cheshmi, Saeed Soori, Bangtian Liu, Avery Laird, Arya Rafii, Victor Kamel, Ray Hung, Kasra Jahankhani, Lucy Farcnik, Milad Khanchi, and Amirhossein Elmi for the countless conversations, collaborations, and moments of friendship.

To everyone who has contributed to this work, whether through direct collaboration or simply through friendship and encouragement, thank you.

\end{acknowledgements}

%% ---- Persian Acknowledgements (separate page) ----
\ifpersianack
\thispagestyle{plain}
\begin{persian}
\footnotesize
\begin{center}
{\large\bfseries سپاسگزاری}
\end{center}
\vspace{0.5em}

در آغاز، مراتب عمیق‌ترین سپاس و قدردانی خود را تقدیم خانواده‌ام می‌کنم. از پدر عزیزم، غلامحسین، و مادر مهربانم، پوران، که عشق بی‌دریغ، فداکاری‌ها و دلگرمی‌های بی‌وقفه‌شان بنیان تمامی دستاوردهای من بوده است. از خواهرم، مریم، که همواره پشتیبان من بوده و هیچ‌گاه از ایمان داشتن به من دست نکشیده است. این رساله به همان اندازه که از آنِ من است، از آنِ شماست.

از استاد راهنمایم، پروفسور مریم مهری دهنوی، به خاطر فراهم آوردن محیط و امکاناتی که انجام این پژوهش را میسّر ساخت، صمیمانه سپاسگزارم.

از اعضای محترم هیئت داوران رساله‌ام، پروفسور آنجلا دمکی براون، پروفسور دن روی، پروفسور ناندیتا ویجی‌کومار و پروفسور مورات اردوغدو، به خاطر بازخوردهای ارزشمند و پرسش‌های ژرف‌اندیشانه‌شان که موجب استحکام این اثر گردید، کمال تشکر و امتنان را دارم. همچنین از پروفسور دن آلیستار به خاطر پذیرش نقش ارزیاب خارجی و وقتی که برای بررسی دقیق این رساله صرف نمودند، قدردانی می‌کنم.

مراتب سپاس ویژه‌ام را نثار امیر یزدان‌بخش می‌کنم که در بخش عمده‌ای از مسیر پژوهشی‌ام، مربی، همکار و سرچشمهٔ الهام من بوده است. راهنمایی‌ها و سخاوتمندی ایشان در اختصاص وقت، نگرش من به حلّ مسائل را عمیقاً دگرگون ساخته است. همچنین از ژائو ژانگ به خاطر همکاری ارزنده و گفتگوهای سازندهٔ فراوانی که میانمان رد و بدل شد، سپاسگزارم.

از اساتید دوران کارشناسی‌ام، پروفسور مریم صباغیان و پروفسور امیرمسعود ربیعی، که نخستین جرقه‌های اشتیاق به پژوهش را در من برافروختند و مرا در این مسیر قرار دادند، قدردانی می‌نمایم. راهنمایی‌های نخستین ایشان زیربنای تمامی آنچه در پی آمد را بنا نهاد.

خوشبختانه افتخار هم‌آزمایشگاهی با دوستان و همکارانی استثنایی را داشته‌ام که این مسیر را هم از نظر علمی پربار و هم لذّت‌بخش ساختند. از بهروز زارع‌بوانی، لوکاس ویلکینسون، یونس حوری، کاظم چشمی، سعید سوری، بنگتیان لیو، اوری لرد، آریا رفیعی، ویکتور کامل، ری هانگ، کسری جهانخانی، لوسی فارچنیک، میلاد خانچی و امیرحسین علمی به خاطر گفتگوها، همکاری‌ها و لحظات رفاقت بی‌شمار سپاسگزارم.

از تمامی کسانی که در این اثر سهمی داشته‌اند، چه از طریق همکاری مستقیم و چه صرفاً با دوستی و دلگرمی، صمیمانه قدردانی می‌کنم.

\end{persian}
\clearpage
\fi

\begin{dedication}
    \begin{center}
        {\fontsize{14}{18}\selectfont\textit{Dedicated to my parents,}}\\[1em]
        {\calligra\fontsize{36}{44}\selectfont Gholamhossein Mozaffari}\\[0.5em]
        {\fontsize{14}{18}\selectfont\textit{and}}\\[0.5em]
        {\calligra\fontsize{36}{44}\selectfont Pouran Shaban Ashini}\\[1em]
        {\fontsize{14}{18}\selectfont\textit{for their endless love and sacrifice.}}
        %% ---- Persian Dedication ----
        \ifpersianack
        \vspace{3em}
        \begin{persian}
        \begin{center}
        {\fontsize{14}{18}\selectfont تقدیم به پدر و مادرم،}\\[1em]
        {\fontsize{28}{36}\selectfont غلامحسین مظفری}\\[0.5em]
        {\fontsize{14}{18}\selectfont و}\\[0.5em]
        {\fontsize{28}{36}\selectfont پوران شبان عشینی}\\[1em]
        {\fontsize{14}{18}\selectfont به پاس عشق و فداکاری بی‌پایانشان.}
        \end{center}
        \end{persian}
        \fi
    \end{center}
\end{dedication}

  \tableofcontents
  \listoftables
  \listoffigures

  \mainmatter
  \chapter{Introduction}
\label{chapter:intro}

Large Language Models (LLMs) have become foundational tools in modern artificial intelligence, demonstrating remarkable capabilities in text generation, reasoning, and multi-modal tasks \cite{gpt3, llama3, gemma3}. However, these capabilities come at a significant cost. Training state-of-the-art models consumes enormous computational, memory, and environmental resources \cite{patterson2021carbon, gpt3}, and deploying them for inference remains a major challenge due to their massive memory footprint and high computational demands.

To mitigate these overheads, various model compression techniques have been proposed \cite{sparsity_survey}. Historically, these methods have often been applied in isolation, focusing on a single tool such as \textbf{sparsity} (removing parameters) \cite{sparsegpt}, \textbf{quantization} (reducing parameter precision) \cite{optq}, or \textbf{low-rank approximations} (factoring parameter matrices) \cite{lora}. This isolated approach is inherently limiting, as it fails to address the multi-faceted nature of the efficiency bottleneck.

This thesis argues that these methods must be applied jointly. We introduce a conceptual framework, which we term the \textbf{Compression Trinity}, built upon the three fundamental pillars of sparsity, quantization, and low-rank approximations. These methods are highly complementary. While all three pillars contribute to reducing overall memory and compute overheads, they play distinct roles in balancing hardware constraints with model expressivity. Specifically, sparsity and quantization directly target the primary bottlenecks of modern hardware: sparsity reduces the \textbf{computational} load, while quantization minimizes \textbf{memory bandwidth} requirements. Low-rank approximations, in turn, act as the critical algorithmic counterweight. By efficiently projecting parameters into lower-dimensional spaces, they restore lost accuracy and model capacity without reintroducing the hardware overheads that sparsity and quantization eliminated. We demonstrate that the joint application of this Compression Trinity across the different phases of an LLM's life-cycle is the key to unlocking new frontiers of efficiency.

To understand the context for these contributions, we must first consider the distinct stages of an LLM's development and deployment.

\section{LLM Life-cycle Stages}
\label{sec:intro_lifecycle}

The life-cycle of a large language model can be broadly divided into two primary stages\footnote{Other stages exist, such as Supervised Fine-Tuning (SFT) and alignment (e.g., RLHF, DPO). We focus on pretraining and inference as they represent the primary computational bottlenecks in the LLM life-cycle.}:

\begin{itemize} 
    \item \textbf{Pretraining:} This is the most computationally expensive phase, where the model is trained from scratch on massive, web-scale datasets to learn general-purpose language representations \cite{pretraining, bert}. 
    \item \textbf{Inference:} This is the deployment stage, where the trained model is used to generate predictions for new user inputs. In many real-world applications, inference must be performed with low latency and on resource-constrained hardware. 
    
\end{itemize}

Each stage presents unique opportunities for acceleration. During the \textbf{pretraining} phase, this can be achieved by either accelerating the training process itself or by improving the \textbf{optimizer} to converge faster. For the \textbf{inference} stage, \textbf{post-training compression} can be applied to make the deployed model more efficient. These opportunities, though different on the surface, share a common computational bottleneck, dense matrix-matrix multiplications, and can therefore share the same fundamental principles for compression, which we will cover next.

\section{Compression Techniques}
\label{sec:intro_techniques}

The core of our approach relies on the three main techniques of the Compression Trinity: sparsity, quantization, and low-rank approximation. \textbf{Let us now define the three pillars of the Compression Trinity.}

\subsection{Sparsity}
\label{subsec:intro_sparsity}

Sparsity is a compression technique that involves identifying and removing (i.e., setting to zero) the least important weights in a neural network, thereby reducing the total number of parameters and floating-point operations (FLOPs) \cite{optimal_brain_damage, optimal_brain_surgeon}.

This technique is broadly categorized into two types. \textbf{Unstructured sparsity} removes arbitrary, individual weights from a matrix. While it offers high flexibility and can often achieve high compression ratios with minimal accuracy loss, it is notoriously difficult to accelerate in practice. Its irregular memory access patterns are not \textbf{hardware-friendly}, meaning they cannot be executed efficiently on modern parallel accelerators like GPUs, which are designed to process data in large, regular chunks \cite{lucas}. Conversely, \textbf{structured sparsity} removes entire blocks of parameters, such as full rows, columns, or filter channels \cite{block_pruning1, channel_pruning1}. This regular structure is hardware-friendly but often damages model accuracy significantly, as it removes entire features indiscriminately.

A new family of \textbf{semi-structured sparsity} patterns has emerged to bridge this gap. A prominent example is \textbf{N:M sparsity}, which enforces that $N$ out of every $M$ consecutive weights are non-zero (e.g., 2:4 sparsity) \cite{n_m3_dominosearch, 2_4_pretraining}. This pattern is flexible enough to preserve accuracy while being regular enough for hardware acceleration on modern GPUs \cite{sparse_tensor_cores}.

However, finding the optimal sparsity pattern and updating the remaining non-zero weights to compensate for the removed elements remains a challenging task \cite{optimal_brain_compression, sparsegpt}. To illustrate the severity of this challenge, we can compare sparsity against quantization. \autoref{tab:sparsity_vs_quant} demonstrates that even at a modest 2x compression ratio (50\% sparsity), removing connections damages the model more than aggressive 4-bit quantization, which offers 4x compression. This counter-intuitive result, that a method providing \textit{less} compression yields \textit{lower} accuracy, highlights that sparsity is the most destructive pillar of the Trinity, necessitating the dedicated optimization strategies we propose in \autoref{chapter:optima} and \autoref{chapter:patch}.

\begin{table}[ht]
\centering
\caption{\textbf{The Sparsity Paradox:} Comparison of LLaMA-2-7B zero-shot accuracy under standard Post-Training Compression. Sparsity yields lower compression ratios (2x) yet results in significantly higher accuracy loss compared to Quantization (4x), highlighting its destructive nature.}
\label{tab:sparsity_vs_quant}
\begin{tabular}{l c c c}
\toprule
\textbf{Method} & \textbf{Compression Ratio} & \textbf{Bit-width / Density} & \textbf{Avg Accuracy} \\
\midrule
Dense Baseline & 1x & FP16 / 100\% & 54.61\% \\
\midrule
4-bit Quantization\textsuperscript{*} & \textbf{4x} & INT4 / 100\% & \textbf{53.63\%} \\
2:4 Sparsity\textsuperscript{\textdagger} & 2x & FP16 / 50\% & 48.62\% \\
\bottomrule
\end{tabular}
\vspace{0.2cm}
\footnotesize{\\\textsuperscript{*} Best among AbsMax and OPTQ \\ \textsuperscript{\textdagger} Best among SparseGPT, Wanda, Thanos, ProxSparse, and MaskLLM}
\end{table}

\subsection{Quantization}
\label{subsec:intro_quantization}

Quantization reduces the numerical precision of the numbers used to represent the model's weights and, in some cases, activations. For example, parameters are typically trained in 32-bit (FP32) or 16-bit (FP16/BF16) floating-point formats, but quantization can compress them down to 8-bit integers (INT8), 4-bit integers (INT4), or even lower bit-widths \cite{int8, optq}.

This reduction in precision has two primary benefits: it saves memory (e.g., 4-bit quantization yields an 8x memory reduction over 32-bit weights) and allows for the use of faster, specialized compute units (like INT8 tensor cores) that can perform integer arithmetic much faster than floating-point operations.

The main challenge in quantization is that the representation capabilities of the numbers are reduced exponentially with the bit-width. This can lead to large accuracy degradation in low bit-width schemes, especially in the presence of outlier values, which are common in LLMs \cite{awq}. Finding the best way to map high-precision weights to a low-bit representation and updating those weights to minimize the resulting error is a non-trivial task \cite{optq}. As with sparsity, only relying on quantization limits the total compression ratio of the models, and other methods should be combined with it to push efficiency further.

\subsection{Low-rank Approximations}
\label{subsec:intro_lowrank}

Low-rank approximations are based on the observation that the large weight matrices in LLMs are often over-parameterized and have a low "intrinsic rank." This redundancy can be exploited by decomposing a large weight matrix $W \in \mathbb{R}^{m \times k}$ into the product of two smaller, "thin" matrices, $L \in \mathbb{R}^{m \times r}$ and $R \in \mathbb{R}^{r \times k}$, where the rank $r \ll m, k$. This technique, popularized by Low-Rank Adaptation (LoRA) \cite{lora}, can reduce the memory and compute overhead of LLMs significantly.

However, low-rank approximations are not very effective in compressing matrices that are inherently high-rank, and applying them too aggressively can lead to significant errors. Relying on low-rank approximation alone is insufficient, as it cannot capture the fine-grained, high-rank information that sparsity or high-precision quantization can preserve.

\section{The Failure of Isolated Compression}
\label{sec:intro_failure}

As mentioned in \autoref{sec:intro_lifecycle}, modern deployment scenarios impose strict constraints on memory and latency. While the individual pillars of compression---sparsity and quantization---are well-studied, existing literature often treats them as independent solutions.

However, empirical evidence suggests that pushing any single pillar to the extreme results in catastrophic accuracy loss. \autoref{tab:intro_failure} illustrates this breakdown on the LLaMA-2-7B model. When we attempt to achieve an 8x compression ratio using only quantization (2-bit) or only sparsity (87.5\%), the model's reasoning capabilities collapse, with accuracy dropping to near-random chance ($\approx$ 31\%).

\begin{table}[h]
\centering
\caption{Average Zero-shot Accuracy of LLaMA-2-7B on 8 tasks (MMLU, PIQA, ARC-Easy, ARC-Challenge, WinoGrande, OpenBookQA, RACE, HellaSwag) at 8x compression ratio. Single-pillar methods fail to retain capabilities, while multi-pillar methods (Compression Trinity) recover accuracy.}
\label{tab:intro_failure}
\begin{tabular}{l c}
\toprule
\textbf{Method} & \textbf{Average Accuracy} \\
\midrule
Dense Baseline (FP16) & 54.61\% \\
\midrule
\textit{Single-Pillar Compression (8x Ratio)} & \\
2-bit Quantization\textsuperscript{*} & 31.81\% \\
87.5\% Unstructured Sparsity\textsuperscript{\textdagger} & 31.24\% \\
\midrule
\textit{Multi-Pillar Compression (8x Ratio)} & \\
4-bit Quantization + 2:4 Sparsity & \textbf{47.97\%} \\
4-bit Quantization + 50\% Unstructured Sparsity & \textbf{52.38\%} \\
\bottomrule
\end{tabular}
\vspace{0.2cm}
\footnotesize{\\\textsuperscript{*} Best among AbsMax and OPTQ \\ \textsuperscript{\textdagger} Best among SparseGPT, Wanda, and Thanos}
\end{table}

In contrast, the table demonstrates that a \textbf{hybrid approach}, combining moderate 4-bit quantization with moderate 50\% sparsity, recovers the majority of the accuracy (52.38\%) while achieving the same compression ratio. This observation forms the core motivation of this thesis: compression is not a singular optimization problem, but a multi-dimensional balancing act.

\section{Thesis Contributions and Roadmap}
\label{sec:intro_contributions}

As discussed in \autoref{sec:intro_techniques}, each of the compression techniques, sparsity, quantization, and low-rank approximation, cannot effectively compress LLMs alone and will hit an accuracy or efficiency wall at some point. This thesis argues and demonstrates that the \textit{joint application} of the Compression Trinity is the key to unlocking new frontiers of efficiency. We show that by combining hardware-friendly sparsity and aggressive quantization, we can achieve massive reductions in compute and memory. We then use low-rank approximations as a controllable, highly efficient method to add back a small number of dense parameters, compensating for the joint compression error and restoring model accuracy.

Before detailing the novel methods that prove this thesis, \autoref{chapter:background} will first provide a comprehensive technical background.

The thesis is structured to explore the Compression Trinity across the full life-cycle of an LLM. We begin by applying the Trinity to the Pretraining phase (\autoref{chapter:mkor} and \autoref{chapter:slope}). We then perform a deep dive into the Sparsity pillar, the most destructive component of the Trinity, exploring both layer-wise (\autoref{chapter:optima}) and end-to-end (\autoref{chapter:patch}) optimization regimes. Finally, we integrate all three pillars for a holistic Post-Training Compression solution (\autoref{chapter:slim}). These contributions are detailed as follows:

\begin{itemize}
    \item \autoref{chapter:mkor} \textbf{MKOR:} We lay the foundation for the Compression Trinity in the expensive pretraining phase. \textsc{MKOR} (Momentum-Enabled Kronecker-Factor-Based Optimizer Using Rank-1 Updates) is a novel second-order optimizer that leverages all three pillars. It approximates the second-order information as a block-diagonal \textbf{sparse} matrix, and then leverages \textbf{rank-1 updates} (a low-rank approximation) to compute the inverse of its covariance matrices. Crucially, this joint formulation is exceptionally stable, allowing the inverse factors to be computed in a \textbf{quantized} 16-bit format, whereas other methods require 32-bit numbers for numerical stability. By applying the Trinity to the optimizer's internal computations, we reduce the complexity of second-order updates from $\mathcal{O}(d^3)$ to $\mathcal{O}(d^2)$ and communication from $\mathcal{O}(d^2)$ to $\mathcal{O}(d)$, where $d$ is the hidden dimension of the model. As a result, \textsc{MKOR} accelerates pretraining by outperforming state-of-the-art optimizers like KFAC by up to 1.85x on BERT-Large training.

    \item \autoref{chapter:slope} \textbf{\slope:} We further accelerate the pretraining phase by applying the Compression Trinity directly to the linear layers. \slope (A Double-Pruned Sparse Plus Lazy Low-rank Adapter Pretraining method) introduces a framework that jointly applies sparsity and low-rank approximations from the start. It accelerates sparse pretraining by introducing a novel double-pruned backward pass that enables N:M sparsity acceleration in both forward and backward passes. To recover accuracy lost from sparsity, we introduce low-rank adapters only during the final 1\% of pretraining iterations. This "lazy" insertion minimizes overhead while maximizing accuracy. By creating a base model that is already sparse and low-rank, \slope produces a model that is not only efficient for inference (1.34x speedup) but is also an ideal and stable candidate for the final pillar, quantization, which we apply in the post-training stage. This approach accelerates end-to-end training and inference of models like OPT-66B by up to 1.14x and 1.34x, respectively, and reduces training memory by 0.77x.

    \item \autoref{chapter:optima} \textbf{\optima:} We transition to the post-training phase, addressing the strict scenario where end-to-end training is infeasible. \optima (Optimal One-shot Pruning via Quadratic Programming) focuses on perfecting the \textbf{sparsity} pillar under a "zero-training" constraint. It formulates the one-shot, post-pruning weight update as a series of independent, row-wise Quadratic Programs (QPs) that share a common layer Hessian. This allows us to find the per-row globally optimal update given the Hessian, minimizing the reconstruction error from pruning. By creating the most accurate and stable sparse model possible, \optima serves as a critical enabling step, producing a high-fidelity model that can withstand the subsequent application of aggressive quantization and low-rank approximations. \optima acts as a drop-in replacement for the update step in methods like Wanda or SparseGPT, consistently improving zero-shot performance by up to 3.97\% absolute accuracy without any fine-tuning.

    \item \autoref{chapter:patch} \textbf{\patch:} We address the scenario where a \textbf{fine-tuning} budget is available to push performance further. \patch (Pruning with a Learnable Tile-level Configuration) overcomes the limitations of the layer-wise mask detection used in \textsc{OPTIMA}. It introduces a hybrid sparsity framework that partitions weight matrices into tiles, assigning each tile to be either \textit{dense} or \textit{2:4 sparse} via a learnable mask. This creates a continuous, effective sparsity ratio between 0\% and 50\%, balancing accuracy in critical regions with acceleration elsewhere. While \patch focuses on enhancing this single pillar, it is designed to be fully composable with the other pillars of the Trinity. We explicitly demonstrate that it can be combined with joint quantization and low-rank approximation methods, such as \slim, to further boost the accuracy and stability of the final, jointly compressed model. On LLaMA-2 7B, the \patch framework alone delivers 1.18x-1.38x end-to-end speedup while improving accuracy by up to 2.96\% compared to state-of-the-art 2:4 pruning.

    \item \autoref{chapter:slim} \textbf{\slim:} We present the complete fulfillment of the Compression Trinity in a one-shot setting. \slim (One-shot Quantization and Sparsity with Low-rank Approximation) holistically integrates all three pillars to solve the "compounded error" problem. It applies aggressive, hardware-friendly quantization and semi-structured sparsity, then compensates for the combined error using a novel saliency function that allows us to \textit{mathematically compute} optimal low-rank adapter values in one shot. This joint approach improves accuracy by up to 5.66\% on LLaMA-2-7B (combining 4-bit quantization and 2:4 sparsity) and achieves up to 4.3x layer-wise speedup.

\end{itemize}

Finally, \autoref{chapter:conclusion} concludes the thesis by summarizing its key findings, reflecting on the impact of the Compression Trinity framework, and discussing potential avenues for future research. The code, pre-trained checkpoints, and interactive visualizations for the methods presented in this thesis are centralized at our research hub\footnote{\url{https://www.cs.toronto.edu/~mmozaffari/compression-trinity/}}.
  \chapter{Background}
\label{chapter:background}

This chapter establishes the technical foundations necessary to understand the contributions of this thesis. We begin by identifying the computational bottleneck that dominates modern LLM workloads: the dense linear layers within the transformer architecture (\autoref{sec:transformer_bottleneck}). We then examine the hardware constraints that govern the execution of these layers on modern GPUs, introducing the memory hierarchy, specialized compute units, and the Roofline model that together determine whether a workload is limited by compute or by memory bandwidth (\autoref{sec:hardware_foundations}). With both the algorithmic bottleneck and the physical constraints established, we formalize the two orthogonal strategies for acceleration, reducing the number of training iterations through better optimization and reducing the cost of each iteration through hardware-aware compression (\autoref{sec:solution_space}). We explore each strategy in turn: \autoref{sec:algorithmic_efficiency} surveys the landscape of first- and second-order optimizers, motivating the need for structured approximations that make curvature information tractable at scale; \autoref{sec:hardware_efficiency_regimes} characterizes the distinct computational regimes of LLM training and inference, revealing why different life-cycle stages demand different compression techniques. Finally, \autoref{sec:compression_trinity} and \autoref{sec:joint_approach} introduce the Compression Trinity as the unifying solution framework for this thesis and argue that the joint application of sparsity, quantization, and low-rank approximations is essential to overcome the limitations of any single technique applied in isolation.

\section{The Transformer Bottleneck: Linear Layers}
\label{sec:transformer_bottleneck}

The Transformer has become the foundational architecture for virtually all modern Large Language Models (LLMs), demonstrating unparalleled scaling and performance on a wide array of language tasks \cite{transformer}. In practice, an LLM is constructed by stacking a large number of identical Transformer blocks, often numbering in the dozens or even hundreds. This repetitive, block-based design means that the computational profile of a single block is representative of the model's entire computational load. Understanding the specific operations within this block is therefore the first step toward identifying the primary opportunities for optimization.

\begin{figure}[tb]
  \centering
  \includegraphics[width=\textwidth]{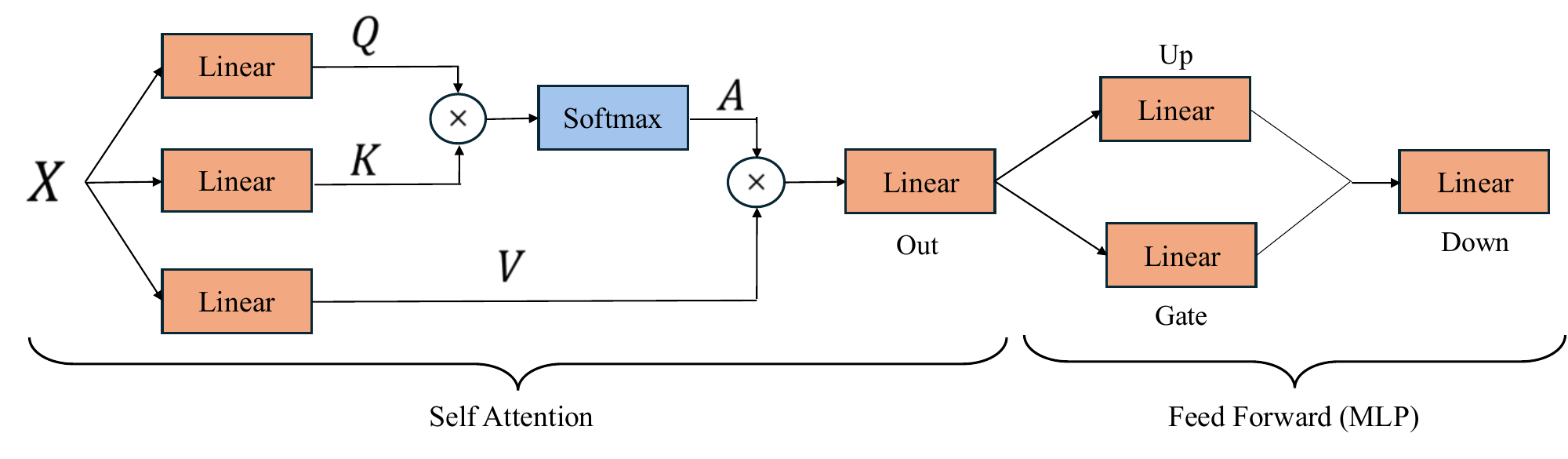}
  \caption{The compute graph of a standard Transformer block, highlighting the Self-Attention and Feed-Forward Network (FFN) sub-layers.}
  \label{fig:transformer_block}
\end{figure}

A standard Transformer block, as illustrated in \autoref{fig:transformer_block}, is composed of two primary sub-components. The first is the Self-Attention mechanism, which allows the model to weigh the importance of different tokens in a sequence relative to each other. The second is a position-wise Feed-Forward Network (FFN), which is typically a two- or three-layer Multi-Layer Perceptron (MLP) that provides the majority of the model's representational capacity. In modern architectures, this FFN often takes the form of a SwiGLU variant \cite{swiglu}. These two components work in tandem, with the attention mechanism handling the aggregation of sequential information and the FFN processing that information at each token's position.

Connecting this architectural graph to its underlying mathematical operations reveals a critical insight: both sub-components are dominated by linear layers, which are implemented as dense matrix-matrix multiplications (GEMMs). The Self-Attention mechanism, for instance, computes its Query (Q), Key (K), and Value (V) representations through three independent linear layers, and a final Output (O) projection is applied after the attention scores are aggregated. Similarly, the SwiGLU FFN is composed of an Up-projection layer, a Gate layer, and a Down-projection layer. Consequently, the vast majority of computations and parameters in the Transformer block are contained within these dense matrix multiplications.

Each linear layer, parameterized by a weight matrix $W \in \mathbb{R}^{d_{\text{out}} \times d_{\text{in}}}$, participates in three distinct matrix multiplications during a single training step. Given an input activation matrix $X \in \mathbb{R}^{b \times d_{\text{in}}}$, where $b$ is the effective batch size (batch size $\times$ sequence length), the \textbf{forward pass} computes the layer's output as $Y = XW^T$. During backpropagation, two additional GEMMs are required: the \textbf{weight gradient} computation, $\nabla_W \mathcal{L} = (\nabla_Y \mathcal{L})^T X$, which determines how the weights should be updated, and the \textbf{input gradient} computation, $\nabla_X \mathcal{L} = (\nabla_Y \mathcal{L}) W$, which propagates the error signal to the preceding layer. Crucially, while the forward pass multiplies the input by $W^T$, the input gradient computation multiplies by $W$ itself. This transpose relationship poses a fundamental challenge for structured compression: a sparsity pattern that is hardware-friendly along the rows of $W$ (as required for the forward pass) may not be hardware-friendly along its columns (as required for the backward pass). This asymmetry is a central obstacle that we address directly in \autoref{chapter:slope}.

\begin{figure}[htbp]
  \centering
  \includegraphics[]{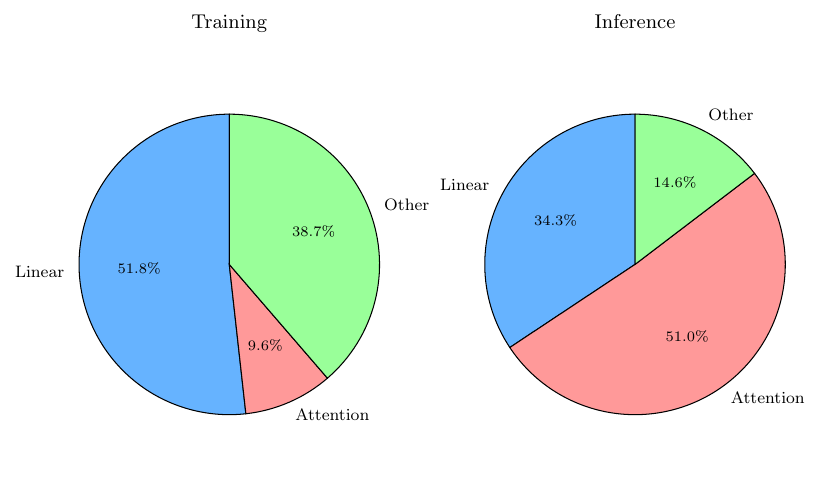}
  \caption{Computational time breakdown for LLaMA-3.1-8B during training and inference, as profiled on a single NVIDIA H100 GPU. The "linear" component is the largest single bottleneck. The batch size, input sequence length, and generation sequence length are set to 4, 1024, and 1024 respectively.}
  \label{fig:llm_time_breakdown}
\end{figure}

This architectural analysis is confirmed by empirical performance profiling, as shown in \autoref{fig:llm_time_breakdown}. The "linear" component is by itself the largest single computational bottleneck, consuming approximately 51.8\% of the total training time for a model like LLaMA-3.1-8B. The "Attention" component, which contains the self-attention computations with dynamic, data-dependent operations excluding the linear layers, accounts for another 9.6\%. While the Attention mechanism's unique properties present their own optimization challenges, this thesis will focus on the linear layers. As the largest and most dominant bottleneck, composed of standard static-weight GEMMs, these linear layers represent one of the most critical and impactful targets for optimization.

The evidence from both the architectural design and the performance profile establishes a clear conclusion: accelerating the dense linear layers is the central challenge for LLM efficiency. However, identifying the mathematical operations is only half the picture. To understand \textit{why} these operations become bottlenecks, one must analyze the physical constraints of the hardware on which they execute. The following section will introduce the fundamental principles of GPU architecture and the Roofline model that govern the performance of these linear layers.

\section{Hardware Constraints and the Roofline Model}
\label{sec:hardware_foundations}

While the Transformer architecture defines the operations to be performed, the execution speed is dictated by the underlying hardware. Modern LLMs are trained and deployed almost exclusively on Graphics Processing Units (GPUs), which are massive throughput-oriented processors. To understand the efficiency bottlenecks discussed in this thesis, we must first establish a model of how these devices process data, specifically focusing on the memory hierarchy, specialized compute units, and the theoretical limits defined by the Roofline model.

\subsection{Memory Hierarchy and Tensor Cores}
\label{subsec:memory_and_tensor_cores}

The computational pipeline of a GPU is constrained by two primary resources: \textbf{Memory Bandwidth} (how fast data can be moved) and \textbf{Compute Throughput} (how fast data can be processed).

\paragraph{Memory Hierarchy:} Data movement in a GPU follows a hierarchy of speed and capacity. The bulk of model parameters and activations reside in High Bandwidth Memory (HBM), which offers high capacity (e.g., 80GB on an NVIDIA H100) but relatively high latency and limited bandwidth compared to on-chip memory. To perform an operation, data must be moved from HBM to the smaller, faster L2 cache, and finally to the Streaming Multiprocessors' (SM) shared memory and registers (SRAM). The cost of moving data from HBM is orders of magnitude higher than moving it within the chip. Consequently, algorithms that reuse loaded data multiple times (high data locality) are significantly more efficient than those that stream data for single use.

\paragraph{Tensor Cores vs. CUDA Cores:} Historically, GPUs relied on general-purpose "CUDA Cores" for arithmetic. However, the rise of Deep Learning necessitated specialized hardware. Modern architectures (e.g., NVIDIA Ampere and Hopper) feature \textbf{Tensor Cores}, specialized execution units designed essentially to perform one operation: matrix-multiply-and-accumulate ($D = A \times B + C$) in a single clock cycle. 

While standard CUDA cores operate on scalars or small vectors, Tensor Cores consume entire $8 \times 16$ or $16 \times 16$ matrices per instruction \cite{nvidia_a100_2020, nvidia_h100_2022}. This specialization allows Tensor Cores to deliver throughputs over an order of magnitude higher than CUDA cores. For example, on the NVIDIA H100, Tensor Cores introduce support for the FP8 data format, doubling the peak throughput compared to the standard BF16 \cite{bfloat} format used in previous generations like the A100. This massive disparity means that any algorithm not utilizing Tensor Cores efficiently leaves the vast majority of the GPU's potential performance on the table.

\subsection{The Roofline Model}
\label{subsec:roofline_model}

\begin{figure}[tb]
  \centering
  \includegraphics[width=\textwidth]{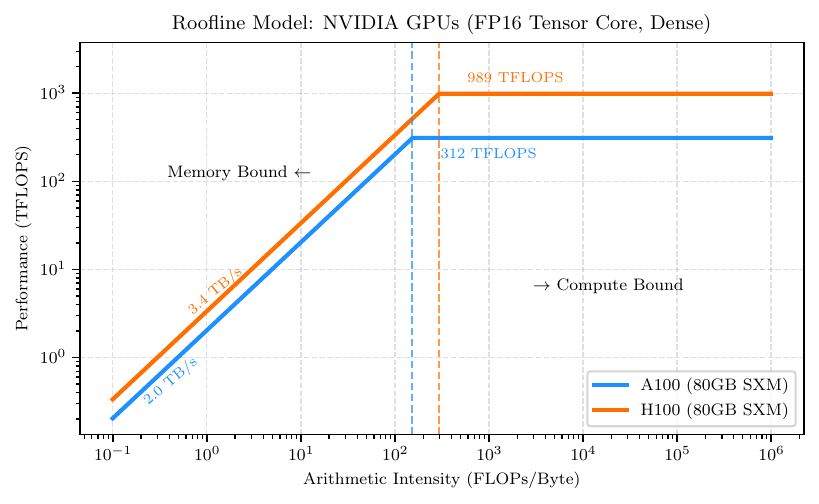}
  \caption{Roofline Model comparison between NVIDIA A100 and H100 GPUs. The H100 (orange) offers significantly higher peak compute (FLOPs). However, memory bandwidth has not scaled proportionally, shifting the "knee" of the curve to the right. This implies that algorithms on H100 require a higher arithmetic intensity to escape the memory-bound region compared to the A100.}
  \label{fig:roofline}
\end{figure}

The interaction between the memory bandwidth and compute throughput is formalized by the \textbf{Roofline Model} \cite{roofline}. This model visualizes the theoretical peak performance of an algorithm based on its \textbf{Arithmetic Intensity} ($I$), defined as the number of floating-point operations (FLOPs) performed for every byte of data transferred from memory:
\begin{equation}
    I = \frac{\text{Total FLOPs}}{\text{Total Bytes Transferred}}
\end{equation}
As illustrated in \autoref{fig:roofline}, the Roofline model divides performance into two distinct regions:
\begin{itemize}
    \item \textbf{Memory-Bound Region (Sloped Line):} When arithmetic intensity is low, the GPU's compute units starve while waiting for data from HBM. In this region, performance is strictly limited by memory bandwidth. Improving performance here requires moving fewer bytes (e.g., via Quantization).
    \item \textbf{Compute-Bound Region (Flat Line):} When arithmetic intensity is high, data is loaded once and reused many times, keeping the compute units fully saturated. In this region, performance is limited by the peak FLOPs of the Tensor Cores. Improving performance here requires doing fewer operations (e.g., via Sparsity).
\end{itemize}

The "knee" of the curve represents the transition point. \autoref{fig:roofline} highlights a critical trend in hardware evolution: the gap between compute and memory is widening. The NVIDIA H100 boasts significantly higher peak FLOPs than the A100 (especially with FP8), pushing the "roof" higher. However, memory bandwidth has scaled much more slowly. This shifts the knee to the right, meaning that modern GPUs require increasingly higher arithmetic intensity to achieve peak utilization. This hardware reality fundamentally shapes the strategies required for optimization: Pretraining (high intensity) sits firmly under the compute roof, while Inference Decoding (low intensity) is trapped under the memory slope.

\subsection{Sparse Tensor Core Acceleration}
\label{subsec:sparse_tensor_cores}

While standard Tensor Cores provide massive throughput for dense matrix multiplications, they perform redundant calculations if the weight matrix contains zeros. To address this, modern NVIDIA architectures introduced \textbf{Sparse Tensor Cores} designed to accelerate fine-grained structured sparsity, specifically the 2:4 pattern.

\paragraph{Mechanism and Format:} The 2:4 structured sparsity format enforces a rigid constraint: in every contiguous block of 4 elements along the reduction dimension, at least 2 elements must be zero. This allows the hardware to compress the sparse matrix by 50\%, storing only the non-zero values in a packed format alongside small metadata indices that record the original positions of the retained elements. During execution, the Sparse Tensor Core reads these indices to select only the relevant entries from the corresponding dense operand, skipping all multiplications that would involve a zero. Because the core performs the same operation in the same number of clock cycles but processes only half as many non-zero elements, it effectively doubles the theoretical peak throughput compared to the equivalent dense precision.

\paragraph{The Backward Pass Challenge:} While 2:4 sparsity offers massive theoretical gains, applying it to training is non-trivial. The Sparse Tensor Core requires the sparsity to exist along the \textit{reduction dimension}. In the forward pass, this aligns naturally. However, in the backward pass, we must multiply by the transpose of the weights, misaligning the sparsity pattern relative to the hardware's expected read order. This structural alignment challenge is a primary motivation for the custom pretraining strategies developed in \autoref{chapter:slope} (\slope).

\section{The Solution Space: Two Frontiers of Acceleration}
\label{sec:solution_space}

Given the dominance of Linear layers identified in \autoref{sec:transformer_bottleneck} and the rigid hardware constraints defined in \autoref{sec:hardware_foundations}, we can now formalize the optimization landscape. To minimize the total time required to train an LLM or process a request, we can decompose the cost into two multiplicative factors:

\begin{equation}
    \text{Total Time} = \underbrace{(\text{Number of Iterations})}_{\text{Sample Efficiency}} \times \underbrace{(\text{Time Per Iteration})}_{\text{Hardware Efficiency}}
\end{equation}

This decomposition reveals two orthogonal frontiers for acceleration, each requiring distinct strategies:

\begin{enumerate}
    \item \textbf{Strategy 1: Sample Efficiency (Reducing the Number of Iterations).} This strategy is exclusive to the training regime. If we can improve the optimization algorithm to converge in fewer steps, we reduce the total time even if the cost of each step remains constant. This motivates the study of advanced optimizers.
    
    \item \textbf{Strategy 2: Hardware Efficiency (Reducing the Time Per Iteration).} This strategy applies to both training and inference. To reduce the cost of a single iteration, we must attack the hardware bottlenecks identified in the Roofline model: we must either reduce the FLOPs (for compute-bound operations) or reduce the memory traffic (for memory-bound operations).
\end{enumerate}

The following sections will explore these two frontiers in detail, starting with Algorithmic Efficiency via advanced optimizers, followed by Hardware Efficiency via the different operating regimes of LLMs.

\section{Strategy 1: Algorithmic Efficiency}
\label{sec:algorithmic_efficiency}

Having defined the physical arena in which our models execute, we turn to the first frontier of acceleration: \textbf{Sample Efficiency}. If the hardware limits how fast we can compute a single update, we must strive to compute fewer updates overall. This brings us to the domain of optimization algorithms, which dictate the path a model takes through the loss landscape from random initialization to convergence.

\subsection{First-Order Methods}
Standard LLM training relies almost exclusively on first-order optimization methods, particularly adaptive gradient variants such as Adam \cite{adam} and AdamW \cite{adamw}. These methods utilize the gradient vector $g = \nabla \mathcal{L}(\theta)$, which represents the slope of the loss function. Geometrically, first-order methods approximate the loss landscape locally as a hyperplane (a linear approximation).

While computationally efficient—requiring only $\mathcal{O}(d)$ memory and compute for $d$ parameters—this linear approximation ignores the \textit{curvature} of the landscape. In the highly non-convex and ill-conditioned optimization landscapes typical of deep neural networks, this blindness to curvature leads to two primary inefficiencies. In "narrow valley" regions where the curvature is high in one direction and low in another, first-order methods tend to oscillate across the valley rather than moving down its floor, wasting iterations. Additionally, without knowledge of the local scale (curvature), determining the optimal step size (learning rate) is difficult, often requiring extensive tuning and warm-up schedules.

\subsection{Second-Order Methods}
Second-order methods address these limitations by incorporating the Hessian matrix $H = \nabla^2 \mathcal{L}(\theta)$, which contains the second-order partial derivatives. By using the Hessian, these methods approximate the loss locally as a quadratic function (a "bowl") rather than a plane. This allows for the computation of the Newton step:
\begin{equation}
    \Delta \theta = -H^{-1}g
\end{equation}
The Newton step uses the inverse Hessian to automatically rescale the gradient. It takes larger steps in directions of low curvature (flat plateaus) and smaller, more cautious steps in directions of high curvature (steep cliffs). Theoretically, this property, known as affine invariance, allows second-order methods to converge in significantly fewer iterations than their first-order counterparts.

\subsection{The Computational Barrier and Structured Approximations}
Despite their theoretical superiority, exact second-order methods are intractable for LLMs due to the sheer size of the Hessian. For a model with $d$ parameters, $H$ is a $d \times d$ matrix. For a modest 7B parameter model, storing $H$ would require exabytes of memory, and inverting it ($\mathcal{O}(d^3)$) is computationally impossible. This presents a classic efficiency trade-off: second-order methods offer high sample efficiency but suffer from catastrophic hardware inefficiency.

To make second-order optimization feasible, we must rely on \textbf{Structured Approximations}. The most prominent approach in deep learning is Kronecker-Factored Approximate Curvature (KFAC) \cite{kfac}.\footnote{There are other lines of work such as Shampoo \cite{shampoo} and Muon \cite{muon} that approximate the Hessian matrix using alternative structured factorizations. The Compression Trinity framework is in principle applicable to these methods as well; however, we focus on the KFAC family in this thesis as a representative and widely adopted testbed for demonstrating the benefits of joint sparsity, quantization, and low-rank approximations within second-order optimization.} KFAC approximates the Fisher Information Matrix (a proxy for the Hessian) not as a single dense matrix, but layer-wise. It assumes the Hessian for a given layer can be approximated as the Kronecker product of two much smaller matrices, $H \approx A \otimes G$, where $A$ relates to the layer's inputs and $G$ to its output gradients.

Inverting this Kronecker product is efficient because $(A \otimes G)^{-1} = A^{-1} \otimes G^{-1}$. This reduces the inversion cost from $\mathcal{O}(d^3)$ to the cubic size of the layer's width (e.g., $\mathcal{O}(width^3)$). However, even with KFAC, the memory cost of maintaining these curvature factors remains prohibitive for modern LLMs, often tripling the memory footprint compared to Adam \citep{kfac, kaisa}.

This unsolved challenge serves as the motivation for \autoref{chapter:mkor} (MKOR). We argue that Low-Rank factorization (like KFAC) is merely the starting point. To truly bridge the gap, achieving the sample efficiency of second-order methods with the hardware efficiency of first-order methods, we must apply the full Compression Trinity to the optimizer itself, combining Low-Rank updates with Sparsity and Quantization on the optimizer states.

\section{Strategy 2: Hardware Efficiency and LLM Regimes}
\label{sec:hardware_efficiency_regimes}

While advanced optimizers attack the "Sample Efficiency" frontier to shorten training, they offer no benefit during inference, where the number of steps is fixed by the user's generation length. To accelerate inference and to further speed up training, we must attack the second frontier: \textbf{Hardware Efficiency}.

However, "efficiency" is not a static target. As predicted by the Roofline model (\autoref{subsec:roofline_model}), the bottleneck shifts dramatically depending on the operational regime. LLM execution is bifurcated into two distinct computational profiles: the \textbf{Compute-Bound} regime (dominating Training and Inference Prefill) and the \textbf{Memory-Bound} regime (dominating Inference Decoding).

\subsection{The Compute-Bound Regimes: Training and Prefill}
The Training phase\footnote{Throughout this thesis, 
we use ``training'' and ``pretraining'' interchangeably when referring to the computational regime. While \emph{pretraining} technically denotes the initial large-scale training phase described in \autoref{sec:intro_lifecycle}, the computational profile, large batches of dense matrix multiplications over many tokens, is shared by all training-like stages 
(including supervised fine-tuning). Our compression techniques apply equally to all such stages; we default to ``training'' when discussing hardware characteristics and ``pretraining'' when emphasizing the life-cycle context.}  and the Inference Prefill phase share a fundamental characteristic: \textbf{massive parallelism}.
In Training, the model processes large batches of sequences simultaneously. Similarly, in the Prefill phase of inference (processing the user's prompt), the model computes attention and feed-forward outputs for all input tokens at once.

In these scenarios, the Arithmetic Intensity is high. The weight matrices are loaded from HBM to the chip once and reused across thousands of tokens (large batch size $\times$ sequence length). Consequently, both Training and Prefill sit firmly on the flat, \textbf{Compute-Bound} plateau of the Roofline model. In this region, the GPU's compute units are fully saturated, and memory bandwidth is not the limiting factor.

Therefore, optimizing Training and Prefill requires strategies that strictly reduce the total number of operations (FLOPs). This makes \textbf{Sparsity} the premier accelerator for this regime. By skipping calculations for zero-valued weights, sparsity directly lowers the compute ceiling required to process the batch, translating to faster training steps and lower prompt latency.

\subsection{The Memory-Bound Regime: Inference Decoding}
Once the prompt is processed, the model enters the Decode phase, generating one token at a time. This phase is inherently sequential; the output of step $t$ is required to compute step $t+1$, preventing parallelization across the sequence dimension.

In this regime, the Arithmetic Intensity collapses. To generate a single token (or a small batch of tokens), the GPU must load the entire model (often 100GB+ for large models) from HBM to the chip, perform a single matrix-vector multiplication, and then discard the weights. The data reuse is minimal. Consequently, the Decode phase falls deep into the \textbf{Memory-Bound} slope of the Roofline model.

Here, the compute units (Tensor Cores) sit idle for the vast majority of execution time, starving for data. Reducing FLOPs (via Sparsity) provides diminishing returns because computation is not the bottleneck. Instead, acceleration is strictly defined by how fast data can be moved. This makes \textbf{Quantization} the dominant accelerator for decoding. By reducing the bit-width of the weights (e.g., from 16-bit to 4-bit), we reduce the data volume by $75\%$, effectively quadrupling the speed at which the model can be fed to the compute units.

The contrast between Training and Prefill, which require FLOP reduction, and Decoding, which requires Bandwidth reduction, reinforces the need for the \textbf{Compression Trinity}. No single compression technique can address both bottlenecks simultaneously: sparsity alone leaves the memory-bound decode phase untouched, while quantization alone cannot accelerate the compute-bound training and prefill phases. It is precisely the joint application of complementary techniques, sparsity to cut FLOPs, quantization to cut data movement, and low-rank approximations to recover lost accuracy, that enables a unified efficiency strategy across the entire LLM lifecycle.

\section{The Compression Trinity as a Solution Framework}
\label{sec:compression_trinity}

To address the specific bottlenecks and strategies identified for both pretraining and inference, we introduce the "Compression Trinity,\footnote{Other compression methods, such as Knowledge 
Distillation \cite{knowledge_distillation}, in which a 
smaller \emph{student} model is trained to replicate 
the behavior of a larger \emph{teacher}, is not included in this list because it 
operates at a fundamentally different level of 
abstraction: while sparsity, quantization, and low-rank 
approximation are all transformations applied to an 
existing model's weight matrices, distillation is a 
training procedure that produces an entirely new model, 
often with a different architecture, and requires 
access to the full training pipeline. This places it 
outside the resource-constrained regime targeted by 
much of this thesis. That said, the two approaches 
are fully composable, a distilled model can serve as 
input to our compression pipeline, and their integration is a promising 
direction for future work.}" the conceptual framework for this thesis. This framework is built upon the three fundamental pillars of model compression: Sparsity, Quantization, and Low-Rank Approximations. Rather than viewing these as independent techniques, we posit that they are highly complementary tools that, when applied jointly, provide a comprehensive solution to the challenges of LLM efficiency. This section will introduce each pillar and map it to the two acceleration strategies defined in \autoref{sec:algorithmic_efficiency} and \autoref{sec:hardware_efficiency_regimes}.

The first pillar, \textbf{Sparsity}, is the process of identifying and removing (pruning) the least important parameters from a weight matrix $W$, thereby reducing its effective size. This technique can be broadly categorized by the pattern of removed weights: \textit{unstructured} sparsity removes individual weights, \textit{structured} sparsity removes entire blocks (e.g., rows or columns), and \textit{semi-structured} N:M sparsity enforces a fine-grained pattern, such as 2 non-zero weights out of every 4. Sparsity is the primary implementation of \textbf{Compute Bound Optimizations (Reduce FLOPs)}. By reducing the number of non-zero parameters, it directly lowers the arithmetic cost, making it ideal for compute-bound regimes like pretraining. Simultaneously, by shrinking the storage requirement, it also assists with the Memory Bound Regime (Reduce Data Movement).

The second pillar, \textbf{Quantization}, is the process of reducing the numerical precision of the weights $W$ (and sometimes activations) from high-precision formats like 32-bit floating point (FP32) to low-precision formats like 8-bit or 4-bit integers (INT8 or INT4). To maintain accuracy, especially in the presence of outlier values common in LLMs \cite{int8}, this is often applied on a per-group basis. Quantization is the most powerful tool for \textbf{Memory Bound Regimes (Reduce Data Movement)}. By reducing the bit-width of $W$ from 16 (for BF16) to 4, it cuts the memory bandwidth requirement by 75\%. This directly attacks the bottleneck of the memory-bound decode phase. While specialized hardware (e.g., FP8 Tensor Cores) allows quantization to also accelerate computation, its dominant benefit remains the massive reduction in memory traffic.

The third pillar, \textbf{Low-Rank Approximations}, is based on the hypothesis that the large weight matrices in LLMs are over-parameterized and have a low intrinsic rank \cite{Aghajanyan2021}. This pillar exploits this redundancy by representing a large matrix as the product of two smaller, "thin" matrices (e.g., $W \approx LR$ or $\Delta W = LR$). This pillar is the most versatile of the three, acting as a bridge between Hardware Efficiency and Sample Efficiency.

For \textbf{Hardware Efficiency (Strategy 2)}, representing $W$ with far fewer parameters (e.g., $d \times r + r \times d$, where $r \ll d$) reduces both the FLOPs and the memory footprint. Additionally, as an efficient means to mitigate accuracy degradation from sparsity and quantization, low-rank approximations provide a computationally cheap mechanism for representing fully dense matrices with unrestricted element values.

Finally, for \textbf{Strategy 1 (Reduce Iterations)}, the pillars work together. While Low-Rank approximations typically provide the mathematical structure necessary to approximate curvature (e.g., factorizing the Hessian matrix), making these advanced optimizers computationally practical often requires the full Compression Trinity. As we will demonstrate with MKOR (\autoref{chapter:mkor}), solely relying on low-rank structure is often insufficient for fitting complex optimizer states into GPU memory. Instead, we must apply Sparsity and Quantization \textit{on top of} the low-rank factors. Thus, Strategy 1 is not the domain of a single pillar, but rather the ultimate synthesis where all three pillars are combined to enable smarter, faster optimization.

\section{The Case for a Joint Approach} 
\label{sec:joint_approach}

The core argument of this thesis is that applying these pillars in isolation is an inherently limited approach. Each technique, when pushed to its extreme, hits a fundamental wall: aggressive sparsity causes catastrophic accuracy degradation, aggressive quantization fails due to the sensitivity of outlier parameters, and low-rank approximation alone cannot capture the full rank information necessary for a model's capabilities.

However, when viewed through the lens of our identified strategies, these pillars become complementary pieces of a unified puzzle. Sparsity maximizes FLOP reduction. Quantization maximizes Bandwidth reduction. Low-Rank Approximation provides a mathematical structure to recover accuracy and enable advanced optimization.

It is important to clarify, however, that while the \textit{ultimate} goal of this thesis is the joint application of the Compression Trinity, achieving this combination requires that each individual pillar be robust enough to support the others. If a single pillar is brittle, combining it with others leads to compounded errors and performance collapse. Therefore, not all chapters in this thesis focus on the simultaneous application of all three pillars. Instead, we adopt a staged approach: we dedicate specific chapters (specifically \optima and \patch) to pushing the boundaries of the \textbf{Sparsity} pillar in isolation. These contributions are necessary foundational steps, transforming sparsity from a fragile technique into a robust building block capable of withstanding the additional pressure of joint Quantization and Low-Rank approximation in the final unification (e.g., \slim).

This thesis will demonstrate how this joint framework can be tailored to solve the distinct challenges of the LLM life-cycle:

\begin{enumerate}
    \item \textbf{Pretraining (Compute \& Sample Efficiency):} To solve the compute-bound pretraining problem, we advocate for a two-pronged approach. We use Strategy 1 by developing an advanced optimizer that leverages all three pillars of the Trinity to approximate curvature and accelerate convergence (as explored in \textsc{MKOR} \cite{mkor}, \autoref{chapter:mkor}). Simultaneously, we use Strategy 2 by jointly applying Sparsity and Low-Rank Approximations to the weights during training (as explored in \textsc{SLoPe} \cite{slope}, \autoref{chapter:slope}).

    \item \textbf{Inference (Memory Efficiency \& Accuracy):} For the post-training inference problem, the goal is to holistically apply all three pillars. We first establish a stable foundation by perfecting structured sparsity (as explored in \textsc{OPTIMA}, \autoref{chapter:optima}) and adapting it for modern hardware (as explored in \textsc{PATCH}, \autoref{chapter:patch}). With this foundation, we finally demonstrate the complete fulfillment of the Trinity: a one-shot method that jointly applies Sparsity, Quantization, and Low-Rank approximation to simultaneously attack compute, memory, and accuracy recovery (as explored in \textsc{SLiM}, \autoref{chapter:slim}).
\end{enumerate}

In conclusion, this chapter has established the core problem (Linear layers), the physical constraints (Roofline model and Memory Hierarchy), the algorithmic opportunity (Sample Efficiency), and the Compression Trinity as our unified solution framework. The following chapters will now present the novel contributions of this thesis, demonstrating the effectiveness of this joint framework in unlocking new frontiers of efficiency.

  \chapter{MKOR: Momentum-Enabled Kronecker-Factor-Based Optimizer Using Rank-1 Updates}
\label{chapter:mkor}

\paragraph{Publication and Contributions.}
The content of this chapter is based on the paper 
``MKOR: Momentum-Enabled Kronecker-Factor-Based 
Optimizer Using Rank-1 Updates'' \cite{mkor}, 
published at the Conference on Neural Information 
Processing Systems (NeurIPS), 2023. This work was 
conducted in collaboration with Sikan Li, Zhao Zhang, 
and Maryam Mehri Dehnavi. Mohammad Mozaffari conceived 
the algorithm, led the implementation, and designed 
and executed the experiments. Sikan Li assisted with 
conducting experiments. Zhao Zhang and Maryam Mehri 
Dehnavi supervised the project and contributed to the 
writing and revision of the manuscript.

\section{Introduction}
\label{section:mkor_intro}

As established in \autoref{chapter:intro}, the pretraining phase of Large Language Models (LLMs) is dominated by massive computational costs. To mitigate this, we employ the second strategy of the Compression Trinity: reducing the total number of training iterations required for convergence. Second-order optimization methods have gained significant attention for this purpose, as they utilize curvature information, specifically the inverse of the Hessian matrix, to precondition gradients, thereby achieving higher convergence rates than first-order counterparts like SGD or Adam. However, these methods face a critical scalability wall. Since the size of the Hessian scales quadratically with the model parameters, computing and storing the exact Hessian and its inverse is computationally intractable for modern deep neural networks (DNNs), necessitating the use of approximation techniques.

Beyond computational complexity, the memory footprint of optimizer states presents a prohibitive bottleneck at the scale of LLMs. Standard first-order optimizers like Adam already require maintaining multiple state tensors (e.g., momentum and variance) for every model parameter, often consuming more GPU memory than the model weights themselves. Second-order methods exacerbate this issue by requiring the storage of curvature information, such as the Hessian or its factors. For models with billions of parameters, the memory required to store these high-precision curvature matrices frequently exceeds the capacity of modern hardware accelerators. Consequently, applying second-order optimization to LLMs requires not only approximating the curvature to reduce computational costs but also aggressively compressing them to fit within memory constraints.

Existing approximation methods attempt to make second-order optimization feasible but typically address only one aspect of the efficiency bottleneck. One common approach is Natural Gradient Descent (NGD)~\cite{ngd}, which substitutes the Hessian with the Fisher Information Matrix (FIM)~\cite{fim}. To handle the memory limitations of large models, algorithms like Kronecker-Factored Approximate Curvature (KFAC)~\cite{kfac} approximate the FIM using \textbf{block-diagonal sparsity}, where each block corresponds to a layer. However, inverting these blocks remains computationally expensive, scaling with $\mathcal{O}(d^3)$ where $d$ is the layer dimension. Consequently, KFAC implementations~\cite{dist_kfac1, dist_kfac2, dist_kfac_conv1, dist_kfac_conv2, dist_kfac_parallel_comp_and_commu, kaisa} must update the curvature information infrequently (e.g., every 100-1000 iterations), which damages convergence rates. Alternative methods like SNGD~\cite{sngd1, sngd2, hylo} and KBFGS~\cite{kbfgs} attempt to shift the complexity to the batch dimension ($\mathcal{O}(b^3)$ or $\mathcal{O}(bd^2)$). While effective for small batches, these fail in Transformer models~\cite{transformer} where the effective batch size scales with sequence lengths that can reach thousands of tokens~\cite{lra}. Thus, existing methods hit a wall: they are either too computationally heavy due to matrix inversion or fail to scale with sequence length.

To overcome these limitations, we apply the Compression Trinity directly to the optimizer's internal computations. We present MKOR, a \textbf{M}omentum-Enabled \textbf{K}ronecker-Factorization-Based \textbf{O}ptimizer with \textbf{R}ank-1 Updates. MKOR unifies the \textbf{Sparsity} pillar, inherited through block-diagonalization, with the \textbf{Low-Rank Approximation} pillar, which approximates the inverse of the covariance blocks using rank-1 updates via the Sherman-Morrison identity. This formulation fundamentally alters the efficiency landscape, reducing the inversion complexity from $\mathcal{O}(d^3)$ to $\mathcal{O}(d^2)$ while simultaneously alleviating the communication bottleneck. Unlike standard second-order methods that require synchronizing large inverse factors ($\mathcal{O}(d^2)$), MKOR synchronizes only the rank-1 approximation vectors, reducing communication costs to $\mathcal{O}(d)$. These efficiency gains allow MKOR to update second-order information up to 100 times more frequently compared to state-of-the-art implementations like KAISA~\cite{kaisa} and HyLo~\cite{hylo}. Crucially, unlike recent attempts like Eva~\cite{eva} which store vectors but sacrifice momentum, MKOR fully preserves momentum information while achieving $\mathcal{O}(d^2)$ complexity.

The most closely related method to MKOR is Eva 
\cite{eva}, which also targets the scalability 
bottleneck of KFAC by maintaining only rank-1 
approximation vectors rather than full covariance 
factor inverses. This gives Eva a memory overhead 
of only $O(d)$, significantly lower than MKOR's 
$O(d^2)$. However, this aggressive memory reduction 
comes at a fundamental cost: Eva discards the 
accumulated momentum of the inverse factors entirely, 
retaining only the most recent rank-1 snapshot. In 
contrast, MKOR preserves full momentum history in 
its factor inverses via the Sherman-Morrison update 
(\autoref{eq:rank1_left_factor_inversion} and \autoref{eq:rank1_right_factor_inversion}), 
which exponentially averages past curvature information 
through the $\gamma$ decay term. As we demonstrate 
in \autoref{sec:experiments}, this distinction 
has practical consequences: Eva fails to converge to 
the target accuracy on several benchmarks where MKOR 
succeeds, suggesting that the memory savings come at 
the expense of optimization stability. MKOR thus 
occupies a deliberate design point in the 
memory--convergence tradeoff, investing $O(d^2)$ 
memory to retain curvature history that proves 
essential for reliable convergence at scale.

However, low-rank approximation alone is insufficient to fully resolve the scalability challenge. While it reduces the computational cost of inversion, it does not inherently address the memory overhead of storing the KFAC factors. To tackle this, we integrate the \textbf{Quantization} pillar. By performing computations and storage in half-precision, we significantly reduce the memory footprint of the curvature factors. This is non-trivial because standard second-order methods rely on operations like Cholesky decomposition or matrix inversion, which are numerically unstable in low precision. By pivoting to rank-1 updates via the Sherman-Morrison identity, MKOR replaces these unstable operations with simple matrix-vector products. These operations are inherently more robust to quantization noise, enabling us to store and compute curvature in half-precision without divergence.

While second-order methods provide a significant advantage in the initial phase of training, their relative benefit over first-order methods can diminish in later stages. To maximize efficiency, we introduce a hybrid variant, MKOR-H. This method combines the rapid initial convergence of second-order optimization with the low computational overhead of first-order methods. By utilizing a loss-reduction-rate-based switching mechanism, MKOR-H automatically transitions between regimes to ensure optimal resource utilization throughout the entire pretraining process.

Our experiments demonstrate that MKOR successfully validates the efficacy of applying the Compression Trinity to the optimizer. MKOR outperforms state-of-the-art distributed second- and first-order methods by up to $2.57\times$, reducing the training time of BERT-Large-Uncased from 8 hours to 3 hours on 64 A100 GPUs. Additionally, it achieves new state-of-the-art metrics on the GLUE dataset, successfully converging in settings where other second-order methods such as KFAC fail.

\section{Background}
\label{section:background}

Training a neural network involves solving an optimization problem to find the optimal values for a set of weights $\mathcal{W} = \{W^m\}_{m = 1}^{M}$, where $M$ is the number of layers in the network and $W^m$ is a matrix in $\mathbb{R}^{d \times d}$. Second-order methods precondition the weights of the network with the inverse of the Hessian for  better convergence rates.
Block-diagonal approximations of NGD methods  replace the Hessian with the block-diagonal FIM as shown in \autoref{eq:ngd}, where $w^m\in \mathbb{R}^{d^2}$ is the vector representation of  $W^m$, $F^m$ is the block corresponding to that layer and $\mathcal{L}$ is the loss function. Martens \cite{fim_vs_hessian} shows that the FIM  matches the Gauss-Newton matrix under certain conditions.

\vspace{-0.05in}

\begin{equation}
    \label{eq:ngd}
    w^m := w^m - \alpha {(F^m)}^{-1} \nabla_{w^m} \mathcal{L}
\end{equation}

KFAC-based methods reformulate the FIM block as the Kronecker product of two matrices. \autoref{eq:kfac_final} shows the update rule in KFAC, where $\mathcal{L}$ is the loss function and ${(L_t^m)}^{-1}$ and ${(R_t^m)}^{-1}$ are the inverses of the left and right factors, respectively.

\vspace{-0.05in}

\begin{equation}
    \label{eq:kfac_final}
    W^m := W^m - \alpha {(L_t^m)}^{-1} \nabla_{W^m} \mathcal{L} {(R_t^m)}^{-1}
\end{equation}

${(L_t^m)}^{-1}$ and ${(R_t^m)}^{-1}$ in \autoref{eq:kfac_final} are computed using \autoref{eq:left_factor_momentum} and \autoref{eq:right_factor_momentum}, respectively, where $a^m$ is the activation value of a sample at layer $m$, and $g_m = \nabla_{a^{m - 1}} \mathcal{L}$ and $\gamma$ incorporate the momentum feature to avoid extreme changes in the factors.

\begin{equation}
    \label{eq:left_factor_momentum}
    L_t^m = \gamma L_{t - 1}^m + (1 - \gamma) \mathbb{E} [ g_t^m {g_t^m}^T ]
\end{equation}
\begin{equation}
    \label{eq:right_factor_momentum}
    R_t^m = \gamma R_{t - 1}^m + (1 - \gamma) \mathbb{E} [a_t^{m - 1} {a_t^{m - 1}}^T ]
\end{equation}

\begin{figure}[tp]
    \centering
    \centerline{\includegraphics[]{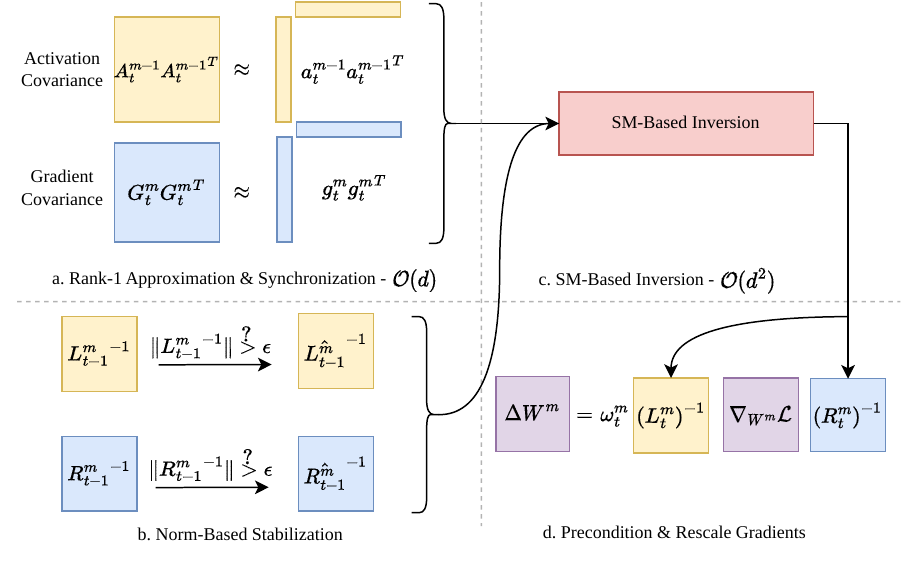}}
    \caption{MKOR  for layer $m$ on a single worker. The inputs of MKOR  are the activations $A_t^m$, the gradients of the loss function with respect to the inputs $G_t^m$, and the gradients of the loss function with respect to the weights $\nabla_{W^m} \mathcal{L}$. The output is the update values $\Delta W^m$.}
    \vspace{-0.1in}
    \label{fig:mkor}
\end{figure}

\section{Methodology}

In this section, we first present the MKOR algorithm, its computation and communication complexity, then present hybrid MKOR (MKOR-H), and finally discuss MKOR's convergence and stability. 

\subsection{The MKOR Algorithm}

\autoref{alg:mkor} summarizes the MKOR optimizer for a single layer and \autoref{fig:mkor} shows the workflow. For each layer (line 1 in \autoref{alg:mkor}) MKOR updates the second-order information and preconditions the gradients, and at the end the backend optimizer updates the weight using the preconditioned gradients (line 14 in \autoref{alg:mkor}).

\begin{algorithm}[tb]
\caption{MKOR Algorithm for a Single Layer $m$}
\label{alg:mkor}
\begin{algorithmic}[1]
\Statex \textbf{\color{KeywordColor}Input:} ${\color{VarColor}A_t^{m - 1}, G_t^m, W_{t-1}^m}$.
\Statex \textbf{\color{KeywordColor}Output:} ${\color{VarColor}W_t^m}$.
\vspace{0.5em}
\If{$m \in$ Second Order Layers}
\State ${\color{VarColor}\mathbf{a_t^{m-1}} \gets \frac{1}{b}\sum_{i=1}^b ({\color{VarColor}A_t^{m-1}})_{:, i}}$ \algorithmiccomment{Approx: $A_t^{m-1}{A_t^{m-1}}^T \approx \mathbf{a_t^{m-1}}\mathbf{a_t^{m-1}}^T$}
\State ${\color{VarColor}\mathbf{g_t^{m}} \gets \frac{1}{b}\sum_{i=1}^b ({\color{VarColor}G_t^{m}})_{:, i}}$ \algorithmiccomment{Approx: $G_t^{m}{G_t^{m}}^T \approx \mathbf{g_t^{m}}\mathbf{g_t^{m}}^T$}
\State ${\color{VarColor}\mathbf{a_t^{m-1}}, \mathbf{g_t^m} \gets}$ AllReduce$({\color{VarColor}\mathbf{a_t^{m-1}}}, {\color{VarColor}\mathbf{g_t^m}})$ \algorithmiccomment{Synchronize Approximations}
\State ${\color{VarColor}{\hat{L{t-1}^m}}^{-1} \gets }$ \text{\textbf{if}} {\color{VarColor} $|{L_{t-1}^m}^{-1}| > \epsilon$ } \text{\textbf{then}} {\color{VarColor} $\zeta {L_{t-1}^m}^{-1} + (1 - \zeta) I$ } \text{\textbf{else}} {\color{VarColor} ${L_{t-1}^m}^{-1}$ }  \algorithmiccomment{Norm-Based Stabilization}
\State ${\color{VarColor}{\hat{R_{t-1}^m}}^{-1} \gets }$ \text{\textbf{if}} {\color{VarColor} $|{R_{t-1}^m}^{-1}| > \epsilon$ } \text{\textbf{then}} {\color{VarColor} $\zeta {R_{t-1}^m}^{-1} + (1 - \zeta) I$ } \text{\textbf{else}} {\color{VarColor} ${R_{t-1}^m}^{-1}$ } 
\State ${\color{VarColor}{L_{t}^m}^{-1} \gets \gamma \hat{L_{t - 1}^m}^{-1} + \frac{(1 - \gamma)}{\gamma^2(1 + \gamma (1 - \gamma) \mathbf{g^m_t}^T\hat{L_{t - 1}^m}^{-1} \mathbf{g^m_t})} \hat{L_{t - 1}^m}^{-1} \mathbf{g^m_t} \mathbf{g^m_t}^T \hat{L_{t - 1}^m}^{-1}}$ \algorithmiccomment{SM-Based Factor Inversion}
\State $\color{VarColor}{R_t^m}^{-1} \gets \gamma \hat{R_{t - 1}^m}^{-1} + \frac{(1 - \gamma)}{\gamma^2(1 + \gamma (1 - \gamma) \mathbf{a^m_t}^T \hat{R_{t - 1}^m}^{-1} \mathbf{a^m_t})} \hat{R_{t - 1}^m}^{-1} \mathbf{a^m_t} \mathbf{a^m_t}^T \hat{R_{t - 1}^m}^{-1}$
\State ${\color{VarColor}\Delta \hat{W^m_t} \gets {L_{t}^m}^{-1} \nabla_{W^m} \mathcal{L} {R_{t}^m}^{-1}}$ \algorithmiccomment{Precondition Gradients}
\State ${\color{VarColor}\Delta W^m_t \gets \frac{|\nabla_{W^m} \mathcal{L} |}{\Delta \hat{W^m_t}} \Delta \hat{W^m_t}}$ \algorithmiccomment{Rescale Gradients}
\Else
\State ${\color{VarColor}\Delta W^m_t \gets\nabla_{W^m} \mathcal{L}}$
\EndIf
\State ${\color{VarColor}W_t^m \gets }$ Optimizer.step$({\color{VarColor}\Delta W^m_t}, {\color{VarColor}W_{t-1}^m})$
\vspace{0.5em}
\Statex \textbf{\color{KeywordColor}Return:} ${\color{VarColor}W_t^m}$.
\end{algorithmic}
\end{algorithm}

\paragraph{Rank-1 Approximation.}
For the rank-1 approximations of the covariance matrices, we use the average of the values across all the samples, i.e. $\mathbf{a_t^{m-1}} = \mathbb{E}[a_t^{m - 1}]$ and $\mathbf{g_t^{m}} = \mathbb{E}[g_t^{m}]$ (lines 2 and 3 in \autoref{alg:mkor} and \autoref{fig:mkor}-a). $(A_t^{m-1})^{-1}_{:, i}$ and $(G_t^m)^{-1}_{:, i}$ show the $i^{th}$ column of $(A_t^{m-1})^{-1}$ and $(G_t^m)^{-1}$ respectively, where $A_t^{m-1}$ and $G_t^{m}$ are the activations and the gradients of layer $m$ respectively.

\paragraph{Norm-Based Stabilizer.}
The values in the factor inverses in second-order methods can become large or vanish due to extremely large or small values in activations and gradients, leading to numerical instabilities and over/underflows. Since the inverse of the factors are directly multiplied by the gradients to find the update values, it can cause oscillations or even divergence. MKOR uses a norm-based stabilizer to detect the numerical instability and addresses it by modifying the inverse of the factors accordingly (lines 5 and 6 in \autoref{alg:mkor} and \autoref{fig:mkor}-b). More details on the norm-based stabilizer are in \autoref{sec:numerical_stability}.

\paragraph{SM-Based Inverter.}
MKOR directly modifies the inverse of the left and right factors using rank-1 updates, while using the momentum for better convergence. If $\mathbb{E}[g^m g^{m^{T}}]$ is approximated using a rank-1 matrix $\mathbf{g^m g^{m^T}}$ and using the Sherman-Morrison identity, \autoref{eq:rank1_left_factor_inversion} is obtained (line 7 in \autoref{alg:mkor} and \autoref{fig:mkor}-c).

\begin{equation}
    \label{eq:rank1_left_factor_inversion}
    {L_{t}^m}^{-1} = \gamma {L_{t - 1}^m}^{-1} + \frac{(1 - \gamma)}{\gamma^2(1 + \gamma (1 - \gamma) \mathbf{g^m_t}^T{L_{t - 1}^m}^{-1} \mathbf{g^m_t})} {L_{t - 1}^m}^{-1} \mathbf{g^m_t} \mathbf{g^m_t}^T {L_{t - 1}^m}^{-1}
\end{equation}

Furthermore, if \autoref{eq:right_factor_momentum} is approximated using $\mathbb{E} [a_t^{m - 1} {a_t^{m - 1}}^T ] \approx \mathbf{a_t^{m} a_t^{{m}^T}}$ with a similar derivation,   \autoref{eq:rank1_right_factor_inversion} is obtained (line 8 in \autoref{alg:mkor} and \autoref{fig:mkor}-c).

\begin{equation}
    \label{eq:rank1_right_factor_inversion}
    {R_t^m}^{-1} = \gamma {R_{t - 1}^m}^{-1} + \frac{(1 - \gamma)}{\gamma^2(1 + \gamma (1 - \gamma) \mathbf{a^m_t}^T {R_{t - 1}^m}^{-1} \mathbf{a^m_t})} {R_{t - 1}^m}^{-1} \mathbf{a^m_t} \mathbf{a^m_t}^T {R_{t - 1}^m}^{-1}
\end{equation}

\paragraph{Rescaling Gradients.}
Preconditioning the gradients using the computed factors can change   gradient norms. Sometimes, these changes interfere with the effect of the learning rate on the training process. To alleviate this and to make learning rate schedulers more effective, the preconditioned gradients are scaled so that their norm matches the original norms (line 10 in \autoref{alg:mkor} and \autoref{fig:mkor}-d).

\begin{table}
    \centering
    \caption{The computation and communication complexity and memory overhead of the state-of-the-art  implementations of the first- and second-order  (second-order optimizers are written in \textbf{bold}). The division by 2 in MKOR is because MKOR uses half-precision computations. The complexity of KFAC-based methods depends on layer dimensions while SNGD methods mostly depend on the batch size. In transformers, due to the scaling of the batch size by the sequence length, batch sizes and layer dimensions are comparable, making both KFAC- and SNGD-based methods   more expensive than SGD.}
    \label{table:complexity}
    \begin{tabularx}{1\textwidth} { 
        | >{\centering\arraybackslash}X 
        | >{\centering\arraybackslash}X
        | >{\centering\arraybackslash}X 
        | >{\centering\arraybackslash}X | }
        \hline
        Optimizer & Computational Complexity & Memory Overhead &Communication Complexity \\
        \hline
        \textbf{MKOR}                & $\mathcal{O}(d^2 + bd)$  & $\mathcal{O}(2d^2/2)$           & $\mathcal{O}(2d/2)$           \\
        \hline
        \textbf{SNGD (HyLo)}         & $\mathcal{O}(b^3)$       & $\mathcal{O}(2bd + b^2)$    & $\mathcal{O}(2bd + b^2)$    \\
        \hline
        \textbf{KFAC (KAISA)}        & $\mathcal{O}(d^3)$       & $\mathcal{O}(4d^2)$         & $\mathcal{O}(4d^2)$         \\
        \hline
        \textbf{Eva}                 & $\mathcal{O}(d^2 + bd)$  & $\mathcal{O}(2d)$           & $\mathcal{O}(2d)$ \\
        \hline
        SGD (Momentum)      & -              & $\mathcal{O}(d^2)$          & -                 \\
        \hline    
        ADAM / LAMB         & -              & $\mathcal{O}(d^2)$          & -                 \\
        \hline
    \end{tabularx}
    
\end{table}

\paragraph{Complexity Analysis.} MKOR reduces the memory, communication, and computation costs for factor inversion. \autoref{table:complexity} compares the overheads of different  optimizers. 
{\textit{(1) Computation Complexity.}} MKOR inverts the left and right factors in \autoref{eq:kfac_final} using  \autoref{eq:rank1_left_factor_inversion} and \autoref{eq:rank1_right_factor_inversion}, both of which can be computed using matrix-vector multiplications, and have $\mathcal{O}(d^2)$ computation complexity, in contrast to KFAC and SNGD methods that need $\mathcal{O}(d^3)$ and $\mathcal{O}(b^3)$ complexity to invert matrices in $\mathbb{R}^{d\times d}$ and $\mathbb{R}^{b\times b}$ respectively.
{\textit{(2) Communication Complexity.}} The only data that is synchronized among different workers in MKOR is the two rank-1 approximations that have $2d$ elements. With quantization, this size can be halved. In KFAC, the activation and gradient covariance matrices and the inversion of left and right factors need to be synchronized between all the workers, leading to $4d^2$ data transfers. In SNGD , the activations and gradients are synchronized, leading to $2bd$ data transfers and the inverted kernels are broadcast, resulting in $b^2$ data transfers. Reducing the communication complexity of MKOR from quadratic to linear results in better performance on large number of workers.
{\textit{ (3) Memory Overhead.}} MKOR needs to store the inverse of the left and right factors and two rank-1 approximation vectors, leading to $2d^2 + 2d$ memory overhead, and using half-precision computations further reduces this. KFAC  stores the activation and gradient covariance matrices and the left and right factors, leading to $4d^2$ memory overhead. SNGD    stores the activations, the gradients, and the kernels they use as second-order information, leading to $2bd + b^2$ memory complexity.

It is worth comparing MKOR's complexity profile 
directly with Eva, as both methods share the same 
$O(d^2 + bd)$ computational complexity and achieve 
linear communication costs. The key distinction lies 
in the memory--accuracy tradeoff. Eva achieves $O(d)$ 
memory by storing only the rank-1 approximation 
vectors and discarding the factor inverses after each 
preconditioning step. MKOR, by contrast, retains the 
full $d \times d$ inverse factors in half-precision, 
resulting in $O(d^2/2)$ memory but enabling momentum 
accumulation across iterations. While this is a 
meaningful increase in memory, it remains substantially 
lower than KFAC's $O(4d^2)$ overhead and, as shown 
in \autoref{sec:experiments}, translates to 
more stable convergence, particularly in settings 
where Eva's lack of momentum leads to suboptimal 
solutions or failure to reach the target accuracy.

\subsection{Hybrid MKOR}
We observed that second-order methods, including MKOR, usually accelerate training  more during the first iterations of the training time, and as the loss flattens, their advantage over their first-order counterparts becomes less noticeable. This is because the second-order information of the loss functions approach identity near convergence points.
Thus we designed a hybrid second- and first-order optimizer with a loss decrease rate-based switching method (MKOR-H). MKOR-H evaluates the changes in the loss function in different iterations and switches back to first-order methods if needed for an efficient trade-off between the costly second-order updates and their benefits for convergence.

\subsection{MKOR Convergence and Stability}
\paragraph{Inversion Frequency.}
Due to the high factor inversion costs in KFAC- and SNGD-based  methods, researchers use the stale factor approach, which updates the inverted factors every $f$ iterations and reuses the results in the other iterations in their preconditioning to reduce the computation and communication costs. The reciprocal of inversion frequency, $f$, varies from a few 100s to a few 1000s. Our experiments show that in average-sized models such as ResNet-50 \cite{resnet}, in an iteration that includes the inversion of factors, the cost of KAISA and HyLo is $150\times$ more than an SGD iteration that reuses stale factors. Furthermore, more than 98\% of the total cost in those iterations are spent on matrix inversion.

The stale factors approach can lead to good preconditioners if the loss function landscape does not vary significantly in each iteration. However, this is a  strong assumption and doesn't necessarily hold in practice. Also, increasing the inversion frequency can benefit the convergence rate of the second-order methods. In addition, our experiments show that using stale factors can  lead to converging to local minima in the loss function and damage the generalization of the model.

% intentionally empty (placeholder kept so the file survives archiving)

\paragraph{Numerical Stability.}
\label{sec:numerical_stability}
In second-order techniques, we need to invert or find the roots of matrices of different sizes, which are usually not full-rank, resulting in numerical issues. 
The KFAC implementation uses singular value decomposition (SVD) of the factors and masks the eigenvalues that are close to zero to deal with singular matrix inversion issues. In practice, the eigenvalues of the left and right factors in KFAC-based methods computed from \autoref{eq:left_factor_momentum} and \autoref{eq:right_factor_momentum} are increased manually by adding $\mu I$ to each of them to improve numerical stability ($\mu > 0$ is called the damping factor), but MKOR doesn't need such numerical fixes. Furthermore, HyLo uses two decomposition methods to sample the batch of inputs, namely KID and KIS. KID requires inverting matrices in $\mathbb{R}^{b \times b}$ of rank $min(b, d)$, thus for     batch sizes  larger than $d$ in a specific layer, the method fails.

Unlike SVD or other iterative methods used for factor inversion, MKOR doesn't suffer from numerical instabilities that rise from large condition numbers. MKOR has a single scalar division, in which the denominator is guaranteed to be non-zero based on \autoref{lemma:positive_definite}, eliminating the numerical over/under-flow possibility and the need for damping factors (required by other second-order methods for computational stability).

\begin{lemma}
    \label{lemma:positive_definite}
    The factors computed using \autoref{eq:rank1_left_factor_inversion} and \autoref{eq:rank1_right_factor_inversion} are all positive-definite.
\end{lemma}

Anil et al. \cite{scalable_shampoo} suggest using double precision representation of numbers to avoid numerical instabilities in inverting or computing the roots of matrices. This approach adds more costs to the matrix inversion and increases the time complexity of the main bottleneck in second-order methods.

MKOR does not need higher precision computations, and can use half-precision floating point operations to reduce  costs significantly. 
This will improve the memory utilization and reduce the communication costs in GPUs by $2\times$ while using cheaper computation blocks for half-precision operations. 
\autoref{lemma:quantization_error} shows an upper bound on the quantization error effect in the MKOR updates.

\begin{lemma}
    \label{lemma:quantization_error}
    Assuming that the maximum quantization error is $\epsilon$, the maximum number in matrices and vectors is $m$, and the dimension of the vectors and matrices are $d$ and $d\times d$ respectively, the quantization error of \autoref{eq:rank1_left_factor_inversion} and \autoref{eq:rank1_right_factor_inversion} is $O((\gamma + 4\frac{(1 - \gamma)}{\gamma^2} m^3 d^2) \epsilon)$
\end{lemma}

\paragraph{Exploding Gradients Problem.}
\label{section:exploding_gradient}
In second-order methods, where the gradients are preconditioned by various factors, the exploding gradient problem is worsened. Our experiments show that in first-order methods, by choosing a learning rate that doesn't lead to divergence in the first few iterations, explosion in gradients almost never occurs. On the other hand, in second-order methods, we observe that the explosion can occur at any iteration, and both KFAC and SNGD implementations are prone to this problem. This can lead to ripples in accuracy and divergence. 

One of the main approaches for solving the exploding gradient problem is choosing small values for the learning rate, limiting the convergence rate significantly. In particular, small learning rates damage the second-order methods and make them almost as performant as their first-order counterparts.

Considering that SGD is more robust against the exploding gradients and taking advantage of the direct control of MKOR on the inverse of the factors, the factors in MKOR are modified to lean toward SGD once the possibility of exploding gradients is detected using \autoref{eq:averaged_left_factor} and \autoref{eq:averaged_right_factor}, where $\zeta$ is a hyperparameter that controls  the amount of information from the original factors that needs to be saved in the new factors.

\begin{equation}
    \label{eq:averaged_left_factor}
    \hat{L_t^m} = \zeta L_t^m + (1 - \zeta) I
\end{equation}

\begin{equation}
    \label{eq:averaged_right_factor}
    \hat{R_t^m} = \zeta R_t^m + (1 - \zeta) I
\end{equation}

 By expanding \autoref{eq:kfac_final} with the new factors, we will get \autoref{eq:averaged_update_rule}, which  reduces the loss based on \autoref{lemma:loss_reduction}. The first term in the right-hand side of \autoref{eq:averaged_update_rule} is the KFAC term, the second and third terms are the left and right preconditioned versions, and the last term is the SGD term.

\begin{equation}
\label{eq:averaged_update_rule}
\begin{aligned}
\hat{L}^{m^{-1}} \nabla_{W^m} \mathcal{L} \hat{R}^{m^{-1}} &= \zeta^2 L^{m^{-1}} \nabla_{W^m} \mathcal{L} R^{m^{-1}} \\
&\quad + \zeta (1 - \zeta) L^{m^{-1}} \nabla_{W^m} \mathcal{L} + \zeta (1 - \zeta) \nabla_{W^m} \mathcal{L} {R^m}^{-1} + (1 - \zeta)^2 \nabla_{W^m} \mathcal{L}
\end{aligned}
\end{equation}

\begin{lemma}
    \label{lemma:loss_reduction}
    Given a differentiable function $\mathcal{L}(w)$ with first-order Taylor series approximation $\mathcal{\hat{L}}(w - \Delta w) = \mathcal{L}(w_0) - \Delta w^T \nabla_w \mathcal{L}(w_0)$ around point $w_0$, assuming that at point $w_0$ the second-order derivative of the function $\mathcal{L}(w)$ is given as $\nabla^2_w \mathcal{L}(w_0) = H = L \otimes R$, where $L$ and $R$ are positive-semi-definite matrices, for a value of $\Delta w = (( \zeta L^{-1} + (1 - \zeta) I) \otimes  (\zeta R^{-1} + (1 - \zeta) I)) \nabla \mathcal{L}(w_0)$, the inequality $\mathcal{\hat{L}}(w_0 - \Delta w) < \mathcal{L}(w_0)$ holds. 
\end{lemma}

While this modification can avoid exploding gradients, overusing it with small values of $\zeta$ will convert MKOR to SGD. MKOR uses a factor norm-based metric that observes the infinity norm of the  factors, and if they are greater than a specific threshold, the process of factor modification will be triggered.

\section{Experimental Results}
\label{sec:experiments}
In this section, we demonstrate the performance of MKOR on a large language model using different benchmarks, and analyze the timing of different components in different first- and second-order algorithms. For results on more models and training sets, please refer to \autoref{app:mkor}.

\niparagraph{Experiment Setup.} For the BERT-Large-Uncased pre-training and fine-tuning experiments, we use up to 64 A100 GPUs on the Polaris~\cite{polaris} cluster, which has 560 nodes each with 4 NVIDIA A-100 GPUs with NVLink interconnects. The rest of the experiments are conducted on the Mist cluster~\cite{computecanada}, with 54 nodes each having 4 NVIDIA V100 GPUs with 32GB memory and NVLink inter-node connections. Each training experiment is conducted 5 times and the median timing is reported; reported accuracies are the median across multiple runs. \autoref{table:models} summarizes the models, datasets, and GPU architectures used in our experiments. For BERT-Large-Uncased pre-training, we use the same hyperparameters as~\cite{nvidia-bert}, with factors in KAISA updated every 50 iterations and factors in MKOR and MKOR-H updated every 10 iterations. For ResNet-50, we follow the hyperparameters from~\cite{kaisa}, with MKOR factors updated every 10 iterations and the learning rate decaying by a factor of 2 at the end of epochs 25, 35, 40, 45, 50, 55, and 56. Our code base is publicly available at \url{https://github.com/Mohammad-Mozaffari/mkor}.

\begin{table}
    \centering
    \caption{List properties of the models, datasets, and settings used in our experiments.}
    \begin{tabularx}{\textwidth} {
        | >{\centering\arraybackslash \hsize=1.6\hsize}X
        | >{\centering\arraybackslash \hsize=1.0\hsize}X
        | >{\centering\arraybackslash \hsize=1.6\hsize}X
        | >{\centering\arraybackslash \hsize=0.6\hsize}X
        | >{\centering\arraybackslash \hsize=0.6\hsize}X
        | >{\centering\arraybackslash \hsize=0.7\hsize}X
        | >{\centering\arraybackslash \hsize=0.9\hsize}X | }
        \hline
        \multicolumn{2}{|c|}{Model}   & \multicolumn{3}{c|}{Dataset}   & \multicolumn{2}{c|}{GPU} \\
        \hline
        Name    & \#Parameters  & Name & Train & Test & Arch  & \# \\
        \hline
        BERT-Large-Uncased & 335.1M & Wikipedia - BookCorpus &   -   & - &   A100 &  64 \\
        \hline
        ResNet-50   &   25.5M   & ImageNet  &   1.2M   &   50k   &   V100    &   64   \\
        \hline
        AlexNet &   20.3M  & \footnotesize{CIFAR-100} & 50K & 10K & V100 & 4 \\
        \hline
        BERT-Base-Cased & 108.9M & SQuAD v1.1 & 87.6K & 10.6K & V100 & 4
        \\
        \hline
        BERT-Large-Cased & 335.1M & IMDB & 25K & 25K & V100 & 4 \\
        \hline

    \end{tabularx}
    \label{table:models}
\end{table}

\paragraph{Large Language Models.} We pre-train BERT-Large Uncased and fine-tune it for different question-answering and text classification tasks. We use a setup similar to KAISA ~\cite{kaisa} for pre-training and fine-tuning. We use Fused LAMB~\cite{lamb} as the state-of-the-art first-order baseline. Similar to KAISA ~\cite{kaisa}, for the pre-training process, we use the English Wikipedia~\cite{english-wikipedia} and the Toronto BookCorpus~\cite{bookcorpus} dataset. These datasets were used in the original BERT pre-training; the latter dataset is not fully available, which results in a small reduction in the baseline accuracies achieved in our experiments from the original BERT results. Following KAISA ~\cite{kaisa}, due to the  time-intensive process of hyperparameter tuning for the first phase of pre-training, we report the effectiveness of MKOR in the second phase of pre-training only while using the checkpoints of the first phase generated using the LAMB optimizer. 
As expected, the computation, communication, and memory complexity of HyLo is  high, and the Khatri-Rao-based Interpolative Decomposition (KID) approximation method,  the main idea of HyLo, cannot be executed because a single sample cannot fit into the 40GB memory of an A100 GPU. In addition, HyLo doesn't support gradient accumulation due to its memory complexity, depending on the batch size; in LLMs such as BERT, the batch sizes are as large as 64k.\footnote{We define the convergence of BERT as the iteration in which the downstream accuracy reaches the first-order baseline. Nevertheless, we continue training until the same iteration for additional ablations and testing whether the second-order baselines can improve the capabilities of the model further.}

For the question answering task, we fine-tune the pre-trained BERT checkpoints on the SQuAD v1.1~\cite{squad} dataset. \autoref{table:squad} shows the F1 Score achieved using different optimizers and compares their convergence rate and speedups.\footnote{All the timings and speedups reported are for the pretraining phase. We omit the fine-tuning time since it's negligible in comparison to the pretraining costs.} Convergence is defined as the number of iterations it takes the model to reach the same accuracy as the first-order optimizer. The vanilla MKOR and KAISA both converge after 1000 iterations, while the LAMB optimizer requires $1,563$ steps. Considering that each step in MKOR is faster than KAISA, MKOR achieves an end-to-end speedup. MKOR-H will converge in 600 steps, reducing the number of steps in LAMB by $2.6\times$, while achieving the same accuracy. In addition, it  achieves $2.57\times$ speedup over the LAMB optimizer and $1.75\times$ speedup over KAISA. As another second-order baseline, we consider Eva, which converges in 1000 iterations, and MKOR-H achieves $1.69\times$ speedup over it.

\begin{table}
    \centering
    \caption{BERT-Large Uncased results on SQuAD v1.1 question answering task}
    \label{table:squad}
    \begin{tabularx}{\textwidth} { 
        | >{\centering\arraybackslash}X 
        | >{\centering\arraybackslash}X 
        | >{\centering\arraybackslash}X 
        | >{\centering\arraybackslash}X 
        | >{\centering\arraybackslash}X
        | >{\centering\arraybackslash}X | }
        \hline
        Metric          & LAMB  & KAISA & MKOR  & MKOR-H & Eva \\
        \hline
        F1              & 90.44 & 90.44 & 90.50 & 90.64 & 90.55     \\
        \hline
        \# Iterations   & 1,563 & 1,000  & 1,000  & 600 & 1,000     \\
        \hline
        Time (h)            & 7.97     & 5.71     & 5.25     & 3.10 & 5.24     \\
        \hline
        Speedup ($\times$)        & 1.00   & 1.39     & 1.51     & 2.57  & 1.52    \\
        \hline
        
    \end{tabularx}
\end{table}

For classification tasks, we fine-tune BERT on the GLUE~\cite{glue} dataset. \autoref{table:glue_speedup} compares the results for different classification tasks in the GLUE dataset. MKOR with 1500 steps achieves a new state-of-the-art accuracy in GLUE dataset on BERT-Large Uncased, and MKOR and MKOR-H with 600 steps achieve the same average metric as the baseline, while reducing the number of steps by a factor of $2.6\times$. MKOR and MKOR-H both achieve $2.57\times$ end-to-end speedup. After training KAISA for 1,563 steps,  the model does not converge to the baseline average accuracy, while slowing down the convergence by $0.89\times$. Eva requires 1000 steps to converge to the target average metric, being $1.69\times$ slower than MKOR-H with 600 steps and $1.24\%$ less accurate than MKOR with 1500 steps (it is noteworthy that the accuracy of the model plateaus when using more iterations with Eva).

\begin{figure}
    \centering
    BERT-Large-Uncased Training Loss
    \centerline{
        \includegraphics[scale=1]{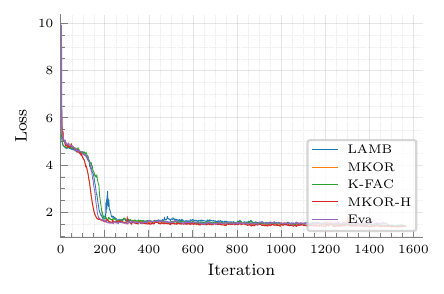}
        \includegraphics[scale=1]{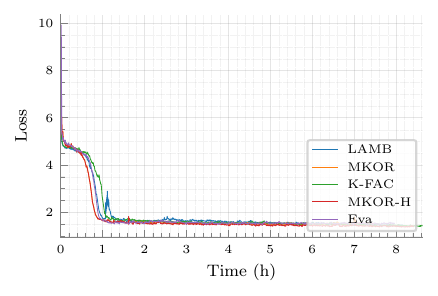}
    }
    \caption{The pre-training loss of BERT-Large-Uncased using different optimizers.}
    \vspace{-0.1in}
    \label{fig:mkor_bert_loss}
\end{figure}

Per \autoref{fig:mkor_bert_loss}, which shows the pre-training error during the training of BERT, MKOR decreases the error in fewer iterations in comparison to KAISA, Eva, and LAMB, leading to faster convergence. From \autoref{table:squad} and \autoref{table:glue_speedup}, MKOR-H converges in only 600 steps.

\begin{table}
    \centering
    \caption{BERT-Large Uncased results on the GLUE classification tasks. We report the average of the metrics of different GLUE tasks (accuracy, F1 score, etc) for easier comparison.
    }
    \label{table:glue_speedup}
    \begin{tabularx}{1\textwidth}{
        | >{\centering\arraybackslash}p{2.8cm}
        | >{\centering\arraybackslash}X
        | >{\centering\arraybackslash}X
        | >{\centering\arraybackslash}X
        | >{\centering\arraybackslash}X
        | >{\centering\arraybackslash}X
        | >{\centering\arraybackslash}X |}
        \hline
        Optimizer & LAMB & KAISA & MKOR & MKOR & MKOR-H & Eva \\
        \hline
        Iterations & 1,563 & 1,563 & 1,500 & 600 & 600 & 1000 \\
        \hline
        Time (h) & 7.97 & 8.93 & 7.88 & 3.10 & 3.10 & 5.24 \\
        \hline
        Speedup ($\times$) & 1.00 & 0.89 & 1.01 & 2.57 & 2.57 & 1.52 \\
        \hline
        Average Metric & 0.8023 & 0.796 & 0.8214 & 0.8078 & 0.811 & 0.809 \\
        \hline
    \end{tabularx}
\end{table}

\paragraph{ResNet-50 Experiments.}
We train ResNet-50, a convolutional neural network with more than 25M parameters, on ImageNet, an image classification task with more than 1.2M samples. The same setup is used in~\cite{kaisa}, and  SGD is used as the first-order baseline. 

The target accuracy in this experiment is 75.9\%. MKOR converges to this target accuracy in 57 epochs, while SGD, the first-order baseline, achieves this accuracy in 88 epochs. MKOR achieves $1.49\times$ speedup over SGD. KAISA, the second-order baseline converges in 54 epochs, but due to its expensive steps, MKOR still converges $1.04\times$ faster than KAISA. We do not compare to HyLo because HyLo is not able to achieve the target accuracy for ResNet (it reaches 75.6\% with tuning as reported in~\cite{hylo} and our experiments confirm it). As shown, the effect of complexity reduction and improvement in performance in MKOR is less obvious in ResNet because the model dimension ($d$) is smaller compared to LLMs such as BERT. Please see \autoref{table:complexity} for comparison of complexity between methods. 

We could not reproduce the ResNet-50 results of Eva ~\cite{eva} on ImageNet because the hyperparameters are not reported. We tried to tune Eva on multiple settings and none converged to desired accuracy. It is important to note Eva is not comparing results with the most efficient implementation of KFAC. The KFAC version used in Eva is from ~\cite{kfac}, dated to 2015. A number of followup works, mentioned in \autoref{section:mkor_intro}, have provided faster implementations of KFAC. We use KAISA ~\cite{kaisa}, the state-of-the-art implementation of KFAC. Also from discussions with KAISA authors and our own experiments the optimal inversion frequency for KFAC is 200. Eva uses an inversion frequency of 50 for KFAC, which makes KFAC slower.

\begin{figure}[ht]
    ResNet-50 Test Accuracy
    \centerline{
        \includegraphics[scale=1]{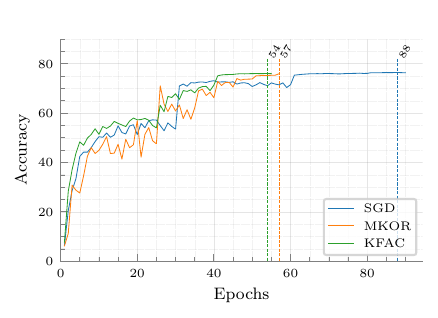}
        \includegraphics[scale=1]{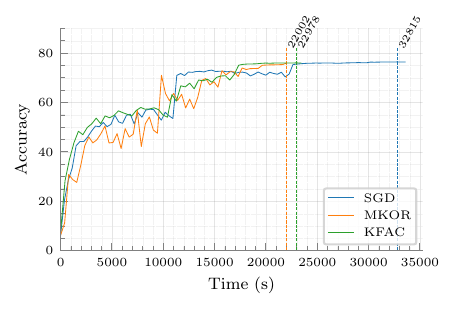}
    }
    (a){\hspace{.5\linewidth}}
    (b){\hspace{.5\linewidth}}
    \centering
    \caption{Test accuracy of ResNet-50 on ImageNet for MKOR, KAISA, and SGD on 64 GPUs.}
    \label{fig:resnet}
\end{figure}

\begin{figure}[ht]
    \centering
    \centerline{
        \includegraphics[scale=1.15]{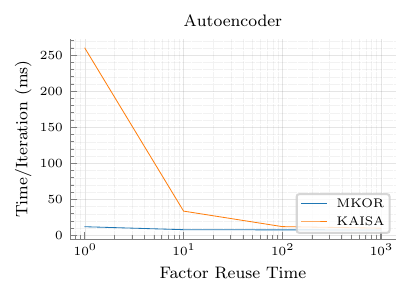}
        \includegraphics[scale=1.15]{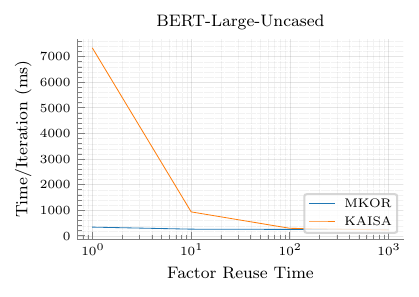}
    }
    (a) - Average Time per Iteration
    \\
    \centerline{
    \includegraphics[scale=1.15]{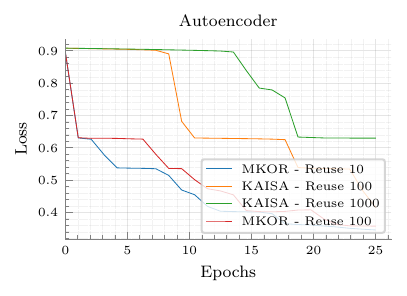}
    \includegraphics[scale=1.15]{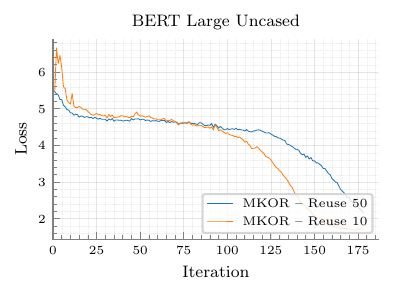}
    }
    (b) - Test Loss
    \caption{The sensitivity of MKOR and KAISA for BERT-Large-Uncased and an Autoencoder model (\textbf{a}) and the effect of inversion frequency on the convergence properties of these models (\textbf{b}).}
    \vspace{-0.3in}
     \label{fig:inversion_freq}
\end{figure}

\paragraph{Inversion Frequency.}
Due to the low computation complexity of the updates on MKOR, the factor inversion frequency ($f$) in MKOR is in the range of 10.
\autoref{fig:inversion_freq}-a shows that while the average iteration cost in KAISA is heavily dependent on the inversion frequency, MKOR's cost is almost independent of the inversion frequency. Also \autoref{fig:inversion_freq}-b shows that increasing the inversion frequency leads to higher convergence rate. In addition, using stale factors may result in converging to a local minima. Hence, in MKOR we increase the convergence rate by updating the factors more frequently, without affecting the per-iteration cost, leading to end-to-end speedups in training. We use a simple autoencoder~\cite{autoencoder} on CIFAR-100~\cite{cifar} in this experiment.

\begin{figure}[tb]
    \centering
    \centerline{
        \includegraphics[scale=1.1]{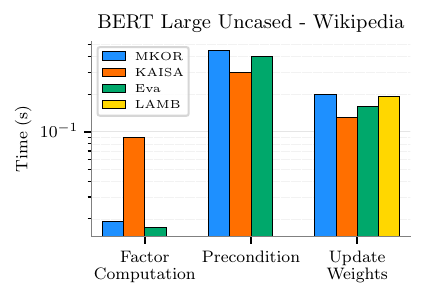}
        \includegraphics[scale=1.1]{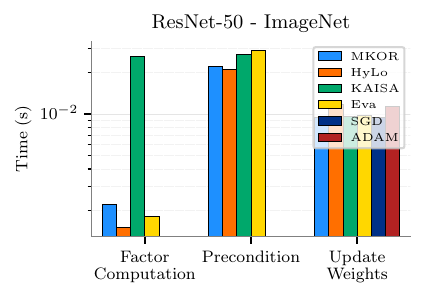}
    }
    \hspace{0.07\linewidth}(a) \hspace{.47\linewidth}
    (b) 
    \caption{Per-step breakdown of different optimizers on BERT-Large-Uncased (\textbf{a}) and ResNet-50 (\textbf{b}). The times reported in these graphs reflect only the optimizer computations. The majority of the training time is spent on the model’s forward and backward passes, which are identical across all optimizers and are not included here.}
    \vspace{-0.2in}
    \label{fig:time_breakdown}
\end{figure}

\paragraph{Performance Analysis.} We compare the performance of different parts of the optimizers to illustrate the bottlenecks and advantages of different methods. The training process for an optimizer has three steps: factor computation, precondition, and update weights. \autoref{fig:time_breakdown} shows the time spent on each task in different optimizers on two  models; BERT-Large-Uncased, a transformer-based LLM with large sequence length and ResNet-50, a CNN. Since first-order optimizers such as SGD, ADAM, and LAMB don't require factorization and preconditioning, their optimization time is only spent in updating the weights. In ResNet-50, since the model size is larger compared to the batch size, the factor computation and inversion is more expensive for KAISA compared to HyLo. This cost is significantly reduced in MKOR.

For BERT-Large-Uncased, because of the large size of the model, the factor inversion time for KAISA is large. Also, due to the large sequence length value in this model, the kernel inversion time for HyLo is comparable to KAISA's inversion time.  But as expected, because of its low computational complexity, the aforementioned cost in our method is much smaller than the total training time, leading to speedups. It is important to note that HyLo diverges in this training process, hence convergence time is not reported for HyLo.

The preconditioning and weight updates for the different methods are similar; hence, not much variation is observed.

\paragraph{Memory Overheads.}

The memory overheads of MKOR in comparison to other optimizers are reported in \autoref{table:Memory}. It can be observed that all the second-order methods have significant memory overheads compared to the first-order methods, but MKOR's overhead is up to 1.5$\times$ lower than KFAC/KAISA.

\begin{table}
    \centering
    \caption{Per-GPU memory usage (in GB) for MKOR, KFAC/KAISA, LAMB, and SGD on BERT-Large-Uncased pre-training and ResNet-50 training on ImageNet.
    }
    \label{table:Memory}
    \begin{tabularx}{1\textwidth} {
        | >{\centering\arraybackslash}X
        | >{\centering\arraybackslash}X
        | >{\centering\arraybackslash}X
        | >{\centering\arraybackslash}X
        | >{\centering\arraybackslash}X | }
        \hline
        Model    & MKOR & KFAC/KAISA   & LAMB   & SGD \\
        \hline
        ResNet-50         & 3.88      & 5.83       & -         & 3.01                 \\
        \hline
        BERT        & 23.34      & 29.97       & 12.80         & -                  \\
        \hline
    \end{tabularx}
\end{table}

\begin{figure}
    \centering
    \centerline{
        \includegraphics[scale=1.15]{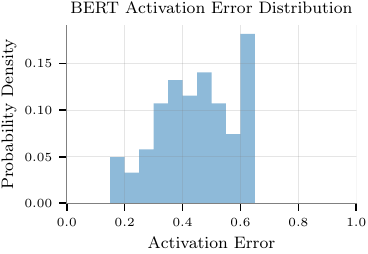}
        \includegraphics[scale=1.15]{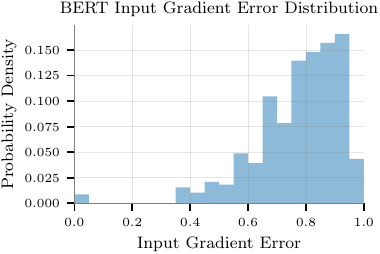}
    }
    \hspace{.05\linewidth}
    (a) \hspace{.42\linewidth}
    (b) 
    \centerline{        
        \includegraphics[scale=1.15]{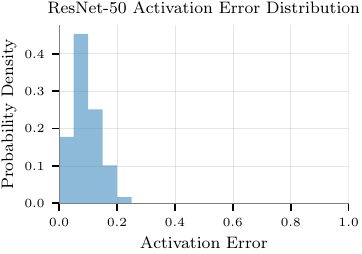}
        \includegraphics[scale=1.15]{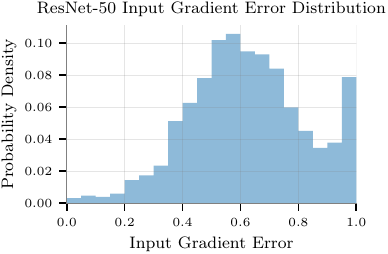}
    }
    \centering
    \hspace*{.05\linewidth}
    (c) \hspace{.42\linewidth}
    (d) 
    \caption{Rank-1 error for activation and input gradient covariance matrices for BERT-Large-Uncased pre-training (\textbf{a, b}) and ResNet-50 on ImageNet (\textbf{c, d}).}
    \vspace{-0.2in}
    \label{fig:rank_1_error}
\end{figure}

\paragraph{Approximation Error Experimental Results.}
Due to the low-rank properties of the covariance matrices, MKOR utilizes rank-1 approximations of the covariance matrices to accelerate the computations and communication in KFAC-based optimizers. Here, we aim to theoretically and experimentally support this choice.
As shown in \autoref{fig:rank_1_error}, our experiments show that the covariance matrices can be approximated with rank-1 matrices with low error and higher rank approximations are unnecessary in practice. \autoref{fig:rank_1_error} shows the error distribution of the optimal rank-1 approximation methods of the covariance matrices in ResNet-50 and BERT-Large-Uncased pre-training. Our extensive tests on well-known benchmarks show this property holds for all models and we have not come across a benchmark that does not have low-rank covariance matrices.

\paragraph{Approximation Error Analysis and Extension to Higher Ranks.} Small batch sizes and over parameterization of   networks will lead to low-rank covariance matrices in DNNs. Let’s consider the covariance matrix $C=XX^T$, where $C \in R^{d \times d}$ is the covariance matrix and $X \in R^{d \times b}$ is a matrix in which each column corresponds to a single sample and $d$ and $b$ are the sample dimension and the per-GPU batch size respectively. Rank of the covariance matrix  is $min(b, d)$. If the per-GPU batch sizes are small, the covariance matrices in each GPU will be low-rank. Rank-1 approximation methods can work well in these scenarios. If the batch sizes in each GPU are large, we observe that the covariance matrices will stay low-rank. The underlying reason for this observation is that current neural networks are over-parameterized, and as a result, different features in the covariance matrices of the activations and the output gradients won’t be linearly independent, resulting in low-rank covariance matrices. 

\paragraph{Extending MKOR to Higher Ranks:} Furthermore, one can extend MKOR to use higher-rank covariance matrices. Let’s assume that $C = \sum_{i=1}^{r}{c_i c_i^T}$ where $r$ is the rank of the covariance matrix $C$. We can apply SMW identity to compute $C_1^{new} = (C^{old} + c_1 c_1^T)^{-1}$ with $O(d^2)$ computational complexity. Then we can compute $C_2^{new} = (C_1^{new} + c_2 c_2^T)^{-1}$ using SMW identity with $O(d^2)$ computational complexity. We can continue the same pattern by computing $C_i^{new} = (C_{i-1}^{new} + c_i c_i^T)^{-1}$. The total computation complexity of this process will be $O(rd^2)$. We should add this cost to the cost of computing the low-rank approximation of $C$ which requires an SVD. Using SVD kills the main advantage of using low-rank computations, since the computational complexity of applying SVD is the same as inverting the factors directly. We could not find any cheaper way to compute low-rank approximations of the covariance matrices, except for the rank-1 approximation used in this chapter. 

\vspace{-0.15in}

\section{Conclusion}
\label{section:mkor_conclusion}

In this chapter, we presented MKOR, a scalable second-order optimizer that effectively executes the second strategy of the Compression Trinity: accelerating pretraining by reducing the number of required iterations. By applying the Trinity's pillars directly to the optimizer's internal mechanics, MKOR overcomes the historical bottlenecks of second-order methods. We leveraged low-rank approximations via rank-1 updates to reduce inversion complexity from $\mathcal{O}(d^3)$ to $\mathcal{O}(d^2)$, and employed stability mechanisms to enable lower-precision communication, reducing overheads to $\mathcal{O}(d)$. Our experiments confirm that MKOR significantly outperforms state-of-the-art first- and second-order optimizers, delivering up to $2.57\times$ faster training for large language models.

However, accelerating convergence is only half of the pretraining equation. While MKOR reduces the total number of steps, the computational cost \textit{per step} remains dominated by the dense matrix multiplications inherent to the Transformer architecture. To unlock the full potential of the Compression Trinity during the pretraining phase, we must also address Strategy 1: reducing the fundamental FLOPs and memory traffic of the linear layers themselves. In the next chapter, we introduce \slope, which extends the Trinity from the optimizer to the model weights, jointly applying sparsity and lazy low-rank adapters to accelerate the training dynamics without sacrificing model quality.

\newcommand{\slopespace}{\textsc{SLoPe}\xspace}
\newcommand{\transposable}{FST}

\newcommand\mohammad[1]{\noindent{{#1}}}
\newcommand{\xx}[1]{\texttt{#1}\xspace}
\newcommand\newtext[1]{\noindent{{#1}}}
\newcommand\mymapsto{\mathrel{\ooalign{$\rightarrow$\cr
  \kern-.15ex\raise.275ex\hbox{\scalebox{1}[0.522]{$\mid$}}\cr}}}

\newcommand{\trainspeedup}{1.25}
\newcommand{\inferencespeedup}{1.54}
\newcommand{\trainmemory}{0.63}
\newcommand{\inferencememory}{0.61}

\chapter{\slope: Double-Pruned Sparse Plus Lazy Low-Rank Adapter Pretraining of LLMs}
\label{chapter:slope}

\paragraph{Publication and Contributions.}
The content of this chapter is based on the paper 
``SLOPE: Double-Pruned Sparse Plus Lazy Low-Rank 
Adapter Pretraining of LLMs'' \cite{slope}, published at the Thirteenth International Conference on Learning Representations (ICLR) 2025. 
This work was conducted in collaboration with 
Amir Yazdanbakhsh, Zhao Zhang, and Maryam Mehri 
Dehnavi. Mohammad Mozaffari was the lead contributor, 
responsible for the algorithm design, implementation, 
and experimental evaluation. Amir Yazdanbakhsh and 
Maryam Mehri Dehnavi supervised the project and 
contributed to the writing and revision of the 
manuscript. Zhao Zhang provided additional supervisory 
guidance.

\section{Introduction}
\label{section:slope_intro}

Following our exploration of optimizer-level acceleration in \autoref{chapter:mkor}, we now turn to the first strategy of the Compression Trinity: accelerating the computational cost of each individual training iteration. Large Language Models (LLMs) require massive resources for their life-cycle stages, specifically pretraining~\cite{pretraining} on high-quality text~\cite{pile, openwebtext} and fine-tuning on downstream tasks~\cite{glue, squad}. These phases are dominated by three intensive matrix multiplications per layer: the forward pass, the backward pass for input gradients, and the backward pass for weight gradients. To execute Strategy 1 (Accelerating Per-Iteration Computation), we must reduce the FLOPs and memory traffic for all three operations. The first pillar of our Trinity, sparsity, offers a path forward~\cite{sparsity_survey}. While unstructured sparsity lacks hardware support~\cite{lucas} and rigid block-structured sparsity damages accuracy~\cite{block_pruning1, channel_pruning1, pixelated_butterfly}, semi-structured N:M sparsity (e.g., 2:4, where 2 out of 4 consecutive elements are set to zero) strikes a balance. It is flexible enough to preserve model quality while being structured enough for acceleration via NVIDIA's Sparse Tensor Cores~\cite{sparse_tensor_cores}, with algorithms rapidly evolving for these patterns~\cite{n_m4,n_m5_step,gradient_flow}.

However, applying N:M sparsity to the training phase faces a critical "transposability" bottleneck. While N:M sparsity successfully accelerates the forward pass, it fails to accelerate the backward pass because the row-wise N:M structure is destroyed when the weight matrix is transposed. Prior attempts to address this have focused on finding "transposable masks" that maintain structure in both orientations~\cite{n_m1,n_m2,2_4_pretraining}. Unfortunately, these methods often require expensive search algorithms or enforce rigid constraints that significantly reduce model accuracy. Paradoxically, the overhead of these complex mask searches can result in severe training slow-downs, up to $8.4\times$~\cite{2_4_pretraining}. Alternative approaches that change the sparsity mask dynamically~\cite{adaptive_pruning, nm_scratch, n_m4, n_m5_step} also introduce computational overheads and waste resources training weights that are eventually pruned.

To strengthen the sparsity pillar for pretraining, we propose a novel \textbf{double-pruned backward pass} formulation with theoretical convergence guarantees. Instead of enforcing the restrictive condition that a mask must be inherently transposable, our approach allows the forward pass to use a standard N:M mask. In the backward pass, we transpose the weight matrix first and then impose a new N:M sparsity pattern. This formulation allows the weight matrices to exhibit a much wider range of sparsity patterns compared to rigid transposable masks, leading to significantly improved accuracy while enabling acceleration in both directions.

While resolving the compute bottleneck, aggressive sparsity can still lead to an accuracy gap compared to dense models. To bridge this gap without sacrificing efficiency, we integrate the third pillar of the Trinity: Low-Rank Approximations. Previous methods often resort to dense fine-tuning to recover accuracy~\cite{spdf, 2_4_pretraining}, but this converts the model back to a dense state, negating all memory and compute savings during inference. The intuition behind this integration lies in the observation that, while the parameter space of LLMs is vast, learning effectively occurs on a manifold of much lower intrinsic dimension~\cite{intrinsic_dim, lora}. This suggests that full-rank updates are not strictly necessary for recovering the expressivity lost to pruning.

However, combining these pillars is non-trivial. Theoretically, a low-rank adapter $LR$ is a dense matrix; adding it to a sparse weight matrix $W$ ($W' = W + LR$) would cause "fill-in," where strictly zero elements become non-zero, effectively destroying the sparsity pattern and its associated hardware benefits. To harness the power of both without this collision, we
propose \textbf{Lazy Low-Rank Adapters}. We treat the sparse weights and dense adapters as parallel computational paths rather than a merged tensor. Furthermore, unlike standard adapters, ours are "lazy" because they are introduced only during the final 1\% of pretraining iterations. Our experiments show that these adapters converge noticeably faster compared to the original model parameters at the same parameter count. This approach improves the accuracy of the models while ensuring the base model remains sparse and efficient for deployment.

\begin{figure*}[t]
\begin{center}
\centerline{\includegraphics[width=0.95\textwidth]{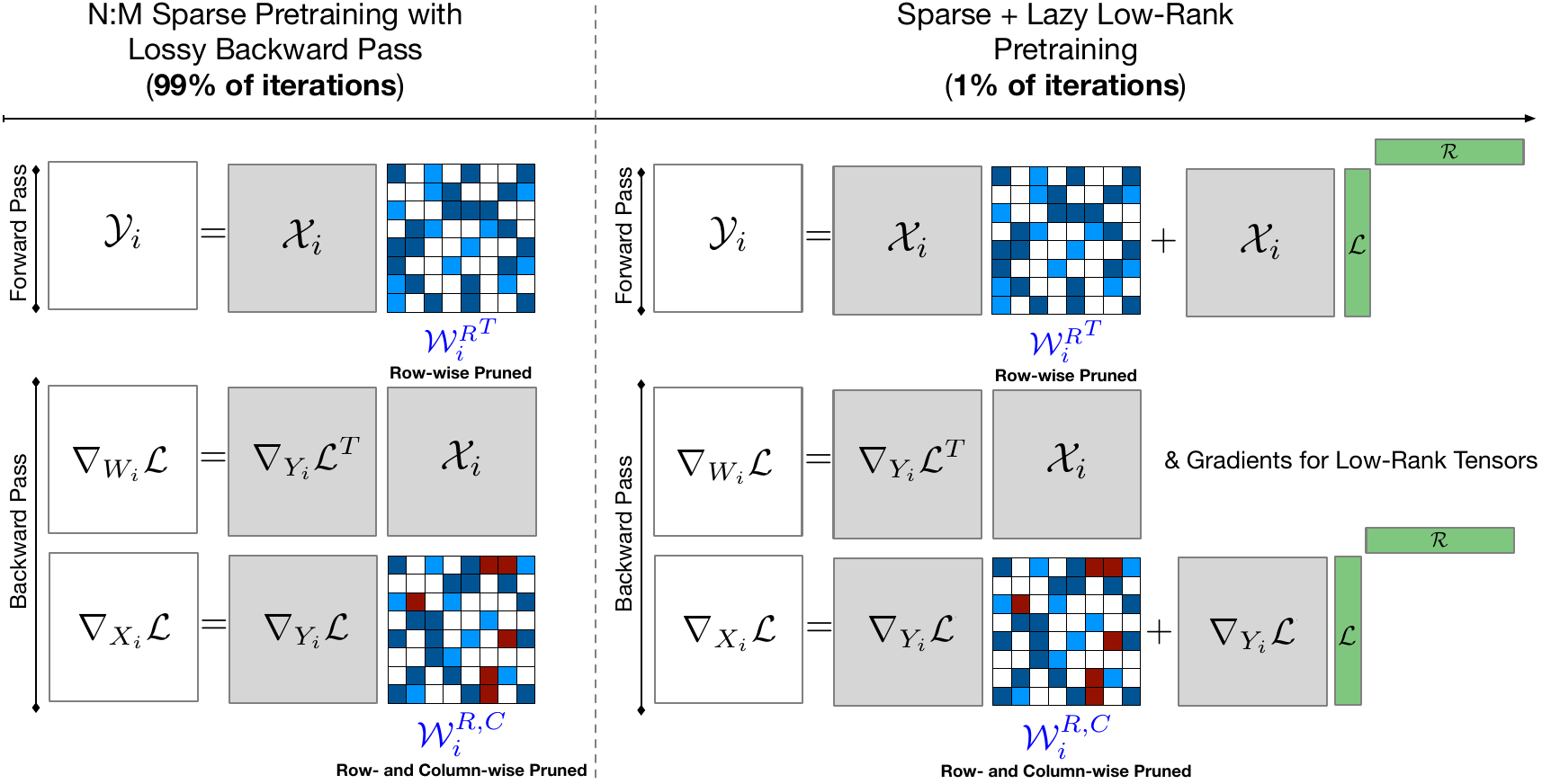}}
\caption{The sparse training pipeline in \slopespace.
Here, $\mathcal{X}$, $\mathcal{Y}$, and $\mathcal{W}$ denote the input, output, and the weight tensors for a specific layer, respectively.
$\nabla_{\cdot}\mathcal{L}$ represents the gradient of the loss function. $\mathcal{L}$ and $\mathcal{R}$ are the low-rank terms that are introduced only in the final 1\% iterations.
Superscript $R$ shows row-wise pruning using $N$:$M$ scheme and $R,C$ shows both column and row-wise $N$:$M$ sparsification, leading to extra imposed zeros.
Blue elements represent non-zero values, while white elements represent pruned values, and red elements indicate additional zeros introduced during the backward pass.
}
\label{fig:pipeline}
\end{center}
\end{figure*}

We present \slope, a Double-Pruned \textbf{S}parse Plus Lazy \textbf{Lo}w-rank Adapter \textbf{P}r\textbf{e}training method for LLMs that jointly leverages these pillars. Key contributions of \slopespace are:
\begin{itemize}[leftmargin=*]
\item \textbf{Double-Pruned backward pass} $\rightarrow$ We propose to transpose an already sparsified N:M weight matrix (forward pass) before imposing another round of N:M sparsity (backward pass). This improves model quality and eliminates the overhead of searching for transposable masks.
\item \textbf{Lazy Low-Rank adapters} $\rightarrow$ We utilize the low-rank pillar to recover accuracy by introducing additional parameters with minimal compute and memory overheads, strictly for the last 1\% of pretraining iterations (see \autoref{fig:pipeline}).
\item \textbf{Optimized CUDA kernels} $\rightarrow$ We jointly optimize NVIDIA 2:4 sparse kernels and low-rank calls through efficient tiling and scheduling. Our highly-optimized CUDA kernels result in ${\trainspeedup}\times$ end-to-end training speedup and ${\inferencespeedup}\times$ inference speedup on LLMs with billions of parameters, while reducing training and inference memory footprints by up to ${\trainmemory}\times$ and ${\inferencememory}\times$, respectively.
\end{itemize}

\section{Additional Related Work}
\paragraph{Model pruning.}
Pruning the models has been one of the most effective methods to reduce the complexity of LLMs \cite{sparsity_survey}. One can pretrain the LLMs sparsely \cite{lottery_tickets} or the pruning can happen after a dense pretraining \cite{optimal_brain_surgeon, optimal_brain_damage}, possibly followed by a fine-tuning stage to recover part of the lost accuracy \cite{fine_tuning1, fine_tuning2}.
Pruning the models after pretraining can be costly \cite{movement_pruning, fine_tuning3} and typically fails to maintain their accuracy \cite{sparsegpt, wanda}.
While the sparse pretraining methods improve the accuracy of the model, they either use unstructured sparsity patterns that cannot be accelerated with the current hardware \cite{spdf} or have significant overheads when searching for and applying their structured sparse masks \cite{n_m1, n_m2, n_m3_dominosearch}.

\paragraph{Low-rank adapters.}
Low-rank adapters have emerged as a promising method to reduce the fine-tuning costs associated with pre-trained LLMs and enable more efficient task switching \cite{lora}.
Different quantization and initialization schemes have been proposed to reduce their overheads in LLM fine-tuning \cite{qlora, lqlora}.
Adding low-rank factors to sparse matrices is a low-weight mechanism widely used to improve the accuracy of approximations of dense matrices \cite{sparse_plus_low_rank}.
In machine learning, the sparse plus low-rank approximations are limited to attention heads \cite{fmmformer, scatterbrain} and pruning after pretraining \cite{rosa, losparse}, and the sparse plus low-rank pretraining has not been investigated. \newtext{Additionally, the sparse plus low-rank fine-tuning work does not provide acceleration in both forward and backward pass of the fine-tuning process. Furthermore, the low-rank adapters in these works are added at the beginning of the fine-tuning process, adding extra overheads to the fine-tuning process.}
\section{Sparse Plus Low-rank Pretraining of LLMs}
\label{sec:slope}
~\autoref{eq:forward}, \autoref{eq:backward_weight}, and \autoref{eq:backward_input} depict the formulas for the forward and backward pass of the $i$-th linear layer in a neural network. Here, the weight tensor is denoted as $\mathcal{W}_i \in \mathbb{R}^{d_{out} \times d_{in}}$ and the input tensor is denoted as $\mathcal{X}_i \in \mathbb{R}^{b \times d_{in}}$. The forward pass generates an output tensor represented as $\mathcal{Y}_i \in \mathbb{R}^{b\times d_{out}}$. In all equations, $d_{in}$ and $d_{out}$ refer to the input and output dimensions of the respective layer and $b$ refers to the batch size.

\vspace{-10pt}
\begin{equation}
    \label{eq:forward}
    \mathrm{FWD}\mymapsto \mathcal{Y}_i = \mathcal{X}_i \mathcal{W}_i^T
\end{equation}
\begin{equation}
    \label{eq:backward_weight}
    \mathrm{BWD-1}\mymapsto \nabla_{W_i} \mathcal{L} = \nabla_{Y_i} \mathcal{L}^T \mathcal{X}_i
\end{equation}
\begin{equation}
    \label{eq:backward_input}
    \mathrm{BWD-2}\mymapsto \nabla_{X_i} \mathcal{L} = \nabla_{Y_i} \mathcal{L} \mathcal{W}_i
\end{equation}
\vspace{-10pt}

The dimension along which N:M pruning occurs corresponds to the reduction dimension in Matrix-Matrix multiplication. Without this restriction, the sparse Matrix-Matrix operation can not be accelerated on GPU~\cite{cuspaselt_reduction_dim}.
With this restriction in mind, to leverage weight sparsity in forward and backward pass, one needs to prune elements along the columns of $\mathcal{W}_i^T$ in \autoref{eq:forward} (FWD) and $\mathcal{W}_i$ in \autoref{eq:backward_input}.
To satisfy this requirement, it is necessary to prune elements of the weight tensor $\mathcal{W}_i$ along both row and column dimensions.
\subsection{Double-pruned Backward Pass}
\label{section:sparse_pretraining}
Various approaches can be used to exploit N:M sparsity during both the forward and backward passes. For example, one may prune the activation tensor $\mathcal{X}_i$
in FWD along the row dimension and $\mathcal{W}_i$ in BWD-2 along the column dimension.
Although diverse combinations exist for pruning, our focus in this study is primarily on the sparsification of weight tensors for two reasons: (a) the sparsification of weight tensors directly impacts the resource required for model storage and serving, and (b) our initial findings indicate that pruning weight tensors during both forward and backward passes has a comparatively lesser adverse impact on the overall end-to-end model quality. More details on our experiments can be found in \autoref{section:pruning_matrices}.
As such, we present a \textit{double-pruned backward pass} formulation that can productively accelerate FWD and BWD-2 computations.

In addition, we prove that such materialization of pruned weight tensors, despite being lossy\footnote{We term this formulation ``\textit{lossy}'' because the weight matrix undergoes information loss during the backward pass compared to its state in the forward pass.}, exhibits convergence properties.
For the rest of this chapter, we represent the weight tensor subjected to row-wise pruning as $\mathcal{W}_i^R$, while the concurrent row-wise and column-wise pruning (double-pruned) is presented as $\mathcal{W}_i^{R,C}$.
We rewrite the training equations to accommodate these modifications, with proposed changes highlighted in {\color{blue}{blue}}:

\begin{equation}
    \label{eq:sparse_forward}
    \mathrm{FWD}\mymapsto \mathcal{Y}_i = \mathcal{X}_i {\color{blue}{{{\mathcal{W}_i^R}}^T}}
\end{equation}
\begin{equation}
    \label{eq:sparse_backward_weight}
    \mathrm{BWD-1}\mymapsto \nabla_{W_i} \mathcal{L} = \nabla_{Y_i} \mathcal{L}^T \mathcal{X}_i
\end{equation}
\begin{equation}
    \label{eq:sparse_backward_input}
    \mathrm{BWD-2}\mymapsto \nabla_{X_i} \mathcal{L} = \nabla_{Y_i} \mathcal{L} {\color{blue}{{\mathcal{W}_i^{R,C}}}}
\end{equation}

Using this formulation for training, we can accelerate both forward and backward passes owing to the existence of N:M sparsity along both dimensions of weight tensors (see \autoref{fig:pipeline}).

\niparagraph{Memory Footprint Analysis.}
Inducing N:M structured sparsity not only improves computational efficiency of GEMM operations but also reduces the memory footprint for storing sparse tensors.
It is noteworthy, however, that the storage of auxiliary meta-data becomes necessary, containing information about the locations of non-zero elements in a supporting matrix.
\autoref{eq:index_bits} delineates the requisite number of bits for storing the indices in the N:M sparsity format, where $\lceil . \rceil$ denotes the ceiling function. We present the detailed results on the memory footprint reduction in~\autoref{section:results}.

\begin{equation}
    \label{eq:index_bits}
    n^{N:M}_{index} = \left\lceil {log\left({\binom{M}{N}}\right)} \right\rceil
\end{equation}

\niparagraph{Convergence Analysis.}
\autoref{lemma:sparsity_ratio} (proof in ~\autoref{lemma_proof}) shows the additional sparsity resulting from double pruning to an initially row-wise N:M pruned matrix.
Following this lemma, we quantify the increased sparsity induced by double pruning with 1:2, 2:4, and 2:8 sparsity patterns as $12.5\%$, $9.375\%$, and $3.39\%$, respectively.
This observation underscores that as the value of M in N:M increases, the surplus of zero elements in a double-pruned matrix diminishes.
This reduction in zero elements consequently implies a decrease in computational errors, enhancing the robustness of the computations.
We expound further insights into this phenomenon in~\autoref{section:lossy_backward_effect}.

\begin{lemma}
    \label{lemma:sparsity_ratio}
    Consider a randomly initialized matrix $A$. Following our notations, we denote the row-wise pruned version of $A$ by $A^R$ and the joint column- and row-wise pruned version of $A$ by $A^{R, C}$. We use $D(.)$ to present the density ratio of a matrix. \autoref{eq:sparsity_ratio} shows the additional zero elements in matrix $A$ that are introduced by double-pruning, where $s = \frac{N}{M}$.
    \begin{equation}
        \label{eq:sparsity_ratio}
        D(A^R) - D(A^{R, C}) = \sum_{j = N + 1}^M {\binom{M}{j}} s^j (1-s)^{M - j} \frac{j - N}{M}
    \end{equation}
\end{lemma}

\autoref{theorem:convergence} states that the dynamic alteration of the column-wise mask in~\autoref{eq:sparse_backward_weight} during each training iteration does not exert a detrimental impact on the convergence of the optimizer.
This phenomenon can be attributed to the equivalence between the left-hand side of \autoref{eq:backward_convergence}, which corresponds to \autoref{eq:backward_input} [BWD-2], and the averaging effect achieved through multiple training iterations of backpropagation with distinct sparsity masks.
However, for arbitrary values of N and M, \autoref{eq:sparse_forward} and \autoref{eq:sparse_backward_weight} can be used in the training with convergence guarantee (proof in ~\autoref{lemma_proof}).
The sparsity mask is chosen randomly at initialization, i.e. all the weights have the same probability of being zero or non-zero. This is because at initialization the location of weights with larger magnitude is arbitrary. After choosing the sparsity mask at initialization, we keep the mask fixed throughout the entire training process. This policy ensures that each element in the weight has the same probability of being non-zero at initialization and satisfies the random mask assumption in \autoref{lemma:sparsity_ratio}.

\begin{theorem}
    \label{theorem:convergence}
    Assuming a loss function $\mathcal{L(W_i, X_i})$ for a random sample $X_i$, and considering a random mask $M_i$, \autoref{eq:backward_convergence} holds, where $E[.]$ is the expectation operator and $\odot$ is the element-wise multiplication.
    \begin{equation}
        \label{eq:backward_convergence}
        E_{X_i}[\nabla_{X_i}\mathcal{L}(W_i, X_i)] = 
        \frac{M}{N} E_{M_i}[E_{X_i}[\nabla_{Y_i}\mathcal{L}(W_i, X_i) (M \odot W_i)]]
    \end{equation}
\end{theorem}

\subsection{Lazy Low-rank Adapters}
\label{section:lazy_low_rank}
Pruning weight tensors in FWD and BWD-2 computations is desirable for computational efficiency but may have detrimental impact on quality.
To mitigate this adverse impact on model quality, we augment the doubly-pruned weight matrix with a low-rank matrix.
The decomposition of the doubly-pruned weight matrix, combined with the low-rank matrix, maintains the computational efficiency of sparse matrix-matrix multiplication during forward and backward passes.
Simultaneously, this approach holds promise in alleviating the adverse effects of double pruning on overall model quality.

Considering the dense weight matrix, denoted by $W_{dense} \in \mathbb{R}^{d_{out} \times d_{in}}$, \autoref{eq:sparse_plus_low_rank} illustrates the proposed matrix decomposition.
In this expression, $W_{sparse} \in \mathbb{R}^{d_{out} \times d_{in}}$ signifies a doubly-pruned matrix and $\mathcal{L} \in \mathbb{R}^{d_{out} \times r}$ and $\mathcal{R} \in \mathbb{R}^{r \times d_{in}}$ are components of the low-rank approximation.
The variable $r$ denotes the rank of this low-rank approximation and
$r$ functions as a hyperparameter that controls the trade-offs between memory footprint, computational efficiency, and model quality.

\vspace{-15pt}
\begin{equation}
    \label{eq:sparse_plus_low_rank}
    \mathcal{W}_{dense} = \mathcal{W}_{sparse} + \mathcal{L}\mathcal{R}
\end{equation}
\vspace{-10pt}

The matrix decomposition of doubly-pruned matrix combined with a low-rank matrix approximation reduces the memory footprint of $\mathcal{W}$ from $d_{in} d_{out}$ to $d_{in} d_{out} \frac{N}{M} + (d_{in} + d_{out}) r$, where $r << min(d_{in}, d_{out})$.
The computational complexity of dense Matrix-Matrix multiplication, however, changes from $b d_{in} d_{out}$ to $b d_{in} d_{out} \frac{N}{M} + b( d_{in} + d_{out})r$.
Given the substantially smaller value of $r$ in comparison to $b$, $d_{in}$, and $d_{out}$, our formulation effectively reduces both memory footprint and computational complexity of Matrix-Matrix multiplication by a factor of $\frac{M}{N}\times$.

We empirically show that the convergence rate of low-rank adapters surpasses that of sparse weights. We attribute this behavior to the notably lower parameter counts inherent in low-rank adapters.
Leveraging this observation, we incorporate low-rank adapters exclusively during the final 1\% of the training iterations.
This confined usage of low-rank adapters results in additional reduction of training cost, specifically in terms of total number of operations.
We term the proposed usage of low-rank adapters in the final steps of the training as \textit{lazy low-rank adapters} (see \autoref{fig:pipeline}).
\subsection{Sparse Kernels}
\label{section:kernels}
cuSPARSELt is a CUDA library designed explicitly for sparse Matrix-Matrix multiplication, where one operand undergoes pruning with the 2:4 sparsity pattern.
However, this library does not offer APIs for other algebraic routines such as addition and assignment for sparse tensors.
We now delve into the details of different kernels for training and overview our implementation methodology.

\autoref{alg:training} shows the training process of a single linear layer taken from an attention-based model.
We assume the use of weight decay in the optimizers, and subsequently design the requisite sparse APIs to facilitate the optimizer operations.
The training starts with matrix initialization (\underline{line 2}) and setting up sparse formats to store weight tensors and their corresponding transpose (\underline{line 3 and 4}).
Then, for every mini-batch in the training set, we compute the forward pass following \autoref{eq:sparse_forward} (\underline{line 8}).
As part of the backward pass, the derivative of the loss function with respect to the output activation is computed (\underline{line 10}). Subsequently, the gradients of the loss function with respect to the input activation (\underline{line 11}) and the weight tensor (\underline{line 12}) are computed using \autoref{eq:sparse_backward_weight} and \autoref{eq:backward_weight}, respectively.
In order to circumvent the necessity of updating weights with zero values and mitigate the associated memory footprint overhead, we employ a strategy wherein we mask the gradients for pruned weights.
The computed values are stored in a sparse format (\underline{line 13}).
Next, in order to implement weight decay in the optimizer and mitigate the impact of gradient scaling, we compute the value of $\frac{1}{\gamma}\nabla_{W} \mathcal{L} + \alpha W$ (\underline{line 15}). Here, $\alpha$ is the weight decay applied in the optimizer, while $\gamma$ denotes the gradient scaling factor for numerical stability during the half-precision backward pass.
The updated values for the weight tensor are calculated according to the optimizer update rule (\underline{line 16}).
Finally, the value of weight tensor and its transpose are updated directly in a sparse format (\underline{line 17} and \underline{line 18}).
More details about the implementation of the custom kernels used in \autoref{alg:training} can be found in \autoref{section:implementation_details}.

\begin{algorithm}[t]
\caption{\textbf{Accelerated Sparse Pretraining Algorithm for a Linear Layer}}
\label{alg:training}
\begin{algorithmic}[1]
\State \textbf{\color{KeywordColor}Input:} Weight $\color{VarColor}\mathbf{W}$, training set $\color{VarColor}\mathbb{D}$, weight decay $\color{VarColor}\alpha$, gradient scaling factor $\color{VarColor}\gamma$.
\State \textbf{\color{KeywordColor}Output:} Updated weight $\color{VarColor}{\mathbf{W}{\text{new}}}$.
\vspace{0.5em}
\State $\texttt{backend.init()}$ \algorithmiccomment{Initialize backend}
\State ${\color{VarColor}\text{WSparseTranspose} \gets} \texttt{backend.setup}({\color{VarColor}\mathbf{W}^T})$ \algorithmiccomment{Setup transpose for sparse matrix multiplication}
\State ${\color{VarColor}\text{WSparse} \gets} \texttt{backend.setup}({\color{VarColor}\mathbf{W}})$ \algorithmiccomment{Setup sparse weight matrix}
\State ${\color{VarColor}\text{sparseMask} \gets} ({\color{VarColor}\text{WSparse} \neq 0})$ \algorithmiccomment{Element-wise mask for sparsity}
\For{\textbf{each training example} $(\color{VarColor}\mathbf{X}, \color{VarColor}\hat{\mathbf{Y}}) \in \color{VarColor}\mathbb{D}$}
\State \textbf{\color{KeywordColor}Forward Pass:}
\State ${\color{VarColor}\mathbf{Y} \gets} \texttt{backend.spmm}({\color{VarColor}\mathbf{X}}, {\color{VarColor}\text{WSparseTranspose}})$
\vspace{0.2em}
\State \textbf{\color{KeywordColor}Backward Pass:}
\State $\color{VarColor}\nabla_{\mathbf{Y}} \mathcal{L} \gets \color{VarColor}\text{gradOutput}$ \algorithmiccomment{Gradient w.r.t. output}
\State ${\color{VarColor}\text{gradInput} \gets} \texttt{backend.spmm}({\color{VarColor}\text{gradOutput}}, {\color{VarColor}\text{WSparse}})$
\State ${\color{VarColor}\text{gradWeight} \gets} \texttt{backend.matmul}({\color{VarColor}\text{gradOutput}^T}, {\color{VarColor}\mathbf{X}})$
\State ${\color{VarColor}\text{gradWeightSparse} \gets} \texttt{backend.pruneAndCompress}({\color{VarColor}\text{gradWeight}}, {\color{VarColor}\text{sparseMask}})$
\vspace{0.2em}
\State \textbf{\color{KeywordColor}Optimizer with Weight Decay:}
\State ${\color{VarColor}\mathbf{g} \gets} \texttt{backend.sparseAdd}({\color{VarColor}\text{gradWeightSparse}}, {\color{VarColor}\text{WSparse}}, {\frac{1}{\color{VarColor}\gamma}}, {\color{VarColor}\alpha})$
\State ${\color{VarColor}\mathbf{W}{\text{new}} \gets} \texttt{optimizer.updateWeight}({\color{VarColor}\mathbf{g}})$
\State $\texttt{backend.updateSparseMatrix}({\color{VarColor}\text{WSparse}}, {\color{VarColor}\mathbf{W}{\text{new}}})$
\State $\texttt{backend.updateSparseMatrix}({\color{VarColor}\text{WSparseTranspose}}, {\color{VarColor}\mathbf{W}{\text{new}}^T})$
\EndFor
\vspace{0.5em}
\State \textbf{\color{KeywordColor}Return:} $\color{VarColor}\mathbf{W}{\text{new}}$.
\end{algorithmic}
\end{algorithm}
\subsection{\slopespace Runtime Optimization}
\label{section:scheduling}
While \slopespace improves the training and inference of LLMs by introducing sparse weights and low-rank adapters, a na\"ive implementation can hinder its full performance improvement.
Specifically, cuSPARSELt \cite{cusparselt} SpMM kernels exhibit sensitivity to input and weight tensor shapes, and introducing low-rank adapters at inference can increase the number of calls during the forward pass of each linear layer.
This section covers our approach to optimize \slopespace's implementation and further improve model performance.

\niparagraph{Efficient tiling of upsample tensors.}
~\autoref{fig:lora_convergence}-\textbf{(a)} showcases the speedup achieved by the cuSPARSELt backend across a range of tensor shapes commonly used in LLMs.
While the speedup of SpMM in downsample tensors increases gradually as their sizes increase, the speedup of upsample tensors drops off at around hidden dimension = 4000.
To overcome this limitation, we tile the upsample tensor into multiple smaller matrices of equal size, each of which benefits from improved speedup when multiplied by the input using 2:4 sparsity. By tuning the size of the tiles, we discovered that the best performance can be achieved by using square tiles.
The results of these multiplications are then concatenated. This optimization, as detailed in ~\autoref{section:scheduling_results}, leads to a \xx{12\%} improvement in inference speed and a \xx{4\%} increase in training speed with \slopespace.

\niparagraph{Efficient kernel for combined SpMM+low-rank adapters.}
A straightforward implementation of low-rank adapters requires four kernel calls: one for sparse matrix multiplication, two for low-rank computations, and one for adding the results.
In addition, our experiments demonstrate that multiplying matrices with low-rank adapters does not scale proportionally with the adapter's rank, leading to significant overheads due to their low arithmetic intensity (see \autoref{section:lora_underutilization}).
To address this, we introduce two optimizations: \textit{(1)} concatenating the downsample tensor to the sparse weight tensor, reducing kernel calls and increasing arithmetic intensity as in \autoref{eq:concat}-left, and \textit{(2)} leveraging a cuBLAS fused matrix multiplication and addition kernel, minimizing cache access and kernel calls as in \autoref{eq:concat}-right.
As demonstrated in ~\autoref{section:low_rank_optimization_speedup}, these optimizations collectively contribute to a speedup improvement of up to 6\% in the end-to-end inference speed.
\begin{equation}
    \label{eq:concat}
     [\mathcal{Y}_1 | \mathcal{Y}_2] = \mathcal{X} [\mathcal{W}^T | \mathcal{L}]; \hspace{40pt} \mathcal{Y} = \mathcal{Y}_2 \mathcal{R} + \mathcal{Y}_1
\end{equation}

\section{Experimental Results}
\label{section:results}
\newif\ifperplexity
\perplexitytrue
\newcommand{\gptmetric}{%
  \ifperplexity
    perplexity %
  \else
    loss %
  \fi
}
This section evaluates the efficacy of \slopespace in accelerating the pretraining while achieving memory savings.
Due to the substantial computational resources required for LLM pretraining, our accuracy evaluation is primarily focused on smaller-scale LLMs up to 774M parameters.
However, the speedup and memory reduction results extend to a wider range of models, from 2.6B up to 66B parameters.

\niparagraph{Experiment Setup.} Our experiments were conducted on the Narval and Mist clusters at Compute Canada~\cite{computecanada} and the Lonestar 6 cluster at the Texas Advanced Computing Center~\cite{lonestar6}. Each Narval node is equipped with four Nvidia A100 GPUs (40GB), Mist nodes feature four Nvidia V100 GPUs (32GB), and Lonestar 6 nodes have three Nvidia A100 GPUs (40GB). For accuracy experiments, we emulated 2:4 and N:M sparsity using custom-designed, low-overhead CUDA kernels to prune weights in both the forward and backward passes, utilizing a mixture of available resources across clusters since model accuracy is not hardware-dependent. Speedup and memory saving experiments were conducted on a single A100 GPU in the Narval cluster over 1000 iterations, reporting the median to mitigate outlier effects; memory reduction experiments were run five times with the median reported. We employed the default hyperparameters from the NVIDIA BERT codebase~\cite{nvidia-bert} and the FlashAttention GPT codebase~\cite{flashattention, flashattention2}. Training BERT-Large-Uncased required approximately 32 hours on 64 A100-64GB GPUs, while pretraining GPT2-Small/Large took 32 and 111 hours on 64 V100-32GB GPUs, respectively.
\subsection{End-to-end Speedup and Memory Saving: Pretraining and Inference}
\label{section:speedup_results}
We evaluate the speedup and memory reduction by \slopespace during pretraining and inference across LLMs with different model parameter sizes.
To demonstrate the scalability and efficiency of our method, we conducted extensive benchmarking on OPT (2.6\,B to 66\,B), LLaMA-3-8B and Mistral-v0.3-7B models. In all the experiments, we have enabled FlashAttention-2 \citep{flashattention2} (~\autoref{sec:flash_attention} presents detailed ablation study on the impact of FlashAttention).
To mitigate the impact of outliers, we conducted 1,000 iterations for each speedup experiment and reported the median value.
For the memory reduction experiments, we performed five independent runs and similarly reported the median outcome. These methodologies were chosen to provide a more reliable measure of central tendency in our results
\footnote{It is noteworthy that for benchmarking speedup and memory savings, which require comparatively fewer computational resources than comprehensive pretraining accuracy experiments, we utilized the OPT, LLaMA-3, and Mistral-v0.3 model families. These families were selected due to their diverse range of model parameter sizes, allowing for a more thorough study of performance across different scales.}.

 We compared our method against dense pretraining and inference directly in PyTorch, which uses efficient cuBLAS backend. As the sparse pretraining benchmark, we compare our work against Fully Sparse Training (\transposable) \cite{2_4_pretraining}, the state-of-the-art 2:4 pretraining method and the only semi-structured sparse pretraining work that provides end-to-end speedups.
Note that methods targeting LLM pretraining with N:M sparsity often suffer from inefficiency due to mask search overheads and/or compression setup.
~\autoref{section:transposable_slow_down} and ~\autoref{section:setup_overhead} detail the profiling in Bi-Mask~\cite{n_m2} and \transposable~\cite{2_4_pretraining}, which similarly use N:M sparsity on both forward and backward passes.

 Notably, our approach, \slopespace, diverges significantly from recent work Fully Sparse Training (\transposable) \cite{2_4_pretraining} in three key aspects. Firstly, \textit{we comprehensively prune all weights in the model, encompassing both MLP and Self-Attention modules}, whereas \transposable\,only prunes weights in the MLP modules. Secondly, \transposable\,employs dynamic transposable weights, which introduce additional computation and memory overhead during training. Thirdly, \transposable\, necessitates dense fine-tuning (\(\sim\)17\% of pretraining), thereby negating their speedup advantages during inference. In contrast, our approach achieves efficient and accurate large language models during both training and inference without such limitations.

\niparagraph{\slopespace Speedup for Pretraining and Inference.}
~\autoref{table:speedup} summarizes the speedups achieved by our method during both training and inference.
Since over 99\% of training occurs without low-rank adapters, the training speedup is largely independent of the adapter rank.
Conversely, inference speedup is directly influenced by the adapter rank.
Given the varying hidden dimensions across different model sizes, we report the inference speedup for various adapter rank ratios: $\frac{adapter-rank}{hidden-dimension}$.
\begin{table}[t]
\caption{Comparative analysis of end-to-end pretraining and inference speedup ($\times$) comparison between \slopespace and the latest work (\transposable) on accelerating pretraining with 2:4 sparsity (ICML 2024)~\cite{2_4_pretraining}. The baseline is dense PyTorch implementation of the models with CUBLAS backend. Note that the lack of inference speedup in \transposable\, is because of the final dense pretraining during the final iterations, resulting in a dense model for inference. E-SR-STE stands for Extended SR-STE.
}
\label{table:speedup}
\centering
\begin{scriptsize}
\begin{sc}
\begin{tabular}{c|c|c|ccc}
\toprule
\multirow{2}{*}{\textbf{Model}} & \multirow{2}{*}{\textbf{Method}} & \textbf{\textbf{Training}} & \multicolumn{3}{c}{\textbf{Inference}} \\
&& No Adapter ($r$ = 0) & No Adapter ($r$ = 0) & 1.56\% Adapter & 6.25\% Adapter\\
\midrule
\multirow{2}{*}{OPT-66B} & \slopespace & \textbf{1.20} &	\textbf{1.46} &	\textbf{1.43} &	\textbf{1.40} \\
& \transposable & 1.06 & 1.00 & 1.00 & 1.00
\\
\midrule
\multirow{2}{*}{OPT-30B} & \slopespace & \textbf{1.22} &	\textbf{1.53} &	\textbf{1.53} &	\textbf{1.50} \\
& \transposable & 1.07 & 1.00 & 1.00 & 1.00
\\
\midrule
\multirow{2}{*}{OPT-13B} & \slopespace & \textbf{1.25} &	\textbf{1.54} &	\textbf{1.39} &	\textbf{1.36} \\
& \transposable & 1.10 & 1.00 & 1.00 & 1.00 
\\
\midrule
\multirow{2}{*}{OPT-6.6B} & \slopespace & \textbf{1.21}	&   \textbf{1.46} &	\textbf{1.46} &	\textbf{1.43} \\
& \transposable & 1.11 & 1.00 & 1.00 & 1.00
\\
\midrule
\multirow{2}{*}{OPT-2.6B} & \slopespace & \textbf{1.13} &	\textbf{1.31} &	\textbf{1.25} &	\textbf{1.18} \\
& \transposable & 1.09 & 1.00 & 1.00 & 1.00 
\\
\midrule
\multirow{2}{*}{LLaMA-3-8B} & \slopespace & \textbf{1.16} & \textbf{1.35} & \textbf{1.33} & \textbf{1.32}
\\
& \transposable & 1.09 & 1.00 & 1.00 & 1.00
\\
\midrule
\multirow{2}{*}{Mistral-v0.3-7B} & \slopespace & \textbf{1.15} & \textbf{1.34} & \textbf{1.32} & \textbf{1.31}
\\
& \transposable & 1.07 & 1.00 & 1.00 & 1.00
\\
\bottomrule
\end{tabular}
\end{sc}
\end{scriptsize}
\end{table}
\begin{table}[t]
\caption{
Comparative analysis of end-to-end memory reductions ($\times$) during training and inference between \slopespace and the latest work (\transposable) on accelerating pretraining with 2:4 sparsity (ICML 2024)~\cite{2_4_pretraining}. Values greater than $1.00\times$ show memory overhead.
}
\label{table:memory_reduction}
\centering
\begin{scriptsize}
\begin{sc}
\begin{tabular}{c|c|c|ccc}
\toprule
\multirow{2}{*}{\textbf{Model}} & \multirow{2}{*}{\textbf{Method}} & \textbf{\textbf{Training}} & \multicolumn{3}{c}{\textbf{Inference}} \\
&& No Adapter ($r$ = 0) & No Adapter ($r$ = 0) & 1.56\% Adapter & 6.25\% Adapter\\
\midrule
\multirow{2}{*}{OPT-66B} & \slopespace & \textbf{0.67} &	\textbf{0.63} &	\textbf{0.65} &	\textbf{0.70} \\
& \transposable & 1.27 & 1.00 & 1.00 & 1.00 \\
\midrule
\multirow{2}{*}{OPT-30B} & \slopespace & \textbf{0.67} &	\textbf{0.61} &	\textbf{0.63} &	\textbf{0.69} \\
& \transposable & 1.17 & 1.00 & 1.00 & 1.00\\
\midrule
\multirow{2}{*}{OPT-13B} & \slopespace & \textbf{0.68} &	\textbf{0.51} &	\textbf{0.62} &	\textbf{0.68} \\
& \transposable & 1.16 & 1.00 & 1.00 & 1.00 \\
\midrule
\multirow{2}{*}{OPT-6.6B} & \slopespace & \textbf{0.68}	&   \textbf{0.60} &	\textbf{0.62} &	\textbf{0.68} \\
& \transposable & 1.19 & 1.00 & 1.00 & 1.00\\
\midrule
\multirow{2}{*}{OPT-2.6B} & \slopespace & \textbf{0.67} &	\textbf{0.62} &	\textbf{0.64} &	\textbf{0.70} \\
& \transposable & 1.18 & 1.00 & 1.00 & 1.00 \\
\midrule
\multirow{2}{*}{LLaMA-3-8B} & \slopespace & \textbf{0.63} & \textbf{0.66} & \textbf{0.69} & \textbf{0.71}
\\
& \transposable & 1.17 & 1.00 & 1.00 & 1.00
\\
\midrule
\multirow{2}{*}{Mistral-v0.3-7B} & \slopespace & \textbf{0.68} & \textbf{0.66} & \textbf{0.69} & \textbf{0.65}
\\
& \transposable & 1.15 & 1.00 & 1.00 & 1.00
\\
\bottomrule
\end{tabular}
\end{sc}
\end{scriptsize}
\vspace*{-10pt}
\end{table}

~\autoref{fig:lora_convergence}-\textbf{(a)} illustrates that cuSPARSELt achieves higher speedups for large matrices until it reaches its maximum performance capacity ($2\times$).
A similar trend is observed in the pretraining and inference speedups of the models.
For small matrices used in low-rank adapters, the lower arithmetic intensity of low-rank adapter multiplication results in higher overhead relative to sparse multiplication.
This is because low arithmetic intensity limits the full utilization of GPU resources, leading to inefficiencies.

\niparagraph{\slopespace Memory Reduction in Pretraining and Inference.}
\newtext{
For training, the memory consumption of a dense model includes weights, gradients, and optimizer states, amounting to $4 \times 16$ bits for weights, $4 \times 16$ bits for gradients, and $2 \times 4 \times 32$ bits for optimizer states.
The sparse model, however, stores non-zero weights and indices twice (for both weights and transposed weights), along with a binary mask, gradients, and reduced optimizer states.
This adds up to $2 \times (16+3)$ bits (weights and transposed weights), $4 \times 8$ bits (binary mask), $2 \times 16$ bits (gradients), and $2 \times 2 \times 32$ bits (optimizer states).
Consequently, the memory footprint during training is reduced by 68\%.
For inference, a dense model requires storing weights with a total memory cost of $4 \times 16$ bits.
In contrast, our sparse model optimizes memory usage by storing only the non-zero weights and their indices, resulting in $2 \times 16$ bits for non-zeros and three bits for indices (see \autoref{eq:index_bits}).
This leads to a 54\% reduction in memory usage during inference.
}

~\autoref{table:memory_reduction} presents the memory reduction for different low-rank adapter ranks and OPT, LLaMA-2, and Mistral model variants.
The memory reduction is slightly less than the theoretical expectation, primarily because of additional memory usage from other model components, such as layer norms, and dense model parameters.
\subsection{Pretraining Accuracy Results}
\label{section:accuracy_results}
To assess the impact of \slopespace on model accuracy, we conducted pretraining experiments across various models and datasets.
In all experiments, the classification heads and the first linear layer following the input are dense.

\niparagraph{{GPT2 (Small/Large)}.} We pretrained both the small (117\,M parameters) and large (774\,M parameters) variants of GPT2~\cite{gpt2} on the OpenWebText dataset~\cite{openwebtext}.
For a fair comparison, we evaluate the models on MMLU~\cite{mmlu}, Arc Challenge~\cite{arc}, and OpenBookQA~\cite{openbookqa} zero-shot tasks implemented in Language Model Evaluation Harness~\cite{lmharness}. Additionally, we evaluate the validation perplexity of the models following the same experimental settings described in FlashAttention~\cite{flashattention,flashattention2}.
We compare \slopespace against two state-of-the-art sparse pretraining methods, including (a) Wanda~\cite{wanda}~$\rightarrow$~a one-shot pruning technique, (b) Extended SR-STE~\cite{nm_scratch, 2_4_pretraining}~$\rightarrow$~a dynamic mask pretraining method for N:M sparsity, which serves as the foundation of follow-up work~\cite{n_m1, n_m2, 2_4_pretraining}. \newtext{Please note that SR-STE only supports stochastic gradient descent optimization, and \transposable~ extended it to other optimizers. We use the extension provided by \transposable~ in our work, and call it Extended SR-STE. The difference between Extended SR-STE and \transposable~ is that \transposable~ requires dense pretraining (fine-tuning) in the last 17\% of pretraining and only prunes the MLP layers of the model, while SR-STE is fully sparse and prunes both the MLP and the Self-Attention layers of the model.}
\begin{figure}[ht]
\begin{center}
\ifperplexity
    \centerline{\includegraphics[width=0.55\columnwidth]{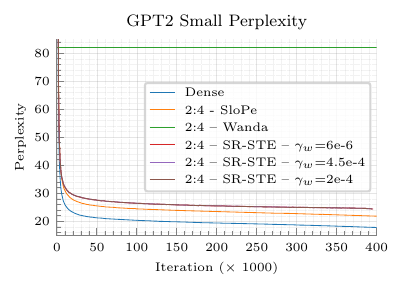}
    \includegraphics[width=0.55\columnwidth]{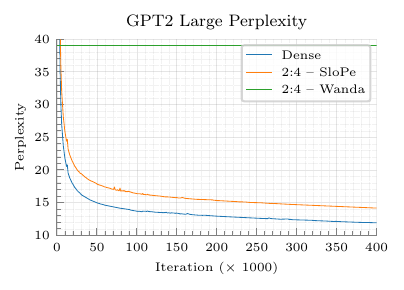}}
\else 

    \centerline{\includegraphics{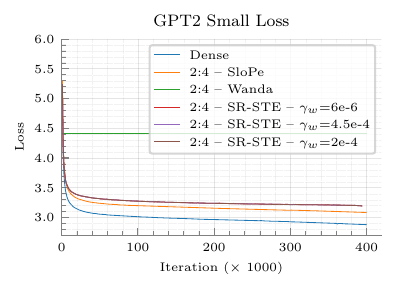}
    \includegraphics{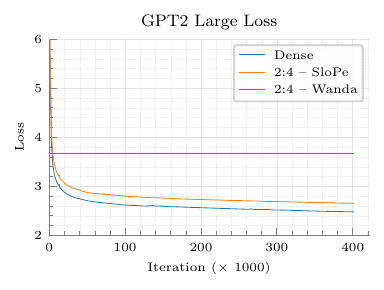}}
\fi
\caption{Validation \gptmetric of GPT2-Small and GPT2-Large on OpenWebText. $\gamma_w$ shows the value of the decay factor parameter in Extended SR-STE (\transposable).}
\label{fig:gpt_perplexity}

\vspace{-10pt}
\end{center}
\end{figure}

\begin{table}[t]
\caption{GPT2-Small accuracy results on zero-shot tasks. Adapter rank is the ratio of the low-rank adapter  to the hidden dimension of the model. For Extended SR-STE, we have used a decay factor of $6\times 10^{-6}$, since it resulted in the lowest perplexity in OpenWebText. The best performing sparse configuration is highlighted in bold.}
\label{table:gpt_zero_shot}

\resizebox{\textwidth}{!}{%

\centering
\begin{small}
\begin{sc}
\begin{tabular}{c|c|cccccccc}
\toprule
\multirow{2}{*}{Method} & Adapter & \multirow{2}{*}{MMLU $\uparrow$} & Arc  & Open- & Wino- & Hella- & \multirow{2}{*}{MathQA$\uparrow$} & \multirow{2}{*}{PiQA$\uparrow$} & \multirow{2}{*}{Race$\uparrow$}
\\
& Rank & & Challenge$\uparrow$ & BookQA$\uparrow$ & Grande$\uparrow$  & Swag$\uparrow$ 
\\
\midrule
Dense & N/A & 22.9 & 20.7 & 16.2 & 50.6 & 28.5 & 21.8 & 59.8 & 28.4
\\
\midrule
 \multirow{3}{*}{\slopespace} & 2.1\% & 23.0 & 19.3 & \textbf{16.4} & \textbf{50.8} & \textbf{27.5} & 20.8 & \textbf{57.6} & \textbf{27.2}
\\
& 0.05\% & 23.0 & \textbf{19.4} & 16.2 & 50.5 & 27.4 & 20.8 & 57.5 & 27.1
\\
& 0 & 23.0 & 19.3 & 16.0 & 50.1 & 27.5 & 20.8 & 57.4 & 27.1
\\
\midrule
 Extended & 2.1\% & \textbf{24.2} & 18.3 & 14.2  & 47.5 & 26.9 & \textbf{21.4} & 55.2 & 24.2
\\
\multirow{2}{*}{SR-STE}& 0.05\% & 24.1 & 18.4 & 14.2 & 47.5 & 26.8 & 21.2 & 54.5 & 24.2
\\
& 0 & 24.1 & 18.3 & 12.6 & 47.5 & 26.9 & 21.2 & 54.8 & 24.0
\\
\bottomrule
\end{tabular}
\end{sc}
\end{small}
}
\vspace{-15pt}
\end{table}

~\autoref{fig:gpt_perplexity} compares the validation \gptmetric\footnote{Perplexity is a 
standard metric for evaluating language models. Intuitively, perplexity measures how 
``surprised'' the model is by the held-out text: 
lower values indicate better predictive accuracy. 
A perplexity of $k$ can be loosely interpreted as 
the model being as uncertain as if it were choosing 
uniformly among $k$ candidates at each step 
\cite{chen1998evaluation}.} and zero-shot accuracy of GPT2-Small and GPT2-Large across a range of sparse pretraining methods with different hyperparameters. We have additionally added lazy low-rank adapters to Extended SR-STE~\cite{nm_scratch} to show the effectiveness of our approach in other methods and also compare both methods with more similar settings.
While a gap in \gptmetric consistently exists between sparse and dense models, \slopespace achieves a lower perplexity compared to Wanda~\cite{wanda} and Extended SR-STE.
Additionally, \autoref{table:gpt_zero_shot} summarizes the achieved accuracy of the models on zero-shot tasks, showing that \slopespace is consistently achieving a higher accuracy in comparison to Extended SR-STE. Moreover, adding lazy low-rank adapters can benefit both static and dynamic training methods.
This improved accuracy stems from \slope's efficient allocation of the training budget.
Specifically, Extended SR-STE, with its dynamic pruning masks, expends a significant portion of its training budget (e.g. gradient updates) updating weights that may be ultimately pruned and not used at inference, leading to wasted resources. ~\autoref{section:rs_ste} provides further details and supporting evidence for this observation. \newtext{Additional validation results for GPT experiments on GLUE dataset are also provided in \autoref{sec:gpt_glue} and \autoref{sec:gpt2_half}.}

\niparagraph{BERT-Large-Uncased.}
We pretrain BERT-Large-Uncased~\cite{bert} (355\,M parameters) and fine-tune it for various question-answering and text classification tasks, following a similar approach to~\cite{nvidia-bert, mkor, kaisa} for both pretraining and fine-tuning. ~\autoref{section:bert_loss} provides details on the pretraining and fine-tuning process.
We evaluate the performance of BERT-Large-Uncased on the SQuAD v1.1~\cite{squad} and GLUE~\cite{glue} tasks. We report the average metric score for GLUE and present the task-specific metrics in ~\autoref{section:glue_results}. Please note that in all the experiments corresponding to BERT-Large-Uncased, when using Wanda, we have fine-tuned the model after pruning to improve the accuracy of the models, since using Wanda alone led to extremely low accuracy results.

\niparagraph{Effects of Low-rank Adapters.}
To understand the impact of low-rank adapters on pretraining performance, we conducted ablations using low-rank adapter ranks of 4, 16, and 64 for 1\% of the total number of iterations.
These ranks represent up to \xx{6.25\%} of the model's hidden dimension.
~\autoref{table:bert_ranks} shows the results of these settings on SQuAD and GLUE downstream tasks. We present per-task metrics for GLUE in ~\autoref{section:glue_results}.
As expected, adding low-rank adapters improves the model's final accuracy across all tasks.
Additionally, higher ranks improve the model's performance at the cost of increased computational requirements.
It is also worth noting that incorporating low-rank adapters only in the final iterations (1\% of total iterations) is sufficient to recover pretraining accuracy.

\begin{figure}[ht]
\begin{center}
\centerline{\includegraphics[width=0.55\columnwidth]{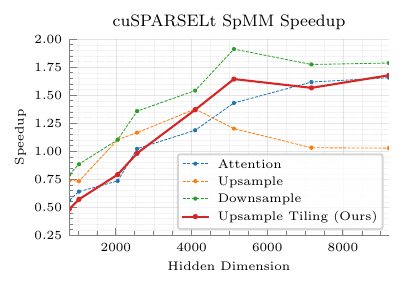} \includegraphics[width=0.55\columnwidth]{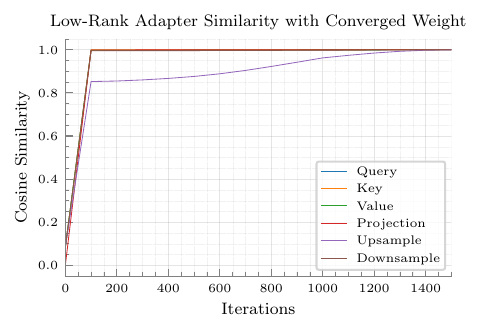}}
\textbf{(a)} \hspace{0.4\columnwidth} \textbf{(b)}
\caption{\textbf{(a)} The speedup achieved using cuSPARSELt backend in PyTorch for Attention ($d_{out} = d_{in}$), Upsample ($d_{out} = 4d_{in}$) and Downsample ($d_{out} = \frac{d_{in}}{4}$) matrices with a batch size of 2048. \textbf{(b)} The cosine similarity of the low-rank adapters and the converged adapters for different layers in the model. The cosine similarities are averaged among the 24 layers of BERT-Large-Uncased.}
\label{fig:lora_convergence}
\end{center}
\end{figure}

\niparagraph{Convergence Rate of Low-rank Adapters.}
We hypothesized that low-rank adapters would converge faster due to their significantly fewer learnable parameters. To test this, we introduced low-rank adapters in the second phase of BERT-Large-Uncased pretraining and monitored their convergence rate. ~\autoref{fig:lora_convergence} shows the cosine similarity of the adapters, with the downsample adapter converging rapidly within 100 iterations and the upsample adapter converging slightly slower. Despite this, limiting training to 100 iterations still yields comparable results on downstream tasks.

\begin{table}[t]
\caption[SQuAD-v1.1 accuracy and GLUE results on BERT-Large-Uncased with different adapter ranks]{SQuAD-v1.1 accuracy and GLUE results on BERT-Large-Uncased with different adapter ranks. GLUE results are reported as the average metric score across all tasks. $r$ denotes the ratio of the low-rank adapter to the hidden dimension (1024).}
\label{table:bert_ranks}
\centering
\begin{small}
\begin{sc}
\begin{tabular}{cccccc}
\toprule
Dataset   & Dense & $r=0$ & $r=0.39\%$ & $r=1.56\%$ & $r=6.25\%$ \\
\midrule
SQuAD     & 90.44 & 89.1 & 89.1 & 89.2 & 89.5 \\
GLUE      & 80.22 & 77.4 & 77.7 & 77.8 & 78.2 \\
\bottomrule
\end{tabular}
\end{sc}
\end{small}
\end{table}

\niparagraph{Effects of Mixed N:M sparsity.}
To study the sensitivity of different blocks to varying sparsity ratios and to assess their relative importance, we experiment across a range of configurations: (a) [2:4-2:4]~$\rightarrow$~uniformly applying 2:4 sparsity across all layers (b) [2:4-2:8]~$\rightarrow$~applying 2:4 sparsity pattern to the first 12 blocks and a 2:8 sparsity pattern to the last 12 blocks and (c) [2:8-2:4]~$\rightarrow$~we reverse the sparsity ratios for the first and last 12 blocks.
Note that, to reduce computational costs, we use the same dense checkpoint for Phase-1 in all settings and a low-rank adapter of rank 40 for all models.
We also replicate this experiment using Wanda~\cite{wanda} and report the comparison results.

\begin{table}[htpb]
\caption{SQuAD-v1.1 accuracy results on BERT-Large-Uncased for different sparsity settings.}
\label{table:mixed_nm}
\centering
\begin{small}
\begin{sc}
\begin{tabular}{ccccc}
\toprule
Sparsity Pattern   & SQuAD & SQuAD & GLUE & GLUE \\
(First 12 blocks - Last 12 blocks) & \slopespace & Wanda & \slopespace & Wanda \\
\midrule
2:4-2:4 & \textbf{90.17} & 89.93 & \textbf{79.08} & 78.84 \\
2:4-2:8 & \textbf{89.85} & 89.55 & \textbf{79.03} & 77.24 \\
2:8-2:4 & \textbf{89.67} & 86.57 & \textbf{75.92} & 69.08 \\
\bottomrule
\end{tabular}
\end{sc}
\end{small}
\end{table}
~\autoref{table:mixed_nm} summarizes the GLUE and SQuAD results for these settings.
As the results show, increasing the sparsity ratio reduces the accuracy of the model on all tasks.
But when the first 12 blocks of the model are pruned, the accuracy drop is significantly higher, especially on the GLUE dataset.
We conclude that the first blocks of the model are more sensitive to sparsity during pretraining, but one can sparsify the last blocks of LLMs more aggressively.
We observe a similar pattern in Wanda results as well, but Wanda performs consistently worse than \slopespace in these cases.

\niparagraph{Effects of sparsification on different modules.}
Each block in LLMs consists of a self-attention module and an MLP module, each containing multiple linear layers. We have analyzed the sensitivity of \slopespace to pruning each of those modules. Our results in \autoref{section:module_sensitivity} demonstrate that \slopespace can sustain competitive quality results while pruning all modules in the model.

\subsection{Ablation Studies}

\begin{figure}[ht]
\begin{center}
{\includegraphics[scale=1]{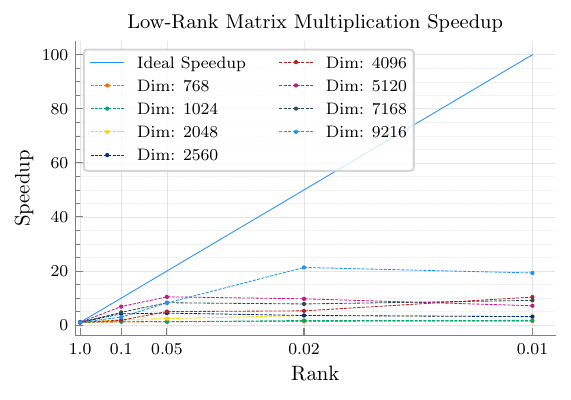}}
\caption{The speedup achieved by low-rank adapters in comparison to a dense matrix-multiplication.}
\label{fig:lora_speedup}
\end{center}
\end{figure}

\textbf{Low-Rank Adapter Performance: Scaling and Arithmetic Intensity.}
\label{section:lora_underutilization}
As discussed in \autoref{section:scheduling}, the computation time of low-rank adapters does \emph{not} scale linearly with their rank. This section provides experimental results to illustrate this behavior in more detail.
The computational complexity of low-rank matrix multiplications is $\mathcal{O}(b r d)$, where $b$, $r$, and $d$ represent the batch size, low-rank, and input/output dimensions of the layer, respectively.
Based on this complexity, we expect the computation time to be a linear function of $r$.  In other words, reducing $r$ by a factor of $\alpha$ should result in a corresponding $\alpha$-fold reduction in computation time.
However, in practice, this linearity does not hold. This deviation arises because the assumption underlying this expectation -- that matrix multiplication is compute-bound -- is not always true.  Specifically, the arithmetic intensity of the operation can fall below the machine's balance point, as described in the Roofline model~\cite{roofline} in \autoref{subsec:roofline_model}.
\autoref{fig:lora_speedup} shows the speedup achieved for different low-rank values using PyTorch's matrix multiplication function, which relies on the CUBLAS backend~\cite{cuda}.
The figure demonstrates that the achieved speedups are significantly lower than the ideal linear scaling, particularly when reducing the rank. Moreover, it is evident that as the matrix dimensions increase, the gap between the ideal speedup and the observed speedup diminishes.  This behavior can be attributed to the increased arithmetic intensity for larger matrices, leading to better utilization of tensor cores.

\paragraph{Efficient Low-rank Adapter Implementation.}
\label{section:low_rank_optimization_speedup}
As discussed in ~\autoref{section:scheduling}, a na\"ive implementation of low-rank adapters can lead to significant performance overheads due to the increased number of kernel launches and the low arithmetic intensity of their multiplications.
To address these issues, we introduced two key optimizations: (1) concatenating one of the low-rank adapters with the sparse weights, and (2) fusing the multiplication of the other low-rank adapter with the subsequent result addition.
These optimizations reduce kernel calls and increase arithmetic intensity, leading to more efficient utilization of GPU resources. \autoref{table:low_rank_optimization_speedup} summarizes the speedup improvements achieved with these optimizations, demonstrating an inference speedup increase of up to 6\%.
\begin{table}[t]
\caption{End-to-end speedup ($\times$) before and after efficient implementation of low-rank adapters.}
\label{table:low_rank_optimization_speedup}
\begin{center}
\begin{small}
\begin{sc}
\begin{tabular}{ccccc}
\toprule
\multirow{2}{*}{Model} & \multicolumn{2}{c}{1.56\% Adapter} & \multicolumn{2}{c}{6.25\% Adapter} \\
\cmidrule(lr){2-3} \cmidrule(lr){4-5}
        & Before & After & Before & After \\
\midrule
OPT-66B &	1.15 & 1.20 &	1.12 & 1.19 \\
OPT-30B &	1.13 & 1.18 &	1.10 & 1.16 \\
OPT-13B &	1.11 & 1.10 &	1.09 & 1.10 \\
OPT-6.6B &	1.07 & 1.12 &	1.06 & 1.11 \\
OPT-2.6B &	1.01 & 1.06 &	0.97 & 1.00 \\
\bottomrule
\end{tabular}
\end{sc}
\end{small}
\end{center}
\end{table}

\paragraph{Efficient Weight Tiling Implementation.}
\label{section:scheduling_results}
We observed that the dimensions and aspect ratios of matrices significantly influence system speedup (~\autoref{section:scheduling}).
To mitigate this, we implemented a matrix tiling strategy, dividing upsample matrices into multiple square matrices.
This approach significantly improves performance, as shown in \autoref{table:scheduling}.
Our results demonstrate that matrix tiling can enhance training speed by up to 4\% and inference speed by up to 12\%, highlighting its effectiveness in optimizing system performance.
\begin{table}[t]
\caption{End-to-end speedup ($\times$) before and after splitting the upsample matrix. In both cases, the optimization discussed in \autoref{table:low_rank_optimization_speedup} is used.}
\label{table:scheduling}
\begin{center}
\begin{small}
\begin{sc}
\begin{tabular}{ccccccccc}
\toprule
\multirow{2}{*}{Model} & \multicolumn{2}{c}{Training} & \multicolumn{2}{c}{\makecell{Inference \\ No Adapter}} & \multicolumn{2}{c}{\makecell{Inference \\ 1.56\% Adapter}} & \multicolumn{2}{c}{\makecell{Inference \\ 6.25\% Adapter}} \\
\cmidrule(lr){2-3} \cmidrule(lr){4-5} \cmidrule(lr){6-7} \cmidrule(lr){8-9}
        & Before & After & Before & After & Before & After & Before & After \\
\midrule
OPT-66B & 1.10 & 1.13 &	1.22 & 1.34 &	1.20 & 1.31 &	1.19 & 1.30 \\
OPT-30B & 1.09 & 1.14 &	1.23 & 1.32 &	1.18 & 1.28 &	1.16 & 1.27 \\
OPT-13B & 1.10 & 1.12 &	1.23 & 1.30 &	1.10 & 1.30 &	1.10 & 1.12 \\
OPT-6.6B & 1.08 & 1.08 &	1.21 & 1.19 &	1.12 & 1.13 &	1.11 & 1.12 \\
OPT-2.6B & 1.03 & 1.02 &	1.02 & 1.07 &	1.06 & 1.05 &	1.00 & 1.00 \\
\bottomrule
\end{tabular}
\end{sc}
\end{small}
\end{center}
\end{table}

\paragraph{\slopespace Sensitivity to Pruning Different Modules in Transformer.}
\label{section:module_sensitivity}
LLMs typically consist of two main modules: the MLP and the self-attention.
The attention module's weights are represented as a matrix in $\mathbb{R}^{d \times 3d}$, while the MLP uses weights in $\mathbb{R}^{d \times 4d}$ and $\mathbb{R}^{4d \times d}$, where $d$ denotes the hidden dimension.
To investigate the impact of sparsity on these modules, we conducted two experiments during Phase-2 of BERT-Large-Uncased pretraining: (a) [\textsc{MLP}]~$\rightarrow$~pruning only MLP modules, and (b) [\textsc{MLP + Self-Attention}]~$\rightarrow$~pruning both MLP and self-attention modules.
\autoref{table:different_modules} presents the SQuAD and GLUE results for these settings.
As expected, we observe a consistent, albeit slight, decrease in model quality as more modules are sparsified. The marginal decrease in performance suggests that models are relatively insensitive to the specific modules being pruned when using our \slopespace pretraining method. This observation underscores the robustness of our approach and its ability to maintain competitive quality across diverse sparsity configurations.
\begin{table}[t]
\caption{SQuADv1.1 results on BERT-Large-Uncased for different pruned modules.}
\label{table:different_modules}
\centering
\begin{small}
\begin{sc}
\begin{tabular}{ccc}
\toprule
Pruned Modules                  & SQuAD & GLUE \\
\midrule
Dense                           & 90.44 & 80.22 \\
MLP                             & 90.28  & 79.03 \\
MLP + Self-Attention            & 89.35  & 77.72\\
\bottomrule
\end{tabular}
\end{sc}
\end{small}
\end{table}
\section{Conclusion}
\label{section:slope_conclusion}

In this chapter, we presented \slope, a method that successfully executes the first strategy of the Compression Trinity: accelerating the computational cost of each pretraining iteration. By innovatively combining the sparsity pillar (via the double-pruned backward pass) and the low-rank pillar (via lazy adapters), \slope overcomes the rigidity of traditional sparse training. It delivers efficient N:M sparsity acceleration in both forward and backward passes while recovering model capacity through targeted low-rank updates. Our results demonstrate that this joint approach achieves up to \trainspeedup$\times$ speedup in pretraining and \inferencespeedup$\times$ speedup in inference, while reducing memory footprints by \trainmemory$\times$ and \inferencememory$\times$, respectively.

Together with MKOR (\autoref{chapter:mkor}), these contributions conclude our exploration of the Pretraining life-cycle stage. We have demonstrated that the Compression Trinity can effectively accelerate training by attacking the problem from two orthogonal angles: reducing the number of iterations via a Trinity-enhanced optimizer (MKOR), and reducing the cost per iteration via Trinity-enhanced weight structures (\slope).

The narrative now shifts to the second major stage of the LLM life-cycle: Post-Training Compression for Inference. While \slope produces efficient sparse models, the ultimate goal of the Trinity is to jointly apply all three pillars, i.e., Sparsity, Low-Rank, \textit{and} Quantization, to maximize inference efficiency on commodity hardware. However, applying these aggressive compression techniques simultaneously to a pre-trained, static model introduces a new challenge: compounded error. Before we can achieve the full Trinity in a one-shot inference setting, we must first establish a stable foundation. The next chapter introduces \optima, where we rigorously perfect the sparsity pillar to withstand the pressures of joint compression.

\chapter{\optima: Optimal One-Shot Pruning for LLMs via Quadratic Programming Reconstruction}
\label{chapter:optima}

\paragraph{Publication and Contributions.}
The content of this chapter is based on the paper
``OPTIMA: Optimal One-Shot Pruning for LLMs via
Quadratic Programming Reconstruction,'' \citep{optima} 2025. This
work was conducted in collaboration with Samuel
Kushnir, Amir Yazdanbakhsh, and Maryam Mehri
Dehnavi. Mohammad Mozaffari conceived the project,
led the implementation, and designed and executed the
experiments. Samuel Kushnir contributed to the
design of the algorithm. Amir Yazdanbakhsh and
Maryam Mehri Dehnavi supervised the project and
contributed to the writing and revision of the
manuscript.

\definecolor{shade}{RGB}{220, 230, 255}
\setlist[itemize]{
  noitemsep,
  topsep=0pt,
  leftmargin=*
}
\newcommand{\columndistance}{1.25cm}

\definecolor{optimaTeal}{RGB}{0,128,128}
\definecolor{optimaOrange}{RGB}{204,85,0}
\definecolor{optimaPurple}{RGB}{102,51,153}
\definecolor{optimaGray}{RGB}{110,110,110}
\definecolor{optimaBg}{RGB}{245,245,245}

\definecolor{KeywordColor}{RGB}{0, 102, 153}
\definecolor{VarColor}{RGB}{128, 0, 128}
\definecolor{CommentColor}{RGB}{0, 128, 0}

\renewcommand{\algorithmiccomment}[1]{\hfill {\color{CommentColor}\scriptsize$\triangleright$ #1}}

\algrenewcommand\alglinenumber[1]{\scriptsize #1}

\newcommand{\doublemidrule}{\midrule[\heavyrulewidth]
                            \midrule[\heavyrulewidth]}

\section{Introduction}
\label{sec:optima_intro}

Having addressed the computational bottlenecks of pretraining in \autoref{chapter:mkor} and \autoref{chapter:slope}, we now turn our attention to the second major stage of the LLM life-cycle: inference. As discussed in \autoref{chapter:background}, efficient inference is primarily constrained by memory bandwidth. While the Compression Trinity advocates for the joint application of sparsity, quantization, and low-rank approximations, blindly applying these methods to a pre-trained model carries significant risk. This is particularly true for \textbf{sparsity}, which we identified in \autoref{sec:intro_failure} as the most structurally destructive pillar of the Trinity. Unlike quantization, which preserves the network's topology, sparsity deletes connections entirely. If this structural skeleton is flawed, no amount of subsequent quantization or low-rank adaptation can recover the lost information.

Therefore, before we can integrate the full Trinity, we must first maximize the accuracy of the sparse foundation. In this chapter, we operate under a strict \textbf{resource-constrained regime}: we assume the practitioner requires a "one-shot" solution with no access to the full training pipeline or budget for backpropagation. The goal is to determine the mathematical limit of reconstruction accuracy achievable using only a small calibration dataset and static, layer-wise optimization.

Post-training one-shot pruning~\cite{sparsity_survey}, which removes parameters from a pretrained model using only a small calibration dataset, offers a potential solution. However, current methods are often forced to choose between efficiency and reconstruction optimality. We can categorize existing approaches into two tiers. The first tier consists of fast, metric-based selectors like Wanda~\cite{wanda}, magnitude pruning~\cite{magnitudepruning}, and ProxSparse~\cite{proxsparse}. While computationally cheap, they perform no weight updates to compensate for removed connections, treating weights as independent and ignoring the correlations captured by the loss landscape curvature. Conversely, principled second-order approaches like Optimal Brain Surgeon~\cite{obs} theoretically recover accuracy but are computationally infeasible at modern LLM scales. As a result, the second tier includes methods like SparseGPT~\cite{sparsegpt} and Thanos~\cite{thanos}, which attempt to adjust the remaining weights. However, to maintain speed, these methods rely on greedy approximations (e.g., iterative coordinate descent or localized Cholesky updates) rather than solving for the global optimum. Consequently, they leave significant performance on the table, an error that becomes significant when compounded with the quantization noise introduced in later chapters.\footnote{For a more detailed discussion of the related work, see \autoref{sec:optima_related_work}.}

To resolve this, we introduce \optima, a practical one-shot post-training pruning framework that combines layer-wise optimality with accelerator-grade efficiency. Distinct from prior heuristics, we formulate the weight update not as an approximation, but as an exact constrained Convex Quadratic Program (QP). This approach draws a direct parallel to the second-order optimization methods discussed in \autoref{chapter:background} (e.g., KFAC). Just as KFAC leverages the curvature of the loss landscape (via the Fisher Information Matrix) to improve training convergence over first-order methods, \optima utilizes the exact curvature of the layer-wise objective (via the Hessian matrix) to minimize pruning error.

The core of our methodology relies on a precise reformulation of the layer-wise reconstruction step. We observe that after fixing a binary mask for a weight matrix, the layer-wise output reconstruction (least-squares) objective decomposes across columns. We exploit a fundamental algebraic property of Transformer linear layers: while the linear constraints differ for each column (dictated by the mask), every column in the same layer shares the same Hessian matrix $H = X^\top X$. Unlike previous works that approximate this Hessian to simplify computation, we use this exact structure to formulate the update for each column as a small constrained QP. This shared-Hessian structure allows us to guarantee per-column global optimality for the reconstruction objective, strictly outperforming greedy heuristics without making additional assumptions about weight independence.

Realizing this formulation in practice requires careful numerical and systems engineering. We adopt a first-order primal–dual QP solver (rAPDHG~\cite{pdqp}) that is well-suited to our constrained problems, as its critical operations reduce to efficient matrix–vector products with the shared Hessian. We further avoid explicit dense equality matrices by enforcing fixed entries via tight bounds, accumulate layer Hessians incrementally from calibration sequences to save memory, and solve columns in batches so thousands of small QPs are processed in parallel. These implementation choices make \optima not only theoretically principled but also practical to run on a single accelerator.

We evaluate \optima across multiple model families (LLaMA, Gemma, and others) and sparsity regimes, including unstructured and 2:4 semi-structured sparsity. \optima is designed to be modular; it acts as a drop-in replacement for the weight-update step in existing mask selectors (e.g., Wanda, SparseGPT, Thanos), consistently improving their zero-shot performance. Across six zero-shot downstream benchmarks in the Language Model Evaluation Harness, we observe up to 3.97\% absolute gains on downstream tasks without any post-pruning fine-tuning.

In summary, our contributions are:
\begin{itemize}
\item We present a column-wise QP reformulation of the post-training reconstruction problem that yields per-column global optimality under a shared-Hessian model and is provably equivalent to the least-squares objective after mask selection (\autoref{sec:methodology}).
\item We design and implement an accelerator-friendly QP solver pipeline that accumulates a single Hessian per layer, enforces mask constraints via bounds, batches thousands of column QPs, and leverages rAPDHG/MPAX for efficient execution on GPUs/TPUs (detailed in \autoref{alg:layerwise_pruning}).
\item We demonstrate the modularity of \optima, showing it can be used as a drop-in weight-update step with common mask selection algorithms (Wanda, SparseGPT, Thanos), consistently improving their accuracy without fine-tuning (\autoref{sec:optima_results}).
\item We provide extensive empirical evidence and practical measurements. \optima yields substantial average accuracy gains across tasks and model sizes (up to 3.97\%), demonstrates robustness at high sparsity (up to 60\%), and can prune billion-parameter models on a single H100 in less than 40 hours.
\end{itemize}

\begin{figure}
    \centering
    \includegraphics[width=0.95\linewidth]{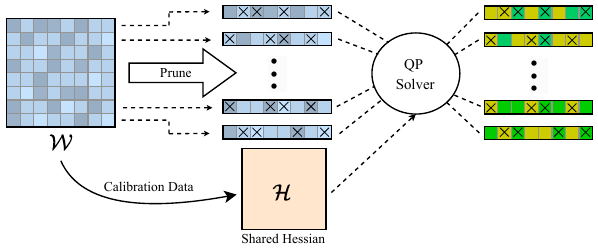}
    \caption{\optima generates a shared Hessian among the different columns of the pruned weight using a small calibration dataset. Then, the weights in different columns will be updated in parallel using a QP solver and the shared Hessian.}
    \label{fig:optima}
\end{figure}

\section{Additional Related Work}
\label{sec:optima_related_work}

Model pruning compresses trained neural networks by eliminating redundant weights, thereby lowering computational and memory requirements during deployment. The field primarily divides into two categories: layer-wise pruning, exemplified by Optimal Brain Surgeon (OBS) \citep{optimal_brain_surgeon}, and end-to-end pruning, represented by Optimal Brain Damage (OBD) \citep{optimal_brain_damage}. We review these approaches in the following subsections, beginning with layer-wise methods.

\paragraph{Layer-wise Model Pruning.} Layer-wise pruning optimizes models by targeting redundancies within individual layers, assuming that local error reductions aggregate to minimize overall model degradation. Optimal Brain Surgeon (OBS) \citep{optimal_brain_surgeon} formalizes this by identifying the least salient weight per layer and adjusting remaining weights to offset its removal \citep{optimal_brain_compression}. However, OBS's computational intensity hinders its application to billion-parameter LLMs, necessitating approximations. SparseGPT \citep{sparsegpt} pioneered scaling OBS to LLMs by framing pruning as sparse regression problems solved approximately, trading some accuracy for efficiency. Thanos \citep{thanos} refines this with multi-column pruning to cut approximation errors. In contrast, Wanda \citep{wanda} employs a saliency metric combining weight magnitudes and activation data from calibration sets, yielding strong results with minimal pruning time. Nonetheless, Wanda lacks mechanisms to update weights post-pruning, opening avenues for enhancements, particularly in end-to-end methods that consider global interactions.

\paragraph{End-to-end Model Pruning.} Unlike layer-wise methods, end-to-end pruning, exemplified by Optimal Brain Damage (OBD) \citep{optimal_brain_damage}, identifies least-important weights globally by leveraging second-order derivatives of the loss function, yielding higher accuracy than OBS. However, computing these derivatives is resource-intensive, demanding approximations \citep{mkor}. WoodFisher \citep{woodfisher} employs Kronecker factorization to approximate the Hessian, easing computation but still faltering at LLM scales. More recently, MaskLLM \citep{maskllm} sidesteps second-order information by recasting pruning as a classification problem solved via standard optimizers like AdamW \citep{adamw}, achieving top performance at 2:4 sparsity. ProxSparse \citep{proxsparse} reduces the costs of MaskLLM by using regularizers instead of training the model on a classification task, trading accuracy for speed. Yet, its optimization demands far exceed those of one-shot pruning, constraining real-world use and highlighting the value of integrating with other compression strategies.

\paragraph{Other Model Compression Methods.} In addition to pruning, several orthogonal techniques enable model compression and can be integrated with pruning for compounded benefits. Quantization reduces parameter precision to lower-bit representations, as surveyed in \citep{quantization_survey, quantization_survey2}, minimizing memory footprint without severe accuracy loss.

Low-rank adapters, such as those in \citep{slim, lqlora, slope}, decompose weight matrices into lower-dimensional factors, while knowledge distillation \citep{knowledge_distillation} transfers knowledge from larger teacher models to compact students. These methods complement pruning by addressing different aspects of redundancy, paving the way for hybrid frameworks in advanced compression research.

\section{Preliminaries}
\label{sec:optima_preliminaries}

Post-training pruning (PTP) compresses pre-trained models without retraining, using a small calibration dataset to produce a sparse model that preserves performance.
To make PTP tractable, the problem is decomposed into independent layer-wise subproblems. For layer $l$, the goal is to find a binary sparsity mask $\mathbf{M}_l$ and updated weights $\hat{\mathbf{W}}_l$ that minimize the output reconstruction error given original weights $\mathbf{W}_l$ and input activations $\mathbf{X}_l$. This task can be formulated as in \autoref{eq:joint_optimization}, where $\odot$ denotes the Hadamard product, and $\mathbf{M}_l$ is a binary tensor of the same shape as $\mathbf{W}_l$ with 0s for pruned weights and 1s for retained ones. \autoref{eq:joint_optimization} is solved sequentially across layers, with $\mathbf{X}_l$ as the pruned output from layer $l-1$. Finding the optimal $\mathbf{M}_l$ is NP-hard, motivating heuristics.

\begin{equation}
    \underset{\mathbf{M}_l, \hat{\mathbf{W}}_l}{\text{argmin}} \|\mathbf{X}_l \mathbf{W}_l - \mathbf{X}_l(\mathbf{M}_l \odot \hat{\mathbf{W}}_l)\|_F^2
    \label{eq:joint_optimization}
\end{equation}

A common heuristic decouples mask selection from weight updates. After selecting $\mathbf{M}_l$ (e.g., by magnitude), the problem simplifies to \autoref{eq:weight_only}, which is a convex least-squares problem, but solving it directly is computationally expensive for large LLM weights.

\begin{equation}
    \min_{\hat{\mathbf{W}}_l} \|\mathbf{X}_l \mathbf{W}_l - \mathbf{X}_l(\mathbf{M}_l \odot \hat{\mathbf{W}}_l)\|_F^2
    \label{eq:weight_only}
\end{equation}

Consequently, many methods employ strategies to circumvent the expensive weight update step. For example, \textbf{Wanda}~\cite{wanda}
avoids weight updates altogether, simply setting the selected weights to zero. However, other methods such as \textbf{SparseGPT} ~\cite{sparsegpt} and \textbf{Thanos} \citep{thanos} adopt a compromise, performing a more complex update but only on a small subset of the weights. These heuristics trade off optimality for computational feasibility.

\section{\optima: Optimal Weight Updates via Quadratic Programming}
\label{sec:methodology}
To overcome the challenges of weight update in LLM pruning, we propose \optima, a novel approach that enables the efficient and optimal update of \textbf{all} remaining weights once the pruning mask $\mathbf{M}_l$ has been chosen.

We achieve this by reformulating the least-squares problem as a set of independent Quadratic Programs (QPs) that can be solved in parallel on hardware accelerators like GPUs or TPUs using iterative methods. Specifically, we derive both a linearly constrained QP formulation and an equivalent unconstrained formulation. While the unconstrained form can be useful for optimizers restricted to such problems or in cases where it can be solved more efficiently, our implementation focuses on the constrained QP formulation, which is more amenable to GPU/TPU acceleration.

\subsection{Reformulation as a Quadratic Program with Linear Constraints}
\label{sec:qp}

As discussed in \autoref{sec:optima_preliminaries}, our goal is to minimize the problem defined in \autoref{eq:weight_only}. The Frobenius norm objective function in \autoref{eq:weight_only} is separable by the columns of the weight matrix.\footnote{Once the mask has been chosen, the weight reconstruction is separable for each column.} We can therefore solve the optimization problem for each column independently.

Let $\mathbf{w}_j$ be the $j$-th column of the original weight matrix $\mathbf{W}_l$, and let $\hat{\mathbf{w}}_j$ be the corresponding column in the updated matrix $\hat{\mathbf{W}}_l$. The mask for this column is $\mathbf{m}_j$. The optimization for this single column can be formulated as in \autoref{eq:single_column}.

\begin{equation}
\min_{\hat{\mathbf{w}}_j} \|\mathbf{X}_l \mathbf{w}_j - \mathbf{X}_l (\mathbf{m}_j \odot \hat{\mathbf{w}}_j)\|_2^2
\label{eq:single_column}
\end{equation}

By defining the change in the weight column as $\Delta\mathbf{w}_j = (\mathbf{m}_j \odot \hat{\mathbf{w}}_j) - \mathbf{w}_j$, the objective can then be rewritten in terms of this change as in \autoref{eq:delta_w} in standard quadratic form.

\begin{equation}
\min_{\Delta\mathbf{w}_j} \|-\mathbf{X}_l \Delta\mathbf{w}_j\|_2^2 = \min_{\Delta\mathbf{w}_j} \Delta\mathbf{w}_j^T (\mathbf{X}_l^T \mathbf{X}_l) \Delta\mathbf{w}_j
\label{eq:delta_w}
\end{equation}

The constraints on $\Delta\mathbf{w}_j$ in \autoref{eq:delta_w} are determined by the mask $\mathbf{m}_j$. Let $\mathcal{S}_j$ be the set of indices where the mask is zero (i.e., weights to be pruned). For each index $i \in \mathcal{S}_j$, the corresponding entry in the updated weight vector, $(\hat{\mathbf{w}}_j)_i$, must be zero. This imposes a linear constraint on the change vector, as shown in \autoref{eq:constraints}.

\begin{equation}
(\mathbf{m}_j \odot \hat{\mathbf{w}}_j)_i = 0 \implies (\Delta\mathbf{w}_j)_i = -(\mathbf{w}_j)_i \quad \forall i \in \mathcal{S}_j
\label{eq:constraints}    
\end{equation}

The entries of $\Delta\mathbf{w}_j$ for the unpruned weights (where $m_{ij}=1$) remain as free variables to be optimized.

For each column $j$ of the weight matrix, we have a QP of the form represented in \autoref{eq:constrained_objective}, where $\mathbf{H} = \mathbf{X}_l^T \mathbf{X}_l$ is the Hessian matrix, which is positive semi-definite and shared across all column-wise problems. The fact that the Hessian is shared among all columns, and only the constraints change, makes it very easy to parallelize on accelerators such as GPUs and TPUs.

\begin{equation}
\begin{aligned}
& \underset{\Delta\mathbf{w}_j}{\text{minimize}}
& & \Delta\mathbf{w}_j^T \mathbf{H} \Delta\mathbf{w}_j \\
& \text{subject to}
& & (\Delta\mathbf{w}_j)_i = -(\mathbf{w}_j)_i, \; \forall i \in \mathcal{S}_j
\end{aligned}
\label{eq:constrained_objective}
\end{equation}

\subsection{Reformulation as an Unconstrained Quadratic Program}

As an alternative to the constrained formulation in \autoref{eq:constrained_objective}, we can reformulate each column-wise problem as an unconstrained quadratic program. This can be useful in settings where solvers are optimized for unconstrained problems or when eliminating constraints enables more efficient optimization. Although our implementation adopts the constrained approach for reasons discussed below, we include the unconstrained version for completeness.

The key idea is to eliminate the equality constraints in \autoref{eq:constraints} by substituting them directly into the objective. For a given column $j$, define $\mathcal{I}_j$ as the set of indices where the mask is one (i.e., unpruned weights), and let $\mathcal{S}_j$ denote the complement set (i.e., pruned weights, where the mask is zero).

We reorder the entries of the change vector $\Delta\mathbf{w}_j$ and the shared Hessian matrix $\mathbf{H} = \mathbf{X}_l^T \mathbf{X}_l$ based on this partitioning, as shown in \autoref{eq:block_partitioning}.

\begin{equation}
\Delta\mathbf{w}_j = \begin{bmatrix} \Delta\mathbf{w}_{\mathcal{I}_j} \\ \Delta\mathbf{w}_{\mathcal{S}_j} \end{bmatrix}, \quad
\mathbf{H} = \begin{bmatrix}
\mathbf{H}_{\mathcal{I}_j\mathcal{I}_j} & \mathbf{H}_{\mathcal{I}_j\mathcal{S}_j} \\
\mathbf{H}_{\mathcal{S}_j\mathcal{I}_j} & \mathbf{H}_{\mathcal{S}_j\mathcal{S}_j}
\end{bmatrix}
\label{eq:block_partitioning}
\end{equation}

As established in \autoref{eq:constraints}, the entries of $\Delta\mathbf{w}_j$ corresponding to $\mathcal{S}_j$ are fixed: $(\Delta\mathbf{w}_j)_i = -(\mathbf{w}_j)_i$ for all $i \in \mathcal{S}_j$. Substituting these fixed values into the quadratic objective yields the expanded form in \autoref{eq:expanded_objective}.

\begin{equation}
\begin{aligned}
\Delta\mathbf{w}_j^T \mathbf{H} \Delta\mathbf{w}_j &=
\Delta\mathbf{w}_{\mathcal{I}_j}^T \mathbf{H}_{\mathcal{I}_j\mathcal{I}_j} \Delta\mathbf{w}_{\mathcal{I}_j} +
2 \Delta\mathbf{w}_{\mathcal{I}_j}^T \mathbf{H}_{\mathcal{I}_j\mathcal{S}_j} \Delta\mathbf{w}_{\mathcal{S}_j} +
\Delta\mathbf{w}_{\mathcal{S}_j}^T \mathbf{H}_{\mathcal{S}_j\mathcal{S}_j} \Delta\mathbf{w}_{\mathcal{S}_j}
\end{aligned}
\label{eq:expanded_objective}
\end{equation}

Since $\Delta\mathbf{w}_{\mathcal{S}_j} = -\mathbf{w}_{\mathcal{S}_j}$, we substitute this to obtain the unconstrained objective in \autoref{eq:unconstrained_objective}.

\begin{equation}
\min_{\Delta\mathbf{w}_{\mathcal{I}_j}} \left(
\Delta\mathbf{w}_{\mathcal{I}_j}^T \mathbf{H}_{\mathcal{I}_j\mathcal{I}_j} \Delta\mathbf{w}_{\mathcal{I}_j}
- 2 \Delta\mathbf{w}_{\mathcal{I}_j}^T \mathbf{H}_{\mathcal{I}_j\mathcal{S}_j} \mathbf{w}_{\mathcal{S}_j}
+ \mathbf{w}_{\mathcal{S}_j}^T \mathbf{H}_{\mathcal{S}_j\mathcal{S}_j} \mathbf{w}_{\mathcal{S}_j}
\right)
\label{eq:unconstrained_objective}
\end{equation}

The final term in \autoref{eq:unconstrained_objective} is constant with respect to the optimization variable $\Delta\mathbf{w}_{\mathcal{I}_j}$ and can therefore be omitted. This results in the unconstrained quadratic program in \autoref{eq:final_unconstrained_qp}.

\begin{equation}
\underset{\Delta\mathbf{w}_{\mathcal{I}_j}}{\text{minimize}} \quad
\Delta\mathbf{w}_{\mathcal{I}_j}^T \mathbf{Q}_j \Delta\mathbf{w}_{\mathcal{I}_j}
+ \mathbf{c}_j^T \Delta\mathbf{w}_{\mathcal{I}_j}
\label{eq:final_unconstrained_qp}
\end{equation}

where the problem-specific matrix and vector are defined as:
\begin{equation}
\mathbf{Q}_j = \mathbf{H}_{\mathcal{I}_j\mathcal{I}_j}, \quad
\mathbf{c}_j = -2 \mathbf{H}_{\mathcal{I}_j\mathcal{S}_j} \mathbf{w}_{\mathcal{S}_j}
\label{eq:unconstrained_qp_params}
\end{equation}

This formulation eliminates the need for explicit constraints, but introduces column-dependent variation in problem dimensions. Specifically, the size of $\mathbf{Q}_j$ and $\mathbf{c}_j$ varies with the number of unpruned weights in each column. Consequently, the unconstrained QPs have heterogeneous shapes and objectives across columns, making them more difficult to batch and parallelize efficiently on accelerators like GPUs or TPUs. This motivates our choice to adopt the constrained formulation in \autoref{eq:constrained_objective}, where the problem structure is uniform and well-suited for high-throughput parallel execution.

\subsection{Solving the Quadratic Programs}

With the constrained QP formulation established, we now select a solver, whose efficiency is crucial for runtime and scalability on parallel hardware like GPUs and TPUs. Our QP, with its shared Hessian $\mathbf{H}$ and simple bounds, suits specialized modern solvers. We adopt the state-of-the-art Restarted Accelerated Primal-Dual Hybrid Gradient (rAPDHG) algorithm \citep{pdqp}, a first-order method effective here for three reasons: (1) its bottleneck—matrix-vector multiplications with $\mathbf{H}$ and its transpose—runs efficiently on GPUs/TPUs; (2) it achieves provably optimal linear convergence; and (3) a high-performance, open-source JAX-based implementation is available in MPAX \citep{mpax}, designed for GPU/TPU execution. This enables parallel solving of thousands of column-wise QPs, leveraging the shared structure.

\subsection{Efficient Implementation}

Naively implementing the optimization problem in \autoref{eq:constrained_objective} is computationally expensive and incurs substantial memory overhead. These costs, however, can be greatly reduced through a series of optimization techniques. In the following, we describe the strategies we employ to solve the QPs efficiently on a single GPU, even for very large LLMs. Additionally, a detailed algorithm of our implementation is provided in \autoref{alg:layerwise_pruning}.

\begin{algorithm}[t]
\caption{\textbf{Layer-wise Pruning with Batched Column-wise Quadratic Programming}}
\label{alg:layerwise_pruning}
\begin{algorithmic}[1]
\Statex \textbf{\color{KeywordColor}Input:} Pre-trained LLM $\color{VarColor}\mathcal{M}$, calibration data $\color{VarColor}\mathbf{X}$, pruning masks $\color{VarColor}\mathcal{M}_{\text{ask}}$, QP solver $\color{VarColor}\mathcal{S}$, batch size $\color{VarColor}B$.
\Statex \textbf{\color{KeywordColor}Output:} Pruned and updated LLM $\color{VarColor}\hat{\mathcal{M}}$, updated masks $\color{VarColor}\hat{\mathcal{M}}_{\text{ask}}$.

\vspace{0.5em}
\For{\textbf{each layer} $\color{VarColor}L$ in the LLM $\color{VarColor}\mathcal{M}$}
    \State Initialize Hessian estimate $\color{VarColor}\mathbf{H} \gets 0$. \algorithmiccomment{\normalsize{Initialize covariance matrix}}
    \vspace{0.2em}
    \For{\textbf{each calibration sample} $\color{VarColor}x \in \color{VarColor}\mathbf{X}$}
        \State $\color{VarColor}y \gets L(\color{VarColor}x)$ \algorithmiccomment{\normalsize{Forward pass for one sequence}}
        \State $\color{VarColor}\mathbf{H} \gets \color{VarColor}\mathbf{H} + \color{VarColor}y^T \color{VarColor}y$ \algorithmiccomment{\normalsize{Accumulate covariance}}
    \EndFor
    \State Store intermediate inputs $\{\color{VarColor}\mathbf{X}_{\mathbf{W}} \mid \color{VarColor}\mathbf{W} \in \color{VarColor}L\}$ from a forward pass of $\color{VarColor}L(\color{VarColor}\mathbf{X})$.

    \vspace{0.3em}
    \For{\textbf{each weight matrix} $\color{VarColor}\mathbf{W}$ in layer $\color{VarColor}L$}
        \State Retrieve corresponding mask $\color{VarColor}\mathbf{M} \in \color{VarColor}\mathcal{M}_{\text{ask}}$.
        \State Partition the columns of $\color{VarColor}\mathbf{W}$ into batches of size $\color{VarColor}B$.
        \For{\textbf{each batch of columns} $\{\color{VarColor}\mathbf{w}_j\}_{j=1}^B$ \textbf{in parallel}}
            \For{\textbf{each column} $\color{VarColor}\mathbf{w}_j$ in the batch}
                \State $\color{VarColor}\mathcal{S}_j \gets \{ i \mid \color{VarColor}\mathbf{M}_{j,i} = 0 \}$ \algorithmiccomment{\normalsize{Indices of pruned entries}}
                \State Define QP:
                \begin{equation}
                \begin{aligned}
                    &\min_{\color{VarColor}\Delta \mathbf{w}_j} \color{VarColor}\Delta \mathbf{w}_j^T \color{VarColor}\mathbf{H} \color{VarColor}\Delta \mathbf{w}_j \\
                    &\text{s.t. } (\color{VarColor}\Delta \mathbf{w}_j)_i = -(\color{VarColor}\mathbf{w}_j)_i, \; \forall i \in \color{VarColor}\mathcal{S}_j
                \end{aligned}
                \end{equation}
            \EndFor
            \State $\color{VarColor}\{\Delta \mathbf{w}_j\}_{j=1}^B \gets \color{VarColor}\mathcal{S}(\color{VarColor}\mathbf{H}, \{\color{VarColor}\mathbf{w}_j\}_{j=1}^B, \{\color{VarColor}\mathcal{S}_j\}_{j=1}^B)$
            \State Update weights: $\color{VarColor}\mathbf{w}_j \gets \color{VarColor}\mathbf{w}_j + \color{VarColor}\Delta \mathbf{w}_j, \quad \forall j$
        \EndFor
    \EndFor
    \State $\color{VarColor}\mathbf{X} \gets \color{VarColor}L(\color{VarColor}\mathbf{X})$ \algorithmiccomment{\normalsize{Update activations for next layer}}
\EndFor

\vspace{0.5em}
\Statex \textbf{\color{KeywordColor}Return:} Updated model $\color{VarColor}\hat{\mathcal{M}}$, updated masks $\color{VarColor}\hat{\mathcal{M}}_{\text{ask}}$.
\end{algorithmic}
\end{algorithm}

\paragraph{Equality Constraints.} Directly encoding the constraints from \autoref{eq:constraints} into the standard quadratic objective leads to a prohibitively large matrix of equalities, even though these constraints merely fix individual variables to constant values. To avoid constructing such large matrices, we instead enforce the constraints by setting upper and lower bounds on the corresponding variables. In particular, fixing the bounds of $(\Delta w_j)_i$ to $-(w_j)_i$ effectively locks the variable to the desired value, without incurring the overhead of explicit equality matrices.

\paragraph{Batching QP Problems.} In memory-limited scenarios, the optimization problems for all columns of the weight matrices may not fit on a single GPU. To address this, we employ a batching strategy that solves a subset of QP problems at a time. This approach reduces memory overhead while still leveraging the efficiency of solving multiple QPs in parallel. As a result, our method enables pruning of large LLMs even on a single GPU.

\paragraph{Hessian calculation.} For each layer, the Hessian matrix can be estimated as the covariance of the dense model’s outputs across multiple sequences. Suppose the output tensor is $Y \in \mathbb{R}^{b \times s \times d}$, where $b$ is the number of sequences, $s$ is the sequence length, and $d$ is the output dimension of the layer. To compute the covariance directly, we would first reshape $Y$ into $\hat{Y} \in \mathbb{R}^{bs \times d}$, effectively stacking all tokens from all sequences into a single matrix, and then evaluate $\hat{Y}^T \hat{Y}$.

While this formulation is straightforward, it requires storing the full $Y$ in accelerator memory, which becomes prohibitively expensive for large $b$ and $s$, often causing out-of-memory errors. To make the computation feasible, we observe that the covariance can be accumulated incrementally. Specifically, $Y$ can be decomposed into $b$ smaller matrices, $y_i \in \mathbb{R}^{s \times d}$, each corresponding to the output of a single sequence. Instead of materializing $\hat{Y}$, we compute $y_i^T y_i$ for each sequence separately and sum the results as in $H \approx \sum_{i=1}^b{y_i^T y_i}$. This decomposition yields the same result as computing $\hat{Y}^T \hat{Y}$ directly, but avoids the need to store the entire $Y$ at once, making the approach scalable to very large LLMs.

\section{Experiments}
\label{sec:optima_results}

\paragraph{Model, datasets, and evaluation.} We evaluate \optima on LLaMA 3.1, LLaMA 3.2 \citep{llama3}, Gemma 2 \citep{gemma2}, and Gemma 3 \citep{gemma3} family of models. Model accuracy is assessed on a range of zero-shot downstream tasks, including MMLU \citep{mmlu}, Piqa \citep{piqa}, Arc-Easy, Arc-Challenge \citep{arc}, WinoGrande \citep{winogrande}, and OpenBookQA \citep{openbookqa}, all of which are commonly used to evaluate LLM compression \citep{slim, wanda}. For zero-shot evaluations, we utilize the Language Model Evaluation Harness \citep{lmharness} framework. In line with prior work \citep{wanda, sparsegpt, slim}, we also report the perplexity of the models on a language modeling task on the WikiText2 \citep{wikitext2} dataset.
\begin{figure}[t]
    \centering
    \includegraphics[width=0.95\linewidth]{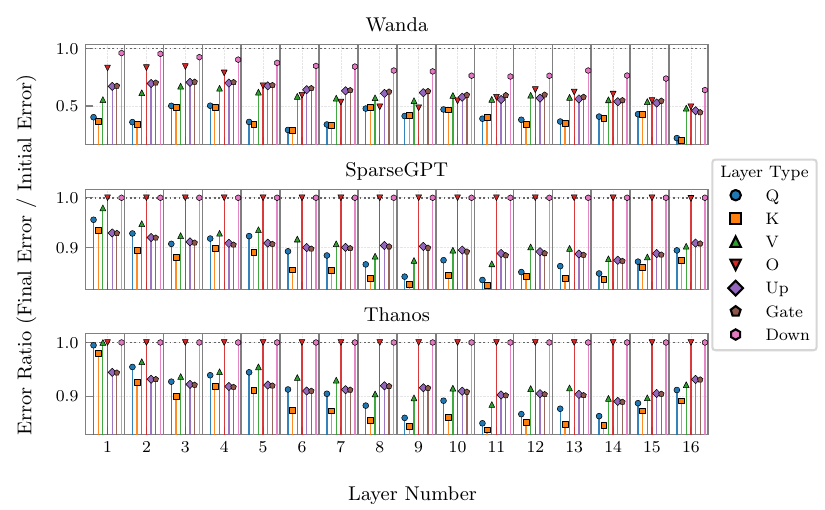}
    \caption{Relative error reduction on \optima in comparison to Wanda, SparseGPT, and Thanos for LLaMA-3.2 1B.}
    \label{fig:layerwise_error}
\end{figure}

\paragraph{Baselines.} We compare \optima against state-of-the-art one-shot pruning methods, including Wanda \citep{wanda}, SparseGPT \citep{sparsegpt}, Thanos \citep{thanos}, and ProxSparse \citep{proxsparse} and show how \optima can improve the performance of all these pruning methods across different models and datasets.
The sensitivity of \optima to the calibration dataset size can be found in \autoref{app:calibration_size}. In terms of memory reductions and speedup, our method is guaranteed to achieve the same performance as other pruning methods such as Wanda and SparseGPT, since the sparsity pattern in these methods stays intact.

\niparagraph{Experiment Setup.} Following previous work \citep{sparsegpt, wanda, slim, thanos}, we use 128 samples, each with 2048 tokens from the C4 dataset \citep{c4} for calibration. We set the relative and absolute tolerance of the rAPDHG QP solver in MPAX to 0.01 and the maximum number of iterations to 100,000. If the optimizer does not converge within this budget for most problems, or the final error of a layer is larger than the initial error, \optima skips updating that layer. \autoref{tab:hyperparameters} summarizes the key hyperparameters. For all baselines, we either use their publicly available checkpoints or reproduce results with default hyperparameters.

\begin{table}[bt]
\centering
\caption{Key hyperparameters used in \optima.}
\label{tab:hyperparameters}
\begin{tabular}{lc}
\toprule
\textbf{Hyperparameter} & \textbf{Value} \\
\midrule
Calibration Samples & 128 \\
Tokens per Sample & 2048 \\
Dataset for Calibration & C4 \\
Relative Tolerance (rAPDHG) & 0.01 \\
Absolute Tolerance (rAPDHG) & 0.01 \\
Maximum Iterations (rAPDHG) & 100,000 \\
ADAM Learning Rate & \{$10^{-2}, 10^{-3}, 10^{-4}, 10^{-5}$\} \\
ADAM Weight Decay & 0 \\
\bottomrule
\end{tabular}
\end{table}

\paragraph{Model Quality.} We evaluate the accuracy of \optima and other state-of-the-art pruning methods across 2:4 and unstructured sparsity benchmarks. Wanda is a mask selection algorithm that does not provide any weight update mechanism for the weights. SparseGPT and Thanos, on the other hand, update the weight values in addition to searching for the best mask. We couple \optima weight update with the masks generated using each of these methods and compare the resulting performance of the models.

\autoref{tab:50_unstructured} summarizes the performance metrics for Wanda, SparseGPT, and Thanos with and without the \optima update mechanism for 50\% unstructured sparsity. It can be seen that models pruned with \optima weight update scheme consistently outperform the methods using weight update methods, providing up to 1.80\% average accuracy improvement across six downstream tasks (Gemma-3-1B).

\begin{table}[tb]
\centering
\setlength{\tabcolsep}{2pt}
\resizebox{\textwidth}{!}{
\begin{tabular}{l l l c c c c c c c | c}
\toprule
\multirow{2}{*}{\parbox[t]{1.5cm}{\centering Model}} &
\multirow{2}{*}{\parbox[t]{1.3cm}{\centering Mask\\Selection}} &
\multirow{2}{*}{\parbox[t]{1cm}{\centering Weight\\Update}} &
\multirow{2}{*}{Perplexity} &
\multicolumn{7}{c}{Metrics (\%)} \\
\cmidrule{5-11}
 & & & &
   \parbox[t]{\columndistance}{\centering \textcolor{blue!70!black}{MMLU}} &
   \parbox[t]{\columndistance}{\centering \textcolor{blue!70!black}{PIQA}} &
   \parbox[t]{\columndistance}{\centering \textcolor{blue!70!black}{Arc-E}} &
   \parbox[t]{\columndistance}{\centering \textcolor{blue!70!black}{Arc-C}} &
   \parbox[t]{\columndistance}{\centering \textcolor{blue!70!black}{Wino}} &
   \parbox[t]{\columndistance}{\centering \textcolor{blue!70!black}{OpenQA}} &
   \parbox[t]{\columndistance}{\centering \textbf{\textcolor{teal}{Average}}} \\
\midrule
\multirow{7}{*}{\parbox[t]{1.5cm}{\centering LLaMA\\3.1 8B}} & Dense & -- &  5.84 & 63.57 & 80.09 & 81.44 & 51.37 & 73.48 & 33.40 & 63.89 \\
 & Wanda & -- &  9.64 & 47.79 & 75.68 & 72.56 & 40.70 & 70.09 & 27.40 & 55.70 \\
 & \cellcolor{shade}{Wanda} & \cellcolor{shade}{\optima} & \cellcolor{shade}{9.37} & \cellcolor{shade}{\textbf{48.85}} & \cellcolor{shade}{\textbf{76.71}} & \cellcolor{shade}{\textbf{73.82}} & \cellcolor{shade}{\textbf{42.32}} & \cellcolor{shade}{\textbf{70.32}} & \cellcolor{shade}{\textbf{28.20}} & \cellcolor{shade}{\textbf{56.70}} \\
 & SparseGPT & SparseGPT &  9.30 & \textbf{51.32} & 76.19 & 73.02 & 41.27 & \textbf{70.88} & \textbf{29.40} & 57.01 \\
 & \cellcolor{shade}{SparseGPT} & \cellcolor{shade}{\optima} & \cellcolor{shade}{9.33} & \cellcolor{shade}{49.31} & \cellcolor{shade}{\textbf{76.61}} & \cellcolor{shade}{\textbf{74.28}} & \cellcolor{shade}{\textbf{\textbf{42.83}}} & \cellcolor{shade}{\textbf{70.88}} & \cellcolor{shade}{28.20} & \cellcolor{shade}{\textbf{57.02}} \\
 & Thanos & Thanos & 9.27 & \textbf{50.36} & \textbf{77.04} & \textbf{74.92} & \textbf{42.58} & \textbf{70.96} & \textbf{30.00} & \textbf{57.64} \\
 & \cellcolor{shade}{Thanos} & \cellcolor{shade}{\optima} & \cellcolor{shade}{9.35} & \cellcolor{shade}{50.17} & \cellcolor{shade}{76.50} & \cellcolor{shade}{74.16} & \cellcolor{shade}{41.89} & \cellcolor{shade}{70.24} & \cellcolor{shade}{28.40} & \cellcolor{shade}{56.89} \\
 \doublemidrule
\multirow{7}{*}{\parbox[t]{1.5cm}{\centering LLaMA\\3.2 1B}} & Dense & -- & 9.75 & 36.92 & 74.27 & 65.53 & 31.31 & 60.30 & 26.20 & 49.09
\\
 & Wanda & -- & 23.51 & 26.35 & 65.18 & 52.10 & 23.81 & 54.62 & 18.00 & 40.01 \\
 & \cellcolor{shade}{Wanda} & \cellcolor{shade}{\optima} & \cellcolor{shade}{18.84} & \cellcolor{shade}{\textbf{27.69}} & \cellcolor{shade}{\textbf{67.08}} & \cellcolor{shade}{\textbf{52.61}} & \cellcolor{shade}{\textbf{24.74}} & \cellcolor{shade}{\textbf{55.64}} & \cellcolor{shade}{\textbf{20.20}} & \cellcolor{shade}{\textbf{41.33}} \\
 & SparseGPT & SparseGPT & 18.84 & 25.71 & 67.85 & 54.29 & \textbf{26.54} & \textbf{57.70} & 22.00 & 42.35 \\
 & \cellcolor{shade}{SparseGPT} & \cellcolor{shade}{\optima} & \cellcolor{shade}{18.09} & \cellcolor{shade}{\textbf{26.95}} & \cellcolor{shade}{\textbf{68.01}} & \cellcolor{shade}{\textbf{54.59}} & \cellcolor{shade}{25.85} & \cellcolor{shade}{56.91} & \cellcolor{shade}{\textbf{24.00}} & \cellcolor{shade}{\textbf{42.72}} \\
 & Thanos & Thanos & 19.70 & 25.37 & 67.63 & 52.99 & \textbf{27.13} & 54.38 & \textbf{22.20} & 41.62 \\
 & \cellcolor{shade}{Thanos} & \cellcolor{shade}{\optima} & \cellcolor{shade}{18.77} & \cellcolor{shade}{\textbf{25.99}} & \cellcolor{shade}{\textbf{68.23}} & \cellcolor{shade}{\textbf{53.49}} & \cellcolor{shade}{26.45} & \cellcolor{shade}{\textbf{55.88}} & \cellcolor{shade}{21.60} & \cellcolor{shade}{\textbf{41.94}} \\
\doublemidrule
\multirow{7}{*}{\parbox[t]{1.5cm}{\centering LLaMA\\3.2 3B}} & Dense & -- & 7.81 & 54.13 & 76.55 & 74.28 & 42.75 & 69.38 & 30.60 & 57.95 \\
 & Wanda & -- & 12.92 & 40.79 & 72.03 & 65.45 & 32.34 & 63.69 & 25.40 & 49.95 \\
 & \cellcolor{shade}{Wanda} & \cellcolor{shade}{\optima} & \cellcolor{shade}{12.24} & \cellcolor{shade}{\textbf{43.11}} & \cellcolor{shade}{\textbf{72.47}} & \cellcolor{shade}{\textbf{66.50}} & \cellcolor{shade}{\textbf{33.53}} & \cellcolor{shade}{\textbf{66.38}} & \cellcolor{shade}{\textbf{26.20}} & \cellcolor{shade}{\textbf{51.37}} \\
& SparseGPT & SparseGPT & 12.32 & 37.96 & \textbf{73.45} & 65.19 & 33.02 & 66.38 & 25.20 & 50.20 \\
 & \cellcolor{shade}{SparseGPT} & \cellcolor{shade}{\optima} & \cellcolor{shade}{12.43} & \cellcolor{shade}{\textbf{40.54}} & \cellcolor{shade}{\textbf{73.45}} & \cellcolor{shade}{\textbf{66.37}} & \cellcolor{shade}{\textbf{35.07}} & \cellcolor{shade}{\textbf{66.69}} & \cellcolor{shade}{\textbf{26.20}} & \cellcolor{shade}{\textbf{51.39}} \\
& Thanos & Thanos & 12.26 & 40.11 & 72.80 & 64.77 & 32.85 & \textbf{67.72} & 26.60 & 50.81 \\
 & \cellcolor{shade}{Thanos} & \cellcolor{shade}{\optima} & \cellcolor{shade}{12.40} & \cellcolor{shade}{\textbf{41.51}} & \cellcolor{shade}{\textbf{73.23}} & \cellcolor{shade}{\textbf{65.07}} & \cellcolor{shade}{\textbf{34.39}} & \cellcolor{shade}{67.25} & \cellcolor{shade}{\textbf{27.00}} & \cellcolor{shade}{\textbf{51.41}} \\
\doublemidrule
\multirow{7}{*}{\parbox[t]{1.5cm}{\centering Gemma\\3 1B}} & Dense & -- & 14.17 & 24.95 & 74.81 & 71.93 & 35.41 & 58.72 & 28.80 & 49.10 \\
 & Wanda & -- & {32.96} & 22.97 & 67.19 & 61.03 & 26.37 & 55.72 & 20.00 & 42.21 \\
 & \cellcolor{shade}{Wanda} & \cellcolor{shade}{\optima} & \cellcolor{shade}{28.90} & \cellcolor{shade}{\textbf{23.96}} & \cellcolor{shade}{\textbf{69.48}} & \cellcolor{shade}{\textbf{62.84}} & \cellcolor{shade}{\textbf{28.58}} & \cellcolor{shade}{\textbf{56.83}} & \cellcolor{shade}{\textbf{22.40}} & \cellcolor{shade}{\textbf{44.01}} \\
 & SparseGPT & SparseGPT & {28.34} & 24.85 & 68.88 & \textbf{60.94} & 26.62 & 55.49 & 21.40 & 43.03 \\
 & \cellcolor{shade}{SparseGPT} & \cellcolor{shade}{\optima} & \cellcolor{shade}{27.35} & \cellcolor{shade}{\textbf{25.73}} & \cellcolor{shade}{\textbf{69.75}} & \cellcolor{shade}{60.90} & \cellcolor{shade}{\textbf{27.82}} & \cellcolor{shade}{\textbf{56.35}} & \cellcolor{shade}{\textbf{22.00}} & \cellcolor{shade}{\textbf{43.76}} \\
 & Thanos & Thanos & {28.65} & 23.09 & \textbf{69.75} & 62.16 & \textbf{27.99} & \textbf{56.51} & \textbf{23.80} & 43.88 \\
 & \cellcolor{shade}{Thanos} & \cellcolor{shade}{\optima} & \cellcolor{shade}{28.14} & \cellcolor{shade}{\textbf{24.70}} & \cellcolor{shade}{69.64} & \cellcolor{shade}{\textbf{63.43}} & \cellcolor{shade}{27.39} & \cellcolor{shade}{55.96} & \cellcolor{shade}{23.20} & \cellcolor{shade}{\textbf{44.05}} \\
\doublemidrule
\multirow{7}{*}{\parbox[t]{1.5cm}{\centering Gemma\\2 2B}} & Dense & -- & 68.69 & 49.33 & 78.24 & 80.22 & 46.93 & 68.82 & 31.40 & 59.16 \\
 & Wanda & -- & {327.45} & 34.17 & \textbf{74.16} & 69.78 & \textbf{34.30} & \textbf{62.83} & \textbf{26.40} & \textbf{50.27} \\
 & \cellcolor{shade}{Wanda} & \cellcolor{shade}{\optima} & \cellcolor{shade}{215.63} & \cellcolor{shade}{\textbf{34.86}} & \cellcolor{shade}{73.99} & \cellcolor{shade}{\textbf{71.38}} & \cellcolor{shade}{32.59} & \cellcolor{shade}{61.96} & \cellcolor{shade}{25.80} & \cellcolor{shade}{50.10} \\
 & SparseGPT & SparseGPT & 234.68 & 35.59 & 73.61 & 69.99 & 34.22 & \textbf{65.82} & \textbf{28.20} & 51.24 \\
 & \cellcolor{shade}{SparseGPT} & \cellcolor{shade}{\optima} & \cellcolor{shade}{{241.09}} & \cellcolor{shade}{\textbf{37.59}} & \cellcolor{shade}{\textbf{73.83}} & \cellcolor{shade}{\textbf{70.62}} & \cellcolor{shade}{\textbf{35.07}} & \cellcolor{shade}{64.72} & \cellcolor{shade}{27.80} & \cellcolor{shade}{\textbf{51.60}} \\
 & Thanos & Thanos & {276.97} & 30.62 & 73.18 & 67.72 & 33.62 & 63.22 & \textbf{26.80} & 49.19 \\
 & \cellcolor{shade}{Thanos} & \cellcolor{shade}{\optima} & \cellcolor{shade}{250.15} & \cellcolor{shade}{\textbf{32.72}} & \cellcolor{shade}{\textbf{73.72}} & \cellcolor{shade}{\textbf{68.81}} & \cellcolor{shade}{\textbf{34.13}} & \cellcolor{shade}{\textbf{63.85}} & \cellcolor{shade}{26.40} & \cellcolor{shade}{\textbf{49.94}} \\
\bottomrule
\end{tabular}
}
\caption{Model perplexity on WikiText2 and accuracy on zero-shot downstream tasks for 50\% unstructured sparsity. \optima consistently improves the accuracy of the models across different tasks.}
\label{tab:50_unstructured}
\end{table}

\autoref{tab:2_4} presents the results of pruning transformer models using 2:4 semi-structured sparsity. In these experiments, we applied pruning exclusively to the weight matrices in the multilayer perceptron (MLP) components, leaving the self-attention layers dense. This approach yielded sparse models with an overall sparsity of 38\% to 41\%. We adopted this selective pruning strategy to maintain model accuracy above a practical threshold, as 2:4 sparsity significantly impacts performance, potentially rendering fully sparse models ineffective. Our results demonstrate that our proposed \optima update mechanism consistently outperforms other methods under 2:4 sparsity, achieving superior accuracy.

\begin{table}[tb]
\centering
\setlength{\tabcolsep}{2pt}
\resizebox{\textwidth}{!}{
\begin{tabular}{l l l c c c c c c c | c}
\toprule
\multirow{2}{*}{\parbox[t]{1.5cm}{\centering Model}} &
\multirow{2}{*}{\parbox[t]{1.3cm}{\centering Mask\\Selection}} &
\multirow{2}{*}{\parbox[t]{1cm}{\centering Weight\\Update}} &
\multirow{2}{*}{Perplexity} &
\multicolumn{7}{c}{Metrics (\%)} \\
\cmidrule{5-11}
 & & & &
   \parbox[t]{\columndistance}{\centering \textcolor{blue!70!black}{MMLU}} &
   \parbox[t]{\columndistance}{\centering \textcolor{blue!70!black}{PIQA}} &
   \parbox[t]{\columndistance}{\centering \textcolor{blue!70!black}{Arc-E}} &
   \parbox[t]{\columndistance}{\centering \textcolor{blue!70!black}{Arc-C}} &
   \parbox[t]{\columndistance}{\centering \textcolor{blue!70!black}{Wino}} &
   \parbox[t]{\columndistance}{\centering \textcolor{blue!70!black}{OpenQA}} &
   \parbox[t]{\columndistance}{\centering \textbf{\textcolor{teal}{Average}}} \\
\midrule
\multirow{7}{*}{\parbox[t]{1.5cm}{\centering LLaMA\\3.1 8B}} & Dense & -- & 5.84 & 63.57 & 80.09 & 81.44 & 51.37 & 73.48 & 33.40 & 63.89 \\
 & Wanda & -- & {13.54} & 43.42 & 73.18 & 69.23 & 35.32 & 67.32 & \textbf{25.80} & 52.38 \\
 & \cellcolor{shade}Wanda & \cellcolor{shade}\optima & \cellcolor{shade}12.58 & \cellcolor{shade}{\textbf{45.45}} & \cellcolor{shade}{\textbf{73.39}} & \cellcolor{shade}{\textbf{69.57}} & \cellcolor{shade}{\textbf{36.18}} & \cellcolor{shade}{\textbf{68.90}} & \cellcolor{shade}{25.20} & \cellcolor{shade}{\textbf{53.12}} \\
 & SparseGPT & SparseGPT & 12.37 & 45.62 & \textbf{73.83} & 69.15 & 35.84 & 69.22 & 25.60 & 53.21 \\
 & \cellcolor{shade}SparseGPT & \cellcolor{shade}\optima & \cellcolor{shade}{{12.54}} & \cellcolor{shade}\textbf{46.04} & \cellcolor{shade}{73.72} & \cellcolor{shade}{\textbf{69.95}} & \cellcolor{shade}\textbf{36.77} & \cellcolor{shade}\textbf{69.61} & \cellcolor{shade}\textbf{27.00} & \cellcolor{shade}\textbf{53.85} \\
 & Thanos & Thanos & 12.66 & 44.39 & 73.94 & 69.57 & 36.18 & \textbf{68.90} & 25.20 & 53.03 \\
 & \cellcolor{shade}Thanos & \cellcolor{shade}\optima & \cellcolor{shade}{{12.80}} & \cellcolor{shade}{\textbf{44.41}} & \cellcolor{shade}\textbf{74.05} & \cellcolor{shade}\textbf{69.95} & \cellcolor{shade}{\textbf{36.43}} & \cellcolor{shade}{68.59} & \cellcolor{shade}{\textbf{25.60}} & \cellcolor{shade}{\textbf{53.17}} \\
 \doublemidrule
\multirow{9}{*}{\parbox[t]{1.5cm}{\centering LLaMA\\3.2 1B}} & Dense & -- & 9.75 & 36.92 & 74.27 & 65.53 & 31.31 & 60.30 & 26.20 & 49.09 \\
 & Wanda & -- & 30.43 & 23.32 & 63.55 & 47.56 & \textbf{23.63} & 55.25 & 15.00 & 38.05 \\
 & \cellcolor{shade}Wanda & \cellcolor{shade}\optima & \cellcolor{shade}{{48.23}} & \cellcolor{shade}\textbf{24.80} & \cellcolor{shade}\textbf{66.10} & \cellcolor{shade}\textbf{58.04} & \cellcolor{shade}{23.55} & \cellcolor{shade}{55.25} & \cellcolor{shade}\textbf{19.80} & \cellcolor{shade}\textbf{41.26} \\
 & SparseGPT & SparseGPT & {21.98} & 23.05 & 65.45 & 52.15 & 25.17 & \textbf{57.62} & 17.60 & 40.17 \\
 & \cellcolor{shade}SparseGPT & \cellcolor{shade}\optima & \cellcolor{shade}{21.40} & \cellcolor{shade}{\textbf{23.40}} & \cellcolor{shade}{\textbf{65.72}} & \cellcolor{shade}{\textbf{52.78}} & \cellcolor{shade}\textbf{25.51} & \cellcolor{shade}{57.06} & \cellcolor{shade}{\textbf{18.60}} & \cellcolor{shade}{\textbf{40.51}} \\
 & Thanos & Thanos & {22.80} & \textbf{24.09} & \textbf{65.67} & 51.68 & \textbf{25.00} & 52.96 & \textbf{17.60} & 39.50 \\
 & \cellcolor{shade}Thanos & \cellcolor{shade}\optima & \cellcolor{shade}{22.26} & \cellcolor{shade}{23.41} & \cellcolor{shade}{65.34} & \cellcolor{shade}{\textbf{52.22}} & \cellcolor{shade}{23.72} & \cellcolor{shade}{\textbf{55.96}} & \cellcolor{shade}{16.80} & \cellcolor{shade}{\textbf{39.58}} \\
 & ProxSparse & -- & {41.95} & \textbf{23.64} & 61.21 & 42.38 & 22.53 & 53.67 & 16.00 & 36.57 \\
 & \cellcolor{shade}ProxSparse & \cellcolor{shade}\optima & \cellcolor{shade}{28.53} & \cellcolor{shade}{23.07} & \cellcolor{shade}{\textbf{63.38}} & \cellcolor{shade}{\textbf{47.90}} & \cellcolor{shade}{22.53} & \cellcolor{shade}{\textbf{54.78}} & \cellcolor{shade}{\textbf{16.40}} & \cellcolor{shade}{\textbf{38.01}} \\
\doublemidrule
\multirow{9}{*}{\parbox[t]{1.5cm}{\centering LLaMA\\3.2 3B}} & Dense & -- & 7.81 & 54.13 & 76.55 & 74.28 & 42.75 & 69.38 & 30.60 & 57.95 \\
 & Wanda & -- & {18.51} & 34.30 & 70.73 & 60.69 & 30.72 & 61.17 & \textbf{24.80} & 47.07 \\
 & \cellcolor{shade}Wanda & \cellcolor{shade}\optima & \cellcolor{shade}16.64 & \cellcolor{shade}{\textbf{37.15}} & \cellcolor{shade}{\textbf{70.78}} & \cellcolor{shade}{\textbf{61.95}} & \cellcolor{shade}{\textbf{31.14}} & \cellcolor{shade}{\textbf{62.51}} & \cellcolor{shade}24.60 & \cellcolor{shade}{\textbf{48.02}} \\
 & SparseGPT & SparseGPT & 16.19 & 36.13 & 70.29 & 63.01 & 30.46 & \textbf{64.72} & 25.00 & 48.27 \\
 & \cellcolor{shade}SparseGPT & \cellcolor{shade}\optima & \cellcolor{shade}{{16.36}} & \cellcolor{shade}\textbf{38.03} & \cellcolor{shade}\textbf{70.84} & \cellcolor{shade}\textbf{63.17} & \cellcolor{shade}\textbf{32.17} & \cellcolor{shade}{63.69} & \cellcolor{shade}\textbf{25.60} & \cellcolor{shade}\textbf{48.92} \\
 & Thanos & Thanos & 16.24 & 35.55 & 70.35 & 61.28 & 29.78 & \textbf{63.30} & 24.20 & 47.41 \\
 & \cellcolor{shade}Thanos & \cellcolor{shade}\optima & \cellcolor{shade}{{16.49}} & \cellcolor{shade}{\textbf{35.72}} & \cellcolor{shade}{\textbf{70.62}} & \cellcolor{shade}{\textbf{62.04}} & \cellcolor{shade}{\textbf{30.97}} & \cellcolor{shade}{63.22} & \cellcolor{shade}\textbf{25.60} & \cellcolor{shade}{\textbf{48.03}} \\
 & ProxSparse & -- & {19.50} & 24.66 & 68.12 & 56.31 & 27.82 & 58.56 & 20.00 & 42.58 \\
 & \cellcolor{shade}ProxSparse & \cellcolor{shade}\optima & \cellcolor{shade}{18.28} & \cellcolor{shade}{\textbf{31.76}} & \cellcolor{shade}{\textbf{69.53}} & \cellcolor{shade}{\textbf{60.27}} & \cellcolor{shade}{\textbf{28.84}} & \cellcolor{shade}{\textbf{60.30}} & \cellcolor{shade}{\textbf{20.60}} & \cellcolor{shade}{\textbf{45.22}} \\
\doublemidrule
\multirow{9}{*}{\parbox[t]{1.5cm}{\centering Gemma\\3 1B}} & Dense & -- & 14.17 & 24.95 & 74.81 & 71.93 & 35.41 & 58.72 & 28.80 & 49.10 \\
 & Wanda & -- & {60.74} & \textbf{23.74} & \textbf{65.51} & \textbf{56.78} & 22.35 & 52.72 & \textbf{19.80} & \textbf{40.15} \\
 & \cellcolor{shade}Wanda & \cellcolor{shade}\optima & \cellcolor{shade}23.25 & \cellcolor{shade}23.25 & \cellcolor{shade}63.38 & \cellcolor{shade}51.14 & \cellcolor{shade}\textbf{24.06} & \cellcolor{shade}{\textbf{54.30}} & \cellcolor{shade}18.20 & \cellcolor{shade}39.06 \\
 & SparseGPT & SparseGPT & {44.87} & 24.83 & \textbf{66.76} & 57.70 & 23.29 & \textbf{55.96} & 19.40 & 41.32 \\
 & \cellcolor{shade}SparseGPT & \cellcolor{shade}\optima & \cellcolor{shade}{42.66} & \cellcolor{shade}{\textbf{25.11}} & \cellcolor{shade}{66.27} & \cellcolor{shade}{\textbf{58.96}} & \cellcolor{shade}{\textbf{23.89}} & \cellcolor{shade}{55.80} & \cellcolor{shade}{\textbf{20.60}} & \cellcolor{shade}\textbf{41.77} \\
 & Thanos & Thanos & {48.50} & 25.23 & 65.89 & \textbf{59.30} & 23.12 & 53.59 & \textbf{20.80} & 41.32 \\
 & \cellcolor{shade}Thanos & \cellcolor{shade}\optima & \cellcolor{shade}{44.91} & \cellcolor{shade}\textbf{25.83} & \cellcolor{shade}{\textbf{66.00}} & \cellcolor{shade}{58.63} & \cellcolor{shade}{\textbf{23.29}} & \cellcolor{shade}{\textbf{54.70}} & \cellcolor{shade}{20.00} & \cellcolor{shade}{\textbf{41.41}} \\
 & ProxSparse & -- & 41.02 & 23.01 & \textbf{66.00} & \textbf{54.34} & 22.44 & \textbf{55.88} & \textbf{20.20} & \textbf{40.31} \\
 & \cellcolor{shade}ProxSparse & \cellcolor{shade}\optima & \cellcolor{shade}{{52.99}} & \cellcolor{shade}{\textbf{24.13}} & \cellcolor{shade}{64.74} & \cellcolor{shade}{53.70} & \cellcolor{shade}{\textbf{22.61}} & \cellcolor{shade}{52.25} & \cellcolor{shade}{17.00} & \cellcolor{shade}{39.07} \\
\doublemidrule
\multirow{9}{*}{\parbox[t]{1.5cm}{\centering Gemma\\2 2B}} & Dense & -- & 68.69 & 49.33 & 78.24 & 80.22 & 46.93 & 68.82 & 31.40 & 59.16 \\
 & Wanda & -- & \textbf{421.01} & 34.34 & 71.33 & 68.10 & 30.97 & 61.40 & \textbf{26.40} & 48.76 \\
 & \cellcolor{shade}Wanda & \cellcolor{shade}\optima & \cellcolor{shade}229.69 & \cellcolor{shade}{\textbf{34.44}} & \cellcolor{shade}{\textbf{71.87}} & \cellcolor{shade}\textbf{68.90} & \cellcolor{shade}{\textbf{33.87}} & \cellcolor{shade}{\textbf{62.27}} & \cellcolor{shade}{25.00} & \cellcolor{shade}{\textbf{49.39}} \\
 & SparseGPT & SparseGPT & \textbf{251.71} & \textbf{32.84} & 71.76 & \textbf{68.73} & \textbf{32.42} & 61.88 & 23.40 & 48.51 \\
 & \cellcolor{shade}SparseGPT & \cellcolor{shade}\optima & \cellcolor{shade}{227.99} & \cellcolor{shade}{32.77} & \cellcolor{shade}{71.76} & \cellcolor{shade}{67.47} & \cellcolor{shade}{32.17} & \cellcolor{shade}\textbf{63.38} & \cellcolor{shade}{\textbf{24.40}} & \cellcolor{shade}{\textbf{48.66}} \\
 & Thanos & Thanos & \textbf{256.58} & 31.02 & 70.73 & \textbf{67.72} & 32.08 & \textbf{62.51} & 24.80 & 48.14 \\
 & \cellcolor{shade}Thanos & \cellcolor{shade}\optima & \cellcolor{shade}{239.20} & \cellcolor{shade}{\textbf{32.58}} & \cellcolor{shade}{\textbf{71.16}} & \cellcolor{shade}{67.47} & \cellcolor{shade}{\textbf{32.25}} & \cellcolor{shade}{60.85} & \cellcolor{shade}{\textbf{25.20}} & \cellcolor{shade}{\textbf{48.25}} \\
 & ProxSparse & -- & 176.03 & 37.19 & \textbf{71.98} & 67.55 & \textbf{34.47} & 61.48 & \textbf{25.00} & 49.61 \\
 & \cellcolor{shade}ProxSparse & \cellcolor{shade}\optima & \cellcolor{shade}{\textbf{254.03}} & \cellcolor{shade}\textbf{38.27} & \cellcolor{shade}{71.27} & \cellcolor{shade}{\textbf{68.60}} & \cellcolor{shade}{33.53} & \cellcolor{shade}{\textbf{61.88}} & \cellcolor{shade}{24.60} & \cellcolor{shade}\textbf{49.69} \\
\bottomrule
\end{tabular}
}
\caption{Model perplexity on WikiText2 and accuracy on zero-shot downstream tasks for 2:4 sparsity. In this experiment, only the layers in the MLP part of the transformer are pruned, and the self-attention layers are dense, resulting in an end-to-end sparsity ratio of 38\% to 41\%. \optima consistently improves the accuracy of the models across different tasks. Please note that ProxSparse pruning is limited to 2:4 sparsity, and hence our unstructured sparsity experiments do not include it.}
\label{tab:2_4}
\vspace{-10pt}
\end{table}

\paragraph{Higher Sparsity Ratios.} To assess the robustness of \optima at more aggressive compression levels, we extend our evaluation to 60\% unstructured sparsity. \autoref{tab:60_unstructured} presents the perplexity and zero-shot accuracy metrics across the same models and tasks. \optima continues to deliver consistent improvements over the baseline pruning methods, with average accuracy gains of up to 2.53\% across the downstream tasks (LLaMA-3.2-1B).

\begin{table}[tb]
\centering
\setlength{\tabcolsep}{2pt}
\resizebox{\textwidth}{!}{
\begin{tabular}{l l l c c c c c c c | c}
\toprule
\multirow{2}{*}{\parbox[t]{1.5cm}{\centering Model}} & 
\multirow{2}{*}{\parbox[t]{1.3cm}{\centering Mask\\Selection}} & 
\multirow{2}{*}{\parbox[t]{1cm}{\centering Weight\\Update}} & 
\multirow{2}{*}{Perplexity} & 
\multicolumn{7}{c}{Metrics (\%)} \\
\cmidrule{5-11}
 & & & & 
   \parbox[t]{\columndistance}{\centering \textcolor{blue!70!black}{MMLU}} & 
   \parbox[t]{\columndistance}{\centering \textcolor{blue!70!black}{PIQA}} & 
   \parbox[t]{\columndistance}{\centering \textcolor{blue!70!black}{Arc-E}} & 
   \parbox[t]{\columndistance}{\centering \textcolor{blue!70!black}{Arc-C}} & 
   \parbox[t]{\columndistance}{\centering \textcolor{blue!70!black}{Wino}} & 
   \parbox[t]{\columndistance}{\centering \textcolor{blue!70!black}{OpenQA}} & 
   \parbox[t]{\columndistance}{\centering \textbf{\textcolor{teal}{Average}}} \\
\midrule
\multirow{7}{*}{\parbox[t]{1.5cm}{\centering LLaMA\\3.1 8B}} & Dense & -- & 5.84 & 63.57 & 80.09 & 81.44 & 51.37 & 73.48 & 33.40 & 63.89 \\
 & Wanda & -- & {21.65} & 31.98 & 69.53 & 61.11 & 27.30 & 61.09 & 21.40 & 45.40 \\
 & \cellcolor{shade}Wanda & \cellcolor{shade}\optima & \cellcolor{shade}{17.56} & \cellcolor{shade}{\textbf{33.96}} & \cellcolor{shade}{\textbf{71.60}} & \cellcolor{shade}{\textbf{63.76}} & \cellcolor{shade}{\textbf{29.35}} & \cellcolor{shade}{\textbf{66.06}} & \cellcolor{shade}{\textbf{22.60}} & \cellcolor{shade}{\textbf{47.89}} \\
 & SparseGPT & SparseGPT & 15.44 & \textbf{35.32} & 71.55 & 62.88 & 31.66 & \textbf{68.19} & 24.20 & \textbf{48.96} \\
 & \cellcolor{shade}SparseGPT & \cellcolor{shade}\optima & \cellcolor{shade}{{15.64}} & \cellcolor{shade}{32.44} & \cellcolor{shade}{\textbf{71.87}} & \cellcolor{shade}{\textbf{63.97}} & \cellcolor{shade}{\textbf{33.11}} & \cellcolor{shade}{67.56} & \cellcolor{shade}\textbf{24.60} & \cellcolor{shade}{48.93} \\
 & Thanos & Thanos & 15.91 & \textbf{35.22} & \textbf{72.09} & \textbf{65.28} & \textbf{33.19} & 67.40 & \textbf{23.40} & \textbf{49.43} \\
 & \cellcolor{shade}Thanos & \cellcolor{shade}\optima & \cellcolor{shade}{{16.09}} & \cellcolor{shade}{34.48} & \cellcolor{shade}{72.03} & \cellcolor{shade}{64.69} & \cellcolor{shade}{33.02} & \cellcolor{shade}\textbf{68.51} & \cellcolor{shade}{22.80} & \cellcolor{shade}{49.25} \\
\doublemidrule
\multirow{7}{*}{\parbox[t]{1.5cm}{\centering LLaMA\\3.2 1B}} & Dense & -- & 9.75 & 36.92 & 74.27 & 65.53 & 31.31 & 60.30 & 26.20 & 49.09 \\
 & Wanda & -- & {71.53} & 22.95 & 59.68 & 39.48 & 18.77 & 50.43 & 12.20 & 33.92 \\
 & \cellcolor{shade}Wanda & \cellcolor{shade}\optima & \cellcolor{shade}{41.50} & \cellcolor{shade}\textbf{23.52} & \cellcolor{shade}{\textbf{62.62}} & \cellcolor{shade}\textbf{44.53} & \cellcolor{shade}{\textbf{20.65}} & \cellcolor{shade}{\textbf{52.57}} & \cellcolor{shade}{\textbf{14.80}} & \cellcolor{shade}{\textbf{36.45}} \\
 & SparseGPT & SparseGPT & {48.00} & \textbf{23.02} & 62.08 & 43.48 & \textbf{21.76} & 52.09 & 17.40 & 36.64 \\
 & \cellcolor{shade}SparseGPT & \cellcolor{shade}\optima & \cellcolor{shade}{38.05} & \cellcolor{shade}{22.95} & \cellcolor{shade}\textbf{63.38} & \cellcolor{shade}{\textbf{43.52}} & \cellcolor{shade}{20.48} & \cellcolor{shade}{\textbf{53.28}} & \cellcolor{shade}\textbf{19.60} & \cellcolor{shade}\textbf{37.20} \\
 & Thanos & Thanos & {46.78} & \textbf{23.25} & 62.57 & 44.49 & 21.59 & 53.20 & 16.60 & 36.95 \\
 & \cellcolor{shade}Thanos & \cellcolor{shade}\optima & \cellcolor{shade}{40.54} & \cellcolor{shade}{23.02} & \cellcolor{shade}{\textbf{62.95}} & \cellcolor{shade}\textbf{44.53} & \cellcolor{shade}{\textbf{21.67}} & \cellcolor{shade}\textbf{53.91} & \cellcolor{shade}{\textbf{17.40}} & \cellcolor{shade}\textbf{37.25} \\
\doublemidrule
\multirow{7}{*}{\parbox[t]{1.5cm}{\centering LLaMA\\3.2 3B}} & Dense & -- & 7.81 & 54.13 & 76.55 & 74.28 & 42.75 & 69.38 & 30.60 & 57.95 \\
 & Wanda & -- & {31.13} & 25.53 & 65.23 & 47.90 & 22.70 & 55.25 & 16.00 & 38.77 \\
 & \cellcolor{shade}Wanda & \cellcolor{shade}\optima & \cellcolor{shade}{23.56} & \cellcolor{shade}{\textbf{31.20}} & \cellcolor{shade}{\textbf{67.41}} & \cellcolor{shade}{\textbf{53.96}} & \cellcolor{shade}{\textbf{24.57}} & \cellcolor{shade}{\textbf{59.51}} & \cellcolor{shade}{\textbf{19.80}} & \cellcolor{shade}{\textbf{42.74}} \\
 & SparseGPT & SparseGPT & 22.00 & \textbf{31.27} & \textbf{69.37} & 53.66 & \textbf{26.02} & 61.33 & \textbf{21.00} & \textbf{43.78} \\
 & \cellcolor{shade}SparseGPT & \cellcolor{shade}\optima & \cellcolor{shade}{{22.67}} & \cellcolor{shade}{29.58} & \cellcolor{shade}{68.77} & \cellcolor{shade}{\textbf{54.80}} & \cellcolor{shade}{24.74} & \cellcolor{shade}\textbf{62.35} & \cellcolor{shade}{20.60} & \cellcolor{shade}{43.47} \\
 & Thanos & Thanos & {22.48} & 29.23 & 67.63 & 55.01 & \textbf{26.02} & 57.85 & 19.20 & 42.49 \\
 & \cellcolor{shade}Thanos & \cellcolor{shade}\optima & \cellcolor{shade}{22.28} & \cellcolor{shade}\textbf{31.43} & \cellcolor{shade}{\textbf{67.90}} & \cellcolor{shade}\textbf{55.26} & \cellcolor{shade}{24.91} & \cellcolor{shade}{\textbf{59.67}} & \cellcolor{shade}{\textbf{20.60}} & \cellcolor{shade}{\textbf{43.30}} \\
\doublemidrule
\multirow{7}{*}{\parbox[t]{1.5cm}{\centering Gemma\\3 1B}} & Dense & -- & 14.17 & 24.95 & 74.81 & 71.93 & 35.41 & 58.72 & 28.80 & 49.10 \\
 & Wanda & -- & {90.48} & 23.04 & 62.19 & 49.75 & 18.60 & 50.99 & 15.20 & 36.63 \\
 & \cellcolor{shade}Wanda & \cellcolor{shade}\optima & \cellcolor{shade}64.79 & \cellcolor{shade}{\textbf{23.34}} & \cellcolor{shade}{\textbf{64.09}} & \cellcolor{shade}{\textbf{52.86}} & \cellcolor{shade}{\textbf{20.48}} & \cellcolor{shade}{\textbf{51.93}} & \cellcolor{shade}{\textbf{16.40}} & \cellcolor{shade}{\textbf{38.18}} \\
 & SparseGPT & SparseGPT & {60.91} & \textbf{24.58} & 65.34 & 51.98 & 21.93 & 51.14 & 16.60 & 38.60 \\
 & \cellcolor{shade}SparseGPT & \cellcolor{shade}\optima & \cellcolor{shade}{56.27} & \cellcolor{shade}{23.72} & \cellcolor{shade}\textbf{66.21} & \cellcolor{shade}{\textbf{52.44}} & \cellcolor{shade}\textbf{22.53} & \cellcolor{shade}{\textbf{52.96}} & \cellcolor{shade}{\textbf{17.60}} & \cellcolor{shade}{\textbf{39.24}} \\
 & Thanos & Thanos & {62.22} & \textbf{24.62} & 64.53 & 52.86 & 20.65 & 52.17 & 18.80 & 38.94 \\
 & \cellcolor{shade}Thanos & \cellcolor{shade}\optima & \cellcolor{shade}{56.78} & \cellcolor{shade}{24.44} & \cellcolor{shade}{\textbf{64.85}} & \cellcolor{shade}\textbf{55.18} & \cellcolor{shade}{\textbf{22.01}} & \cellcolor{shade}\textbf{54.85} & \cellcolor{shade}\textbf{19.80} & \cellcolor{shade}\textbf{40.19} \\
\doublemidrule
\multirow{7}{*}{\parbox[t]{1.5cm}{\centering Gemma\\2 2B}} & Dense & -- & 68.69 & 49.33 & 78.24 & 80.22 & 46.93 & 68.82 & 31.40 & 59.16 \\
 & Wanda & -- & {757.47} & 23.36 & 65.78 & 56.10 & 21.59 & 52.64 & 19.80 & 39.88 \\
 & \cellcolor{shade}Wanda & \cellcolor{shade}\optima & \cellcolor{shade}435.10 & \cellcolor{shade}{\textbf{24.37}} & \cellcolor{shade}{\textbf{66.59}} & \cellcolor{shade}\textbf{58.50} & \cellcolor{shade}{\textbf{21.93}} & \cellcolor{shade}{\textbf{57.38}} & \cellcolor{shade}{\textbf{20.00}} & \cellcolor{shade}{\textbf{41.46}} \\
 & SparseGPT & SparseGPT & {488.25} & 24.49 & 68.50 & 57.45 & 25.00 & \textbf{58.96} & \textbf{25.00} & 43.23 \\
 & \cellcolor{shade}SparseGPT & \cellcolor{shade}\optima & \cellcolor{shade}{451.46} & \cellcolor{shade}\textbf{25.89} & \cellcolor{shade}\textbf{68.88} & \cellcolor{shade}\textbf{58.50} & \cellcolor{shade}\textbf{26.28} & \cellcolor{shade}{58.01} & \cellcolor{shade}{24.20} & \cellcolor{shade}\textbf{43.63} \\
 & Thanos & Thanos & {523.61} & \textbf{23.69} & \textbf{68.23} & \textbf{58.12} & \textbf{23.89} & 58.33 & \textbf{21.20} & \textbf{42.24} \\
 & \cellcolor{shade}Thanos & \cellcolor{shade}\optima & \cellcolor{shade}{497.75} & \cellcolor{shade}{23.12} & \cellcolor{shade}{67.74} & \cellcolor{shade}{57.07} & \cellcolor{shade}{23.38} & \cellcolor{shade}\textbf{59.27} & \cellcolor{shade}{20.60} & \cellcolor{shade}{41.86} \\
\bottomrule
\end{tabular}
}
\caption{Model perplexity on WikiText2 and accuracy on zero-shot downstream tasks for 60\% unstructured sparsity. \optima consistently improves the accuracy of the models across different tasks.}
\label{tab:60_unstructured}
\end{table}

These enhancements are particularly notable at higher sparsity ratios, where pruning a larger portion of weights introduces greater reconstruction error. By optimally readjusting the remaining weights through our QP formulation, \optima effectively mitigates this error, leading to lower perplexity and higher downstream performance compared to Wanda, SparseGPT, or Thanos individually. For example, on LLaMA-3.2-3B, \optima increases Wanda's average accuracy from 38.77\% to 42.74\%, highlighting its ability to preserve model utility under extreme sparsity conditions.

\label{app:qwen}
\paragraph{Extended Evaluation on the Qwen-2.5 Model Family.}
To further validate the robustness and generalizability of \optima, we conduct additional experiments on the Qwen-2.5 family of models, with sizes ranging from 0.5B to 14B parameters. These models were not included in the preceding analysis, and this evaluation serves to confirm that \optima's benefits apply across different model architectures.

We evaluate performance across three distinct settings, mirroring the main experiments: 50\% unstructured sparsity (\autoref{tab:qwen_50_unstructured}), 60\% unstructured sparsity (\autoref{tab:qwen_60_unstructured}), and 2:4 semi-structured sparsity (\autoref{tab:qwen_2_4_sparsity}). 

\paragraph{Unstructured Sparsity (50\% and 60\%).}
At 50\% unstructured sparsity (\autoref{tab:qwen_50_unstructured}), \optima consistently improves zero-shot performance across all Qwen-2.5 model sizes and for all mask selection methods (Wanda, SparseGPT, and Thanos). For example, on the Qwen-2.5 3B model, \optima boosts the average accuracy of Wanda from 54.02\% to 55.33\% and SparseGPT from 54.70\% to 55.69\%. These gains demonstrate that our \optima reconstruction successfully recovers accuracy lost during the pruning step.

The advantages of \optima are even more pronounced at the more aggressive 60\% sparsity ratio, as shown in \autoref{tab:qwen_60_unstructured}. At this level, pruning introduces a more significant reconstruction error, providing a greater opportunity for \optima to recover performance. This is especially clear on the Qwen-2.5 3B model, where \optima improves Wanda's average accuracy from 43.67\% to 47.86\% (a 4.19\% absolute gain) and Thanos's from 48.45\% to 49.98\% (a 1.53\% gain).

\paragraph{Semi-Structured Sparsity (2:4).}
In the 2:4 semi-structured sparsity setting (\autoref{tab:qwen_2_4_sparsity}), where pruning is applied only to the MLP layers, \optima provides clear improvements for most models, particularly in the 1.5B and 3B range. For instance, it improves the average accuracy of the 3B model pruned with Wanda from 49.48\% to 50.63\% and the 1.5B model from 46.01\% to 47.26\%.

On the larger 7B and 14B models, the results are more varied, with performance differing based on the underlying mask selector. This suggests a complex interaction between mask selection heuristics and \optima reconstruction for structured sparsity at this scale, which could be a valuable avenue for future investigation.

Overall, these experiments on the Qwen-2.5 family reinforce the findings from the preceding sections. They confirm that \optima is a broadly applicable and effective method for enhancing model accuracy post-pruning, delivering its most significant and consistent gains in high-sparsity unstructured regimes.

\begin{table}[tb]
\centering
\setlength{\tabcolsep}{2pt}
\resizebox{\textwidth}{!}{
\begin{tabular}{l l l c c c c c c c | c}
\toprule
\multirow{2}{*}{\parbox[t]{1.8cm}{\centering Model}} &
\multirow{2}{*}{\parbox[t]{1.3cm}{\centering Mask\\Selection}} &
\multirow{2}{*}{\parbox[t]{1cm}{\centering Weight\\Update}} &
\multirow{2}{*}{Perplexity} &
\multicolumn{7}{c}{Metrics (\%)} \\
\cmidrule{5-11}
 & & & &
  \parbox[t]{\columndistance}{\centering \textcolor{blue!70!black}{MMLU}} &
  \parbox[t]{\columndistance}{\centering \textcolor{blue!70!black}{PIQA}} &
  \parbox[t]{\columndistance}{\centering \textcolor{blue!70!black}{Arc-E}} &
  \parbox[t]{\columndistance}{\centering \textcolor{blue!70!black}{Arc-C}} &
  \parbox[t]{\columndistance}{\centering \textcolor{blue!70!black}{Wino}} &
  \parbox[t]{\columndistance}{\centering \textcolor{blue!70!black}{OpenQA}} &
  \parbox[t]{\columndistance}{\centering \textbf{\textcolor{teal}{Average}}} \\
\midrule

\multirow{7}{*}{\parbox[t]{1.8cm}{\centering Qwen 2.5\\0.5B}} & Dense & -- & 13.08 & 47.36 & 69.97 & 64.18 & 29.18 & 55.80 & 24.40 & 48.48 \\
 & Wanda & -- & 24.00 & \textbf{30.52} & 64.09 & 57.41 & 24.06 & 54.38 & 19.80 & 41.71 \\
 & \cellcolor{shade}{Wanda} & \cellcolor{shade}{\optima} & \cellcolor{shade}{\textbf{22.70}} & \cellcolor{shade}{26.14} & \cellcolor{shade}{\textbf{64.58}} & \cellcolor{shade}{\textbf{57.79}} & \cellcolor{shade}{\textbf{25.26}} & \cellcolor{shade}{\textbf{56.04}} & \cellcolor{shade}{\textbf{22.00}} & \cellcolor{shade}{\textbf{41.97}} \\
 & SparseGPT & SparseGPT & 20.33 & \textbf{29.38} & 64.74 & 56.52 & 24.15 & \textbf{56.20} & \textbf{20.60} & \textbf{41.93} \\
 & \cellcolor{shade}{SparseGPT} & \cellcolor{shade}{\optima} & \cellcolor{shade}{\textbf{19.54}} & \cellcolor{shade}{27.68} & \cellcolor{shade}{\textbf{65.13}} & \cellcolor{shade}{\textbf{56.99}} & \cellcolor{shade}{\textbf{24.66}} & \cellcolor{shade}{55.33} & \cellcolor{shade}{\textbf{20.60}} & \cellcolor{shade}{41.73} \\
 & Thanos & Thanos & 20.85 & 28.94 & \textbf{65.40} & 55.93 & \textbf{24.40} & \textbf{56.35} & 21.60 & 42.10 \\
 & \cellcolor{shade}{Thanos} & \cellcolor{shade}{\optima} & \cellcolor{shade}{\textbf{20.41}} & \cellcolor{shade}{\textbf{30.00}} & \cellcolor{shade}{64.69} & \cellcolor{shade}{\textbf{56.10}} & \cellcolor{shade}{\textbf{24.40}} & \cellcolor{shade}{55.41} & \cellcolor{shade}{\textbf{22.20}} & \cellcolor{shade}{\textbf{42.13}} \\
\doublemidrule

\multirow{7}{*}{\parbox[t]{1.8cm}{\centering Qwen 2.5\\1.5B}} & Dense & -- & 9.28 & 59.70 & 75.73 & 75.34 & 40.96 & 63.14 & 32.20 & 57.84 \\
 & Wanda & -- & 14.45 & 44.76 & 71.22 & \textbf{66.62} & 31.74 & 59.91 & \textbf{24.80} & 49.84 \\
 & \cellcolor{shade}{Wanda} & \cellcolor{shade}{\optima} & \cellcolor{shade}{\textbf{12.85}} & \cellcolor{shade}{\textbf{45.61}} & \cellcolor{shade}{\textbf{72.36}} & \cellcolor{shade}{\textbf{66.62}} & \cellcolor{shade}{\textbf{32.34}} & \cellcolor{shade}{\textbf{61.80}} & \cellcolor{shade}{24.60} & \cellcolor{shade}{\textbf{50.55}} \\
 & SparseGPT & SparseGPT & 13.09 & 46.80 & 71.65 & \textbf{66.75} & \textbf{33.62} & \textbf{62.27} & 25.60 & \textbf{51.12} \\
 & \cellcolor{shade}{SparseGPT} & \cellcolor{shade}{\optima} & \cellcolor{shade}{\textbf{12.76}} & \cellcolor{shade}{\textbf{46.96}} & \cellcolor{shade}{\textbf{71.82}} & \cellcolor{shade}{65.45} & \cellcolor{shade}{33.02} & \cellcolor{shade}{61.80} & \cellcolor{shade}{\textbf{26.20}} & \cellcolor{shade}{50.87} \\
 & Thanos & Thanos & 13.17 & \textbf{48.40} & 71.76 & 66.84 & \textbf{33.70} & \textbf{62.83} & \textbf{27.20} & \textbf{51.79} \\
 & \cellcolor{shade}{Thanos} & \cellcolor{shade}{\optima} & \cellcolor{shade}{\textbf{12.89}} & \cellcolor{shade}{48.21} & \cellcolor{shade}{\textbf{72.03}} & \cellcolor{shade}{\textbf{67.26}} & \cellcolor{shade}{33.53} & \cellcolor{shade}{62.04} & \cellcolor{shade}{26.20} & \cellcolor{shade}{51.55} \\
\doublemidrule

\multirow{7}{*}{\parbox[t]{1.8cm}{\centering Qwen 2.5\\3B}} & Dense & -- & 8.03 & 65.00 & 78.35 & 77.31 & 44.88 & 68.43 & 29.20 & 60.53 \\
 & Wanda & -- & 11.39 & 49.09 & 73.23 & 71.46 & \textbf{38.48} & 65.43 & 26.40 & 54.02 \\
 & \cellcolor{shade}{Wanda} & \cellcolor{shade}{\optima} & \cellcolor{shade}{\textbf{10.59}} & \cellcolor{shade}{\textbf{52.00}} & \cellcolor{shade}{\textbf{74.37}} & \cellcolor{shade}{\textbf{72.18}} & \cellcolor{shade}{38.05} & \cellcolor{shade}{\textbf{66.77}} & \cellcolor{shade}{\textbf{28.60}} & \cellcolor{shade}{\textbf{55.33}} \\
 & SparseGPT & SparseGPT & 10.74 & 52.49 & 74.65 & \textbf{71.34} & 36.86 & 64.64 & 28.20 & 54.70 \\
 & \cellcolor{shade}{SparseGPT} & \cellcolor{shade}{\optima} & \cellcolor{shade}{\textbf{10.57}} & \cellcolor{shade}{\textbf{53.92}} & \cellcolor{shade}{\textbf{75.35}} & \cellcolor{shade}{70.83} & \cellcolor{shade}{\textbf{38.31}} & \cellcolor{shade}{\textbf{66.14}} & \cellcolor{shade}{\textbf{29.60}} & \cellcolor{shade}{\textbf{55.69}} \\
 & Thanos & Thanos & 10.64 & \textbf{52.61} & \textbf{75.52} & \textbf{70.54} & 36.69 & 66.61 & \textbf{28.40} & \textbf{55.06} \\
 & \cellcolor{shade}{Thanos} & \cellcolor{shade}{\optima} & \cellcolor{shade}{\textbf{10.52}} & \cellcolor{shade}{52.11} & \cellcolor{shade}{75.46} & \cellcolor{shade}{70.12} & \cellcolor{shade}{\textbf{37.29}} & \cellcolor{shade}{\textbf{66.69}} & \cellcolor{shade}{28.20} & \cellcolor{shade}{54.98} \\
\doublemidrule

\multirow{7}{*}{\parbox[t]{1.8cm}{\centering Qwen 2.5\\7B}} & Dense & -- & 6.85 & 71.76 & 78.73 & 80.51 & 48.38 & 72.61 & 33.40 & 64.23 \\
 & Wanda & -- & 8.62 & 65.89 & 77.31 & 75.08 & 40.53 & 70.17 & \textbf{30.80} & 59.96 \\
 & \cellcolor{shade}{Wanda} & \cellcolor{shade}{\optima} & \cellcolor{shade}{\textbf{8.33}} & \cellcolor{shade}{\textbf{66.17}} & \cellcolor{shade}{\textbf{77.69}} & \cellcolor{shade}{\textbf{76.43}} & \cellcolor{shade}{\textbf{42.66}} & \cellcolor{shade}{\textbf{71.27}} & \cellcolor{shade}{30.60} & \cellcolor{shade}{\textbf{60.80}} \\
 & SparseGPT & SparseGPT & 8.42 & \textbf{66.09} & \textbf{78.07} & 75.34 & 42.75 & 71.11 & 31.00 & 60.73 \\
 & \cellcolor{shade}{SparseGPT} & \cellcolor{shade}{\optima} & \cellcolor{shade}{\textbf{8.36}} & \cellcolor{shade}{65.78} & \cellcolor{shade}{77.64} & \cellcolor{shade}{\textbf{75.63}} & \cellcolor{shade}{\textbf{42.92}} & \cellcolor{shade}{\textbf{71.51}} & \cellcolor{shade}{\textbf{31.60}} & \cellcolor{shade}{\textbf{60.85}} \\
 & Thanos & Thanos & 8.49 & 66.21 & \textbf{77.86} & 74.71 & 42.32 & 70.17 & 30.40 & 60.28 \\
 & \cellcolor{shade}{Thanos} & \cellcolor{shade}{\optima} & \cellcolor{shade}{\textbf{8.46}} & \cellcolor{shade}{\textbf{66.23}} & \cellcolor{shade}{77.58} & \cellcolor{shade}{\textbf{76.22}} & \cellcolor{shade}{\textbf{44.45}} & \cellcolor{shade}{\textbf{71.19}} & \cellcolor{shade}{\textbf{31.20}} & \cellcolor{shade}{\textbf{61.15}} \\
\doublemidrule

\multirow{7}{*}{\parbox[t]{1.8cm}{\centering Qwen 2.5\\14B}} & Dense & -- & 5.30 & 77.62 & 81.28 & 82.24 & 55.80 & 75.14 & 34.40 & 67.75 \\
 & Wanda & -- & 7.30 & \textbf{69.84} & 79.16 & 81.02 & 51.28 & 73.72 & \textbf{34.60} & 64.94 \\
 & \cellcolor{shade}{Wanda} & \cellcolor{shade}{\optima} & \cellcolor{shade}{\textbf{7.18}} & \cellcolor{shade}{69.29} & \cellcolor{shade}{\textbf{79.43}} & \cellcolor{shade}{\textbf{81.19}} & \cellcolor{shade}{\textbf{52.30}} & \cellcolor{shade}{\textbf{73.80}} & \cellcolor{shade}{33.80} & \cellcolor{shade}{\textbf{64.97}} \\
 & SparseGPT & SparseGPT & 7.24 & \textbf{69.83} & \textbf{79.60} & 80.98 & 51.02 & 72.93 & 32.80 & 64.53 \\
 & \cellcolor{shade}{SparseGPT} & \cellcolor{shade}{\optima} & \cellcolor{shade}{\textbf{7.14}} & \cellcolor{shade}{69.71} & \cellcolor{shade}{79.54} & \cellcolor{shade}{\textbf{81.19}} & \cellcolor{shade}{\textbf{51.79}} & \cellcolor{shade}{\textbf{73.80}} & \cellcolor{shade}{\textbf{33.60}} & \cellcolor{shade}{\textbf{64.94}} \\
 & Thanos & Thanos & 7.25 & \textbf{70.57} & \textbf{79.87} & 80.18 & 49.15 & 73.09 & 32.20 & 64.17 \\
 & \cellcolor{shade}{Thanos} & \cellcolor{shade}{\optima} & \cellcolor{shade}{\textbf{7.19}} & \cellcolor{shade}{70.16} & \cellcolor{shade}{79.60} & \cellcolor{shade}{\textbf{81.57}} & \cellcolor{shade}{\textbf{51.37}} & \cellcolor{shade}{\textbf{73.48}} & \cellcolor{shade}{\textbf{33.00}} & \cellcolor{shade}{\textbf{64.86}} \\
\bottomrule
\end{tabular}
}
\caption{Qwen-2.5 family perplexity on WikiText2 and accuracy on zero-shot downstream tasks for 50\% unstructured sparsity. \optima consistently improves the accuracy of the models across different tasks.}
\label{tab:qwen_50_unstructured}
\end{table}

\begin{table}[tb]
\centering
\setlength{\tabcolsep}{2pt}
\resizebox{\textwidth}{!}{
\begin{tabular}{l l l c c c c c c c | c}
\toprule
\multirow{2}{*}{\parbox[t]{1.8cm}{\centering Model}} &
\multirow{2}{*}{\parbox[t]{1.3cm}{\centering Mask\\Selection}} &
\multirow{2}{*}{\parbox[t]{1cm}{\centering Weight\\Update}} &
\multirow{2}{*}{Perplexity} &
\multicolumn{7}{c}{Metrics (\%)} \\
\cmidrule{5-11}
 & & & &
  \parbox[t]{\columndistance}{\centering \textcolor{blue!70!black}{MMLU}} &
  \parbox[t]{\columndistance}{\centering \textcolor{blue!70!black}{PIQA}} &
  \parbox[t]{\columndistance}{\centering \textcolor{blue!70!black}{Arc-E}} &
  \parbox[t]{\columndistance}{\centering \textcolor{blue!70!black}{Arc-C}} &
  \parbox[t]{\columndistance}{\centering \textcolor{blue!70!black}{Wino}} &
  \parbox[t]{\columndistance}{\centering \textcolor{blue!70!black}{OpenQA}} &
  \parbox[t]{\columndistance}{\centering \textbf{\textcolor{teal}{Average}}} \\
\midrule

\multirow{7}{*}{\parbox[t]{1.8cm}{\centering Qwen 2.5\\0.5B}} & Dense & -- & 13.08 & 47.36 & 69.97 & 64.18 & 29.18 & 55.80 & 24.40 & 48.48 \\
 & Wanda & -- & 83.42 & 23.02 & 59.96 & 43.81 & 18.09 & 50.28 & 12.80 & 34.66 \\
 & \cellcolor{shade}{Wanda} & \cellcolor{shade}{\optima} & \cellcolor{shade}{\textbf{51.97}} & \cellcolor{shade}{\textbf{23.16}} & \cellcolor{shade}{\textbf{60.72}} & \cellcolor{shade}{\textbf{46.25}} & \cellcolor{shade}{\textbf{20.14}} & \cellcolor{shade}{\textbf{51.78}} & \cellcolor{shade}{\textbf{16.40}} & \cellcolor{shade}{\textbf{36.41}} \\
 & SparseGPT & SparseGPT & 40.56 & 22.90 & 61.59 & 48.40 & 21.25 & 52.80 & 16.80 & 37.29 \\
 & \cellcolor{shade}{SparseGPT} & \cellcolor{shade}{\optima} & \cellcolor{shade}{\textbf{36.77}} & \cellcolor{shade}{\textbf{23.06}} & \cellcolor{shade}{\textbf{62.13}} & \cellcolor{shade}{\textbf{48.74}} & \cellcolor{shade}{\textbf{21.33}} & \cellcolor{shade}{\textbf{53.99}} & \cellcolor{shade}{\textbf{17.40}} & \cellcolor{shade}{\textbf{37.77}} \\
 & Thanos & Thanos & 44.29 & \textbf{23.78} & \textbf{62.02} & \textbf{48.65} & 21.33 & 52.25 & 17.80 & 37.64 \\
 & \cellcolor{shade}{Thanos} & \cellcolor{shade}{\optima} & \cellcolor{shade}{\textbf{41.92}} & \cellcolor{shade}{23.59} & \cellcolor{shade}{61.86} & \cellcolor{shade}{46.80} & \cellcolor{shade}{\textbf{22.35}} & \cellcolor{shade}{\textbf{53.75}} & \cellcolor{shade}{\textbf{19.60}} & \cellcolor{shade}{\textbf{37.99}} \\
\doublemidrule

\multirow{7}{*}{\parbox[t]{1.8cm}{\centering Qwen 2.5\\1.5B}} & Dense & -- & 9.28 & 59.70 & 75.73 & 75.34 & 40.96 & 63.14 & 32.20 & 57.84 \\
 & Wanda & -- & 58.38 & 27.25 & 65.18 & 54.50 & 24.74 & 53.04 & 17.20 & 40.32 \\
 & \cellcolor{shade}{Wanda} & \cellcolor{shade}{\optima} & \cellcolor{shade}{\textbf{23.81}} & \cellcolor{shade}{\textbf{30.99}} & \cellcolor{shade}{\textbf{66.87}} & \cellcolor{shade}{\textbf{56.44}} & \cellcolor{shade}{\textbf{24.91}} & \cellcolor{shade}{\textbf{56.83}} & \cellcolor{shade}{\textbf{18.40}} & \cellcolor{shade}{\textbf{42.41}} \\
 & SparseGPT & SparseGPT & 21.92 & \textbf{33.56} & 67.36 & \textbf{58.08} & \textbf{27.47} & 57.14 & 21.60 & \textbf{44.20} \\
 & \cellcolor{shade}{SparseGPT} & \cellcolor{shade}{\optima} & \cellcolor{shade}{\textbf{19.35}} & \cellcolor{shade}{31.44} & \cellcolor{shade}{\textbf{67.79}} & \cellcolor{shade}{56.27} & \cellcolor{shade}{27.22} & \cellcolor{shade}{\textbf{59.27}} & \cellcolor{shade}{\textbf{22.40}} & \cellcolor{shade}{44.07} \\
 & Thanos & Thanos & 27.07 & 33.66 & \textbf{67.63} & 57.49 & \textbf{27.73} & 56.67 & 20.60 & 43.96 \\
 & \cellcolor{shade}{Thanos} & \cellcolor{shade}{\optima} & \cellcolor{shade}{\textbf{23.64}} & \cellcolor{shade}{\textbf{35.97}} & \cellcolor{shade}{67.14} & \cellcolor{shade}{\textbf{57.66}} & \cellcolor{shade}{26.19} & \cellcolor{shade}{\textbf{58.09}} & \cellcolor{shade}{\textbf{20.80}} & \cellcolor{shade}{\textbf{44.31}} \\
\doublemidrule

\multirow{7}{*}{\parbox[t]{1.8cm}{\centering Qwen 2.5\\3B}} & Dense & -- & 8.03 & 65.00 & 78.35 & 77.31 & 44.88 & 68.43 & 29.20 & 60.53 \\
 & Wanda & -- & 22.06 & 28.07 & 67.14 & 60.86 & 27.39 & 58.17 & 20.40 & 43.67 \\
 & \cellcolor{shade}{Wanda} & \cellcolor{shade}{\optima} & \cellcolor{shade}{\textbf{15.67}} & \cellcolor{shade}{\textbf{37.22}} & \cellcolor{shade}{\textbf{70.24}} & \cellcolor{shade}{\textbf{63.55}} & \cellcolor{shade}{\textbf{30.89}} & \cellcolor{shade}{\textbf{61.64}} & \cellcolor{shade}{\textbf{23.60}} & \cellcolor{shade}{\textbf{47.86}} \\
 & SparseGPT & SparseGPT & 14.82 & \textbf{43.16} & 71.60 & 64.35 & 32.59 & 63.30 & 23.20 & 49.70 \\
 & \cellcolor{shade}{SparseGPT} & \cellcolor{shade}{\optima} & \cellcolor{shade}{\textbf{14.50}} & \cellcolor{shade}{40.25} & \cellcolor{shade}{\textbf{72.20}} & \cellcolor{shade}{\textbf{64.90}} & \cellcolor{shade}{\textbf{33.62}} & \cellcolor{shade}{\textbf{63.69}} & \cellcolor{shade}{\textbf{24.00}} & \cellcolor{shade}{\textbf{49.78}} \\
 & Thanos & Thanos & 14.90 & 40.76 & \textbf{71.38} & 63.26 & 30.63 & 61.25 & 23.40 & 48.45 \\
 & \cellcolor{shade}{Thanos} & \cellcolor{shade}{\optima} & \cellcolor{shade}{\textbf{14.42}} & \cellcolor{shade}{\textbf{42.58}} & \cellcolor{shade}{71.27} & \cellcolor{shade}{\textbf{64.73}} & \cellcolor{shade}{\textbf{32.68}} & \cellcolor{shade}{\textbf{63.61}} & \cellcolor{shade}{\textbf{25.00}} & \cellcolor{shade}{\textbf{49.98}} \\
\doublemidrule

\multirow{7}{*}{\parbox[t]{1.8cm}{\centering Qwen 2.5\\7B}} & Dense & -- & 6.85 & 71.76 & 78.73 & 80.51 & 48.38 & 72.61 & 33.40 & 64.23 \\
 & Wanda & -- & 14.09 & 54.58 & 72.03 & 71.68 & 37.03 & 66.46 & 25.40 & 54.53 \\
 & \cellcolor{shade}{Wanda} & \cellcolor{shade}{\optima} & \cellcolor{shade}{\textbf{11.15}} & \cellcolor{shade}{\textbf{55.49}} & \cellcolor{shade}{\textbf{73.99}} & \cellcolor{shade}{\textbf{73.86}} & \cellcolor{shade}{\textbf{37.88}} & \cellcolor{shade}{\textbf{67.96}} & \cellcolor{shade}{\textbf{26.20}} & \cellcolor{shade}{\textbf{55.90}} \\
 & SparseGPT & SparseGPT & 10.86 & \textbf{56.63} & 74.92 & 73.36 & 40.61 & \textbf{67.25} & 25.80 & 56.43 \\
 & \cellcolor{shade}{SparseGPT} & \cellcolor{shade}{\optima} & \cellcolor{shade}{\textbf{10.53}} & \cellcolor{shade}{55.55} & \cellcolor{shade}{\textbf{75.46}} & \cellcolor{shade}{\textbf{73.78}} & \cellcolor{shade}{\textbf{40.70}} & \cellcolor{shade}{66.93} & \cellcolor{shade}{\textbf{26.60}} & \cellcolor{shade}{\textbf{56.50}} \\
 & Thanos & Thanos & 11.07 & \textbf{59.54} & 74.70 & \textbf{73.44} & \textbf{40.44} & 69.22 & 26.40 & \textbf{57.29} \\
 & \cellcolor{shade}{Thanos} & \cellcolor{shade}{\optima} & \cellcolor{shade}{\textbf{10.74}} & \cellcolor{shade}{58.90} & \cellcolor{shade}{\textbf{75.35}} & \cellcolor{shade}{72.69} & \cellcolor{shade}{40.10} & \cellcolor{shade}{\textbf{69.46}} & \cellcolor{shade}{\textbf{27.00}} & \cellcolor{shade}{57.25} \\
\doublemidrule

\multirow{7}{*}{\parbox[t]{1.8cm}{\centering Qwen 2.5\\14B}} & Dense & -- & 5.30 & 77.62 & 81.28 & 82.24 & 55.80 & 75.14 & 34.40 & 67.75 \\
 & Wanda & -- & 11.16 & 61.38 & 75.41 & 74.12 & \textbf{42.15} & 71.51 & 29.20 & 58.96 \\
 & \cellcolor{shade}{Wanda} & \cellcolor{shade}{\optima} & \cellcolor{shade}{\textbf{9.69}} & \cellcolor{shade}{\textbf{61.74}} & \cellcolor{shade}{\textbf{75.57}} & \cellcolor{shade}{\textbf{75.34}} & \cellcolor{shade}{41.98} & \cellcolor{shade}{\textbf{73.09}} & \cellcolor{shade}{\textbf{29.40}} & \cellcolor{shade}{\textbf{59.52}} \\
 & SparseGPT & SparseGPT & 9.22 & \textbf{62.83} & 76.66 & 76.18 & 44.45 & \textbf{72.14} & \textbf{29.60} & \textbf{60.31} \\
 & \cellcolor{shade}{SparseGPT} & \cellcolor{shade}{\optima} & \cellcolor{shade}{\textbf{8.97}} & \cellcolor{shade}{62.22} & \cellcolor{shade}{\textbf{76.93}} & \cellcolor{shade}{\textbf{76.47}} & \cellcolor{shade}{\textbf{44.54}} & \cellcolor{shade}{71.67} & \cellcolor{shade}{29.00} & \cellcolor{shade}{60.14} \\
 & Thanos & Thanos & 9.14 & \textbf{63.03} & \textbf{77.20} & 76.05 & 43.77 & 71.98 & 29.80 & \textbf{60.31} \\
 & \cellcolor{shade}{Thanos} & \cellcolor{shade}{\optima} & \cellcolor{shade}{\textbf{8.99}} & \cellcolor{shade}{60.30} & \cellcolor{shade}{76.39} & \cellcolor{shade}{\textbf{76.30}} & \cellcolor{shade}{\textbf{43.77}} & \cellcolor{shade}{\textbf{72.14}} & \cellcolor{shade}{\textbf{30.60}} & \cellcolor{shade}{59.92} \\
\bottomrule
\end{tabular}
}
\caption{Qwen-2.5 perplexity on WikiText2 and accuracy on zero-shot downstream tasks for 60\% unstructured sparsity. \optima consistently improves the accuracy of the models across different tasks.}
\label{tab:qwen_60_unstructured}
\end{table}

\begin{table}[tb]
\centering
\setlength{\tabcolsep}{2pt}
\resizebox{\textwidth}{!}{
\begin{tabular}{l l l c c c c c c c | c}
\toprule
\multirow{2}{*}{\parbox[t]{1.8cm}{\centering Model}} &
\multirow{2}{*}{\parbox[t]{1.3cm}{\centering Mask\\Selection}} &
\multirow{2}{*}{\parbox[t]{1cm}{\centering Weight\\Update}} &
\multirow{2}{*}{Perplexity} &
\multicolumn{7}{c}{Metrics (\%)} \\
\cmidrule{5-11}
 & & & &
  \parbox[t]{\columndistance}{\centering \textcolor{blue!70!black}{MMLU}} &
  \parbox[t]{\columndistance}{\centering \textcolor{blue!70!black}{PIQA}} &
  \parbox[t]{\columndistance}{\centering \textcolor{blue!70!black}{Arc-E}} &
  \parbox[t]{\columndistance}{\centering \textcolor{blue!70!black}{Arc-C}} &
  \parbox[t]{\columndistance}{\centering \textcolor{blue!70!black}{Wino}} &
  \parbox[t]{\columndistance}{\centering \textcolor{blue!70!black}{OpenQA}} &
  \parbox[t]{\columndistance}{\centering \textbf{\textcolor{teal}{Average}}} \\
\midrule

\multirow{7}{*}{\parbox[t]{1.8cm}{\centering Qwen 2.5\\0.5B}} & Dense & -- & 13.08 & 47.36 & 69.97 & 64.18 & 29.18 & 55.80 & 24.40 & 48.48 \\
 & Wanda & -- & 41.30 & \textbf{27.88} & 61.75 & 48.99 & \textbf{23.38} & 52.72 & 14.00 & 38.12 \\
 & \cellcolor{shade}{Wanda} & \cellcolor{shade}{\optima} & \cellcolor{shade}{\textbf{27.61}} & \cellcolor{shade}{25.57} & \cellcolor{shade}{\textbf{63.76}} & \cellcolor{shade}{\textbf{51.18}} & \cellcolor{shade}{22.18} & \cellcolor{shade}{\textbf{53.28}} & \cellcolor{shade}{\textbf{15.80}} & \cellcolor{shade}{\textbf{38.63}} \\
 & SparseGPT & SparseGPT & 27.15 & \textbf{24.83} & 62.79 & 49.41 & 22.35 & 52.33 & \textbf{17.20} & 38.15 \\
 & \cellcolor{shade}{SparseGPT} & \cellcolor{shade}{\optima} & \cellcolor{shade}{\textbf{25.77}} & \cellcolor{shade}{23.38} & \cellcolor{shade}{\textbf{62.95}} & \cellcolor{shade}{\textbf{51.30}} & \cellcolor{shade}{\textbf{22.95}} & \cellcolor{shade}{\textbf{54.70}} & \cellcolor{shade}{17.00} & \cellcolor{shade}{\textbf{38.71}} \\
 & Thanos & Thanos & 27.58 & \textbf{24.31} & 62.68 & 49.92 & 21.42 & 51.78 & \textbf{16.60} & 37.78 \\
 & \cellcolor{shade}{Thanos} & \cellcolor{shade}{\optima} & \cellcolor{shade}{\textbf{26.26}} & \cellcolor{shade}{23.60} & \cellcolor{shade}{\textbf{62.79}} & \cellcolor{shade}{\textbf{51.22}} & \cellcolor{shade}{\textbf{22.18}} & \cellcolor{shade}{\textbf{54.30}} & \cellcolor{shade}{\textbf{16.60}} & \cellcolor{shade}{\textbf{38.45}} \\
\doublemidrule

\multirow{7}{*}{\parbox[t]{1.8cm}{\centering Qwen 2.5\\1.5B}} & Dense & -- & 9.28 & 59.70 & 75.73 & 75.34 & 40.96 & 63.14 & 32.20 & 57.84 \\
 & Wanda & -- & 21.92 & 39.95 & 67.25 & 61.11 & 28.84 & 58.09 & 20.80 & 46.01 \\
 & \cellcolor{shade}{Wanda} & \cellcolor{shade}{\optima} & \cellcolor{shade}{\textbf{17.14}} & \cellcolor{shade}{\textbf{39.96}} & \cellcolor{shade}{\textbf{69.26}} & \cellcolor{shade}{\textbf{62.33}} & \cellcolor{shade}{\textbf{30.29}} & \cellcolor{shade}{\textbf{58.72}} & \cellcolor{shade}{\textbf{23.00}} & \cellcolor{shade}{\textbf{47.26}} \\
 & SparseGPT & SparseGPT & 17.24 & \textbf{41.05} & 69.64 & \textbf{63.09} & \textbf{30.63} & \textbf{61.17} & \textbf{23.00} & \textbf{48.10} \\
 & \cellcolor{shade}{SparseGPT} & \cellcolor{shade}{\optima} & \cellcolor{shade}{\textbf{16.52}} & \cellcolor{shade}{35.64} & \cellcolor{shade}{\textbf{69.91}} & \cellcolor{shade}{62.75} & \cellcolor{shade}{29.35} & \cellcolor{shade}{60.93} & \cellcolor{shade}{\textbf{23.00}} & \cellcolor{shade}{46.93} \\
 & Thanos & Thanos & 17.56 & \textbf{42.83} & 68.55 & 61.53 & 29.10 & \textbf{58.96} & 21.40 & 47.06 \\
 & \cellcolor{shade}{Thanos} & \cellcolor{shade}{\optima} & \cellcolor{shade}{\textbf{16.83}} & \cellcolor{shade}{40.31} & \cellcolor{shade}{\textbf{69.75}} & \cellcolor{shade}{\textbf{62.67}} & \cellcolor{shade}{\textbf{30.38}} & \cellcolor{shade}{58.56} & \cellcolor{shade}{\textbf{25.00}} & \cellcolor{shade}{\textbf{47.78}} \\
\doublemidrule

\multirow{7}{*}{\parbox[t]{1.8cm}{\centering Qwen 2.5\\3B}} & Dense & -- & 8.03 & 65.00 & 78.35 & 77.31 & 44.88 & 68.43 & 29.20 & 60.53 \\
 & Wanda & -- & 17.14 & \textbf{46.68} & 70.08 & 64.77 & \textbf{31.66} & 61.48 & 22.20 & 49.48 \\
 & \cellcolor{shade}{Wanda} & \cellcolor{shade}{\optima} & \cellcolor{shade}{\textbf{14.08}} & \cellcolor{shade}{46.55} & \cellcolor{shade}{\textbf{71.44}} & \cellcolor{shade}{\textbf{64.90}} & \cellcolor{shade}{31.31} & \cellcolor{shade}{\textbf{64.17}} & \cellcolor{shade}{\textbf{25.40}} & \cellcolor{shade}{\textbf{50.63}} \\
 & SparseGPT & SparseGPT & 14.06 & \textbf{43.79} & \textbf{72.03} & 66.16 & 31.31 & 64.96 & 25.20 & 50.58 \\
 & \cellcolor{shade}{SparseGPT} & \cellcolor{shade}{\optima} & \cellcolor{shade}{\textbf{13.57}} & \cellcolor{shade}{43.36} & \cellcolor{shade}{71.44} & \cellcolor{shade}{\textbf{66.62}} & \cellcolor{shade}{\textbf{31.91}} & \cellcolor{shade}{\textbf{65.27}} & \cellcolor{shade}{\textbf{27.00}} & \cellcolor{shade}{\textbf{50.93}} \\
 & Thanos & Thanos & 14.35 & 41.41 & \textbf{71.49} & 61.70 & 29.27 & 64.01 & 25.20 & 48.85 \\
 & \cellcolor{shade}{Thanos} & \cellcolor{shade}{\optima} & \cellcolor{shade}{\textbf{13.75}} & \cellcolor{shade}{\textbf{43.56}} & \cellcolor{shade}{70.73} & \cellcolor{shade}{\textbf{64.06}} & \cellcolor{shade}{\textbf{30.97}} & \cellcolor{shade}{\textbf{64.09}} & \cellcolor{shade}{\textbf{25.40}} & \cellcolor{shade}{\textbf{49.80}} \\
\doublemidrule

\multirow{7}{*}{\parbox[t]{1.8cm}{\centering Qwen 2.5\\7B}} & Dense & -- & 6.85 & 71.76 & 78.73 & 80.51 & 48.38 & 72.61 & 33.40 & 64.23 \\
 & Wanda & -- & \textbf{11.47} & \textbf{61.03} & \textbf{74.48} & \textbf{75.34} & \textbf{42.15} & \textbf{68.82} & \textbf{27.80} & \textbf{58.27} \\
 & \cellcolor{shade}{Wanda} & \cellcolor{shade}{\optima} & \cellcolor{shade}{11.80} & \cellcolor{shade}{53.90} & \cellcolor{shade}{74.10} & \cellcolor{shade}{71.97} & \cellcolor{shade}{38.05} & \cellcolor{shade}{67.96} & \cellcolor{shade}{26.40} & \cellcolor{shade}{55.40} \\
 & SparseGPT & SparseGPT & \textbf{10.21} & \textbf{60.30} & \textbf{75.57} & \textbf{75.59} & \textbf{41.38} & \textbf{71.51} & \textbf{28.20} & \textbf{58.76} \\
 & \cellcolor{shade}{SparseGPT} & \cellcolor{shade}{\optima} & \cellcolor{shade}{10.92} & \cellcolor{shade}{53.90} & \cellcolor{shade}{74.32} & \cellcolor{shade}{72.43} & \cellcolor{shade}{37.46} & \cellcolor{shade}{69.61} & \cellcolor{shade}{27.80} & \cellcolor{shade}{55.92} \\
 & Thanos & Thanos & \textbf{10.45} & \textbf{60.12} & \textbf{74.54} & \textbf{75.08} & \textbf{41.13} & \textbf{69.93} & \textbf{28.80} & \textbf{58.27} \\
 & \cellcolor{shade}{Thanos} & \cellcolor{shade}{\optima} & \cellcolor{shade}{11.13} & \cellcolor{shade}{54.90} & \cellcolor{shade}{73.94} & \cellcolor{shade}{70.83} & \cellcolor{shade}{35.15} & \cellcolor{shade}{69.06} & \cellcolor{shade}{26.00} & \cellcolor{shade}{54.98} \\
\doublemidrule

\multirow{7}{*}{\parbox[t]{1.8cm}{\centering Qwen 2.5\\14B}} & Dense & -- & 5.30 & 77.62 & 81.28 & 82.24 & 55.80 & 75.14 & 34.40 & 67.75 \\
 & Wanda & -- & 9.70 & 65.82 & 76.99 & 76.89 & 45.39 & 73.56 & 31.80 & 61.74 \\
 & \cellcolor{shade}{Wanda} & \cellcolor{shade}{\optima} & \cellcolor{shade}{\textbf{8.90}} & \cellcolor{shade}{\textbf{67.03}} & \cellcolor{shade}{\textbf{77.31}} & \cellcolor{shade}{\textbf{77.82}} & \cellcolor{shade}{\textbf{46.76}} & \cellcolor{shade}{\textbf{74.27}} & \cellcolor{shade}{\textbf{32.60}} & \cellcolor{shade}{\textbf{62.63}} \\
 & SparseGPT & SparseGPT & 9.02 & \textbf{67.45} & 77.58 & 77.61 & \textbf{44.62} & 73.88 & \textbf{32.60} & \textbf{62.29} \\
 & \cellcolor{shade}{SparseGPT} & \cellcolor{shade}{\optima} & \cellcolor{shade}{\textbf{8.82}} & \cellcolor{shade}{67.33} & \cellcolor{shade}{\textbf{77.58}} & \cellcolor{shade}{\textbf{77.82}} & \cellcolor{shade}{44.28} & \cellcolor{shade}{\textbf{73.88}} & \cellcolor{shade}{32.00} & \cellcolor{shade}{62.15} \\
 & Thanos & Thanos & 9.06 & \textbf{66.31} & 77.64 & \textbf{77.90} & \textbf{46.25} & 72.77 & 31.20 & 62.01 \\
 & \cellcolor{shade}{Thanos} & \cellcolor{shade}{\optima} & \cellcolor{shade}{\textbf{8.92}} & \cellcolor{shade}{66.07} & \cellcolor{shade}{\textbf{77.97}} & \cellcolor{shade}{77.44} & \cellcolor{shade}{45.65} & \cellcolor{shade}{\textbf{72.93}} & \cellcolor{shade}{\textbf{31.80}} & \cellcolor{shade}{61.98} \\
\bottomrule
\end{tabular}
}
\caption{Qwen-2.5 perplexity on WikiText2 and accuracy on zero-shot downstream tasks for 2:4 sparsity. In this experiment, only the layers in the MLP part of the transformer are pruned, and the self-attention layers are dense, resulting in an end-to-end sparsity ratio of 38\% to 41\%. \optima consistently improves the accuracy of the models across different tasks. Please note that ProxSparse pruning is limited to 2:4 sparsity, and hence our unstructured sparsity experiments do not include it.}
\label{tab:qwen_2_4_sparsity}
\end{table}

\label{app:adam}
\paragraph{Comparison with Alternative Optimizers.} 
While our constrained QP solver leverages theoretical guarantees for convergence and optimality, we also compare it against ADAM \citep{adam}, a popular first-order optimizer without such assurances for quadratic problems. We reformulate the weight update as a mean squared error (MSE) minimization problem and use ADAM for solving it. Optimizers such as ADAM do not guarantee convergence, and are sensitive to their hyperparameters. For each layer, we do an exhaustive search with 4 different learning rates ranging from $10^{-2}$ to $10^{-5}$, each with a linear learning rate scheduler and choose the best configuration for final weight update.

\begin{table}[tb]
\centering
\setlength{\tabcolsep}{2pt}
\resizebox{\textwidth}{!}{
\begin{tabular}{l l l c c c c c c c | c}
\toprule
\multirow{2}{*}{\parbox[t]{1.5cm}{\centering Model}} & 
\multirow{2}{*}{\parbox[t]{1.3cm}{\centering Mask\\Selection}} & 
\multirow{2}{*}{\parbox[t]{1cm}{\centering Weight\\Update}} & 
\multirow{2}{*}{Perplexity} & 
\multicolumn{7}{c}{Metrics (\%)} \\
\cmidrule{5-11}
 & & & & 
   \parbox[t]{\columndistance}{\centering \textcolor{blue!70!black}{MMLU}} & 
   \parbox[t]{\columndistance}{\centering \textcolor{blue!70!black}{PIQA}} & 
   \parbox[t]{\columndistance}{\centering \textcolor{blue!70!black}{Arc-E}} & 
   \parbox[t]{\columndistance}{\centering \textcolor{blue!70!black}{Arc-C}} & 
   \parbox[t]{\columndistance}{\centering \textcolor{blue!70!black}{Wino}} & 
   \parbox[t]{\columndistance}{\centering \textcolor{blue!70!black}{OpenQA}} & 
   \parbox[t]{\columndistance}{\centering \textbf{\textcolor{teal}{Average}}} \\
\midrule
\multirow{7}{*}{\parbox[t]{1.5cm}{\centering Gemma\\3 1B}} & Dense & -- & 14.17 & 24.95 & 74.81 & 71.93 & 35.41 & 58.72 & 28.80 & 49.10 \\
 & Wanda & -- & 32.96 & 22.97 & 67.19 & 61.03 & 26.37 & 55.72 & 20.00 & 42.21 \\
 & Wanda & ADAM & 29.25 & 23.16 & 69.04 & 62.71 & 27.73 & 57.46 & 22.20 & 43.72 \\
 & \cellcolor{shade}Wanda & \cellcolor{shade}\optima & \cellcolor{shade}28.90 & \cellcolor{shade}23.96 & \cellcolor{shade}69.48 & \cellcolor{shade}\textbf{62.84} & \cellcolor{shade}\textbf{28.58} & \cellcolor{shade}56.83 & \cellcolor{shade}22.40 & \cellcolor{shade}\textbf{44.01} \\
 & SparseGPT & SparseGPT & 28.34 & 24.85 & 68.88 & 60.94 & 26.62 & 55.49 & 21.40 & 43.03 \\
 & SparseGPT & ADAM & 27.12 & 24.74 & 69.53 & 61.36 & 27.05 & 54.78 & 22.20 & 43.28 \\
 & \cellcolor{shade}SparseGPT & \cellcolor{shade}\optima & \cellcolor{shade}27.35 & \cellcolor{shade}\textbf{25.73} & \cellcolor{shade}\textbf{69.75} & \cellcolor{shade}60.90 & \cellcolor{shade}27.82 & \cellcolor{shade}56.35 & \cellcolor{shade}22.00 & \cellcolor{shade}43.76 \\
\doublemidrule
\multirow{7}{*}{\parbox[t]{1.5cm}{\centering OPT\\125M}} & Dense & -- & 27.67 & 22.85 & 62.84 & 43.56 & 19.45 & 49.88 & 16.40 & 35.83 \\
 & Wanda & -- & 39.50 & 22.92 & 61.15 & 39.94 & 19.88 & 52.17 & 14.00 & 35.01 \\
 & Wanda & ADAM & 205.82 & 25.63 & 57.02 & 34.13 & 17.66 & 50.51 & 13.00 & 32.99 \\
 & \cellcolor{shade}Wanda & \cellcolor{shade}\optima & \cellcolor{shade}35.44 & \cellcolor{shade}23.02 & \cellcolor{shade}61.66 & \cellcolor{shade}\textbf{42.93} & \cellcolor{shade}19.11 & \cellcolor{shade}50.12 & \cellcolor{shade}14.60 & \cellcolor{shade}35.24 \\
 & SparseGPT & SparseGPT & 36.88 & 23.00 & 61.97 & 40.99 & 19.71 & 53.59 & 14.60 & 35.64 \\
 & SparseGPT & ADAM & 224.34 & 23.15 & 56.75 & 35.65 & 17.49 & 47.36 & 12.20 & 32.10 \\
 & \cellcolor{shade}SparseGPT & \cellcolor{shade}\optima & \cellcolor{shade}35.61 & \cellcolor{shade}\textbf{23.85} & \cellcolor{shade}\textbf{62.37} & \cellcolor{shade}42.28 & \cellcolor{shade}\textbf{19.97} & \cellcolor{shade}\textbf{52.25} & \cellcolor{shade}\textbf{15.40} & \cellcolor{shade}\textbf{36.02} \\
\bottomrule
\end{tabular}
}
\caption{Comparison of \optima with other optimizers without convergence guarantees (ADAM). ADAM can lead to suboptimal solutions (Gemma 3 1B) or divergence of the model (OPT 125M).}
\label{tab:qp_vs_adam}
\end{table}

\autoref{tab:qp_vs_adam} illustrates this on Gemma 3 1B and OPT 125M \citep{opt} under 50\% unstructured sparsity. We show two examples in \autoref{tab:qp_vs_adam}, showing that ADAM results in suboptimal solutions. To further test the limitations of optimizers without convergence guarantees, we test ADAM on OPT-125M, and observe that it leads to divergence of the model. On Gemma 3 1B, ADAM yields competitive results in some cases (e.g., slightly lower perplexity for SparseGPT+ADAM at 27.12 versus \optima's 27.35), but \optima achieves higher overall accuracy (e.g., 44.01\% for Wanda+\optima versus 43.72\% for Wanda+ADAM). However, on smaller models like OPT 125M, ADAM exhibits instability, leading to divergence and dramatically higher perplexity (e.g., 205.82 for Wanda+ADAM versus 35.44 for Wanda+\optima). This underscores the risks of using non-specialized optimizers for our column-wise QPs, where suboptimal or unstable solutions can degrade model quality. \optima's use of provably convergent methods like rAPDHG ensures reliable and superior weight updates, making it a more robust choice for post-training pruning.

\paragraph{Layer-wise Error Improvement.} To provide a deeper insight into how \optima improves the accuracy of the models, we compare the layer-wise error of different layers in LLaMA-3.2 1B during pruning with and without \optima. \autoref{fig:layerwise_error} shows the relative output error improvement of all the pruned layers in the model, defined as $\frac{MSE(Y_\text{\optima}, Y_\text{dense})}{MSE(Y_\text{other}, Y_\text{dense})}$, where $MSE$ denotes the mean squared error across the calibration dataset. \autoref{fig:layerwise_error} shows that \optima consistently improves the layer-wise error of other methods, resulting in superior accuracy on the downstream tasks.

\paragraph{Pruning Time Analysis.} To evaluate the computational efficiency of \optima, we measured the time required to prune various language models. The pruning process was conducted on a single NVIDIA H100 GPU with 80GB of memory. Our measurements show that pruning times vary with model size: smaller models like LLaMA 3.2 1B and Gemma 3 1B each required approximately \SI{2.5}{\hour}, Gemma 2 2B took \SI{5.5}{\hour}, LLaMA 3.2 3B needed \SI{7.0}{\hour}, and the larger LLaMA 3.1 8B model required up to \SI{40.0}{\hour}.

The results indicate that pruning time scales with model size, reflecting the computational complexity of \optima's pruning algorithm, which adapts to the architectural differences across models. The consistency in pruning times for models of similar size (e.g., LLaMA 3.2 1B and Gemma 3 1B) highlights the robustness of \optima in handling diverse model architectures efficiently.

\section{Conclusion and Limitations}
\label{sec:optima_conclusion}

In this chapter, we explored the limits of the sparsity pillar under a strict \textbf{resource-constrained regime}. Recognizing that end-to-end training is not always feasible, we developed \optima to determine the mathematical upper bound of reconstruction accuracy possible using only a small calibration dataset. By reformulating post-training weight reconstruction as globally optimal, column-wise Quadratic Programs (QPs) and leveraging the shared-Hessian structure, we achieved massive parallelism on standard GPUs without the need for backpropagation.

Our results demonstrate that this principled approach pays significant dividends. \optima functions as a drop-in weight-update step for common mask selectors, improving zero-shot accuracy across various LLM families by up to 3.97 percentage points without any fine-tuning. Crucially, these gains persist even at high sparsity levels ($\geq 60\%$), proving that it is possible to create a highly sparse model that retains the fidelity of the original dense network.

However, while \optima minimizes the reconstruction error for any given mask, it remains bound by two fundamental limitations:
\begin{enumerate}
    \item \textbf{Layer-wise Pruning Sub-optimality:} It must operate within the constraints of a fixed, pre-determined sparsity pattern (like 2:4), which may not align with the model's true information distribution.
    \item \textbf{The Accuracy Gap:} Even with optimal reconstruction, a gap often remains between the sparse model and the dense baseline, suggesting that sparsity alone—without the aid of other Trinity pillars—has reached its ceiling in the zero-shot regime.
\end{enumerate}

To address the first limitation, the next chapter introduces \textbf{\patch}. We transition from the "no-training" regime of \optima to a "fine-tuning" regime, where we utilize a training budget to learn a flexible, \textbf{hybrid sparsity} structure that dynamically preserves density where it matters most. To address the second limitation (the accuracy gap), we will revisit \optima's findings in \autoref{chapter:slim}, where we demonstrate that re-introducing the \textbf{Low-Rank} pillar can bridge the remaining distance to dense performance.

\definecolor{PatchShade}{gray}{0.95}
\newcommand{\patchtile}{\textsc{PATCH}$^\text{Tile}$\xspace}
\newcommand{\patchjoint}{\textsc{PATCH}$^\text{Joint}$\xspace}
\definecolor{modelblue}{HTML}{646464}
\newcommand{\titlebox}[1]{%
  \tikz[baseline=(M.base)]{%
    \node[
      draw=modelblue,
      fill=modelblue!8,
      rounded corners=2pt,
      inner sep=1pt,
      font=\sffamily\footnotesize,
      baseline
    ] (M) {#1};\unskip
  }%
}
\definecolor{KeywordColor}{RGB}{0, 102, 153}
\definecolor{VarColor}{RGB}{128, 0, 128}
\definecolor{CommentColor}{RGB}{0, 128, 0}

\renewcommand{\algorithmiccomment}[1]{\hfill {\color{CommentColor}\scriptsize$\triangleright$ #1}}

\algrenewcommand\alglinenumber[1]{\scriptsize #1}

\chapter{\patch: Learnable Tile-Level Hybrid Sparsity for LLMs}
\label{chapter:patch}

\section{Statement of Contributions}

The content of this chapter is derived from the paper ``PATCH: Learnable Tile-Level Hybrid Sparsity for LLMs''~\citep{patch}. This research was a collaborative effort with Younes Hourri, and we are co-first authors with equal contribution.

The specific breakdown of contributions is as follows:

\begin{itemize}
    \item \textbf{Mohammad Mozaffari:} Conceived the original concept of learnable hybrid sparsity and implemented the initial version of the codebase. Designed and executed the experiments regarding quantization, Low-Rank Adaptation (LoRA), and the fine-tuning (FT) comparisons.

    \item \textbf{Younes Hourri:} Formulated the mathematical framework for the tile-level probability distributions (specifically \autoref{eq:tiled_mask} and \autoref{eq:tile_merge}), extended the codebase with additional capabilities, and conducted the primary pruning experiments. He was also responsible for the hardware acceleration implementation and throughput analysis.

    \item \textbf{Joint Contributions:} Both authors collaborated closely on the writing and revision of the manuscript.
\end{itemize}

Maryam Mehri Dehnavi held a supervisory position in this work.

\section{Introduction}
\label{sec:patch_intro}

In the previous chapter, we established \optima to maximize the accuracy of sparse models under a strict "no-training" constraint. We demonstrated that for a fixed mask structure, one can mathematically solve for the optimal weights. However, \optima, and indeed any layer-wise pruning and weight reconstruction method, eventually hits an accuracy ceiling imposed by the rigidity of the mask itself. To break this ceiling and fully refine the sparsity pillar, we must relax the resource constraints. In this chapter, we transition from the "zero-training" regime of \optima to a \textbf{learnable regime}, utilizing a training budget to move beyond optimizing values within a fixed pattern and address the pattern itself. We need a method that bridges the gap between the high accuracy of flexible unstructured pruning and the hardware acceleration of rigid structured patterns.

Currently, sparsity techniques operate at two extremes, neither of which is sufficient for the Compression Trinity. Unstructured sparsity allows non-zero elements to appear anywhere, theoretically matching dense model accuracy due to its flexibility in allocation~\citep{wanda, sparsegpt, cerebras_sparsity}. However, its irregular memory access patterns hinder acceleration on modern hardware like GPUs, preventing practical speedups~\citep{flash_llm, spinfer}. Conversely, semi-structured sparsity offers practical acceleration but imposes strict layout expectations. Specifically, we focus on the 2:4 sparse pattern supported by NVIDIA Ampere and Hopper architectures, as detailed in \autoref{chapter:background}. This format requires every block of four contiguous weights to contain at least two zeros to utilize sparse Tensor Cores. This "one-size-fits-all" approach enforces a uniform 50\% sparsity ratio across all layers, failing to account for the varying sensitivity of different network components. This often leads to significant accuracy loss when models are pruned using one-shot methods~\citep{wanda, sparsegpt, thanos, proxsparse} or end-to-end learned masks~\citep{maskllm}. Recent studies confirm that sparsity should be allocated non-uniformly for optimal performance~\citep{owl, rethinkingvaluetransformercomponents, layeradaptivesparsitymagnitudebasedpruning}, yet standard 2:4 sparsity locks the model into a fixed allocation.

To bridge this gap and create a truly flexible sparsity pillar, we propose \textbf{P}runing with a Le\textbf{a}rnable \textbf{T}ile-level \textbf{C}onfiguration for \textbf{H}ybrid Sparsity (\textbf{\patch}). Instead of forcing the entire model to adhere to a rigid sparse structure, \patch introduces a hybrid mask that partitions each weight matrix into hardware-friendly tiles. Through a learnable masking process---enabled by our relaxed training budget---\patch designates each tile as either dense (0\% sparsity) or 2:4 sparse (50\% sparsity). This adaptive approach allows the matrix to realize an effective global sparsity ratio anywhere between 0\% and 50\%, dynamically balancing accuracy in critical regions with hardware-friendly sparsity elsewhere. While the \patch methodology is generalizable to any block-based sparsity pattern (such as 4:8 or custom N:M ratios), we focus our evaluation on 2:4 sparsity as it is the only pattern currently supported by native acceleration on commodity GPUs (see \autoref{chapter:background}).

We enable this flexibility through two distinct optimization strategies. For maximum accuracy, we employ a joint optimization method that tunes both the sparsity pattern within the 2:4 tiles and the tile-level configurations during training. For scenarios with tighter compute budgets, we offer a variant that tunes only the location of the dense tiles while keeping the initial 2:4 mask fixed. Importantly, unlike theoretical hybrid methods that never see deployment, \patch is fully compatible with tile-level sparsity acceleration libraries and compilers such as STOICC~\citep{stoicc}. This makes it the first hybrid sparsity method to demonstrate practical speedups on commodity hardware. For instance, on LLaMA-2 7B running on a consumer-grade A6000 GPU, \patch achieves 1.18$\times$--1.38$\times$ end-to-end speedup over the dense baseline while improving accuracy by 0.37\%--2.96\% compared to the state-of-the-art 2:4 pruning method, MaskLLM. Note that this chapter focuses exclusively on refining the sparsity pillar; the integration of \patch with quantization and low-rank approximation is presented in \autoref{chapter:slim}, where we demonstrate the combined efficacy of the Compression Trinity.

\begin{figure}[t]
\centering\includegraphics[width=\linewidth]{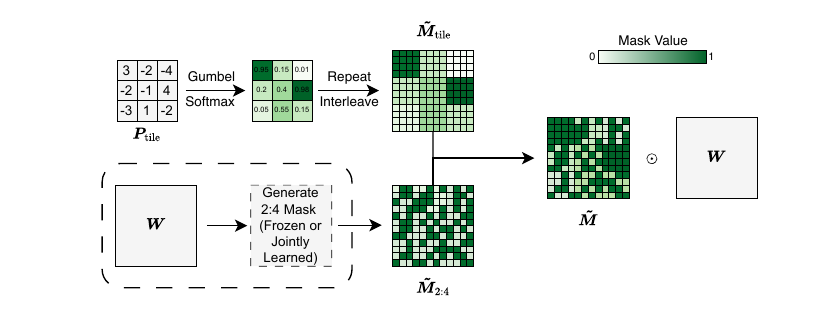}
\caption{Illustration of the PATCH learning process for generating tile-level hybrid masks. Each tile is parameterized by a learnable distribution and sampled with Gumbel Softmax to produce $\Tilde{M}_{\text{tile}}$. The dense probability is expanded and merged with a 2:4 mask $\Tilde{M}_{2:4}$, which can be fixed or jointly learned during training, yielding $\Tilde{M}$. The final mask assigns each tile to remain dense or follow the 2:4 pattern, enabling flexible sparsity across the weight matrix.}
\label{fig:overview}
\end{figure}

\section{Additional Related Work}
\label{sec:related_work}

\subsection{Pruning methods}

Pruning is one of the most widely studied approaches for compressing deep neural networks, with the goal of removing redundant parameters while preserving accuracy. Classical pruning methods can be broadly categorized into \emph{local} (layer-wise) and \emph{global} (end-to-end) strategies.

\paragraph{Local pruning.}
Local approaches prune each layer independently, typically by minimizing reconstruction error within that layer. A seminal example is Optimal Brain Surgeon (OBS) \citep{obs, optimal_brain_compression}, which leverages second-order information to identify and remove weights while updating the remaining parameters to compensate for loss. While highly principled, the quadratic cost of computing and inverting the Hessian makes OBS infeasible for large models. 

Recent work adapts these ideas to LLM-scale pruning. SparseGPT \citep{sparsegpt} formulates layer-wise pruning as a sparse regression problem, enabling efficient approximations of OBS that scale to billion-parameter models. Thanos \citep{thanos} further improves accuracy by employing multi-column approximations to reduce error accumulation. Wanda \citep{wanda}, on the other hand, discards explicit weight updates and instead uses a simple magnitude-activation criterion with calibration data, yielding competitive quality with extremely fast runtimes. Despite their efficiency, local methods often suffer from limited capacity to recover accuracy since pruning decisions ignore cross-layer dependencies.

\paragraph{Global pruning.}
Global approaches aim to jointly optimize pruning decisions across layers, typically leading to better overall trade-offs. Optimal Brain Damage (OBD) \citep{optimal_brain_damage} is an early global method that estimates weight saliency using the diagonal Hessian. Extensions such as WoodFisher \citep{woodfisher} approximate the Hessian via Kronecker factorizations, making computation more tractable but still challenging for modern LLMs \citep{mkor}. 

More recent approaches bypass costly second-order computations. MaskLLM \citep{maskllm} formulates pruning as a binary classification task (keep vs.\ prune) and solves it using standard optimizers such as AdamW \citep{adamw}, achieving strong results even under hardware-friendly structured sparsity (e.g., $2{:}4$). ProxSparse \citep{proxsparse} instead adopts a proximal regularization framework, reducing the overhead of MaskLLM while trading off some pruning accuracy. These works highlight the tension between pruning quality and efficiency: global methods often achieve higher accuracy but remain more computationally expensive than simple one-shot local pruning.

\subsection{Complementary compression techniques}

Beyond pruning, several orthogonal compression techniques are widely used and can be combined with sparsity for additional gains. \emph{Quantization} reduces the bit precision of parameters and activations, e.g., from 32-bit floating point to 8- or 4-bit integers, thereby reducing memory footprint and accelerating inference \citep{quantization_survey, quantization_survey2}. 

\emph{Low-rank adaptation} methods decompose weight matrices into smaller factors, effectively reducing parameter counts while maintaining expressivity. Recent approaches such as LQ-LoRA \citep{lqlora}, SLiM \citep{slim}, and SLoPe \citep{slope} demonstrate that low-rank structures can be used both for efficient fine-tuning and for direct model compression. 

Finally, \emph{knowledge distillation} \citep{knowledge_distillation} transfers knowledge from a large teacher model to a smaller student, yielding compact models that retain much of the teacher’s performance. These methods are complementary to pruning, and hybrid frameworks that integrate sparsity, quantization, and low-rank factorization represent a promising direction for achieving high compression ratios without sacrificing accuracy.

\section{Preliminaries}

\paragraph{Differentiable Sampling.} Sampling from a categorical distribution is inherently non-differentiable, which poses challenges for gradient-based optimization. The Gumbel Softmax~\citep{gumbelsoftmax} addresses this by combining the Gumbel-Max reparameterization trick together with a softmax relaxation. The reparameterization expresses the sampling process by decoupling the deterministic log-probabilities $p \in \mathbb{R}^n$ from the stochastic perturbations $z \in \mathbb{R}^n$ introduced by Gumbel noise, which emulate random draws from the distribution. The subsequent softmax yields a differentiable approximation to categorical sampling:

\begin{equation}
    \mathrm{GS}(p;\, \tau)_k =  \frac{\exp((p_k + z_k) / \tau)}{\sum_j \exp((p_j + z_j) / \tau)}
    \label{eq:gumbel_softmax}
\end{equation}

where $z_k = -\log(-\log(u_k))$ with $u_k \sim \mathrm{Uniform}(0,1)$. The resulting vector $\mathrm{GS}(p;\, \tau) \in \mathbb{R}^n$ is a soft index vector whose entries $\mathrm{GS}(p;\, \tau)_k$ represent the relaxed probability of selecting class $k$.

Additionally, the temperature parameter $\tau$ controls the hardness of the sampled index. Lower values of $\tau$ yield a more peaked distribution, causing $\mathrm{GS}(p)$ to converge to a one-hot vector as $\tau \rightarrow 0$. 

\paragraph{Learnable 2:4 Mask.} MaskLLM~\citep{maskllm} formulates 2:4 mask selection as a learnable probabilistic process over the six possible patterns. The underlying weights remain fixed, while training shifts the categorical distribution to favor masks that preserve better pruning performance. The mask for each four consecutive elements can be parameterized with a vector $p \in \mathbb{R}^{6 \times 1}$. Scaling this vector to a weight matrix  $ W \in \mathbb{R}^{d_1 \times d_2}$ will result in $P_{\text{2:4}} \in \mathbb{R}^{6 \times \frac{d_1 d_2}{4}}$ as the mask search parameters. The resulting mask can be computed as in \autoref{eq:2_4_mask}, where $\Tilde{M}_{\text{2:4}} \in [0,1]^{d_1 \times d_2}$ denotes the 2:4 soft mask, obtained as a weighted average over the candidate masks, and $S \in \mathbb{R}^{6 \times 4}$ is the matrix containing these six candidates as its rows.\footnote{We will refer to a mask value of $1$ as \textit{keeping} the corresponding weight and a value of $0$ as \textit{pruning} it.}

\begin{equation}
    \Tilde M_{\text{2:4}} =  \text{\tt {reshape}}(\mathrm{GS}(P_{\text{2:4}}; \tau, \kappa) \times S,  \mathbb{R}^{d_1 \times d_2})
    \label{eq:2_4_mask}
\end{equation}

A scaling factor $\kappa$ is also introduced in \autoref{eq:gumbel_softmax}, where it multiplies the logits $p$ before adding the Gumbel noise $z$, thereby controlling their relative influence. Small $\kappa$ values let the noise dominate, encouraging exploration across candidate masks, while larger $\kappa$ values amplify the logits and make the sampling more deterministic.

\section{\patch}
\label{sec:patch}

To overcome the rigidity of fixed 50\% 2:4 sparsity, we introduce \patch. \patch learns a structured mask—optimized on top of frozen weights—that is partitioned into tiles, where each tile decides whether its corresponding weights remain dense or are pruned with a 2:4 pattern. This design preserves accuracy in sensitive regions while exploiting hardware-accelerated sparsity elsewhere. Unlike fixed 2:4 sparsity, which enforces the same pattern across all weights, \patch adapts at the tile level by assigning dense tiles to critical regions and sparse tiles elsewhere.

Finding the optimal allocation of dense tiles (value 1) and sparse tiles (2:4 pattern) within a mask is a combinatorially difficult problem, as the number of possible configurations grows rapidly with the number of tiles across the LLM. By also modeling this problem as a probabilistic sampling process, and adjusting the probability of each tile (and the 2:4 patterns within sparse tiles), \patch can efficiently explore the space of configurations and converge toward masks that balance accuracy and sparsity. The mask distributions are learned end-to-end by training the Gumbel–Softmax logits while keeping the model weights frozen. We address this challenge by formulating mask selection as two coupled subproblems: \textit{(1) selecting which tiles are dense or sparse}, and \textit{(2) choosing the 2:4 sparsity pattern within sparse tiles}.

\paragraph{Tile-based pruning of LLMs.} We associate each parameter matrix $W \in \mathbb{R}^{d_1 \times d_2}$ with a grid of tile-level distributions, each parameterized by a learnable logit. Collectively, these form $P_{\text{tile}} \in \mathbb{R}^{\tfrac{d_1}{b_1} \times \tfrac{d_2}{b_2}}$, where each entry specifies the unnormalized score of keeping the corresponding $b_1 \times b_2$ tile fully dense. To create a two-class distribution (keep dense vs.\ prune), we concatenate a fixed zero to each logit, yielding $[P_{\text{tile}}, 0] \in \mathbb{R}^{\tfrac{d_1}{b_1} \times \tfrac{d_2}{b_2} \times 2}$. After applying Gumbel–Softmax, we broadcast the dense probabilities across their respective $b_1 \times b_2$ region (since the weighted average of the two outcomes reduces to $p_{\text{dense}} \cdot 1 + p_{\text{prune}} \cdot 0 = p_{\text{dense}}$), so that all elements of a tile receive the same mask value. Formally,

\begin{equation}
    \Tilde{M}_{\text{tile}} = \mathrm{GS}([P_{\text{tile}},0]; \tau, \kappa)_{:,:,0} \otimes \mathbf{1}.
    \label{eq:tiled_mask}
\end{equation}

This yields the tile-level mask $\Tilde{M}_{\text{tile}} \in [0,1]^{d_1 \times d_2}$ in \autoref{eq:tiled_mask}, where $\mathbf{1} \in \mathbb{R}^{b_1 \times b_2}$ is an all-ones matrix and $\otimes$ denotes the Kronecker product.

\paragraph{Joint optimization with sparse mask.} To fully determine the effective sparsity pattern, the tile-level mask must be combined with the fine-grained 2:4 mask. Assuming that the 2:4 mask $\Tilde{M}_{2:4}$ is generated using \autoref{eq:2_4_mask}, \patch combines it with the tile mask $\Tilde{M}_{\text{tile}}$ as shown in \autoref{eq:tile_merge}. The resulting soft mask interpolates between dense and sparse behavior: values of $\Tilde{M}_{\text{tile}}$ close to one make the tile predominantly dense, while values close to zero shift the tile toward the soft 2:4 mask pattern defined by $\Tilde{M}_{\text{2:4}}$. Thus, $\Tilde{M}$ can be understood as a per-tile weighted average of the dense option and the 2:4 patterns, with $\Tilde{M}_{\text{tile}}$ determining the relative contribution of each. An overview of the process is provided in \autoref{fig:overview}.

\begin{equation}
    \Tilde{M} = \Tilde{M}_{\text{tile}} + \left(1 - \Tilde{M}_{\text{tile}}\right) \odot \Tilde{M}_{\text{2:4}}
    \label{eq:tile_merge}
\end{equation}

\paragraph{Learning masks with targeted sparsity.} \patch uses a novel regularization term to achieve a flexible 0\%–50\% sparsity ratio across the model by controlling the number of dense tiles. Unlike traditional regularization methods like weight decay, which produce non-deterministic sparsity ratios, our term penalizes deviations from the target sparsity, enabling precise control. This global sparsity approach prunes sensitive linear layers less aggressively while setting redundant weight elements to zero, offering greater flexibility than fixed per-layer sparsity. We directly compare global versus per-layer sparsity regularization in \autoref{sec:patch_results}.

\paragraph{Training objective.}
The overall training objective, as shown in \autoref{eq::loss}, of \patch combines three components: the standard modeling loss, 
a sparsity regularization term that enforces the target density of the model $\rho$,  and a weight regularization term (as in MaskLLM) that promotes larger weight magnitudes and gradient propagation.  
Formally,
\begin{equation}
    \mathcal{L} 
    = \mathcal{L}_{LM}\!\left(x ; \Tilde{M}_i \odot W_i\right) 
    + \lambda_1 \left\lVert \frac{\sum_i \Tilde{M}_i}{\sum_i \lVert W_i \rVert_0} - \rho \right\rVert_1 
    - \lambda_2 \frac{\sum_i \lVert \Tilde{M}_i \odot W_i \rVert_2^2}{\sum_i \lVert W_i \rVert_2^2}
    \label{eq::loss}
\end{equation}

Following MaskLLM, we progressively \textit{decrease} $\tau$ and \textit{increase} $\kappa$ during training so that the Gumbel-Softmax distribution converges to a clear one-hot choice of mask by the end of training.

\paragraph{Inference.}  After training, the sign of each logit in $P_{\text{tile}}$ determines the final mask. Since a zero logit is concatenated to represent the sparse class (\autoref{eq:tiled_mask}), positive values correspond to the dense option, while negative values correspond to the sparse option. The complete procedure is outlined in \autoref{alg:patch}.

\begin{algorithm}[t]
\caption{\textbf{Joint Tile \& 2{:}4 Mask Learning}}
\label{alg:patch}

\begin{algorithmic}[1]
\Statex \textbf{\color{KeywordColor}Input:} Weight matrix $\color{VarColor}\mathbf{W}$, tile size $(\color{VarColor}b_1, \color{VarColor}b_2)$, sparsity target $\color{VarColor}\rho$, training steps $\color{VarColor}T$, loss hyperparameters $\color{VarColor}\lambda_1, \color{VarColor}\lambda_2$, temperature schedule $\{\color{VarColor}\tau_t\}_{t=1}^T$, scaling schedule $\{\color{VarColor}\kappa_t\}_{t=1}^T$.
\Statex \textbf{\color{KeywordColor}Output:} Learned pruning masks $\color{VarColor}\mathbf{M}^{\star}$, pruned weights $\color{VarColor}\widehat{\mathbf{W}}$.

\vspace{0.5em}
\State Initialize tile logits $\color{VarColor}\mathbf{P}_{\text{tile}} \in \mathbb{R}^{\frac{d_1}{b_1}\times \frac{d_2}{b_2}}$.
\State Initialize $\color{VarColor}\mathbf{P}_{\text{tile}}$ with one-shot prior.
\State Initialize differentiable 2{:}4 parameters $\color{VarColor}\mathbf{P}_{\text{2:4}} \in \mathbb{R}^{6 \times \frac{d_1 d_2}{4}}$.

\vspace{0.3em}
\For{$t = 1 \; \to \; T$}
    \State $\Tilde{\mathbf{M}}_{\text{tile}} \gets \mathrm{GS}([\mathbf{P}_{\text{tile}}, 0]; \tau_t, \kappa_t)_{:,:,0} \otimes \mathbf{1}_{b_1 \times b_2}$ \algorithmiccomment{\normalsize{Dense soft tile mask}}
    \State $\Tilde{\mathbf{M}}_{\text{2:4}} \gets$ \autoref{eq:2_4_mask} \algorithmiccomment{\normalsize{Differentiable 2:4 mask}}
    \State $\Tilde{\mathbf{M}}_i \gets \Tilde{\mathbf{M}}_{\text{tile}} + (1 - \Tilde{\mathbf{M}}_{\text{tile}}) \odot \Tilde{\mathbf{M}}_{\text{2:4}}$ \algorithmiccomment{\normalsize{Merge masks}}

    \vspace{0.3em}
    \State Compute loss:
    \begin{equation*}
    \begin{aligned}
        \mathcal{L} &= \mathcal{L}_{LM}(x ; \Tilde{\mathbf{M}} \odot \mathbf{W}) \quad + \lambda_1 \left\lVert \frac{\sum_i \Tilde{\mathbf{M}}_i}{\sum_i \lVert \mathbf{W}_i \rVert_0} - \rho \right\rVert_1 \quad - \lambda_2 \frac{\sum_i \lVert \Tilde{\mathbf{M}}_i \odot \mathbf{W}_i \rVert_2^2}{\sum_i \lVert \mathbf{W}_i \rVert_2^2}
    \end{aligned}
    \end{equation*}

    \State Update $\color{VarColor}\mathbf{P}_{\text{tile}}$,$
    \color{VarColor}\mathbf{P}_{\text{2:4}}$ via backpropagation.
\EndFor

\vspace{0.5em}
\State $\mathbf{M}_{\text{tile}}^{\star} \gets \mathbf{1}[\mathbf{P}_{\text{tile}} > 0] \otimes \mathbf{1}_{b_1 \times b_2}$ \algorithmiccomment{\normalsize{Hard tile mask}}
\State $\mathbf{M}_{\text{2:4}}^{\star} \gets$ select best 2:4 mask from $\mathbf{P}_{\text{2:4}}$.
\State $\mathbf{M}_i^{\star} \gets \mathbf{M}_{\text{tile}}^{\star} + (1 - \mathbf{M}_{\text{tile}}^{\star}) \odot \mathbf{M}_{\text{2:4}}^{\star}$.
\State $\widehat{\mathbf{W}} \gets \mathbf{W} \odot \mathbf{M}_i^{\star}$ \algorithmiccomment{\normalsize{Final pruned weights}}

\vspace{0.5em}
\Statex \textbf{\color{KeywordColor}Return:} Learned mask $\color{VarColor}\mathbf{M}^{\star}$, pruned weights $\color{VarColor}\widehat{\mathbf{W}}$.
\end{algorithmic}
\end{algorithm}

\paragraph{Memory efficient \patch.} To further reduce overhead, \patch can be run in a memory-efficient manner by freezing the sparse mask parameters and optimizing only the tile-level decisions. This reduces the number of learnable parameters to $\tfrac{d_1 d_2}{b_1 b_2}$. While this lighter formulation limits mask-selection flexibility and can reduce performance as seen in \autoref{ablation:tile_only}, it makes training feasible under strict memory constraints, such as fitting an 8B model on a single 80GB GPU. We denote this version of \patch by \patchtile and the joint optimization version of \patch by \patchjoint.

\section{Efficient deployment of \patch}

Executing \patch requires handling hybrid sparse–dense tiles, a capability not supported by existing GPU libraries. Current tools either focus exclusively on dense computation (e.g., cuBLAS \citep{cublas}, dense CUTLASS \citep{cutlass}, OpenAI Triton \citep{triton}), or restrict support to fixed $2{:}4$ sparsity (e.g., cuSPARSELt \citep{cusparselt}, sparse CUTLASS). STOICC \citep{stoicc} lifts these limitations by extending Triton with hybrid tile-level sparsity, making it a suitable backend for accelerating \patch.

Similar to Triton, STOICC employs an inspector that benchmarks candidate kernel configurations for each sparsity ratio, identifying the most hardware-efficient tile size for the target GPU. On NVIDIA A100 and A6000 GPUs, our experiments show that the optimal configurations are consistently drawn from $128{\times}128$ or its subdivisions (e.g., $128{\times}64$, $64{\times}128$, $64{\times}64$). In practice, this means that regardless of the sparsity ratio or the layer shape, the chosen $128{\times}128$ granularity guarantees that STOICC’s autotuned tiles can be applied consistently. Unless otherwise specified, we adopt these hardware-friendly tile sizes in all \patch experiments. Further implementation details are provided in \autoref{app::stoicc}.

\section{Experiments}
\label{sec:patch_results}

\paragraph{Model, dataset and evaluation.} We evaluate \patch across diverse transformer architectures, including the Qwen-2.5~\citep{qwen}, Gemma 3~\citep{gemma3}, and LLaMA-2~\citep{llama2} and 3~\citep{llama3} model families, spanning 500M to 8B parameters. Following the dataset size and configurations in MaskLLM \citep{maskllm}, masks are trained for 2000 steps with a batch size of 256 on sequences with a length of 4096 tokens from the SlimPajama dataset~\citep{slimpajama}.

Following previous LLM compression work \citep{slim, maskllm}, we evaluate the models on eight zero-shot downstream tasks: PIQA~\citep{piqa}, ARC-Easy and ARC-Challenge~\citep{arc}, Winogrande~\citep{winogrande}, OpenBookQA~\citep{openbookqa}, RACE~\citep{race}, HellaSwag~\citep{hellaswag}, and MMLU~\citep{mmlu} using the Language Model Evaluation Harness~\citep{eval-harness} framework. Additionally, similar to previous work \citep{slim, sparsegpt, wanda}, we evaluate the models on a language modeling task using the WikiText2~\citep{wiki} dataset with a sequence length of 4096, comparing against established baselines in the following sections.

\paragraph{Baselines.} To evaluate \patch against established 2:4 sparsity pruning techniques, we compare it with the state-of-the-art learnable method MaskLLM~\citep{maskllm}, as well as one-shot methods including Wanda~\citep{wanda}, SparseGPT~\citep{sparsegpt}, Thanos \citep{thanos}, ProxSparse \citep{proxsparse} and magnitude pruning~\citep{magnitudepruning}. For one-shot pruning methods, following the default configurations in each paper, we prune the models over 128 samples from the C4 dataset.

The publicly available MaskLLM pruned checkpoints are limited to LLaMA-2 7B and LLaMA-3.1 8B models. To ensure a fair comparison across all models, we implemented MaskLLM in PyTorch and replicated its results for additional architectures presented in this study.

We faced a similar challenge with ProxSparse as well, where only the LLaMA-2-7B and LLaMA-3.1-8B checkpoints are publicly available. We have pruned other models with their official code base using their default hyperparameters for comparison.

\niparagraph{Experiment Setup.} All masks are trained using the HuggingFace Trainer API \citep{huggingface} for 2000 steps with a global batch size of 256 and a sequence length of 4096, processing 2B tokens from the SlimPajama corpus \citep{slimpajama}. Training is accelerated via data parallelism across a single node with 4 H100 GPUs. In this setup, \patchjoint requires 4.5 and 6 GPU hours on the 0.5B and 1B models, respectively, while \patchtile requires 21 and 24 GPU hours on the 7B and 8B models. The hyperparameters for \patchjoint and \patchtile are summarized in \autoref{table::hyper_params}, tuned on Qwen-2.5-0.5B. For the 2:4 mask parameters, we follow the configuration from MaskLLM \citep{maskllm}.

\begin{table}[H]
    \centering
        \caption{Hyper-parameters used for \patchjoint and \patchtile across sparsity ratios. All hyper parameters were tuned on Qwen-2.5-0.5B.}
    \label{table::hyper_params}
    \resizebox{\linewidth}{!}{
    \begin{tabular}{c c c c c c c c c}
    \toprule
        \bf Sparsity & \bf Method & \bf Optimizer & \bf Logits Init &
        \bf Gumbel Scaling &
        \bf Gumbel &
        \bf Prior(Strength)  &
        \bf Sparse Reg. & \bf Weight Reg. 
 \\
    \midrule
       25\% & \patchjoint  & Adam(0.001)  & $\mathcal{N}(0, 0.014)$ & $25 \rightarrow 350$  & $2 \rightarrow 0.05$ & SparseGPT($3$) & 7 & 10 \\
       35\% & \patchjoint  & Adam(0.001)  & $\mathcal{N}(0, 0.014)$ & $25 \rightarrow 350$  & $2 \rightarrow 0.05$ & SparseGPT($3$) & 7 & 10 \\
       45\% & \patchjoint  & Adam(0.001)  & $\mathcal{N}(0, 0.014)$ & $25 \rightarrow 350$  & $4 \rightarrow 0.05$ & SparseGPT($3$) & 7 & 10 \\
    \midrule
       25\% & \patchtile   & Adam(0.0001) & $\mathcal{N}(0, 0.014)$ & $100 \rightarrow 500$ & $2 \rightarrow 0.05$ & SparseGPT($3$) & 3 & 0.1 \\
       35\% & \patchtile   & Adam(0.0001) & $\mathcal{N}(0, 0.014)$ & $100 \rightarrow 500$ & $2 \rightarrow 0.05$ & SparseGPT($3$) & 3 & 0.1 \\
       45\% & \patchtile   & Adam(0.0001) & $\mathcal{N}(0, 0.014)$ & $100 \rightarrow 500$ & $2 \rightarrow 0.05$ & SparseGPT($3$) & 3 & 0.1 \\
    \bottomrule
    \end{tabular}
    }

\end{table}

\subsection{Model Quality Results}

\paragraph{Joint sparse and dense tile optimization.} For smaller models like Qwen-2.5 0.5B, LLaMA-3.2 1B, and Gemma-3 1B, we apply the joint variant \patchjoint, which simultaneously optimizes dense tile locations and sparsity patterns within sparse tiles. This approach enables effective performance.

The average accuracy of the models across eight zero-shot downstream tasks and their perplexity on the WikiText2 dataset is reported in \autoref{table::joint_patch}. The results demonstrate that \patchjoint provides a flexible tradeoff between sparsity ratio and model quality, narrowing the performance gap to dense models while ensuring hardware-friendly inference. A similar pattern holds for larger models using a memory-efficient variant, as explored next.

\begin{table*}[tb]
\centering
\small
\caption{Model quality (average accuracy across eight zero-shot tasks and perplexity on WikiText2 dataset) for different pruning methods. By jointly optimizing the location of dense tiles and the sparsity pattern within the sparse tiles, \patchjoint allows for a continuous sparsity ratio for the models, providing a flexible tradeoff between sparsity and model quality.}
\setlength{\tabcolsep}{3pt}
\resizebox{\textwidth}{!}{
\begin{tabular}{l l l c c c c c c}
\toprule
\textbf{Sparsity} & \textbf{Method} & \textbf{Pattern} &
\multicolumn{2}{c}{\textbf{Qwen-2.5 0.5B}} &
\multicolumn{2}{c}{\textbf{LLaMA-3.2 1B}} &
\multicolumn{2}{c}{\textbf{Gemma-3 1B}} \\
\cmidrule(lr){4-5} \cmidrule(lr){6-7} \cmidrule(lr){8-9}
& & & \textbf{Acc (\% $\uparrow$)} & \textbf{PPL ($\downarrow$)} & \textbf{Acc (\% $\uparrow$)} & \textbf{PPL ($\downarrow$)} & \textbf{Acc (\% $\uparrow$)} & \textbf{PPL ($\downarrow$)} \\
\midrule
0\%  & Dense     & -      & 46.00 & 12.08 & 47.70 & 9.06  & 47.01 & 11.67 \\
\midrule
50\% & Magnitude & 2:4    & 30.16 & 6734.97 & 29.66 & 563.44 & 31.66 & 5005.56 \\
     & Wanda     & 2:4    & 32.97 & 72.48   & 31.61 & 78.18  & 34.16 & 69.41   \\
     & SparseGPT & 2:4    & 34.81 & 36.59   & 35.55 & 32.73  & 35.58 & 44.59   \\
     & Thanos    & 2:4    & 31.31 & 37.32   & 35.71 & 33.03  & 35.09 & 62.63   \\
     & ProxSparse& 2:4    & 32.05 & 111.05   & 33.55 & 49.33  & 36.63 & 90.50 \\
     & MaskLLM   & 2:4    & 39.33 & 15.22   & 41.04 & 12.93  & 41.84 & 12.82   \\
\midrule
\rowcolor{PatchShade} 45\% & \patchjoint     & Dense/2:4 Tiles & 40.29 & 14.57 & 42.08 & 12.23 & 42.80 & 11.96 \\
\rowcolor{PatchShade} 35\% & \patchjoint     & Dense/2:4 Tiles & 41.15 & 13.84 & 42.72 & 11.67 & 43.30 & 11.48 \\
\rowcolor{PatchShade} 25\% & \patchjoint     & Dense/2:4 Tiles & 42.39 & 13.47 & 43.81 & 11.00 & 44.07 & 11.17 \\
\bottomrule
\end{tabular}
}

\label{table::joint_patch}
\end{table*}

\paragraph{Memory-efficient tile selection.} For larger models such as LLaMA-2 7B and LLaMA-3.1 8B, we employ the memory-efficient variant \patchtile, which freezes the fine-grained sparse weight structure while optimizing dense tile selections.

\autoref{table::tiled_patch} summarizes the average accuracy of the models across eight downstream tasks in addition to their perplexity on the WikiText2 dataset for different sparsity ratios, illustrating that \patchtile delivers a comparable flexible sparsity-quality tradeoff when using a high-quality frozen 2:4 mask.

\begin{table*}[tb]
\centering
\small
\caption{Model quality (average accuracy across eight zero-shot tasks and perplexity on WikiText2 dataset) for different pruning methods. By only optimizing the location of dense tiles while keeping sparsity pattern within the sparse tiles frozen, \patchtile provides a memory efficient variant for \patchjoint,  allowing for a continuous sparsity ratio for the models and providing a flexible tradeoff between sparsity and model quality.}
\label{table::tiled_patch}
\setlength{\tabcolsep}{3pt}
\resizebox{0.8\textwidth}{!}{
\begin{tabular}{l l l c c c c c c}
\toprule
\textbf{Sparsity} & \textbf{Method} & \textbf{Pattern} &
\multicolumn{2}{c}{\textbf{LLaMA-2 7B}} &
\multicolumn{2}{c}{\textbf{LLaMA-3.1 8B}} \\
\cmidrule(lr){4-5} \cmidrule(lr){6-7}
& & & \textbf{Acc (\% $\uparrow$)} & \textbf{PPL ($\downarrow$)} & \textbf{Acc (\% $\uparrow$)} & \textbf{PPL ($\downarrow$)} \\
\midrule
0\%  & Dense     & -      & 54.61 & 5.12 & 60.31 & 5.84 \\
\midrule
50\% & Magnitude & 2:4    & 43.44 & 54.39 & 35.93 & 765.92 \\
     & Wanda     & 2:4    & 44.30 & 11.15 & 41.77 & 21.29 \\
     & SparseGPT & 2:4    & 45.09 & 10.12 & 45.53 & 15.11 \\
     & Thanos    & 2:4    & 44.80 & 11.19 & 45.72 & 16.09 \\
     & ProxSparse& 2:4    & 45.92 & 9.18 & 45.14 & 15.17 \\
     & MaskLLM   & 2:4    & 48.62 & 6.78  & 52.80 & 8.58  \\
\midrule
\rowcolor{PatchShade} 45\% & \patchtile     & Dense/2:4 Tiles & 48.99 & 6.55 & 53.60 & 8.20 \\
\rowcolor{PatchShade} 35\% & \patchtile     & Dense/2:4 Tiles & 50.08 & 6.18 & 55.28 & 7.89 \\
\rowcolor{PatchShade} 25\% & \patchtile     & Dense/2:4 Tiles & 51.58 & 5.86 & 56.48 & 7.34 \\
\bottomrule
\end{tabular}
}

\end{table*}

Overall, across \autoref{table::joint_patch} and \autoref{table::tiled_patch}, \patch consistently surpasses one-shot methods like Wanda, SparseGPT, and magnitude pruning due to its end-to-end training on large corpora. While MaskLLM also trains end-to-end on a large dataset, its fixed 2:4 sparsity ratio limits achievable accuracy and perplexity. In contrast, \patch overcomes this limitation with flexible dense tile allocation, achieving accuracy gains and perplexity reductions from 45\% to 25\% sparsity that progressively align with dense model performance. The full per-task accuracy results are provided in \autoref{app::per_task}.

\label{app::unstructured}
\paragraph{Comparison with unstructured sparsity}
In this section, we compare the quality of the models pruned with \patch against other unstructured sparsity methods. \autoref{table::unstructured} summarizes the average accuracy of the models across eight downstream tasks and the model perplexity on WikiText2 dataset. The results indicate that while unstructured sparsity consistently outperforms the hybrid sparsity, the gap between the two is not significant, showing that \patch is helping to bridge the gap between unstructured sparsity and semi-structured sparsity.

\begin{table*}[ht]
\centering
\caption{Model quality (average accuracy across eight zero-shot tasks and perplexity on WikiText2 dataset) for \patch, Wanda, and SparseGPT. For models with less than or equal to 1B parameters, \patchjoint optimizes both dense tile locations and sparsity patterns, while for larger models \patchtile optimizes only dense tile locations with frozen sparsity patterns, both using Dense/2:4 Tiles pattern allowing continuous sparsity ratios and flexible tradeoffs between sparsity and model quality. Wanda and SparseGPT are unstructured pruning methods.}
\label{table::unstructured}
\small
\setlength{\tabcolsep}{3pt}
\resizebox{\textwidth}{!}{
\begin{tabular}{l l l c c c c c c c c c c}
\toprule
\textbf{Sparsity} & \textbf{Method} & \textbf{Pattern} &
\multicolumn{2}{c}{\textbf{Qwen-2.5 0.5B}} &
\multicolumn{2}{c}{\textbf{LLaMA-3.2 1B}} &
\multicolumn{2}{c}{\textbf{Gemma-3 1B}} &
\multicolumn{2}{c}{\textbf{LLaMA-2 7B}} &
\multicolumn{2}{c}{\textbf{LLaMA-3.1 8B}} \\
\cmidrule(lr){4-5} \cmidrule(lr){6-7} \cmidrule(lr){8-9} \cmidrule(lr){10-11} \cmidrule(lr){12-13}
& & & \textbf{Acc (\% $\uparrow$)} & \textbf{PPL ($\downarrow$)} & \textbf{Acc (\% $\uparrow$)} & \textbf{PPL ($\downarrow$)} & \textbf{Acc (\% $\uparrow$)} & \textbf{PPL ($\downarrow$)} & \textbf{Acc (\% $\uparrow$)} & \textbf{PPL ($\downarrow$)} & \textbf{Acc (\% $\uparrow$)} & \textbf{PPL ($\downarrow$)} \\
\midrule
\rowcolor{PatchShade} 45\% & \patch & Dense/2:4 Tiles & 40.29 & 14.57 & 42.08 & 12.23 & 42.80 & 11.96 & 48.99 & 6.55 & 53.60 & 8.20 \\
45\% & Wanda & Unstructured & 41.45 & 18.81 & 40.76 & 16.56 & 42.87 & 25.38 & 52.72 & 6.36 & 55.67 & 8.24 \\
45\% & SparseGPT & Unstructured & 42.31 & 17.65 & 42.66 & 15.01 & 43.52 & 22.26 & 52.77 & 6.46 & 56.70 & 8.21 \\
\midrule
\rowcolor{PatchShade} 35\% & \patch & Dense/2:4 Tiles & 41.15 & 13.84 & 42.72 & 11.67 & 43.30 & 11.48 & 50.08 & 6.18 & 55.28 & 7.89 \\
35\% & Wanda & Unstructured & 43.46 & 15.04 & 44.60 & 11.95 & 45.50 & 16.98 & 54.37 & 5.87 & 58.68 & 7.02 \\
35\% & SparseGPT & Unstructured & 44.66 & 14.79 & 45.62 & 11.68 & 45.45 & 16.92 & 54.18 & 5.92 & 58.81 & 7.07 \\
\midrule
\rowcolor{PatchShade} 25\% & \patch & Dense/2:4 Tiles & 42.39 & 13.47 & 43.81 & 11.00 & 44.07 & 11.17 & 51.58 & 5.86 & 56.48 & 7.34 \\
25\% & Wanda & Unstructured & 45.70 & 13.70 & 46.50 & 10.46 & 46.56 & 15.14 & 54.60 & 5.65 & 59.80 & 6.54 \\
25\% & SparseGPT & Unstructured & 45.28 & 13.63 & 46.52 & 10.42 & 46.37 & 15.05 & 54.71 & 5.68 & 59.52 & 6.55 \\
\bottomrule
\end{tabular}
}

\end{table*}

\subsection{Understanding the components of \patch}

This subsection examines the design choices driving \patch's performance by analyzing its behavior across various configurations on the Qwen-2.5 0.5B model.

\paragraph{Tile size.} We initially assess the impact of tile size on PATCH's performance, fixing hyperparameters to those optimized for 128$\times$128 tiles. \autoref{ablation:tile_size} reveals that $4 \times 4$ tiles maximize model quality through finer sparse-dense control, though larger tile sizes show minimal variation, suggesting robustness. However, smaller tiles may hinder hardware efficiency, requiring a balance with hardware specifications.

\begin{table}[t]
\centering
\begin{minipage}{0.60\textwidth}
\centering
\centering
\small
\setlength{\tabcolsep}{6pt}
\caption{Impact of \patch's tile size across sparsity levels ($\downarrow$ is better). The effect of tile size on model quality is not significant, showing \patch's robustness against tile size.}
\label{ablation:tile_size}
\resizebox{\textwidth}{!}{
\begin{tabular}{l c c c c c c}
\toprule
\makecell{\textbf{Sparsity} \\ \textbf{(0.5B)}} & \textbf{128} & \textbf{64} & \textbf{32} & \textbf{16} & \textbf{8} & \textbf{4} \\
\midrule
45\% & 14.57 & 14.66 & 14.70 & 14.67 & 14.70 & \textbf{14.55} \\
35\% & 13.84 & 14.08 & 14.15 & 14.03 & 14.01 & \textbf{13.72} \\
25\% & 13.47 & 13.54 & 13.52 & 13.53 & 13.40 & \textbf{13.11} \\
\bottomrule
\end{tabular}
}
\end{minipage}
\begin{minipage}{0.35\textwidth}
\centering
\centering
\small
\setlength{\tabcolsep}{6pt}
\caption{Global sparsity yields better quality by concentrating pruning in less important blocks and preserving density elsewhere ($\downarrow$ is better).}
\label{ablation:global_vs_layer_sparsity}
\resizebox{\textwidth}{!}{
\begin{tabular}{l c c}
\toprule
\makecell{\textbf{Sparsity} \\ \textbf{(0.5B)}} & \textbf{Global} & \textbf{Layer-wise} \\
\midrule
45\% & \textbf{14.57} & 15.17 \\
35\% & \textbf{13.84} & 14.48 \\
25\% & \textbf{13.47} & 13.95 \\
\bottomrule
\end{tabular}
}

\end{minipage}
\end{table}

\paragraph{Joint vs. tile-only mask search.} 
We then analyze the impact of fixing the 2:4 masks and optimizing only tile masks. \autoref{ablation:tile_only} shows that among frozen 2:4 masks, MaskLLM provides the strongest results. On the other hand, one-shot pruning methods perform comparably at higher sparsity levels but diverge at lower sparsity, with SparseGPT emerging as the best overall. When comparing against our full approach, joint optimization of both tile and 2:4 masks consistently outperforms tile-only training across sparsity ratios. Nevertheless, tile-only training remains a practical alternative for larger models in resource-constrained settings, as also reflected in \autoref{table::tiled_patch}.

\begin{table}[t]
\centering
\setlength{\tabcolsep}{6pt}
\caption{Impact  of fixed 2:4 mask selection for \patchtile{}, compared with joint optimization ($\downarrow$ is better). \patchjoint{} achieves the lowest perplexity overall, while for \patchtile{}, MaskLLM provides the best frozen mask.}

\label{ablation:tile_only}
\begin{tabular}{l c c c c | c}
\toprule
\textbf{Sparsity (0.5B)} & \textbf{MaskLLM} & \makecell{\textbf{SparseGPT}\\(w/o weight update)} & \textbf{Wanda} & \textbf{Magnitude} & \textbf{\patchjoint} \\
\midrule
45\% & \textbf{15.06} & 21.84 & 21.83 & 21.33 & 14.57 \\
35\% & \textbf{14.55} & 17.29 & 17.96 & 19.90 & 13.84 \\
25\% & \textbf{14.17} & 14.89 & 15.09 & 16.05 & 13.47 \\
\bottomrule
\end{tabular}
\end{table}

\begin{figure}[tb]
\centering
\includegraphics[width=0.85\linewidth]{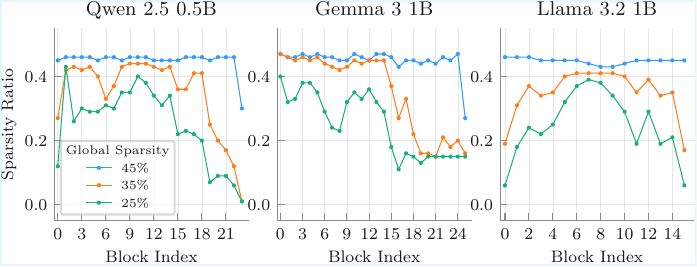}
\caption{Layer-wise sparsity allocation under different global sparsity budgets for various models. \patch achieves the target global sparsity while flexibly distributing pruning across transformer layers.}
\label{fig:sparsity_allocation}
\end{figure}

\paragraph{Sparsity allocation.} We analyze how sparsity is allocated across transformer blocks under a global target. Across models, deeper transformer blocks are pruned far less, while the initial blocks also tend to receive lighter pruning depending on the architecture. By contrast, the middle blocks consistently absorb most of the sparsity, suggesting that they contain more redundancy (\autoref{fig:sparsity_allocation}). We compare this flexible allocation to enforcing sparsity uniformly at the layer level. As shown in \autoref{ablation:global_vs_layer_sparsity}, global targets deliver better results by pruning more aggressively in redundant layers while preserving capacity in sensitive ones. In contrast, layer-wise targets impose uniform sparsity that can over-prune critical components \citep{adaptive1, adaptive2, adaptive3, owl}.

\begin{figure}[tb]
\centering
\includegraphics[width=1\linewidth]{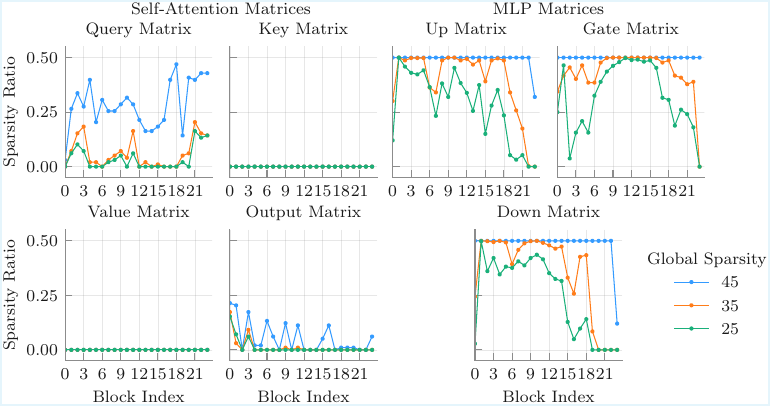}
\caption{Sparsity distribution across Attention and MLP layers under varying global sparsity budgets in Qwen-2.5 0.5B.}
\label{fig:linear_layer_sparsity_allocation_qwen}
\end{figure}

On top of variation across depth, sparsity is also distributed unevenly across the individual linear layers within each transformer block. \autoref{fig:linear_layer_sparsity_allocation_qwen} breaks down the allocation into the query, key, value, and output matrices of the attention module, as well as the up, gate, and down matrices of the MLP for the Qwen 2.5 0.5B model. The up, gate, and down layers absorb most of the sparsity and largely explain the overall allocation pattern seen in \autoref{fig:sparsity_allocation}. In contrast, the attention module is treated as more critical. The key and value matrices are never pruned, while the output matrix shows moderate pruning at higher global sparsity targets. The query matrix is pruned the most, suggesting it is the least important within the attention submodule.

\begin{figure}[htb]
\begin{center}
\includegraphics[width=\linewidth]{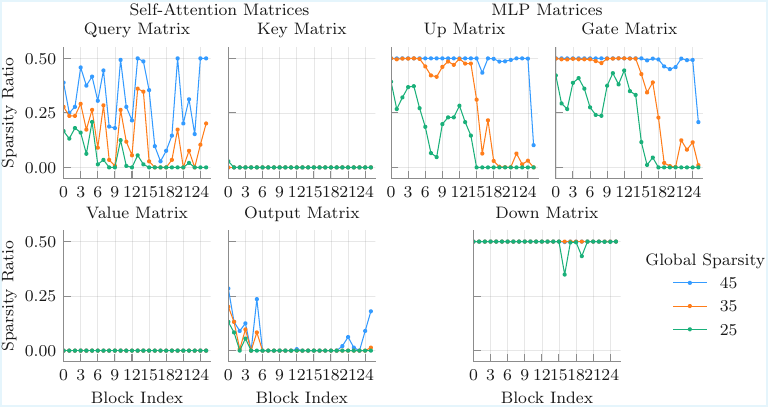}
\end{center}
\caption{Sparsity distribution across Attention and MLP layers under varying global sparsity budgets in Gemma-3 1B.}
\label{fig:linear_layer_sparsity_allocation_gemma}
\end{figure}

\begin{figure}[hbt]
\begin{center}
\includegraphics[width=\linewidth]{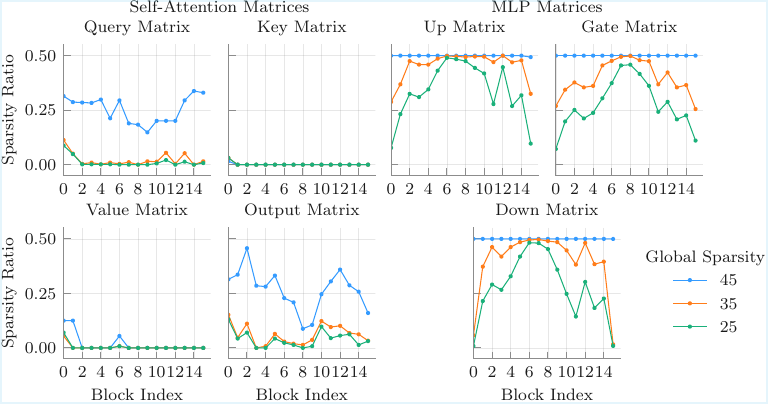}
\end{center}
\caption{Sparsity distribution across Attention and MLP layers under varying global sparsity budgets in LLaMA-3.2 1B.}
\label{fig:linear_layer_sparsity_allocation_llama}
\end{figure}

Additionally, we provide the sparsity distributions for the Gemma-3-1B (\autoref{fig:linear_layer_sparsity_allocation_gemma}) and Llama-3.2-1B (\autoref{fig:linear_layer_sparsity_allocation_llama}) models, as referenced in the main text. Similar to the Qwen-2.5 0.5B model, the patterns observed here indicate that MLP layers (up, gate, and down matrices) are pruned more aggressively, absorbing the majority of sparsity. In contrast, the self-attention layers are treated as more critical, with key and value matrices remaining largely dense or unpruned, while the query matrix experiences the highest pruning within the attention submodule, and the output matrix shows moderate pruning under higher global sparsity targets. This consistent behavior across models underscores the redundancy in MLP components and the sensitivity of attention mechanisms.

\subsection{Speedup and memory savings}

We evaluate the inference efficiency of the LLaMA-2 7B model pruned with \patch using the STOICC \citep{stoicc} compiler. With a batch size of 16 on an A6000 GPU, we observe end-to-end throughput improvements of 1.18$\times$, 1.27$\times$, and 1.38$\times$ at sparsity levels of 25\%, 35\%, and 45\%, respectively, compared to the dense baseline. At the same sparsity levels, the model’s GPU memory footprint during inference is also reduced, dropping to 0.76×, 0.68×, and 0.59× of the fully dense model, respectively. These results underscore the trade-off between accuracy retention and the computational savings enabled by sparsity.

\section{Conclusion and Limitations}
\label{sec:patch_conclusion}

In this chapter, we explored the upper limits of the sparsity pillar when a training budget is available. We introduced \patch, a hybrid sparsity framework that breaks the rigidity of the layer-wise static masks used in \autoref{chapter:optima}. By partitioning weight matrices into tiles designated as either dense or 2:4 sparse, \patch enables adaptive sparsity ratios between 0\% and 50\%, dynamically balancing accuracy in critical regions with hardware acceleration elsewhere. 

Experiments across models up to 8B parameters show that \patch consistently improves accuracy over state-of-the-art 2:4 pruning methods while achieving up to 1.38$\times$ end-to-end speedup on consumer-grade GPUs. These results demonstrate the promise of hybrid sparsity as a practical approach to efficient LLM inference and motivate future work on broader sparsity formats, integration with quantization, and co-design with hardware kernels.

While \patch offers superior accuracy-efficiency trade-offs, it is important to situate it within the broader "compute budget" narrative of this thesis. Unlike \optima, which targeted the zero-training regime, \patch requires a \textbf{fine-tuning budget}. The learnable masking process introduces computational overhead that makes it unsuitable for instant, on-device adaptation. However, as established in \autoref{sec:intro_contributions}, these two chapters represent complementary solutions for different deployment scenarios: \optima maximizes performance when training is impossible, whereas \patch maximizes performance when resources permit.

With \patch, we have now refined the sparsity pillar to its logical conclusion in both static and dynamic regimes. We have made it highly accurate (via hybrid masks) and physically fast (via tile-level kernels). Yet, we have applied this pillar in isolation. As noted in the "Sparsity Paradox" (\autoref{tab:sparsity_vs_quant}), even optimized sparsity eventually hits an accuracy wall that cannot be overcome by simply removing \textit{fewer} weights. To unlock the next frontier of performance, we must stop treating sparsity as a solo actor. In the next chapter, we present the culmination of this thesis: \slim. There, we integrate our optimized sparsity findings with aggressive quantization and low-rank approximations, finally realizing the full potential of the Compression Trinity in a unified, one-shot framework.

\newcommand{\quantization}{\textsc{SLiM}-Quant~}
\newcommand{\quantizationnospace}{\textsc{SLiM}-Quant}
\newcommand{\slimspace}{\textsc{SLiM}~}
\newcommand{\accuracygain}{5.66\%}
\newcommand{\accuracygainft}{1.66\%}
\newcommand{\rtxspeedup}{3.78$\times$}
\newcommand{\aspeedup}{3.75$\times$}

\newcommand{\ayazdan}[1]{\textcolor{red}{AY: #1}}

\chapter{\slim: One-shot Quantization and Sparsity with Low-rank Approximation for LLM Weight Compression}
\label{chapter:slim}

\paragraph{Publication and Contributions.}
The content of this chapter is based on the paper
``SLiM: One-shot Quantized Sparse Plus Low-rank
Approximation of LLMs'' \cite{slim}, published at Forty-Second International Conference on Machine Learning (ICML), 2025. This
work was conducted in collaboration with Amir
Yazdanbakhsh and Maryam Mehri Dehnavi. Mohammad
Mozaffari was the lead contributor, responsible for
the algorithm design, implementation, and experimental
evaluation. Amir Yazdanbakhsh and Maryam Mehri Dehnavi
supervised the project and contributed to the writing
and revision of the manuscript.

\section{Introduction}
\label{sec:slim_intro}

In the preceding chapters, we pushed the \textbf{Sparsity} pillar to its limits. With \optima (\autoref{chapter:optima}), we established the mathematical upper bound for reconstruction with layer-wise pruning and weight update, and with \patch (\autoref{chapter:patch}), we broke the rigidity of those masks using end-to-end learnable hybrid patterns. However, as identified in the "Sparsity Paradox" (\autoref{sec:intro_failure}), sparsity is the most destructive pillar. Even with the optimized skeletons provided by \optima and \patch, a performance gap remains because we are removing information that cannot be fully recovered by weight adjustment alone. Additionally, with a perfected sparse model, relying on a single compression technique limits the potential for efficiency. To bridge this gap and fully conquer the memory bandwidth bottleneck, we must stop treating sparsity in isolation. We must integrate it with the remaining two pillars of the Compression Trinity: \textbf{Quantization} and \textbf{Low-Rank Approximation}.

This chapter presents the culmination of this thesis: \slim, a unified framework that jointly applies hardware-friendly sparsity, quantization, and low-rank approximations in a single one-shot step. Bringing these three pillars together presents a formidable challenge: \textbf{compounded error}. When aggressive sparsity (removing weights) meets aggressive quantization (reducing precision), the errors do not merely add up; they exacerbate each other, leading to a catastrophic drop in model capability~\citep{sparsegpt,omniquant, affinequant, awq, jsq}. Traditional recovery methods rely on costly retraining (e.g., Quantization-Aware Training~\citep{movement_pruning, quantization_aware_training}), which violates the "resource-constrained" principles we explored in \autoref{chapter:optima}. Conversely, existing one-shot methods like SparseGPT~\citep{sparsegpt} struggle to combine structured patterns (like 2:4 sparsity) with low-bit quantization, failing to arrest the accuracy slide.\footnote{For a more detailed discussion of the related work, see~\autoref{sec:slim_related_work}.}

To resolve these limitations and realize the full Trinity without retraining, we propose \slim. We decompose the problem into three synchronized sub-tasks, ensuring that each pillar supports the others rather than conflicting with them.
\begin{enumerate}
    \item \textbf{Quantization:} We prioritize uniform quantization for hardware efficiency. While \slim is compatible with any standard quantization kernel (e.g., Group MinMax or AbsMax), we introduce \quantizationnospace, a probabilistic and tractable reformulation that finds the optimal quantization parameters. This allows users to either leverage existing quantization standards or utilize our optimizer to minimize the error floor before sparsity is even applied.

    \item \textbf{Sparsity:} We apply hardware-friendly pruning to the quantized weights to create the efficient sparse structure. Crucially, the \slim framework is agnostic to the specific mask selection algorithm. This allows us to seamlessly integrate masks generated by Wanda~\citep{wanda}, \optima, \patch, or future state-of-the-art selectors, ensuring the framework remains relevant as pruning metrics evolve.

    \item \textbf{Low-Rank Approximation:} Finally, we deploy the third pillar not just for compression, but as a \textbf{mathematically derived error-correction mechanism}. This is the critical innovation that closes the accuracy gap left by \autoref{chapter:optima} and \autoref{chapter:patch}. We propose \slim-LoRA, a one-shot low-rank adaptation method designed to compensate for the aggregated error. Unlike standard adapters that require iterative training to find optimal values~\citep{lora, rosa, qlora, lqlora, losparse}, we develop a saliency function that is both invertible and additive. These properties enable us to \emph{analytically compute} the optimal low-rank adapter values that minimize the compression-induced error in one shot, eliminating the need for any retraining overhead.
\end{enumerate}

By solving the compounded error problem through this joint formulation, \slim shifts the Pareto frontier of efficiency. It achieves what neither \optima nor \patch could do alone: recovering close to dense model accuracy while maintaining high compression rates. Compared to state-of-the-art methods, \slimspace achieves an average accuracy improvement of 5.66\% on LLaMA-2-7B under 2:4 sparsity and 4-bit quantization. Uniquely, it delivers higher model accuracy at the \emph{same total bit budget} compared to existing techniques (up to 0.5\%) and even outperforms uncompressed dense models at equal parameter budgets (up to 0.6\%). Beyond accuracy, \slimspace demonstrates the practical value of the Trinity, achieving up to \rtxspeedup~ and \aspeedup~ layer-wise speedup on NVIDIA RTX3060 and A100 GPUs, respectively. For cases requiring maximal performance, we also support an optional lightweight PEFT method, providing up to an \accuracygainft~ additional accuracy improvement.

\section{Related work}
\label{sec:slim_related_work}

\slimspace combines model pruning and quantization for compression, complemented by zero-shot low-rank adapters to recover lost accuracy. This section reviews related work on these topics.

\subsection{Pruning}

Eliminating redundant weights reduces computation and memory costs during inference. Optimal Brain Damage (OBD) \citep{optimal_brain_damage} leverages second-order information of the loss function to identify the least important weights but is computationally prohibitive for large language models (LLMs) \citep{mkor}. WoodFisher \citep{woodfisher} approximates the Hessian matrix using Kronecker Factorization to mitigate this overhead but struggles to scale to LLMs.

Optimal Brain Surgeon (OBS) \citep{obs} evaluates weight matrices layer-wise using the layer-wise Hessian matrix to preserve layer outputs. However, the cubic growth in the cost of inverting the layer-wise Hessian with model size renders this approach impractical for LLMs. Optimal Brain Compression (OBC) \citep{optimal_brain_compression} addresses the OBS-defined compression problem using a greedy algorithm, while SparseGPT reformulates it as a sparse regression problem. Wanda introduces a lightweight method based on weight and activation magnitudes to identify unimportant weights without updating their values.

\subsection{Quantization}

Quantizing all elements in a matrix is challenging due to the significant impact of outliers on the model \citep{int8}. Group quantization \citep{group_quantization_1, group_quantization_2} addresses this by quantizing small groups of a weight matrix with a shared quantization parameter, but it introduces challenges discussed in \autoref{appendix:group_quantization_challenges}.

AbsMax \citep{absmax} with round-to-nearest (RTN) is the simplest quantization scheme for matrix elements. OPTQ \citep{optq} minimizes layer-wise error using an approach akin to OBS. AWQ \citep{awq} shifts the challenge of quantizing salient weights to activations, while SmoothQuant \citep{smoothquant} balances quantization error between weights and activations, enabling input quantization. OmniQuant \citep{omniquant} improves accuracy with learnable clipping and channel scaling. AffineQuant leverages equivalent affine transformations to reduce quantization error, and QuaRot \citep{quarot} uses rotations to eliminate outliers during quantization.

Advanced methods like JSQ \citep{jsq} jointly prune and quantize weights to 8 bits but struggle to recover accuracy in low bit-width quantization, limiting their utility.

\subsection{Low-rank Adapters}

Low-rank adapters were first introduced to LLMs to reduce the overhead of fine-tuning \citep{lora, slope}. Q-LoRA \citep{qlora} extended this approach by quantizing weights before fine-tuning, allowing the process to recover accuracy lost during quantization. LQ-LoRA \citep{lqlora} further improved Q-LoRA by initializing the adapters using the SVD of the quantization error. LoSparse \citep{losparse} has a similar approach as LQ-LoRA, but for sparsity, initializing the low-rank adapters to the norm of the pruning error. RoSA \citep{rosa} expands the learning capability of the model by adding both low-rank and sparse adapters to the model. This approach adds an extra sparse matrix multiplication to the inference, increasing the adapter overhead even further. However, all these methods require hundreds of millions of tokens for fine-tuning, making them costly and not comparable to one-shot pruning and quantization methods, or methods that use much shorter fine-tuning phases.

L$^2$QER \citep{lqer} avoids fine-tuning by using one-shot low-rank adapters to mitigate quantization error. However, it performs poorly when combined with sparsity, resulting in a significant accuracy gap between the compressed and dense models.

\subsection{Sparse Plus Low-Rank Matrix Decomposition}
\label{sec:rpca-related}

The decomposition of a matrix into the sum of a sparse component and a
low-rank component is a classical problem in signal processing and
optimization, most prominently studied under the framework of Robust
Principal Component Analysis (RPCA)~\citep{candes2011robust}. Given an
observed matrix $M$, RPCA seeks to recover $M = L_0 + S_0$, where $L_0$
is low-rank and $S_0$ is sparse, by solving the convex program known as
Principal Component Pursuit (PCP):
\begin{equation}
  \min_{L, S} \; \|L\|_* + \lambda \|S\|_1 \quad \text{subject to} \quad L + S = M,
\end{equation}
where $\|\cdot\|_*$ denotes the nuclear norm and $\lambda$ is a
regularization parameter. Cand\`{e}s et
al.~\citep{candes2011robust} proved that under mild incoherence
conditions on the low-rank component, this convex relaxation exactly
recovers both $L_0$ and $S_0$ even when the sparse errors are
arbitrarily large in magnitude. Chandrasekaran et
al.~\citep{chandrasekaran2011rank} provided complementary recovery
guarantees through rank-sparsity incoherence conditions, while Zhou et
al.~\citep{zhou2010stable} extended the theory to the noisy setting
(Stable PCP), showing that the decomposition remains robust when the
observation also contains a small dense noise term, i.e.,
$M = L_0 + S_0 + N_0$.

The structural parallel between RPCA and model compression was first
explored in the context of deep convolutional networks by Yu et
al.~\citep{yu2017compressing}, who decomposed weight matrices into
sparse-plus-low-rank form using greedy bilateral decomposition and
showed that this representation achieves better accuracy-compression
trade-offs than either pure pruning or pure low-rank factorization in
isolation. More recently, the connection to LLM compression has been
made explicit. OATS~\citep{zhang2025oats} formulates post-training weight
compression as a sparse-plus-low-rank decomposition problem, scaling the
weights by the second moment of input embeddings before decomposition to
preserve outlier features. HASSLE-free~\citep{makni2025hassle}
established a unified framework showing that several existing LLM
pruning methods, including Wanda and Magnitude Pruning, can be viewed as
special cases of an alternating minimization procedure for the
sparse-plus-low-rank objective. Concurrently,
3BASiL~\citep{behdin2025basil} proposed a three-block ADMM formulation
that jointly optimizes the sparse and low-rank components, offering
improved convergence guarantees over alternating minimization.

\slim's formulation $W \approx W^C + LR$, where $W^C$ is the sparse (and
quantized) component and $LR$ is the low-rank correction, can be viewed
through the lens of this decomposition tradition. However, \slim differs
from standard RPCA approaches in several important ways. First, the
sparse component in \slim is additionally quantized, introducing a
structured noise that is absent in classical RPCA. Second, rather than
solving for $S$ and $L$ jointly via convex optimization, \slim employs a
sequential pipeline (quantize, then sparsify, then compute the low-rank
correction), which enables the use of off-the-shelf pruning and
quantization algorithms but forgoes the joint optimality guarantees of
PCP. Third, \slim's saliency-weighted SVD (\autoref{eq:lora_svd})
introduces input statistics into the decomposition, a data-dependent
weighting that has no direct analogue in standard RPCA but shares the
spirit of the outlier-aware scaling in OATS~\citep{zhang2025oats}.
Bertsimas et al.~\citep{sparse_plus_low_rank} further studied the
sparse-plus-low-rank decomposition from a discrete optimization
perspective, providing mixed-integer programming formulations that could,
in principle, be adapted to the compression setting.

\section{Preliminaries}
\label{sec:preliminaries}

\textbf{Model Compression.}
Model compression reduces the compute and memory demands of large models while maintaining predictive accuracy by minimizing output differences between compressed and original models. However, directly optimizing these differences across the entire model is computationally infeasible due to the high dimensionality of neural networks. Optimal Brain Surgeon (OBS) \citep{obs} simplifies this challenge by focusing on minimizing output discrepancies layer by layer, using calibration datasets.

OBS applies a layer-wise approach to compress feed-forward layers efficiently. Denoting compressed matrices with a superscript $C$, for a layer with input $\mathcal{X} \in \mathbb{R}^{b\times d_{in}}$, weight $\mathcal{W} \in \mathbb{R}^{d_{in}\times d_{out}}$, and output $\mathcal{Y} \in \mathbb{R}^{b\times d_{out}}$, it minimizes output differences by optimizing \autoref{eq:obs}. This method ensures compression fidelity and has become foundational for many modern compression techniques.

\begin{equation}
    \label{eq:obs}
    \min_{\mathcal{W}^C}{|\mathcal{Y}^C - \mathcal{Y}|^2} = \min_{\mathcal{W}^C}{|\mathcal{X}(\mathcal{W}^C - \mathcal{W})|^2}
\end{equation}

\textbf{Symmetric Quantization.}
Symmetric quantization is a core technique for reducing model size and boosting computational efficiency. It computes the quantized matrix $\mathcal{M}^Q \propto round(\frac{\mathcal{M}}{\alpha})$, where $\alpha$ is a scaling factor based on the range or norm of the matrix. This scaling ensures $\mathcal{M}^Q$ values stay within the representable range, enabling efficient matrix multiplications with minimal overhead. However, its effectiveness depends on selecting $\alpha$ carefully, as this choice significantly impacts precision.

AbsMax, the most common symmetric quantization method, selects $\alpha$ as the matrix's maximum absolute value, ensuring all values remain within the target range. Unfortunately, it is highly sensitive to outliers; a single large value can inflate $\alpha$, reducing the precision of most quantized weights. For zero-centered, bell-curved distributions typical in LLMs, AbsMax maps many weights to zero, leading to significant quantization errors.

Group quantization \cite{group_quantization_1, group_quantization_2} tackles AbsMax's outlier sensitivity by assigning separate scaling factors to subgroups of the weight matrix. This approach captures local variations in weight magnitudes, reducing quantization error for non-uniform distributions. However, storing multiple scaling factors increases memory usage, and subgroup-specific dequantization increases computational complexity, potentially slowing inference. The challenges of using group quantization are discussed in \autoref{appendix:group_quantization_challenges}.

\section{Quantized sparse plus low-rank approximation of LLMs}

\begin{figure}
    \centering
    \includegraphics[width=1\textwidth]{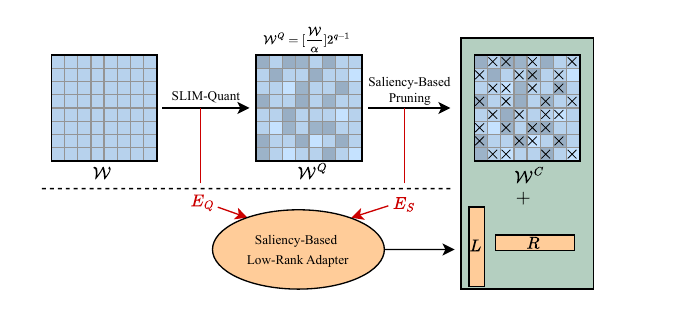}
    \caption{The \slimspace weight compression pipeline consists of three main steps: (1) Quantizing weights using the symmetric \quantization algorithm, producing quantized weights $\mathcal{W^Q}$ and quantization error $E_Q$; (2) Sparsifying quantized weights $\mathcal{W^Q}$ through a pruning method, resulting in compressed weights $\mathcal{W^C}$ and sparsity error $E_S$; (3) Mitigating compression errors through \slimspace saliency-based low-rank approximation, generating left and right low-rank adapters $L$ and $R$. Optionally, these adapters can be fine-tuned with sparse quantized weights frozen to further enhance model accuracy.}
    \label{fig:slim_pipeline}
\end{figure}

To achieve effective compression of LLMs while preserving accuracy, \slimspace combines quantization, pruning, and saliency-based low-rank adapters into an integrated pipeline.
First, \slimspace applies \quantization, a novel scheme designed to minimize quantization error, laying the foundation for subsequent pruning using methods such as Wanda \citep{wanda}. Finally, low-rank adapters are introduced to reduce the impact of compression errors from both quantization and pruning, ensuring minimal accuracy loss. The overall process is illustrated in \autoref{fig:slim_pipeline}, providing a visual summary of how these components interact to achieve effective model compression. In the following sections, we dive into the details of each step, highlighting the innovations and contributions of \quantization, the pruning strategy, and the saliency-based low-rank adapters.

\subsection{\quantization quantization method}

\slimspace adopts symmetric weight quantization due to its low dequantization and memory overhead and ease of implementation. Denoting the quantized matrices by $Q$ superscript, \autoref{eq:symmetric_quantization} shows the symmetric quantization formula for $q$-bit quantization, where $\alpha$ is the quantization scaling parameter and $clip(.)$ operator clips the input to values between $[-1, 1]$.

\begin{equation}
    \label{eq:symmetric_quantization}
    \mathcal{W}^Q = round(clip(\frac{\mathcal{W}}{\alpha})) 2^{q - 1}
\end{equation}

The objective of quantization is to reduce the weight reconstruction error shown in \autoref{eq:symmetric_quantization_objective}, where the $*$ superscript shows the optimal value. But the objective function in \autoref{eq:symmetric_quantization_objective} is not convex, and to the best of our knowledge, does not have a closed form solution.

\begin{equation}
    \label{eq:symmetric_quantization_objective}
    {\alpha}^* = \arg \min_{\alpha}{||\mathcal{W}^Q - \mathcal{W}||^2} = \arg \min_{\alpha}{||round(clip(\frac{\mathcal{W}}{\alpha})) 2^{q - 1} - \mathcal{W}||^2}
\end{equation}

To solve the mean squared error (MSE) problem in \autoref{eq:symmetric_quantization_objective}, we propose a probabilistic reformulation as shown in \autoref{eq:probabilistic_mse}, where $Q(.)$ and $Q^{-1}(.)$ are the quantization and dequantization functions respectively, and $f(.)$ is the probability distribution function (PDF) of the weight elements.

\begin{equation}
    \label{eq:probabilistic_mse}
    \alpha^* = \arg \min_{\alpha}{E_Q} = \arg \min_{\alpha}{||\mathcal{W}^Q - \mathcal{W}||^2}  = \arg \min_{\alpha}{\int_{-\infty}^{\infty}{f(x)|Q^{-1}(Q(x)) - x|^2 dx}}
\end{equation}

By incorporating the quantization formula from \autoref{eq:symmetric_quantization} into \autoref{eq:probabilistic_mse}, we can simplify the integration into the sum of two terms based on the absolute value of the data: the quantization error for absolute values less than $\alpha$ (\autoref{eq:quantization_error}) and the clipping error for absolute values larger than $\alpha$ (\autoref{eq:clipping_error}). Here, $f_{abs}(.)$ represents the probability density function (PDF) of the absolute value of the weights. \autoref{eq:error_summation} presents the simplified version of \autoref{eq:probabilistic_mse}.

\begin{equation}
    \label{eq:quantization_error}
    E_{quant}(\alpha) = \int_0^{\alpha}{f_{abs}(x)|\alpha \times round(\frac{x}{\alpha}) \times 2^{1-q} - x|^2 dx}
\end{equation}

\vspace{-10pt}

\begin{equation}
    \label{eq:clipping_error}
    E_{clip}(\alpha) = \int_{\alpha}^{\infty}{f_{abs}(x)|\alpha - x|^2 dx}
\end{equation}

\vspace{-10pt}

\begin{equation}
    \label{eq:error_summation}
    \alpha^* = \arg \min_{\alpha} E_Q(\alpha) = \arg \min_{\alpha}{E_{quant}(\alpha) + E_{clip}(\alpha)}
\end{equation}

\autoref{eq:error_summation} can be solved theoretically by differentiating the objective function with respect to $\alpha$, provided the probability density function (PDF) of the weight distribution is known. However, the weight distribution of neural networks rarely conforms to standard PDFs. To verify this, we tested various candidate distributions, including Gaussian, Laplace, Pareto, q-Gaussian, and Weibull, as they are commonly used in modeling natural data. Unfortunately, none of these matched the observed weight distributions accurately. This discrepancy underscores the need for a more adaptable method, motivating the data-driven approach we adopt in \quantization.

To address the absence of a closed-form weight PDF, we employ numerical integration on the weight histogram to solve \autoref{eq:error_summation}. To enhance efficiency, we adopt a multi-grid strategy: starting with 10 uniform samples in the range $(0, \max(W))$, the grid is iteratively refined around the region of minimum error. This iterative process converges to the optimal $\alpha$ with minimal computational overhead. The full procedure is detailed in \autoref{alg:quantization}.

\begin{algorithm}[tb]
   \caption{\quantization Algorithm}
   \label{alg:quantization}

    \begin{algorithmic}[1]
    \State \textbf{\color{KeywordColor}Input:} Weight magnitude PDF $\color{VarColor}f_{abs}$, high resolution step size $\color{VarColor}\eta_{high}$, low resolution step size $\color{VarColor}\eta_{low}$, \\ weight matrix $\color{VarColor}\mathcal{W}$, quantization bitwidth $\color{VarColor}q$.
    \State \textbf{\color{KeywordColor}Output:} Quantized weight matrix $\color{VarColor}\mathcal{W}_{quant}$.
    
    \vspace{0.5em}
    \Function{EstimateError($\alpha$)}
        \State $E_{quant}(\alpha) = \int_0^{\alpha}{f_{abs}(x) |\alpha \times \text{round}(\frac{x}{\alpha}) \times 2^{1-q} - x|^2 dx}$
        \State $E_{clip}(\alpha) = \int_{\alpha}^{\infty}{f_{abs}(x)|\alpha - x|^2 dx}$
        \State \textbf{return} $E_{quant} + E_{clip}$
    \EndFunction
    
    \vspace{0.3em}
    \State $\color{VarColor}E \gets $ \texttt{EmptyDictionary}() \algorithmiccomment{\normalsize{Initialize error dictionary}}
    
    \vspace{0.2em}
    \For{\textbf{for} $\color{VarColor}\alpha$ \textbf{in} $\texttt{range}(0, M, \color{VarColor}{\eta_{low}})$}
        \State ${\color{VarColor}E(\alpha) \gets} \texttt{EstimateError}({\color{VarColor}\alpha})$
    \EndFor
    \State ${\color{VarColor}\alpha_{low} \gets} \arg\min_{{\color{VarColor}\alpha}} {\color{VarColor}E(\alpha)}$
    
    \vspace{0.2em}
    \For{\textbf{for} $\color{VarColor}\alpha$ \textbf{in} $\texttt{range}({\color{VarColor}\alpha_{low}} - {\color{VarColor}\eta_{low}}, {\color{VarColor}\alpha_{low}} + {\color{VarColor}\eta_{low}, \color{VarColor}\eta_{high}})$}
        \State ${\color{VarColor}E(\alpha) \gets} \texttt{EstimateError}({\color{VarColor}\alpha})$
    \EndFor
    \State ${\color{VarColor}\alpha^* \gets} \arg\min_{{\color{VarColor}\alpha}} {\color{VarColor}E(\alpha)}$
    \State ${\color{VarColor}\mathcal{W}_{quant} \gets} \text{round}(\text{clip}({\color{VarColor}\frac{\mathcal{W}}{\alpha^*}})) \times {\color{VarColor}2^{q - 1}}$ \algorithmiccomment{\normalsize{Apply optimal quantization}}
    
    \vspace{0.5em}
    \State \textbf{\color{KeywordColor}Return:} $\color{VarColor}\mathcal{W}_{quant}$.
    \end{algorithmic}

\end{algorithm}

\subsection{\slim-LoRA low-rank adapters}

The use of a low-rank adapter to compensate for compression errors can be
situated within the broader framework of sparse-plus-low-rank matrix
decomposition, a problem studied extensively in the Robust PCA
literature~\citep{candes2011robust, chandrasekaran2011rank}. In this
classical formulation, a matrix is expressed as the sum of a sparse
component and a low-rank component, with provable recovery guarantees
under suitable incoherence conditions. Yu et
al.~\citep{yu2017compressing} applied this decomposition paradigm to
compress deep neural network weights, and recent works such as
OATS~\citep{zhang2025oats} and HASSLE-free~\citep{makni2025hassle} have
extended it to LLMs. \slim adapts this principle to the joint compression
setting: rather than solving a single convex program, we derive the
low-rank component analytically from the compression error using a
saliency-weighted SVD, which enables a one-shot solution without
iterative optimization. We detail this procedure below.

After quantizing the model using \quantization, we sparsify it using an off-the-shelf one-shot pruning method such as Wanda. The combined effects of quantization and pruning of a weight matrix can be modeled as additive noise, such that $\mathcal{W}^C = \mathcal{W} + E_Q + E_S$, where $E_Q = \mathcal{W} - \mathcal{W}^Q$ and $E_S = \mathcal{W}^C - \mathcal{W}^Q$ are the quantization and sparsity errors respectively. To mitigate these errors, we introduce low-rank adapters that adjust the compressed weights such that $\mathcal{W} \approx \mathcal{W}^C + \mathcal{L R}$, where $\mathcal{L} \in \mathbb{R}^{d_{in} \times r}$ and $\mathcal{R} \in \mathbb{R}^{r \times d_{out}}$ are the low-rank adapters and $r$ is the adapter rank. 

A straightforward approach minimizes the total error norm between $\mathcal{W}$ and $\mathcal{W}^C$, focusing solely on reducing the error magnitude while ignoring the saliency of individual elements in the weight matrix. We call this method \textbf{Naive-LoRA} \textit{as it overlooks the significance of individual elements in the weight matrix}. However, this method is suboptimal and can be substantially improved.

To address the limitations of Naive-LoRA, we propose a novel low-rank approximation formulation that integrates weight saliency and uses a carefully designed saliency function to determine optimal adapters. The saliency function ($F$) in our formulation needs to satisfy two key properties. First, it needs to be \underline{invertible}, enabling the retrieval of low-rank adapters from their saliency. Second, it must be \underline{additive}, meaning $\forall A, B: F(A + B) = F(A) + F(B)$. The additive property is crucial for isolating the saliency of low-rank adapters from the compressed matrix and distinguishing the saliency of the error from that of the original weights. These properties ensure that the saliency function can effectively isolate and optimize the contribution of low-rank adapters, forming the foundation of our proposed formulation.

Assuming that there exists an additive invertible saliency function $F: {\mathbb{R}}^{d_{in} \times d_{out}} \rightarrow {\mathbb{R}}^{d_{in} \times d_{out}}$, we need to solve \autoref{eq:lora_optimization_problem} to find the optimal adapters. By using the additive property of the saliency function $F(.)$, we can simplify \autoref{eq:lora_optimization_problem} to \autoref{eq:lora_final_objective}.

\vspace{-10pt}

\begin{equation}
    \label{eq:lora_optimization_problem}
    \mathcal{L, R} = \arg \max_{\mathcal{L, R}}{||F(\mathcal{W}^C + \mathcal{LR})||^2} = \arg \min_{\mathcal{L, R}}{||F(\mathcal{W} - (\mathcal{W}^C + \mathcal{LR}))||^2}
\end{equation}

\vspace{-10pt}

\begin{equation}
    \label{eq:lora_final_objective}
     \mathcal{L, R} = \arg \min_{\mathcal{L, R}}{||F(\mathcal{W} - \mathcal{W}^C) - F(\mathcal{LR})||^2}  = \arg \min_{\mathcal{L, R}}{||F(-(E_Q +E_S)) - F(\mathcal{LR})||^2}
\end{equation}

Now, we can find $F(\mathcal{LR})$ by computing the SVD of $F(-(E_Q + E_S))$, and using the invertibility property of $F$, we can obtain the exact value of $\mathcal{L}$ and $\mathcal{R}$.

The saliency function used in \slimspace must satisfy three essential criteria—invertibility, additivity, and the effective utilization of input and weight statistics—to optimize weight importance during compression. Recent works such as Wanda, AWQ, LLM.int8(), and L$^2$QER suggest that the product of the magnitude of the weights and activations is a useful metric for identifying important weights during pruning and quantization. Motivated by this observation, we propose a saliency function formulation for $F$ that meets these criteria and leverages weight-activation interactions for effective compression.

To incorporate input statistics into the saliency function, we define $F(\mathcal{W}) \triangleq diag(\mathbf{x}) \mathcal{W}$, where  $\mathbf{x} \in \mathbb{R}^{d_{in}}$ represents the average absolute value of inputs from a calibration set. This formulation ensures that the saliency function effectively weights the matrix elements based on their significance during compression, facilitating a more accurate approximation. By replacing $F(\mathcal{W})$ in \autoref{eq:lora_final_objective}, the optimization problem transforms into a computationally efficient solution using singular value decomposition, followed by an inverse saliency transformation to derive the left low-rank adapter (\autoref{eq:lora_svd}).

\vspace{-10pt}

\begin{equation}
    \label{eq:lora_wanda_objective_function}
    \mathcal{L, R} = \arg \min_{\mathcal{L, R}}{||-diag(\mathbf{x}) (E_Q + E_S) - diag(\mathbf{x}) \mathcal{LR}||^2}
\end{equation}

\vspace{-10pt}

\begin{equation}
    \label{eq:lora_svd}
    diag(\mathbf{x})\mathcal{L}, \mathcal{R} =  -SVD(diag(\mathbf{x})(E_Q + E_S))
\end{equation}

We refer to this method of computing saliency-based low-rank adapters as \textbf{\slim-LoRA}, a practical and efficient approach tailored for addressing compression errors in large language models. To ensure numerical stability and guarantee the invertibility of the saliency function, an identity matrix with small values can be added to $diag(\mathbf{x})$. This adjustment is equivalent to uniformly shifting all elements of $\mathbf{x}$ and ensures that the saliency function remains robust even when $\mathbf{x}$ contains near-zero elements. \autoref{alg:lora} provides a comprehensive overview of the steps involved in computing saliency-based low-rank adapters using \slim-LoRA, ensuring reproducibility and clarity.

\begin{algorithm}[tb]
   \caption{\slim-LoRA Saliency-based Low-rank Adapter Computation}
   \label{alg:lora}
\begin{algorithmic}[1]
\Statex \textbf{\color{KeywordColor}Input:} Original weight $\color{VarColor}\mathcal{W}$, compressed weight $\color{VarColor}\mathcal{W}^C$, calibration input $\color{VarColor}\mathcal{X}$.
\Statex \textbf{\color{KeywordColor}Output:} Saliency-based low-rank adapters $\color{VarColor}\mathcal{L}, \color{VarColor}\mathcal{R}$.

\vspace{0.5em}
\State $\color{VarColor}E_C \gets E_Q + E_S = \mathcal{W}^C - \mathcal{W}$ \algorithmiccomment{\normalsize{Compute error}}
\State ${\color{VarColor}\tilde{\mathbf{x}}} \gets \texttt{mean}({\color{VarColor}\mathcal{X}})$ \algorithmiccomment{\normalsize{Average over all the samples}}
\State ${\color{VarColor}\mathbf{x} \gets \tilde{\mathbf{x}} +} \texttt{min}({\color{VarColor}|\tilde{\mathbf{x}}|})$ \algorithmiccomment{\normalsize{Shift values to avoid zeros in $\mathbf{x}$}}
\State ${\color{VarColor}\mathcal{S}_C \gets} \texttt{diag}({\color{VarColor}\mathbf{x}}) {\color{VarColor}E_C}$ \algorithmiccomment{\normalsize{Compute error saliency}}
\State ${\color{VarColor}\tilde{\mathcal{L}}, \tilde{\mathcal{R}} \gets} \texttt{SVD}({\color{VarColor}\mathcal{S}_C})$ \algorithmiccomment{\normalsize{Low-rank approximation}}
\State ${\color{VarColor}\mathcal{L} \gets} \texttt{diag}({\color{VarColor}1 / \mathbf{x}}) {\color{VarColor}\tilde{\mathcal{L}}}$ \algorithmiccomment{\normalsize{Converting saliency to weight}}
\State $\color{VarColor}\mathcal{R} \gets \tilde{\mathcal{R}}$

\vspace{0.5em}
\Statex \textbf{\color{KeywordColor}Return:} $\color{VarColor}\mathcal{L}, \color{VarColor}\mathcal{R}$.
\end{algorithmic}
\end{algorithm}

\subsection{Low-rank adapter quantization}
\label{sec:lora_quantization}

While pruning and quantizing the weights significantly reduce the model's computation and memory requirements (\(\sim\)8$\times$ memory footprint reduction), incorporating full-precision low-rank adapters reintroduces overhead, partially offsetting these gains. To address this, we apply 4-bit quantization to compress the adapters. This step ensures that the compression efficiency achieved through weight pruning and quantization is preserved, while maintaining the performance benefits of the low-rank adapters.

Quantizing low-rank adapters poses unique challenges due to the long-tailed distribution of their elements, which limits the effectiveness of advanced non-group quantization methods, such as \quantizationnospace. To address this, we adopt an AbsMax group quantization scheme for the adapters, where groups of 128 elements share the same quantization parameter. By grouping elements, this method effectively captures the distribution's variability while minimizing quantization error, striking a balance between accuracy and compression. This approach not only reduces the adapter overhead by $4\times$ but ensures that their contribution to overall model compression and performance is retained; as demonstrated in our experimental evaluation.

\subsection{Optional Post-Compression Fine-Tuning}

Fine-tuning large language models post-compression has many challenges because the high parameter count and memory demands of traditional methods make them computationally prohibitive. For example, using a simple optimizer such as ADAMW leads to $4\times$ additional memory overhead to store gradient and optimizer states, rendering these approaches impractical for compressed models.
Thus, parameter-efficient fine-tuning is essential for preserving the benefits of compression while avoiding excessive computational and memory costs. This necessity is further highlighted by the results in \autoref{sec:results}, which illustrate the overheads of traditional fine-tuning and the advantages of parameter-efficient alternatives.

To overcome the challenges of fine-tuning compressed models, \slimspace employs parameter-efficient low-rank adapters as the only tunable components during the fine-tuning phase. During this optional phase, \slimspace freezes the sparse and quantized weights, enabling focused fine-tuning solely on the adapters. If the adapters are quantized, \slimspace uses a straight-through estimator (STE) for quantization-aware fine-tuning and reduces its overheads with custom quantization and dequantization kernels implemented in Triton. This parameter-efficient fine-tuning method allows rapid accuracy improvements for the compressed model, requiring only a short fine-tuning phase over thousands of tokens. By limiting the fine-tuning process to a small subset of parameters, \slimspace significantly reduces computational requirements while ensuring the model can adapt effectively to new data or tasks. This approach maintains the benefits of compression while enabling efficient adaptation, as demonstrated by the significant improvements achieved during fine-tuning.

\section{Experimental results}
\label{sec:results}

\niparagraph{Models, Datasets, and Evaluation.} We evaluate \slimspace on the OPT \citep{opt} and LLaMA-2 \citep{llama2} model families, both of which serve as standard baselines in model compression studies \citep{affinequant, sparsegpt, wanda}. Model accuracy is assessed on a range of zero-shot downstream tasks, including MMLU \citep{mmlu}, Piqa \citep{piqa}, Arc-Easy, Arc-Challenge \citep{arc}, WinoGrande \citep{winogrande}, and OpenBookQA \citep{openbookqa}. For zero-shot evaluations, we utilize the Language Model Evaluation Harness \citep{lmharness} framework. In line with prior work \citep{wanda, sparsegpt, affinequant}, we also report the perplexity of the models on a language modeling task on the WikiText2 \citep{wikitext2} dataset, provided in \autoref{appendix:language_modeling}.

\niparagraph{Baselines.} We compare \slimspace against state-of-the-art one-shot pruning methods, including Wanda \citep{wanda}, SparseGPT \citep{sparsegpt}, and Magnitude Pruning \citep{fine_tuning3}, as well as one-shot quantization techniques like OPTQ \citep{optq}, OmniQuant \citep{omniquant}, AffineQuant \citep{affinequant}, L$^2$QER \citep{lqer}, and AbsMax. Additionally, we extend Joint Sparsification and Quantization (JSQ) \citep{jsq} to support 4-bit weight quantization and include it in our experiments. To ensure fairness, we use the optimal hyperparameters reported for each method, or the default hyperparameters if not explicitly reported. For a thorough \underline{\textbf{description of the notations}} used to show the different variants of \slim, please see \autoref{tab:slim_notation} in \autoref{appendix:notation}. The hyperparameters used in our experiments are detailed below.

To the best of our knowledge, L$^2$QER is the only compression method utilizing zero-shot low-rank adapters to enhance model accuracy. Our approach, \slim, significantly diverges from L$^2$QER in several key aspects. First, we employ saliency-based low-rank adapters to mitigate compression loss in \emph{quantized and sparse} models, whereas L$^2$QER is tailored exclusively for quantization, resulting in reduced accuracy when combined with sparsity, as demonstrated in the subsequent sections.
Second, we introduce \quantization, which lowers the overhead and complexity of group quantization compared to methods like L$^2$QER.
Finally, \slimspace compresses and fine-tunes low-rank adapters efficiently to minimize overhead. In contrast, L$^2$QER relies on full-precision low-rank adapters, which incur additional overhead and do not benefit from the parameter-efficient fine-tuning proposed in our work.

\niparagraph{Experiment Setup.} Similar to Wanda, SparseGPT, and OPTQ, \slimspace uses 128 sequences sampled from the C4 \citep{c4} dataset for calibration, and 300,000 tokens from C4 for all fine-tuning experiments. \quantization uses a histogram of weight elements to find the optimal scaling factor, with the number of bins set to $\max(512, \min(\frac{d_{in} \times d_{out}}{1000}, 20{,}000))$ to achieve an accurate approximation. All quantization experiments follow a 4-bit weight-only scheme with a group size of 128, consistent with prior work (OPTQ, OmniQuant, AffineQuant, etc.). For experiments involving Naive-LoRA and \slim-LoRA, the adapter rank is set to 10\% of the model's hidden dimension unless stated otherwise. Fine-tuning is performed with the HuggingFace Trainer \citep{huggingface} using the AdaFactor \citep{adafactor} optimizer with linear learning rate scheduling and default parameters. We use BFloat-16 \citep{bfloat} on NVIDIA A100 GPUs, with a local batch size of 1 and gradient accumulation factor of 64 to reduce memory overhead. Weight updates for sparse and/or quantized weights and corresponding biases are disabled during fine-tuning.

\niparagraph{Accuracy results.} We evaluate the accuracy of \slim and other state-of-the-art pruning and quantization methods across 2:4 and unstructured sparsity benchmarks, highlighting \slim's superiority in \autoref{tab:sparse_quantized}.
SparseGPT and Group OPTQ, designed to work together, achieve competitive performance. For other advanced quantization methods, we pruned models using Wanda and quantized the sparse checkpoints with Group AbsMax, AWQ, OmniQuant, and AffineQuant, reporting the best results (detailed in \autoref{appendix:accuracy_results}). Notably, methods like OmniQuant and AffineQuant struggle to quantize OPT-350M, often resulting in NaN values. Moreover, AWQ, OmniQuant, AffineQuant, and L$^2$QER encounter out-of-memory (OOM) errors when compressing models on a single A100-40GB GPU. While JSQ performs well for the LLaMA-2 family, its difficulty compressing the OPT family limits its broader applicability.

\begin{table*}[htbp]
\caption{Average zero-shot accuracy of LLaMA-2 and OPT models with \textbf{50\% sparsity and 4-bit weight quantization}. \textit{Best Method$^*$} indicates the best quantization method out of Group AbsMax, AWQ, OmniQuant, and AffineQuant. $\uparrow$ indicates better performance.}
\label{tab:sparse_quantized}
\begin{center}
\resizebox{\textwidth}{!}{%
\begin{tabular}{llllllllll}
\multicolumn{1}{c}{Pruning/LoRA}  &\multicolumn{1}{c}{Weight} & \multicolumn{6}{c}{OPT} & \multicolumn{2}{c}{LLaMA-2}
\\
\multicolumn{1}{c}{Method}  &\multicolumn{1}{c}{Quantization} & 125M & 350M & 1.3B & 2.7B & 6.7B & 13B & 7B & 13B
\\ \hline \\[-8pt]
Dense & - & 35.9 & 37.1 & 43.4 & 45.5 & 48.3 & 48.7 & 56.6 & 60.8 
\\ \hline \\[-3pt]
\textbf{2:4 Sparsity} \\
Magnitude & Group AbsMax & 32.19 & 31.94 & 33.82 & 33.43 & 34.81 & 34.68 & 44.64 & 44.18
\\
SparseGPT & Group OPTQ & 33.70 & 33.38 & 38.75 & 40.15 & 44.32 & 45.64 & 45.49 & 51.05
\\
Wanda &  Best Method$^*$ & 33.39 & 32.79 & 38.43 & 40.00 & 43.41 & 44.07 & 44.86 & 48.94
\\
JSQ & JSQ & 31.98 & 31.13 & 36.34 & 31.79 & 41.33 & 37.38 & 45.34 & 49.45
\\
L$^2$QER & Group AbsMax & 33.34 & 31.68 & 36.68 & 38.11 & 41.37 & OOM & 43.77 & OOM
\\ \hline \\[-8pt]
Naive-LoRA & \quantization & 34.28 & 33.38 & 38.36 & 41.21 & 44.91 & 45.25 & 48.45 & 51.94
\\
\slim-LoRA & \quantization & \textbf{34.62} & \textbf{34.36} & \textbf{40.61} & \textbf{42.73} & 45.99 & 46.09 & \textbf{51.15} & \textbf{54.94}
\\
\slim-LoRA$^Q$ & \quantization & 34.43 & 34.30 & 40.11 & 42.37 & \textbf{46.33} & \textbf{46.24} & 51.02 & 53.55
\\ \hline\hline \\[-3pt]
\textbf{50\% Unstructured} \\
Magnitude & Group AbsMax & 33.34 & 33.51 & 32.12 & 39.90 & 36.44 & 32.33 & 47.03 & 51.04
\\
SparseGPT & OPTQ & 35.10 & 35.13 & 38.72 & 43.43 & 46.97 & 47.38 & 51.09 & 55.94
\\
Wanda &  Best Method$^*$ & 35.11 & 33.89 & 41.02 & 42.89 & 46.52 & 46.84 & 53.62 & 56.76
\\
JSQ & JSQ & 32.05 & 31.09 & 39.53 & 33.35 & 41.04 & 31.80 & 52.08 & 57.00
\\
L$^2$QER & Group AbsMax & 34.45 & 34.45 & 38.38 & 41.28 & 45.08 & OOM & 50.60 & OOM
\\ \hline \\[-8pt]
Naive-LoRA & \quantization & 34.77 & 34.23 & 40.40 & 43.37 & 46.64 & 47.30 & 51.52 & 55.33
\\
\slim-LoRA & \quantization & 35.20 & \textbf{35.32} & \textbf{41.85} & 43.48 & 47.08 & \textbf{47.96} & \textbf{54.26} & \textbf{57.85}
\\
\slim-LoRA$^Q$ & \quantization & \textbf{35.35} & 35.13 & 41.74 & \textbf{43.63} & \textbf{47.16} & 47.86 & 54.18 & 57.33
\\ \hline 
\end{tabular}
}
\end{center}
\end{table*}

The progression from Naive-LoRA to \slim-LoRA and \slim-LoRA$^Q$ demonstrates the benefits of incorporating weight saliency into low-rank adapters and applying quantization for reducing overhead. While Naive-LoRA improves model accuracy across different sizes, \slim-LoRA achieves additional gains by effectively leveraging the saliency of the weights in the adapter design. Extending this, \slim-LoRA$^Q$ applies quantization to the low-rank adapters, further minimizing overhead with minimal impact on accuracy, adding negligible improvements or degradation to the accuracy of the model. 

\autoref{tab:fine-tuning-accuracy} demonstrates how lightweight fine-tuning (FT) improves the accuracy of both \slim-LoRA and Naive-LoRA, with \slim-LoRA exhibiting greater gains due to its saliency-aware design. Further details on the fine-tuning process and its overhead are provided in \autoref{appendix:fine_tuning_overhead}, illustrating its practicality for enhancing compressed model performance.

\begin{table}[!t]
\caption{Effects of fine-tuning on the average zero-shot accuracy of LLaMA-2 models with 50\% sparsity and 4-bit weight quantization. $\uparrow$ indicates better performance.}
\label{tab:fine-tuning-accuracy}
\begin{center}
\begin{tabular}{llll}
\multicolumn{1}{c}{Pruning/LoRA}  &\multicolumn{1}{c}{Weight} & \multicolumn{2}{c}{LLaMA-2}
\\
\multicolumn{1}{c}{Method}  &\multicolumn{1}{c}{Quantization} & 7B & 13B
\\ \hline \\[-6pt]
Dense & - & 56.6 & 60.8 
\\ \hline \\[-6pt]
\textbf{50\% 2:4} \\
Naive-LoRA + FT & \quantization & 50.89 & 55.70
\\
\slim-LoRA + FT & \quantization & \textbf{52.12} & \textbf{56.60}
\\
\slim-LoRA$^Q$ + FT & \quantization & 48.31 & 56.50
\\ \hline\hline \\[-6pt]
\textbf{50\% Unstructured}  \\
Naive-LoRA + FT & \quantization & 52.90 & 57.08
\\
\slim-LoRA + FT & \quantization & \textbf{54.69} & \textbf{57.96}
\\
\slim-LoRA$^Q$ + FT & \quantization & 53.57 & 57.78
\\ \hline 
\end{tabular}
\end{center}
\end{table}

\niparagraph{Integration with Weight Update.} \autoref{chapter:optima} proposes a method to compute the optimal per-layer weight updates given a calibration dataset. After determining the low-rank adapter values in \slim, we can find the optimal weight values by solving $\mathcal{W}^{C*} = \arg \min_{\mathcal{W^C}} \|\mathcal{XW}^C - \mathcal{X(W - LR)}\|$ using the QP solvers introduced in \optima. As shown in \autoref{tab:slim-optima-accuracy}, our results indicate that applying the compression trinity with \optima as the weight update and \slim-LoRA as the low-rank adapters can further boost the accuracy of the models. 

\begin{table}[!t]
\caption{Accuracy results of \optima weight update mechanism with \slim-LoRA. $\uparrow$ indicates better performance.}
\label{tab:slim-optima-accuracy}
\begin{center}
\begin{tabular}{llll}
\multicolumn{1}{c}{Pruning/LoRA}  &\multicolumn{1}{c}{Weight} & \multicolumn{2}{c}{LLaMA-2}
\\
\multicolumn{1}{c}{Method}  &\multicolumn{1}{c}{Quantization} & 7B & 13B
\\ \hline \\[-6pt]
Dense & - & 56.6 & 60.8 
\\ \hline \\[-6pt]
\textbf{50\% 2:4} \\
\slim-LoRA$^Q$  & \quantization & 51.02 & 53.55
\\
\slim-LoRA$^Q$ + \optima & \quantization & 51.62 & 53.84
\\ \hline\hline \\[-6pt]
\textbf{50\% Unstructured}  \\
\slim-LoRA$^Q$  & \quantization & 54.18 & 57.33
\\
\slim-LoRA$^Q$ + \optima & \quantization & 54.32 & 57.45
\\ \hline 
\end{tabular}
\end{center}
\end{table}

\niparagraph{Integration with Hybrid Sparsity.} As demonstrated in \autoref{chapter:patch}, \patch improves upon rigid semi-structured sparsity by enabling adaptive, tile-level density. While the standard implementation of \slim presented in this chapter utilizes Wanda~\citep{wanda} for efficient pruning, the modular design of the Compression Trinity allows us to substitute this component with more advanced sparsity operators. In this section, we integrate \patch into the \slim pipeline to evaluate the impact of hybrid sparsity on the fully compressed model.

We apply the \slim-LoRA error correction mechanism on top of the hybrid masks generated by \patch. For these specific experiments, we utilize Group AbsMax quantization to isolate the benefits of the hybrid sparsity pattern when combined with standard quantization schemes. \autoref{table::quantization} reports the results on LLaMA-2 7B and LLaMA-3.1 8B. The results demonstrate that combining the flexible sparsity of \patch with quantization and low-rank approximation enables controllable tradeoffs between compression ratio and model quality. This confirms that the enhancements made to the sparsity pillar in the previous chapter translate directly to improved flexibility in the joint compression setting.

\begin{table*}[tb]
\centering
\small
\setlength{\tabcolsep}{3pt}
\caption{Average accuracy ($\uparrow$ indicates better) across eight zero-shot downstream tasks (including RACE \citep{race} and HellaSwag \citep{hellaswag}) and WikiText2 perplexity ($\downarrow$ indicates better) of compressed models with \textbf{4-bit weight-only quantization}. Please note that using LoRA adds additional parameters to the model.}
\label{table::quantization}
\resizebox{\textwidth}{!}{
\begin{tabular}{l l l l c c c c}
\toprule
\textbf{Sparsity} & \textbf{Method} & \textbf{Pattern} & \textbf{LoRA} &
\multicolumn{2}{c}{\textbf{LLaMA-2-7B}} &
\multicolumn{2}{c}{\textbf{LLaMA-3.1-8B}} \\
\cmidrule(lr){5-6} \cmidrule(lr){7-8}
& & & & \textbf{Acc (\% $\uparrow$)} & \textbf{PPL ($\downarrow$)} & \textbf{Acc (\% $\uparrow$)} & \textbf{PPL ($\downarrow$)} \\
\midrule
0\%  & Dense     & -      & -      & 54.61 & 5.12 & 60.31 & 5.84 \\
\midrule
50\% & MaskLLM   & 2:4    & -      & 47.98 & 7.64 & 51.12 & 9.92 \\
\midrule

45\% & \patchtile     & Dense/2:4 Tiles & -      & 48.19 & 7.34 & 52.47 & 9.68 \\

45\% & \patchtile     & Dense/2:4 Tiles & \textsc{SLiM}-LoRA & 50.71 & 6.83 & 54.04 & 9.12 \\

35\% & \patchtile     & Dense/2:4 Tiles & -      & 49.38 & 6.92 & 53.81 & 9.26 \\

35\% & \patchtile     & Dense/2:4 Tiles & \textsc{SLiM}-LoRA & 51.91 & 6.42 & 55.70 & 8.37 \\

25\% & \patchtile     & Dense/2:4 Tiles & -      & 50.45 & 6.57 & 55.45 & 8.69 \\

25\% & \patchtile     & Dense/2:4 Tiles & \textsc{SLiM}-LoRA & 52.62 & 6.11 & 56.99 & 7.77 \\
\bottomrule
\end{tabular}
}
\end{table*}

\niparagraph{Comparison of large compressed and small dense models.}
This section compares large compressed models with dense models of equivalent parameter size, offering guidelines for configuration selection under hardware constraints. We focus on 2:4 sparsity due to its hardware acceleration support and evaluate the OPT model family, which spans a wide range of sizes for comprehensive analysis.

We analyze model performance by plotting average accuracy against parameter size, calculated as detailed in \autoref{appendix:memory_reduction}.
This visualization enables a direct performance comparison between models with an equal number of bits.

\begin{figure}
    \centering
    \includegraphics[scale=1.1]{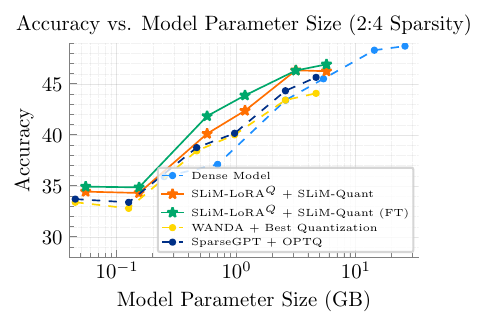}
    \caption{Accuracy results of the OPT family across different compression methods ($\uparrow$ indicates better performance). At equal parameter size, \slimspace outperforms both dense models and other compression techniques, demonstrating that model compression with \slimspace yields superior performance under the same budget.}
    \label{fig:large_compressed_vs_small_dense}
\end{figure}

\autoref{fig:large_compressed_vs_small_dense} presents the accuracy results of the OPT model family across different compression methods.
The x-axis represents the model parameter size in gigabytes, while the y-axis denotes accuracy (higher is better).
The results demonstrate that \slim-LoRA$^Q$, both with and without fine-tuning, consistently outperforms dense models and other compression techniques at the same parameter size.
Notably, compressed models achieve higher accuracy than dense models of equivalent size, highlighting the effectiveness of the proposed method.
This trend underscores the advantage of \slim-LoRA$^Q$ in maximizing model efficiency under strict hardware constraints.

\niparagraph{Speedup.}
Leveraging sparsity and quantization enhances GPU resource utilization, enabling faster model inference. Following Wanda's experimental setup, we evaluate the speedup achieved across different model layers and sizes. Similar to Wanda, AWQ, and QuaRot \citep{quarot}, we focus on consumer-grade GPUs and conduct our experiments on NVIDIA RTX 3060 GPUs. Speedup results for NVIDIA A100 GPUs are provided in \autoref{appendix:speedup}.

\slimspace achieves notable speedups through optimized sparse and quantized matrix multiplication, utilizing Sparse Marlin \citep{marlin} integrated with vLLM \citep{vllm}. For inference, we adopt small batch sizes during decoding, as recommended by prior works \citep{flash_llm, sparta}. Dense Quantized Marlin or PyTorch kernels handle the low-rank adapters based on their quantization status. \autoref{tab:rtx_speedup} highlights the speedup achieved across different LLaMA-2 layers compared to dense, unquantized models. Larger matrices, such as those in self-attention and feed-forward modules, consistently yield greater speedups, aligning with trends detailed in \autoref{appendix:speedup}.

\niparagraph{Sparse-only results.}
\label{appendix:sparse_only}
To evaluate the isolated impact of sparsity on model accuracy, we disable quantization and benchmark Magnitude Pruning, SparseGPT, and Wanda, alongside low-rank approximations like Wanda-SVD and \slimspace. Our experiments assess both 50\% unstructured sparsity and 2:4 structured sparsity patterns.

\autoref{tab:sparse} shows the accuracy results for sparse models. Magnitude Pruning performs the worst, while Wanda and SparseGPT achieve comparable results, with larger accuracy gaps for semi-structured sparsity. Low-rank adapters improve accuracy, with \slimspace leveraging saliency-based approximation for superior performance. A brief fine-tuning phase further boosts the accuracy of low-rank approximations.

\begin{table}[H]
\caption{Average zero-shot accuracy of LLaMA-2 and OPT models with pruning. The quantization is disabled in this experiment. $\uparrow$ indicates better performance.}
\label{tab:sparse}
\begin{center}
\begin{tabular}{lllllllll}
\multicolumn{1}{c}{Pruning/LoRA}  & \multicolumn{6}{c}{OPT} & \multicolumn{2}{c}{LLaMA-2}
\\
\multicolumn{1}{c}{Method}  & 125M & 350M & 1.3B & 2.7B & 6.7B & 13B & 7B & 13B
\\ \hline \\ [-8pt]
Dense & 35.9 & 37.1 & 43.4 & 45.5 & 48.3 & 48.7 & 56.6 & 60.8
\\ \hline \\[-8pt]
\textbf{2:4 Sparsity} \\
Magnitude & 32.6 & 31.8 & 35.4 & 33.9 & 36.4 & 30.7 & 31.2 & 32.0
\\
SparseGPT & 33.8 & 33.2 & 37.7 & 41.3 & 45.2 & 45.6 & 47.3 & 52.3
\\
Wanda & 34.0 & 32.5 & 38.3 & 40.5 & 43.2 & 44.1 & 46.1 & 49.7
\\
\slim-Naive & 34.1 & 34.1 & 40.4 & 42.8 & 46.0 & 45.9 & 51.6 & 55.8
\\
\slim-Naive + FT & 34.8 & 34.5 & 41.3 & 43.4 & \textbf{46.5} & 47.2 & \textbf{52.4} & \textbf{56.9}
\\
\slim-LoRA & 34.5 & 32.9 & 40.7 & 43.1 & 46.4 & 46.3 & 51.4 & 56.1
\\
\slim-LoRA  + FT & \textbf{35.1} & \textbf{34.9} & \textbf{41.5} & \textbf{43.8} & \textbf{46.5} & \textbf{47.3} & 51.6 & 56.4
\\ \hline \\[-8pt]
\textbf{50\% Unstructured} \\
Magnitude & 33.3 & 33.7 & 34.0 & 40.6 & 35.8 & 30.9 & 32.6 & 31.9
\\
SparseGPT & 35.5 & 35.1 & 39.6 & 43.5 & 47.4 & 47.8 & 53.3 & 57.3
\\
Wanda & 35.0 & 34.5 & 41.1 & 42.9 & 46.5 & 46.8 & 52.7 & 57.2
\\
\slim-Naive & 35.3 & 35.2 & 41.9 & 44.1 & 47.5 & 47.8 & 54.9 & 58.5
\\
\slim-Naive + FT & 35.74 & 35.7 & 42.7 & 44.6 & \textbf{47.8} & \textbf{48.4} & 54.9 & 58.7
\\
\slim-LoRA & 35.2 & 35.1 & 42.0 & 44.1 & 47.7 & 48.2 & \textbf{55.0} & \textbf{58.8}
\\
\slim-LoRA  + FT & \textbf{35.9} & \textbf{35.7} & \textbf{42.5} & \textbf{44.7} & 47.7 & \textbf{48.4} & \textbf{55.0} & \textbf{58.8}
\\ \hline
\end{tabular}
\end{center}
\end{table}

\niparagraph{Quantization-only results.}
\label{appendix:quantize_only}
To evaluate the impact of \quantization and low-rank compensation in \slim, we conduct experiments without sparsity, testing quantization schemes like Group AbsMax, OPTQ, AWQ, OmniQuant, AffineQuant, L$^2$QER, and \quantization. To enhance accuracy, we add low-rank adapters to \quantization and Group AbsMax, optimizing either error saliency (\slim-LoRA) or reconstruction error norm (Naive-LoRA). Other quantization methods cannot incorporate low-rank adapters due to conflicting weight/activation update rules.

\autoref{tab:quantized} presents the quantization results. Adding low-rank adapters to Group AbsMax significantly boosts model accuracy, outperforming most advanced methods. While \quantization alone is not designed for high accuracy, its integration with \slimspace variants achieves results comparable to or better than Group AbsMax with low-rank adapters, highlighting the value of co-design in compression methods. Furthermore, a lightweight fine-tuning phase with \quantization delivers state-of-the-art accuracy.

\begin{table}[H]
\caption{Average zero-shot accuracy of LLaMA-2 and OPT models with quantization. The sparsity is disabled in this experiment. $\uparrow$ indicates better performance.}
\label{tab:quantized}
\begin{center}
\resizebox{\textwidth}{!}{%
\begin{tabular}{llllllllll}
\multicolumn{1}{c}{Quantization} & \multicolumn{1}{c}{Low-rank}  & \multicolumn{6}{c}{OPT} & \multicolumn{2}{c}{LLaMA-2}
\\
\multicolumn{1}{c}{Method} & \multicolumn{1}{c}{Adapter} & 125M & 350M & 1.3B & 2.7B & 6.7B & 13B & 7B & 13B
\\ \hline \\
Dense & - & 35.9 & 37.1 & 43.4 & 45.5 & 48.3 & 48.7 & 56.6 & 60.8
\\ \hline \\
OPTQ & - & 35.64 & 36.46 & 42.83 & 44.20 & 47.46 & 48.24 & 53.53 & 59.80
\\
AWQ & - & 36.16 & 31.83 & 42.98 & 45.28 & 48.45 & 48.76 & 53.97 & OOM
\\
OmniQuant & - & 35.46 & NaN & 42.15 & 44.71 & 46.65 & OOM & 54.33 & OOM
\\
AffineQuant & - & 35.73 & NaN & 42.62 & 44.92 & 47.91 & OOM & 54.52 & OOM
\\
\hline
\\
Group AbsMax & - & 35.45 & 36.67 & 42.57 & 44.79 & 48.30 & 48.49 & 55.56 & 60.12
\\
Group AbsMax & L$^2$QER & 34.75 & 35.63 & 40.60 & 44.22 & 46.90 & OOM & 55.95 & OOM
\\
Group AbsMax & \slim-Naive & \textbf{36.30} & 36.58 & 43.07 & 45.13 & 48.26 & 48.72 & 56.23 & 60.53
\\
Group AbsMax & \slim-LoRA & 36.18 & \textbf{36.72} & 42.89 & \textbf{45.65} & \textbf{48.45} & 48.89 & 55.99 & 60.16
\\
\hline
\\
\quantization & - & 31.98 & 36.46 & 36.19 & 40.08 & 45.61 & 38.27 & 31.11 & 30.51
\\
\quantization & \slim-Naive & 35.29 & 36.02 & 42.48 & 45.01 & 47.75 & 48.38 & 55.96 & \textbf{60.85}
\\
\quantization & \slim-LoRA & 35.69 & 36.42 & 42.59 & 45.26 & 48.18 & 48.52 & 56.26 & 60.59
\\
\quantization & \slim-LoRA + FT & 35.91 & 36.61 & \textbf{43.29} & 45.58 & 48.29 & \textbf{49.04} & \textbf{56.51} & 60.65
\\ \hline
\end{tabular}}%
\end{center}
\end{table}

\niparagraph{Additional Experiments.}
We provide additional experiments for a comprehensive evaluation in the appendix.

The \textbf{\underline{Language Modeling Experiments}} (\autoref{appendix:language_modeling}) evaluate \slimspace across sparse and quantized, sparse-only, and quantized-only models on WikiText-2. The results align with the accuracy trends reported in the preceding sections, further validating the effectiveness of \slimspace.

The \textbf{\underline{Fine-tuning Costs}} (\autoref{appendix:fine_tuning_overhead}) show that \slimspace reduces fine-tuning overhead from over 36 days for 13B parameter models to just 14 hours on a single GPU, demonstrating its practicality and efficiency.
\begin{table}[htbp]
\caption{LLaMA-2 family of models speedup ($\times$) using \slimspace compared to original dense unquantized model on NVIDIA RTX-3060. $\uparrow$ shows higher speedup.}
\label{tab:rtx_speedup}
\begin{small}
\begin{center}
\begin{tabular}{llllll}
Model & Batch & LoRA & Self- & Up- & Down
\\
Size & Size & Type & Attention & Projection & Projection
\\ \hline \\[-8pt]
\multirow{6}{*}{7B} & \multirow{2}{*}{16} & FP16 & 1.99 & 2.58 & 3.23
\\
&& INT4 & 1.00 & 2.29 & 2.58
\\
\cline{2-6}
\\[-8pt]
& \multirow{2}{*}{32} & FP16 & 2.06 & 2.53 & 3.40
\\
&& INT4 & 1.13 & 2.27 & 2.97
\\
\cline{2-6}
\\[-8pt]
& \multirow{2}{*}{64} & FP16 & 1.54 & 1.70 & 1.85
\\
&& INT4 & 0.96 & 1.62 & 1.76
\\ \hline\hline \\[-8pt]
\multirow{6}{*}{13B} & \multirow{2}{*}{16} & FP16 & 2.18 & 2.53 & 2.60
\\
&& INT4 & 1.28 & 3.24 & 3.17
\\
\cline{2-6}
\\[-8pt]
& \multirow{2}{*}{32} & FP16 & 2.23 & 2.68 & 2.91
\\
&& INT4 & 1.43 & 2.96 & 3.20
\\
\cline{2-6}
\\[-8pt]
& \multirow{2}{*}{64} & FP16 & 1.38 & 1.78 & 1.67
\\
&& INT4 & 1.21 & 1.69 & 1.65
\\ \hline\hline \\[-8pt]
\multirow{6}{*}{70B} & \multirow{2}{*}{16} & FP16 & 2.18 & 2.86 & 2.75
\\
&& INT4 & 3.11 & 3.99 & 3.79
\\
\cline{2-6}
\\[-8pt]
& \multirow{2}{*}{32} & FP16 & 2.00 & 2.63 & 2.67
\\
&& INT4 & 2.75 & 3.19 & 3.39
\\
\cline{2-6}
\\[-8pt]
& \multirow{2}{*}{64} & FP16 & 1.38 & 1.70 & 1.86
\\
&& INT4 & 1.51 & 1.77 & 1.94
\\ 
\hline 
\end{tabular}
\end{center}
\end{small}
\end{table}

We provide a comparison between \textbf{\underline{Sparsity vs.}} \textbf{\underline{Quantization}} (\autoref{appendix:sparsity_vs_quantization}) to show that combining 50\% sparsity and 4-bit quantization helps achieve better compression results in comparison to solely using 2-bit quantization, while maintaining a similar compression ratio (\(\sim\)$8\times$).

Additional speedup results for \slimspace on NVIDIA A100-40GB GPUs are provided in the \textbf{\underline{Additional}} \textbf{\underline{Speedup Results}} (\autoref{appendix:speedup}). A theoretical analysis of computation and memory reductions can be found in the \textbf{\underline{Computation Reduction Analysis}} (\autoref{appendix:compute_reduction}) and \textbf{\underline{Memory Reduction Analysis}} (\autoref{appendix:memory_reduction}), highlighting the efficiency of \slimspace.

\underline{\textbf{Compression Costs}} (\autoref{sec:compression_cost}) details the time required to compress models of various sizes across different methods. \underline{\textbf{Rank Analysis}} (\autoref{sec:rank_analysis}) explores how rank choices in low-rank adapters impact computational and memory costs, as well as model accuracy. \textbf{\underline{Sparsity Analysis}} (\autoref{appendix:sparsity_analysis}) analyzes the effects of different sparsity ratios on model compression. Lastly, \underline{\textbf{Effects of Calibration Sample Count}} (\autoref{sec:calibration_sample}) evaluates the influence of calibration sample counts on the accuracy of calibration-based methods.

\section{Conclusion}
\label{sec:slim_conclusion}

In this chapter, we presented \slim, the complete fulfillment of the Compression Trinity framework. We began this thesis by identifying the "Sparsity Paradox," the observation that removing weights is structurally destructive and, when applied in isolation, leads to early accuracy collapse. \slim resolves this paradox. By seamlessly integrating optimized uniform quantization (\slim-Quant), hardware-friendly sparsity, and crucially mathematically derived low-rank error correction (\slim-LoRA), we have turned the Low-Rank pillar into a restorative force. It recovers the information lost by the aggressive application of the first two pillars, solving the "compounded error" challenge that has historically hindered joint compression.

An important direction for strengthening the theoretical foundations of
\slim is to draw more explicitly on the guarantees provided by Robust PCA
theory. The classical PCP framework~\citep{candes2011robust} establishes
that the sparse-plus-low-rank decomposition is exactly recoverable under
incoherence conditions, and the Stable PCP
extension~\citep{zhou2010stable} shows that this recovery is robust to
additional dense noise, a property that could be leveraged to account for
quantization error. Translating these guarantees to the LLM compression
setting would require verifying whether the incoherence assumptions hold
for pre-trained weight matrices and characterizing the interaction
between quantization noise, sparsity patterns, and low-rank structure.
Furthermore, replacing \slim's current sequential pipeline with a joint
optimization formulation, as explored by OATS~\citep{zhang2025oats},
HASSLE-free~\citep{makni2025hassle}, and 3BASiL~\citep{behdin2025basil},
could yield tighter error bounds and potentially improve accuracy by
avoiding the suboptimality inherent in sequential decomposition. Such a
formulation would need to incorporate the quantization constraint,
extending the standard sparse-plus-low-rank problem to a ``quantized
sparse-plus-low-rank'' decomposition, an open problem that merits further
investigation.

\slim not only shifts the Pareto frontier, outperforming dense models at equal parameter budgets, but also serves as the final proof of our core thesis: that efficiency is not a singular optimization problem, but a multi-dimensional balancing act. We have traversed the complete arc of this life-cycle: from accelerating pretraining dynamics (\autoref{chapter:mkor}, \autoref{chapter:slope}) to establishing the limits of sparsity in both static (\autoref{chapter:optima}) and dynamic (\autoref{chapter:patch}) regimes, and finally achieving a unified, one-shot solution for deployment (\autoref{chapter:slim}).

The next and final chapter will summarize these contributions, discuss the broader implications of the Compression Trinity for the future of efficient AI, and outline potential avenues for further research.

  \chapter{Conclusion and Future Work}
\label{chapter:conclusion}

\section{Summary of Contributions}

This thesis has argued that the efficiency bottleneck in Large Language Models (LLMs) is not merely a resource constraint, but a methodological failure to integrate complementary compression principles. We have established that the ``efficiency wall'' encountered by isolated techniques is \textbf{methodological rather than fundamental}. By jointly applying the ``Compression Trinity,'' sparsity, quantization, and low-rank approximations, we have demonstrated that the distinct hardware bottlenecks of compute FLOPs, memory bandwidth, and parameter redundancy can be attacked simultaneously.

Crucially, we established that the Trinity is not merely a post-training optimization tool; it is a fundamental framework applicable to the entire life-cycle of the model.

\subsection{The Trinity in Training Dynamics}

We demonstrated that the training dynamics themselves are compressible. By selectively applying pillars of the Trinity during optimization, we proved that high-fidelity, dense updates are not strictly necessary for convergence.

\begin{itemize}
    \item \textbf{Sparse, Low-Rank, and Quantized Optimization (MKOR):} In \autoref{chapter:mkor}, we addressed the prohibitive computational penalty of second-order optimization. MKOR validates the Trinity's utility in training by combining block diagonal \textbf{Sparsity} with \textbf{Low-Rank} approximations (rank-1 updates) and \textbf{Quantization} (to stabilize memory footprint). This approach reduces curvature update complexity from $\mathcal{O}(d^3)$ to $\mathcal{O}(d^2)$, accelerating convergence by up to $2.57\times$ compared to first-order baselines and $1.75\times$ compared to KFAC.
    
    \item \textbf{Sparse and Low-Rank Training (\slope):} In \autoref{chapter:slope}, we validated the concept of ``Lossy Training'' by integrating \textbf{Sparsity} and \textbf{Low-Rank} principles. By enforcing N:M sparsity in the backward pass and delaying low-rank recovery (``lazy'' adapters) to the final 1\% of training, we showed that the training process can withstand significant information loss without sacrificing final model accuracy.
\end{itemize}

\subsection{The Trinity in Post-Training and Inference}

For deployed models, we systematically dismantled the primary barriers to compression, compounded error and structural rigidity, resulting in a unified inference framework.

\begin{itemize}
    \item \textbf{Solving Compounded Error (\optima):} \optima (\autoref{chapter:optima}) utilized global optimization to stabilize the \textbf{Quantization} pillar. By establishing that weight reconstruction must be solved via column-wise Quadratic Programs (QPs) with a shared Hessian, \optima provides the foundational stability necessary to withstand aggressive compression.
    
    \item \textbf{Breaking Rigidity (\patch):} \patch (\autoref{chapter:patch}) advanced the \textbf{Sparsity} pillar by breaking the rigidity of hardware-enforced patterns. By introducing learnable tile-level hybrid sparsity, we proved that sparsity ratios can be continuous and adaptive (0--50\%) rather than discrete, preserving density in information-critical layers.
    
    \item \textbf{Unified Inference (\slim):} The empirical validation of the full framework culminates in \slim (\autoref{chapter:slim}). \slim represents the simultaneous integration of the full Trinity, aggressive \textbf{Quantization}, semi-structured \textbf{Sparsity}, and saliency-based \textbf{Low-Rank} adapters, during the inference phase. It shifts the Pareto frontier of model efficiency, demonstrating that a fully compressed model can improve accuracy by up to 5.66\% over state-of-the-art methods and, in specific configurations, outperform uncompressed dense models at equivalent parameter budgets. The industrial significance of these findings, particularly the necessity of 2:4 sparsity in modern production stacks, was subsequently featured in our technical analysis on the PyTorch blog\footnote{\url{https://pytorch.org/blog/when-quantization-isnt-enough-why-24-sparsity-matters/}}.
\end{itemize}

\section{Exploratory Frameworks and Open Research}

The core pillars of the Compression Trinity, sparsity, quantization, and low-rank approximations, provide a rigorous foundation for model efficiency. However, the application of these principles need not be confined to the rigid structures of traditional academic publication cycles. In parallel with the formal chapters of this thesis, we have developed agile, exploratory frameworks that extend the Trinity into new domains of adaptability and rapid deployment. These works, released as open-research contributions, demonstrate the flexibility of our methodology in addressing emerging challenges in the LLM landscape.

\subsection{BEAM: Blockwise Error Minimization for One-shot Compression of LLMs}
\label{sec:beam}
Standard post-training compression techniques (e.g., GPTQ, Wanda) often hit an accuracy ceiling because they optimize weights locally (layer-wise) without accounting for the non-linear interactions across the full transformer block. Conversely, full model fine-tuning (e.g., LoRA) is computationally expensive and requires curated datasets. 

To bridge this gap, we introduced \textbf{BEAM}\footnote{Full release: \url{https://www.cs.toronto.edu/~mmozaffari/compression-trinity/beam/index.html}}, a framework for one-shot compression that requires no end-to-end retraining. BEAM re-frames the compression problem by treating the intermediate activations of the original, uncompressed model as the ``ground truth'' for the compressed model. By splitting the LLM into independent transformer blocks and optimizing the compressed weights to minimize the feature reconstruction error of each block, BEAM captures the non-linear dependencies lost by simple layer-wise techniques. Crucially, this method is orthogonal to the specific compression type; it serves as a universal refinement stage that can recover up to 4.34\% accuracy on sparse and quantized models using a single GPU in under four hours.

\subsection{LEAP: Learnable End-to-End Adaptive Pruning of LLMs}
\label{sec:leap}
In \autoref{chapter:patch}, we explored hybrid structural sparsity to balance hardware efficiency with information retention. However, imposing \textit{any} structure (even a hybrid one) inherently limits the model's expressivity compared to unstructured pruning. 

\textbf{LEAP}\footnote{Full release: \url{https://www.cs.toronto.edu/~mmozaffari/compression-trinity/leap/index.html}} challenges the assumption that unstructured sparsity must be static or heuristically determined (e.g., magnitude-based pruning). Instead, LEAP introduces a fully differentiable masking mechanism where the binary inclusion of every parameter is treated as a learnable latent variable during training. By relaxing the discrete mask into a continuous probability distribution and applying straight-through estimation, LEAP allows the model to dynamically ``evolve'' its own sparsity pattern end-to-end. This approach reveals that optimal sparsity is not fixed; it shifts during training as the model specializes, suggesting that the ``Trinity'' can eventually include \textit{topology} as a learnable parameter alongside weights.

\subsection{SLICE: Selecting Layer-wise Configurations for Matryoshka-Style LLMs}
\label{sec:slice}
The prevailing paradigm in LLM deployment is ``one size fits all'': a model is compressed to a fixed target (e.g., 4-bit, 50\% sparsity) and deployed. This rigidity is inefficient for dynamic environments where hardware availability changes in real-time.

\textbf{SLICE}\footnote{Full release: \url{https://www.cs.toronto.edu/~mmozaffari/compression-trinity/slice/index.html}} extends the concept of Matryoshka Representation Learning to the compression configuration itself. SLICE trains a single ``super-model'' capable of operating at multiple efficiency tiers simultaneously. By solving for a nested set of configurations (e.g., a 2-bit core nested within a 4-bit shell), SLICE enables elastic deployment: the same model can instantaneously shed layers or precision bits to meet strict latency deadlines, or expand to full capacity when resources permit. This work points toward a future where the Compression Trinity is not a static compilation step, but a dynamic runtime state.

\section{Limitations of the Current Approach}

While the Compression Trinity offers a robust framework, the specific implementations proposed in this thesis are subject to constraints that affect their immediate generalizability and ease of adoption.

\niparagraph{Hardware Coupling and Portability.}
Our methods, particularly \patch and the semi-structured sparsity utilized in \slope and \slim, are currently tightly coupled to the NVIDIA Sparse Tensor Core architecture (2:4 sparsity). While effective on dominant hardware, the transferability of these specific patterns to non-NVIDIA accelerators (e.g., TPUs, AMD MI-series) or general-purpose CPUs remains unproven. Furthermore, the reliance on custom CUDA and Triton kernels introduces a significant software portability barrier, effectively restricting these optimizations to advanced engineering environments.

\niparagraph{Training Instability Risks.}
Although MKOR introduces stabilizers to mitigate exploding gradients, second-order optimizers remain inherently more sensitive to hyperparameters than robust first-order methods like AdamW. The introduction of additional hyperparameters for curvature approximation imposes a tuning burden on practitioners, potentially offsetting the wall-clock speed gains in experimental settings.

\niparagraph{Pipeline Complexity.}
The ``Compression Trinity'' introduces significant engineering overhead. Deploying a pipeline that requires simultaneous quantization, pruning, and low-rank adaptation (as in \slim) is considerably more complex to implement, debug, and maintain than simpler post-training quantization techniques (e.g., INT8/FP8). This complexity represents a barrier to adoption for practitioners seeking ``plug-and-play'' solutions.

\niparagraph{Fairness and Differential Impact on Underrepresented Populations.}
Throughout this thesis, compression quality is measured by aggregate accuracy on standard benchmarks, yet matching the uncompressed model's overall accuracy does not guarantee uniform performance across all subpopulations. Pruning, quantization, and low-rank factorization remove model capacity that may disproportionately encode knowledge about underrepresented groups, a phenomenon that aggregate metrics can mask entirely. For instance, a compressed language model may preserve perplexity on high-resource languages such as English while exhibiting significant degradation on low-resource languages that were sparsely represented in the fine-tuning or calibration datasets. Similarly, in classification tasks, accuracy on minority demographic groups may suffer even when the overall metric remains stable. Because our calibration and evaluation pipelines rely on datasets that predominantly reflect majority populations, we cannot rule out that the proposed compression methods amplify existing biases or introduce new disparities. A rigorous fairness audit, disaggregating performance across languages, dialects, demographic groups, and downstream tasks, is an important direction that lies outside the scope of this work but is essential before deploying compressed models in high-stakes applications.

\niparagraph{Scale of Evaluation.}
Our empirical validation focused on models in the 125M to 70B parameter range (e.g., OPT, LLaMA-2/3). As scaling laws push frontier models into the trillion-parameter regime, emergent behaviors or shifting bottlenecks (e.g., massive cross-node communication overheads) may alter the effectiveness of these compression techniques, a domain this thesis leaves unexplored.

\section{Future Research Directions}

\niparagraph{Compressing the Context Window.}
This thesis focused heavily on Linear layers, which currently dominate compute. However, as sequence lengths grow to 1M+ tokens, the Attention mechanism and Key-Value (KV) cache become the dominant memory bottlenecks. Future work must extend the Trinity to the context window: quantizing dynamic KV states, inducing sparsity in attention patterns (e.g., via Sliding Window or Block-Sparse Attention), and applying low-rank approximations to the attention heads themselves to enable infinite-context reasoning on commodity hardware.

\niparagraph{Hardware-Algorithm Co-design.}
Current sparse formats incur a memory overhead for storing metadata (indices), which can negate compression gains at lower bit-widths. Future compression research cannot occur in a software vacuum; it requires a co-design approach to propose ``hardware-defining'' sparse formats. We envision a move toward algorithmic sparsity where the pattern is deterministic or predicted, eliminating the need for explicit index storage and further reducing the memory footprint.

\niparagraph{The Trinity for Activations.}
While we successfully compressed weights, activation tensors remain a challenge during the prefill phase. Future work should investigate applying {\patch}-like learnable masks to dynamic activation tensors, enabling ``activation sparsity'' that can accelerate the compute-bound prefill phase without requiring full retraining.

\section{Closing Remarks}
Ultimately, this thesis posits that efficiency is not merely a post-hoc optimization, but a fundamental design constraint. We have argued that the ``efficiency wall'' is a methodological artifact, dissolvable by the joint application of sparsity, quantization, and low-rank approximations.

By proving that the ``Compression Trinity'' can be integrated into every stage of the LLM life-cycle, from the training dynamics of \mkor to the inference engines of \slim, we pave the way for a new paradigm of model design. In this paradigm, models are not simply made smaller; they are architected from the ground up to be dense in knowledge yet sparse in computation, rendering high-intelligence AI fundamentally more accessible and sustainable.

  \bibliography{bibliography}

@article{ngd,
	title        = {{Natural Gradient Works Efficiently in Learning}},
	author       = {Amari, Shun-Ichi},
	year         = 1998,
	journal      = {Neural Computation},
	publisher    = {MIT Press},
	volume       = 10,
	number       = 2,
	pages        = {251--276}
}

@article{fim,
	title        = {{On the Locality of the Natural Gradient for Learning in Deep Bayesian Networks}},
	author       = {Ay, Nihat},
	year         = 2020,
	journal      = {Information Geometry},
	publisher    = {Springer},
	pages        = {1--49}
}

@article{gpt3,
	title        = {{Language Models Are Few-Shot Learners}},
	author       = {Brown, Tom and Mann, Benjamin and Ryder, Nick and Subbiah, Melanie and others},
	year         = 2020,
	journal      = {Advances in Neural Information Processing Systems},
	volume       = 33,
	pages        = {1877--1901}
}

@article{alexnet,
	title        = {{ImageNet Classification with Deep Convolutional Neural Networks}},
	author       = {Krizhevsky, Alex and Sutskever, Ilya and Hinton, Geoffrey E},
	year         = 2017,
	journal      = {Communications of the ACM},
	publisher    = {AcM New York, NY, USA},
	volume       = 60,
	number       = 6,
	pages        = {84--90}
}

@inproceedings{hylo,
	title        = {{HyLo: A Hybrid Low-Rank Natural Gradient Descent Method}},
	author       = {Mu, Baorun and Soori, Saeed and Can, Bugra and G{\"u}rb{\"u}zbalaban, Mert and others},
	year         = 2022,
	booktitle    = {Proceedings of the International Conference on High Performance Computing, Networking, Storage and Analysis},
	pages        = {1--16}
}

@article{sngd1,
	title        = {{Efficient Subsampled Gauss-Newton and Natural Gradient Methods for Training Neural Networks}},
	author       = {Ren, Yi and Goldfarb, Donald},
	year         = 2019,
	journal      = {arXiv preprint arXiv:1906.02353}
}

@article{sngd2,
	title        = {{Sketchy Empirical Natural Gradient Methods for Deep Learning}},
	author       = {Yang, Minghan and Xu, Dong and Wen, Zaiwen and Chen, Mengyun and others},
	year         = 2020,
	journal      = {arXiv preprint arXiv:2006.05924}
}

@article{transformer,
	title        = {{Attention Is All You Need}},
	author       = {Vaswani, Ashish and Shazeer, Noam and Parmar, Niki and Uszkoreit, Jakob and others},
	year         = 2017,
	journal      = {Advances in Neural Information Processing Systems},
	volume       = 30
}

@inproceedings{lra,
	title        = {{Long Range Arena: A Benchmark for Efficient Transformers}},
	author       = {Tay, Yi and Dehghani, Mostafa and Abnar, Samira and Shen, Yikang and others},
	year         = 2021,
	booktitle    = {ICLR}
}

@inproceedings{kfac,
	title        = {{Optimizing Neural Networks with Kronecker-Factored Approximate Curvature}},
	author       = {Martens, James and Grosse, Roger},
	year         = 2015,
	booktitle    = {ICML},
	pages        = {2408--2417},
	organization = {PMLR}
}

@article{kbfgs,
	title        = {{Practical Quasi-Newton Methods for Training Deep Neural Networks}},
	author       = {Goldfarb, Donald and Ren, Yi and Bahamou, Achraf},
	year         = 2020,
	journal      = {Advances in Neural Information Processing Systems},
	volume       = 33,
	pages        = {2386--2396}
}

@inproceedings{dist_kfac1,
	title        = {{Distributed Second-Order Optimization Using Kronecker-Factored Approximations}},
	author       = {Ba, Jimmy and Grosse, Roger and Martens, James},
	year         = 2017,
	booktitle    = {ICLR}
}

@inproceedings{dist_kfac2,
	title        = {{Rich Information Is Affordable: A Systematic Performance Analysis of Second-Order Optimization Using K-FAC}},
	author       = {Ueno, Yuichiro and Osawa, Kazuki and Tsuji, Yohei and Naruse, Akira and others},
	year         = 2020,
	booktitle    = {Proceedings of the 26th ACM SIGKDD International Conference on Knowledge Discovery \& Data Mining},
	pages        = {2145--2153}
}

@inproceedings{dist_kfac_conv1,
	title        = {{Convolutional Neural Network Training with Distributed K-FAC}},
	author       = {Pauloski, J Gregory and Zhang, Zhao and Huang, Lei and Xu, Weijia and others},
	year         = 2020,
	booktitle    = {SC20: International Conference for High Performance Computing, Networking, Storage and Analysis},
	pages        = {1--12},
	organization = {IEEE}
}

@inproceedings{dist_kfac_parallel_comp_and_commu,
	title        = {{Accelerating Distributed K-FAC with Smart Parallelism of Computing and Communication Tasks}},
	author       = {Shi, Shaohuai and Zhang, Lin and Li, Bo},
	year         = 2021,
	booktitle    = {2021 IEEE 41st International Conference on Distributed Computing Systems (ICDCS)},
	pages        = {550--560},
	organization = {IEEE}
}

@inproceedings{dist_kfac_conv2,
	title        = {{Large-Scale Distributed Second-Order Optimization Using Kronecker-Factored Approximate Curvature for Deep Convolutional Neural Networks}},
	author       = {Osawa, Kazuki and Tsuji, Yohei and Ueno, Yuichiro and Naruse, Akira and others},
	year         = 2019,
	booktitle    = {Proceedings of the IEEE/CVF Conference on Computer Vision and Pattern Recognition},
	pages        = {12359--12367}
}

@article{sgd,
	title        = {{An Overview of Gradient Descent Optimization Algorithms}},
	author       = {Ruder, Sebastian},
	year         = 2016,
	journal      = {arXiv preprint arXiv:1609.04747}
}

@inproceedings{shampoo,
	title        = {{Shampoo: Preconditioned Stochastic Tensor Optimization}},
	author       = {Gupta, Vineet and Koren, Tomer and Singer, Yoram},
	year         = 2018,
	booktitle    = {ICML},
	pages        = {1842--1850},
	organization = {PMLR}
}

@article{scalable_shampoo,
	title        = {{Scalable Second Order Optimization for Deep Learning}},
	author       = {Anil, Rohan and Gupta, Vineet and Koren, Tomer and Regan, Kevin and others},
	year         = 2020,
	journal      = {arXiv preprint arXiv:2002.09018}
}

@inproceedings{cosine_annealing_lr,
	title        = {{SGDR: Stochastic Gradient Descent with Warm Restarts}},
	author       = {Loshchilov, Ilya and Hutter, Frank},
	year         = 2017,
	booktitle    = {ICLR}
}

@inproceedings{triangular_lr,
	title        = {{Universal Language Model Fine-Tuning for Text Classification}},
	author       = {Howard, Jeremy and Ruder, Sebastian},
	year         = 2018,
	booktitle    = {ACL},
	pages        = {328--339}
}

@inproceedings{resnet,
	title        = {{Deep Residual Learning for Image Recognition}},
	author       = {He, Kaiming and Zhang, Xiangyu and Ren, Shaoqing and Sun, Jian},
	year         = 2016,
	booktitle    = {Proceedings of the IEEE Conference on Computer Vision and Pattern Recognition},
	pages        = {770--778}
}

@inproceedings{bert,
	title        = {{BERT: Pre-Training of Deep Bidirectional Transformers for Language Understanding}},
	author       = {Devlin, Jacob and Chang, Ming-Wei and Lee, Kenton and Toutanova, Kristina},
	year         = 2019,
	booktitle    = {NAACL},
	pages        = {4171--4186}
}

@techreport{cifar,
	title        = {{Learning Multiple Layers of Features from Tiny Images}},
	author       = {Krizhevsky, Alex and Hinton, Geoffrey and others},
	year         = 2009,
	number       = {Tr-2009},
	institution  = {University of Toronto}
}

@inproceedings{imdb,
	title        = {{Learning Word Vectors for Sentiment Analysis}},
	author       = {Maas, Andrew L. and Daly, Raymond E. and Pham, Peter T. and Huang, Dan and others},
	year         = 2011,
	month        = {June},
	booktitle    = {Proceedings of the 49th Annual Meeting of the Association for Computational Linguistics: Human Language Technologies},
	publisher    = {Association for Computational Linguistics},
	address      = {Portland, Oregon, USA},
	pages        = {142--150},
	url          = {http://www.aclweb.org/anthology/P11-1015}
}

@article{squad,
	title        = {{SQuAD: 100,000+ Questions for Machine Comprehension of Text}},
	author       = {Rajpurkar, Pranav and Zhang, Jian and Lopyrev, Konstantin and Liang, Percy},
	year         = 2016,
	journal      = {arXiv preprint arXiv:1606.05250}
}

@inproceedings{optimal_brain_surgeon,
	title        = {{Optimal Brain Surgeon and General Network Pruning}},
	author       = {Hassibi, Babak and Stork, David G and Wolff, Gregory J},
	year         = 1993,
	booktitle    = {IEEE International Conference on Neural Networks},
	pages        = {293--299},
	organization = {IEEE}
}

@article{optimal_brain_damage,
	title        = {{Optimal Brain Damage}},
	author       = {LeCun, Yann and Denker, John and Solla, Sara},
	year         = 1989,
	journal      = {Advances in Neural Information Processing Systems},
	volume       = 2
}

@inproceedings{lamb,
	title        = {{Large Batch Optimization for Deep Learning: Training BERT in 76 Minutes}},
	author       = {You, Yang and Li, Jing and Reddi, Sashank and Hseu, Jonathan and others},
	year         = 2020,
	booktitle    = {ICLR}
}

@misc{nvidia-bert,
	title        = {{NVIDIA Deep Learning Examples}},
	author       = {{NVIDIA Corporation}},
	howpublished = {\url{https://github.com/NVIDIA/DeepLearningExamples}}
}

@misc{english-wikipedia,
	title        = {{Wikipedia Corpus}},
	author       = {{Wikipedia}},
	howpublished = {\url{https://meta.wikimedia.org/wiki/Data_dump_torrents#English_Wikipedia}}
}

@inproceedings{bookcorpus,
	title        = {{Aligning Books and Movies: Towards Story-Like Visual Explanations by Watching Movies and Reading Books}},
	author       = {Zhu, Yukun and Kiros, Ryan and Zemel, Rich and Salakhutdinov, Ruslan and others},
	year         = 2015,
	booktitle    = {Proceedings of the IEEE International Conference on Computer Vision},
	pages        = {19--27}
}

@inproceedings{adam,
	title        = {{Adam: A Method for Stochastic Optimization}},
	author       = {Kingma, Diederik P and Ba, Jimmy},
	year         = 2015,
	booktitle    = {ICLR}
}

@article{fim_vs_hessian,
	title        = {{New Insights and Perspectives on the Natural Gradient Method}},
	author       = {Martens, James},
	year         = 2020,
	journal      = {Journal of Machine Learning Research},
	publisher    = {JMLR.org},
	volume       = 21,
	number       = 1,
	pages        = {5776--5851}
}

@article{autoencoder,
	title        = {{Deep Learning in Neural Networks: An Overview}},
	author       = {Schmidhuber, J{\"u}rgen},
	year         = 2015,
	journal      = {Neural Networks},
	publisher    = {Elsevier},
	volume       = 61,
	pages        = {85--117}
}

@misc{polaris,
	title        = {{Polaris}},
	author       = {{Argonne Leadership Computing Facility}},
	howpublished = {\url{https://www.alcf.anl.gov/polaris}}
}

@misc{eva,
	title        = {{EVA: A General Vectorized Approximation Framework for Second-Order Optimization}},
	author       = {Lin Zhang and Shaohuai Shi and Bo Li},
	year         = 2023,
	eprint       = {2308.02123},
	archiveprefix = {arXiv},
	primaryclass = {cs.LG}
}

@article{llama2,
	title        = {{LLaMA 2: Open Foundation and Fine-Tuned Chat Models}},
	author       = {Touvron, Hugo and Martin, Louis and Stone, Kevin and Albert, Peter and others},
	year         = 2023,
	journal      = {arXiv preprint arXiv:2307.09288}
}

@article{gpt2,
	title        = {{Language Models Are Unsupervised Multitask Learners}},
	author       = {Radford, Alec and Wu, Jeffrey and Child, Rewon and Luan, David and others},
	year         = 2019,
	journal      = {OpenAI Blog},
	volume       = 1,
	number       = 8,
	pages        = 9
}

@article{spdf,
	title        = {{SPDF: Sparse Pre-Training and Dense Fine-Tuning for Large Language Models}},
	author       = {Thangarasa, Vithursan and Gupta, Abhay and Marshall, William and Li, Tianda and others},
	year         = 2023,
	journal      = {arXiv preprint arXiv:2303.10464}
}

@article{n_m1,
	title        = {{Accelerated Sparse Neural Training: A Provable and Efficient Method to Find N:M Transposable Masks}},
	author       = {Hubara, Itay and Chmiel, Brian and Island, Moshe and Banner, Ron and others},
	year         = 2021,
	journal      = {NeurIPS}
}

@article{n_m2,
	title        = {{Bi-Directional Masks for Efficient N:M Sparse Training}},
	author       = {Zhang, Yuxin and Luo, Yiting and Lin, Mingbao and Zhong, Yunshan and others},
	year         = 2023,
	journal      = {arXiv preprint arXiv:2302.06058}
}

@inproceedings{n_m3_dominosearch,
	title        = {{DominoSearch: Find Layer-Wise Fine-Grained N:M Sparse Schemes from Dense Neural Networks}},
	author       = {Sun, Wei and Zhou, Aojun and Stuijk, Sander and Wijnhoven, Rob and others},
	year         = 2021,
	booktitle    = {NeurIPS}
}

@inproceedings{lora,
	title        = {{LoRA: Low-Rank Adaptation of Large Language Models}},
	author       = {Hu, Edward J and Shen, Yelong and Wallis, Phillip and Allen-Zhu, Zeyuan and others},
	year         = 2022,
	booktitle    = {ICLR}
}

@inproceedings{qlora,
	title        = {{QLoRA: Efficient Finetuning of Quantized LLMs}},
	author       = {Dettmers, Tim and Pagnoni, Artidoro and Holtzman, Ari and Zettlemoyer, Luke},
	year         = 2023,
	booktitle    = {NeurIPS}
}

@inproceedings{lqlora,
	title        = {{LQ-LoRA: Low-Rank Plus Quantized Matrix Decomposition for Efficient Language Model Finetuning}},
	author       = {Guo, Han and Greengard, Philip and Xing, Eric P and Kim, Yoon},
	year         = 2024,
	booktitle    = {ICLR}
}

@article{optimal_brain_compression,
	title        = {{Optimal Brain Compression: A Framework for Accurate Post-Training Quantization and Pruning}},
	author       = {Frantar, Elias and Alistarh, Dan},
	year         = 2022,
	journal      = {NeurIPS},
	volume       = 35,
	pages        = {4475--4488}
}

@article{fine_tuning1,
	title        = {{The State of Sparsity in Deep Neural Networks}},
	author       = {Gale, Trevor and Elsen, Erich and Hooker, Sara},
	year         = 2019,
	journal      = {arXiv preprint arXiv:1902.09574}
}

@article{fine_tuning2,
	title        = {{Deep Compression: Compressing Deep Neural Networks with Pruning, Trained Quantization and Huffman Coding}},
	author       = {Han, Song and Mao, Huizi and Dally, William J},
	year         = 2015,
	journal      = {arXiv preprint arXiv:1510.00149}
}

@article{fine_tuning3,
	title        = {{Learning Both Weights and Connections for Efficient Neural Network}},
	author       = {Han, Song and Pool, Jeff and Tran, John and Dally, William},
	year         = 2015,
	journal      = {NeurIPS}
}

@inproceedings{lottery_tickets,
	title        = {{Linear Mode Connectivity and the Lottery Ticket Hypothesis}},
	author       = {Frankle, Jonathan and Dziugaite, Gintare Karolina and Roy, Daniel and Carbin, Michael},
	year         = 2020,
	booktitle    = {ICML}
}

@inproceedings{pixelated_butterfly,
	title        = {{Pixelated Butterfly: Simple and Efficient Sparse Training for Neural Network Models}},
	author       = {Dao, Tri and Chen, Beidi and Liang, Kaizhao and Yang, Jiaming and others},
	year         = 2022,
	booktitle    = {ICLR}
}

@inproceedings{kaisa,
	title        = {{KAISA: An Adaptive Second-Order Optimizer Framework for Deep Neural Networks}},
	author       = {Pauloski, J Gregory and Huang, Qi and Huang, Lei and Venkataraman, Shivaram and others},
	year         = 2021,
	booktitle    = {SC}
}

@inproceedings{mkor,
	title        = {{MKOR: Momentum-Enabled Kronecker-Factor-Based Optimizer Using Rank-1 Updates}},
	author       = {Mohammad Mozaffari and Sikan Li and Zhao Zhang and Maryam Mehri Dehnavi},
	year         = 2023,
	booktitle    = {NeurIPS}
}

@misc{openwebtext,
	title        = {{OpenWebText Corpus}},
	author       = {Aaron Gokaslan and Vanya Cohen and Ellie Pavlick and Stefanie Tellex},
	year         = 2019
}

@inproceedings{flashattention,
	title        = {{FlashAttention: Fast and Memory-Efficient Exact Attention with IO-Awareness}},
	author       = {Dao, Tri and Fu, Daniel Y. and Ermon, Stefano and Rudra, Atri and others},
	year         = 2022,
	booktitle    = {NeurIPS}
}

@inproceedings{flashattention2,
	title        = {{FlashAttention-2: Faster Attention with Better Parallelism and Work Partitioning}},
	author       = {Dao, Tri},
	year         = 2024,
	booktitle    = {ICLR}
}

@inproceedings{adamw,
	title        = {{Decoupled Weight Decay Regularization}},
	author       = {Loshchilov, Ilya and Hutter, Frank},
	year         = 2019,
	booktitle    = {ICLR}
}

@article{pile,
	title        = {{The Pile: An 800GB Dataset of Diverse Text for Language Modeling}},
	author       = {Gao, Leo and Biderman, Stella and Black, Sid and Golding, Laurence and others},
	year         = 2020,
	journal      = {arXiv preprint arXiv:2101.00027}
}

@inproceedings{rosa,
	title        = {{RoSA: Accurate Parameter-Efficient Fine-Tuning via Robust Adaptation}},
	author       = {Nikdan, Mahdi and Tabesh, Soroush and Alistarh, Dan},
	year         = 2024,
	booktitle    = {ICML}
}

@inproceedings{int8,
	title        = {{GPT3.int8(): 8-Bit Matrix Multiplication for Transformers at Scale}},
	author       = {Dettmers, Tim and Lewis, Mike and Belkada, Younes and Zettlemoyer, Luke},
	year         = 2022,
	booktitle    = {NeurIPS},
	publisher    = {Curran Associates, Inc.},
	pages        = {30318--30332},
	editor       = {S. Koyejo and S. Mohamed and A. Agarwal and D. Belgrave and K. Cho and A. Oh}
}

@article{scatterbrain,
	title        = {{Scatterbrain: Unifying Sparse and Low-Rank Attention Approximation}},
	author       = {Chen, Beidi and Dao, Tri and Winsor, Eric and Song, Zhao and others},
	year         = 2021,
	journal      = {arXiv preprint arXiv:2110.15343}
}

@article{knowledge_distillation,
	title        = {{Knowledge Distillation: A Survey}},
	author       = {Gou, Jianping and Yu, Baosheng and Maybank, Stephen J and Tao, Dacheng},
	year         = 2021,
	journal      = {International Journal of Computer Vision},
	publisher    = {Springer},
	volume       = 129,
	number       = 6,
	pages        = {1789--1819}
}

@article{sparsity_survey,
	title        = {{Sparsity in Deep Learning: Pruning and Growth for Efficient Inference and Training in Neural Networks}},
	author       = {Hoefler, Torsten and Alistarh, Dan and Ben-Nun, Tal and Dryden, Nikoli and others},
	year         = 2021,
	journal      = {JMLR}
}

@article{pretraining,
	title        = {{Improving Language Understanding by Generative Pre-Training}},
	author       = {Radford, Alec and Narasimhan, Karthik and Salimans, Tim and Sutskever, Ilya and others},
	year         = 2018,
	journal      = {OpenAI}
}

@inproceedings{glue,
	title        = {{GLUE: A Multi-Task Benchmark and Analysis Platform for Natural Language Understanding}},
	author       = {Wang, Alex and Singh, Amanpreet and Michael, Julian and Hill, Felix and others},
	year         = 2019,
	booktitle    = {ICLR}
}

@article{block_pruning1,
	title        = {{TETRIS: Tile-Matching the Tremendous Irregular Sparsity}},
	author       = {Ji, Yu and Liang, Ling and Deng, Lei and Zhang, Youyang and others},
	year         = 2018,
	journal      = {NeurIPS}
}

@inproceedings{channel_pruning1,
	title        = {{MetaPruning: Meta Learning for Automatic Neural Network Channel Pruning}},
	author       = {Liu, Zechun and Mu, Haoyuan and Zhang, Xiangyu and Guo, Zichao and others},
	year         = 2019,
	booktitle    = {ICCV}
}

@inproceedings{fmmformer,
	title        = {{FMMformer: Efficient and Flexible Transformer via Decomposed Near-Field and Far-Field Attention}},
	author       = {Nguyen, Tan and Suliafu, Vai and Osher, Stanley and Chen, Long and others},
	year         = 2021,
	booktitle    = {NeurIPS}
}

@inproceedings{losparse,
	title        = {{LoSparse: Structured Compression of Large Language Models Based on Low-Rank and Sparse Approximation}},
	author       = {Li, Yixiao and Yu, Yifan and Zhang, Qingru and Liang, Chen and others},
	year         = 2023,
	booktitle    = {ICML}
}

@article{sparse_plus_low_rank,
	title        = {{Sparse Plus Low Rank Matrix Decomposition: A Discrete Optimization Approach}},
	author       = {Bertsimas, Dimitris and Cory-Wright, Ryan and Johnson, Nicholas AG},
	year         = 2023,
	journal      = {JMLR}
}

@article{n_m5_step,
	title        = {{STEP: Learning N:M Structured Sparsity Masks from Scratch with Precondition}},
	author       = {Lu, Yucheng and Agrawal, Shivani and Subramanian, Suvinay and Rybakov, Oleg and others},
	year         = 2023,
	journal      = {arXiv preprint arXiv:2302.01172}
}

@article{n_m4,
	title        = {{Training Recipe for N:M Structured Sparsity with Decaying Pruning Mask}},
	author       = {Kao, Sheng-Chun and Yazdanbakhsh, Amir and Subramanian, Suvinay and Agrawal, Shivani and others},
	year         = 2022,
	journal      = {arXiv preprint arXiv:2209.07617}
}

@misc{computecanada,
	title        = {{Compute Canada}},
	author       = {{Compute Canada}},
	howpublished = {\url{https://computecanada.ca/}}
}

@misc{lonestar6,
	title        = {{Lonestar 6}},
	author       = {{Texas Advanced Computing Center}},
	howpublished = {\url{https://tacc.utexas.edu/systems/lonestar6/}}
}

@article{nm_scratch,
	title        = {{Learning N:M Fine-Grained Structured Sparse Neural Networks from Scratch}},
	author       = {Zhou, Aojun and Ma, Yukun and Zhu, Junnan and Liu, Jianbo and others},
	year         = 2021,
	journal      = {arXiv preprint arXiv:2102.04010}
}

@misc{sparse_tensor_cores,
	title        = {{NVIDIA Ampere Architecture In-Depth}},
	author       = {{NVIDIA Corporation}},
	howpublished = {\url{https://developer.nvidia.com/blog/nvidia-ampere-architecture-in-depth}}
}

@article{lucas,
	title        = {{Register Tiling for Unstructured Sparsity in Neural Network Inference}},
	author       = {Wilkinson, Lucas and Cheshmi, Kazem and Dehnavi, Maryam Mehri},
	journal      = {PLDI},
	year         = 2023
}

@article{adaptive_pruning,
	title        = {{Sparse Networks from Scratch: Faster Training without Losing Performance}},
	author       = {Dettmers, Tim and Zettlemoyer, Luke},
	year         = 2019,
	journal      = {arXiv preprint arXiv:1907.04840}
}

@misc{cuspaselt_reduction_dim,
	title        = {{NVIDIA cuSPARSELt Functions}},
	author       = {{NVIDIA Corporation}},
	howpublished = {\url{https://docs.nvidia.com/cuda/cusparselt/functions.html}}
}

@misc{cuda,
	title        = {{CUDA, Release: 10.2.89}},
	author       = {NVIDIA and Vingelmann, P{\'e}ter and Fitzek, Frank H.P.},
	year         = 2020,
	url          = {https://developer.nvidia.com/cuda-toolkit}
}

@article{opt,
	title        = {{OPT: Open Pre-Trained Transformer Language Models}},
	author       = {Zhang, Susan and Roller, Stephen and Goyal, Naman and Artetxe, Mikel and others},
	year         = 2022,
	journal      = {arXiv preprint arXiv:2205.01068}
}

@inproceedings{2_4_pretraining,
	title        = {{Accelerating Transformer Pre-Training with 2:4 Sparsity}},
	author       = {Hu, Yuezhou and Zhao, Kang and Huang, Weiyu and Chen, Jianfei and others},
	year         = 2024,
	booktitle    = {ICML}
}

@article{roofline,
	title        = {{Roofline: An Insightful Visual Performance Model for Multicore Architectures}},
	author       = {Williams, Samuel and Waterman, Andrew and Patterson, David},
	year         = 2009,
	journal      = {Communications of the ACM},
	publisher    = {ACM New York, NY, USA}
}

@article{perplexity,
	title        = {{Evaluation Metrics for Language Models}},
	author       = {Chen, Stanley F and Beeferman, Douglas and Rosenfeld, Roni},
	year         = 1998,
	journal      = {Carnegie Mellon University}
}

@inproceedings{nm_attention,
	title        = {{Dynamic N:M Fine-Grained Structured Sparse Attention Mechanism}},
	author       = {Chen, Zhaodong and Qu, Zheng and Quan, Yuying and Liu, Liu and others},
	year         = 2023,
	booktitle    = {PPoPP}
}

@inproceedings{perplexity2,
	title        = {{Same Pre-Training Loss, Better Downstream: Implicit Bias Matters for Language Models}},
	author       = {Liu, Hong and Xie, Sang Michael and Li, Zhiyuan and Ma, Tengyu},
	year         = 2023,
	booktitle    = {ICML}
}

@article{perplexity3,
	title        = {{On the Effect of Pretraining Corpora on In-Context Learning by a Large-Scale Language Model}},
	author       = {Shin, Seongjin and Lee, Sang-Woo and Ahn, Hwijeen and Kim, Sungdong and others},
	year         = 2022,
	journal      = {arXiv preprint arXiv:2204.13509}
}

@article{mmlu,
	title        = {{Measuring Massive Multitask Language Understanding}},
	author       = {Hendrycks, Dan and Burns, Collin and Basart, Steven and Zou, Andy and others},
	year         = 2020,
	journal      = {arXiv preprint arXiv:2009.03300}
}

@article{arc,
	title        = {{Think You Have Solved Question Answering? Try ARC, the AI2 Reasoning Challenge}},
	author       = {Clark, Peter and Cowhey, Isaac and Etzioni, Oren and Khot, Tushar and others},
	year         = 2018,
	journal      = {arXiv preprint arXiv:1803.05457}
}

@inproceedings{openbookqa,
	title        = {{Can a Suit of Armor Conduct Electricity? A New Dataset for Open Book Question Answering}},
	author       = {Todor Mihaylov and Peter Clark and Tushar Khot and Ashish Sabharwal},
	year         = 2018,
	booktitle    = {EMNLP}
}

@misc{lmharness,
	title        = {{A Framework for Few-Shot Language Model Evaluation}},
	author       = {Gao, Leo and Tow, Jonathan and Abbasi, Baber and Biderman, Stella and others},
	year         = 2024,
	month        = {07},
	publisher    = {Zenodo},
	doi          = {10.5281/zenodo.12608602},
	url          = {https://zenodo.org/records/12608602},
	version      = {v0.4.3}
}

@article{gemma2,
	title        = {{Gemma 2: Improving Open Language Models at a Practical Size}},
	author       = {Team, Gemma and Riviere, Morgane and Pathak, Shreya and Sessa, Pier Giuseppe and others},
	year         = 2024,
	journal      = {arXiv preprint arXiv:2408.00118}
}

@article{movement_pruning,
	title        = {{Movement Pruning: Adaptive Sparsity by Fine-Tuning}},
	author       = {Sanh, Victor and Wolf, Thomas and Rush, Alexander},
	year         = 2020,
	journal      = {NeurIPS}
}

@inproceedings{quantization_aware_training,
	title        = {{Value-Aware Quantization for Training and Inference of Neural Networks}},
	author       = {Park, Eunhyeok and Yoo, Sungjoo and Vajda, Peter},
	year         = 2018,
	booktitle    = {ECCV}
}

@article{obs,
	title        = {{Optimal Brain Surgeon: Extensions and Performance Comparisons}},
	author       = {Hassibi, Babak and Stork, David and Wolff, Gregory},
	year         = 1993,
	journal      = {NeurIPS}
}

@inproceedings{wanda,
	title        = {{A Simple and Effective Pruning Approach for Large Language Models}},
	author       = {Sun, Mingjie and Liu, Zhuang and Bair, Anna and Kolter, J Zico},
	year         = 2024,
	booktitle    = {ICLR}
}

@inproceedings{sparsegpt,
	title        = {{SparseGPT: Massive Language Models Can Be Accurately Pruned in One-Shot}},
	author       = {Frantar, Elias and Alistarh, Dan},
	year         = 2023,
	booktitle    = {ICML}
}

@inproceedings{optq,
	title        = {{OPTQ: Accurate Quantization for Generative Pre-Trained Transformers}},
	author       = {Frantar, Elias and Ashkboos, Saleh and Hoefler, Torsten and Alistarh, Dan},
	year         = 2022,
	booktitle    = {ICLR}
}

@inproceedings{group_quantization_1,
	title        = {{QSGD: Randomized Quantization for Communication-Efficient Stochastic Gradient Descent}},
	author       = {Alistarh, Dan and Grubic, Demjan and Li, Jerry and Tomioka, Ryota and others},
	year         = 2017,
	booktitle    = {NeurIPS}
}

@article{group_quantization_2,
	title        = {{nuQmm: Quantized MatMul for Efficient Inference of Large-Scale Generative Language Models}},
	author       = {Gunho, Park and Baeseong, Park and Se Jung, Kwon and Byeongwook, Kim and others},
	year         = 2022,
	journal      = {arXiv preprint arXiv:2206.09557}
}

@article{woodfisher,
	title        = {{WoodFisher: Efficient Second-Order Approximation for Neural Network Compression}},
	author       = {Singh, Sidak Pal and Alistarh, Dan},
	year         = 2020,
	journal      = {NeurIPS}
}

@inproceedings{piqa,
	title        = {{PIQA: Reasoning About Physical Commonsense in Natural Language}},
	author       = {Bisk, Yonatan and Zellers, Rowan and Gao, Jianfeng and Choi, Yejin and others},
	year         = 2020,
	booktitle    = {AAAI}
}

@misc{wikitext2,
	title        = {{Pointer Sentinel Mixture Models}},
	author       = {Stephen Merity and Caiming Xiong and James Bradbury and Richard Socher},
	year         = 2016,
	eprint       = {1609.07843},
	archiveprefix = {arXiv},
	primaryclass = {cs.CL}
}

@article{c4,
	title        = {{Exploring the Limits of Transfer Learning with a Unified Text-to-Text Transformer}},
	author       = {Colin Raffel and Noam Shazeer and Adam Roberts and Katherine Lee and others},
	year         = 2019,
	journal      = {arXiv e-prints},
	archiveprefix = {arXiv},
	eprint       = {1910.10683}
}

@inproceedings{absmax,
	title        = {{Quantization and Training of Neural Networks for Efficient Integer-Arithmetic-Only Inference}},
	author       = {Jacob, Benoit and Kligys, Skirmantas and Chen, Bo and Zhu, Menglong and others},
	year         = 2018,
	booktitle    = {CVPR}
}

@misc{cusparselt,
	title        = {{NVIDIA cuSPARSELt}},
	author       = {{NVIDIA Corporation}},
	howpublished = {\url{https://docs.nvidia.com/cuda/cusparselt/index.html}}
}

@misc{cutlass,
	title        = {{CUTLASS 4.2.0: CUDA Templates for Linear Algebra Subroutines}},
	author       = {NVIDIA Corporation},
	year         = 2025,
	note         = {Also see: Kerr, A., Merrill, D., Demouth, J., Tran, J. "CUTLASS: Fast Linear Algebra in CUDA C++", NVIDIA blog, Dec. 2017},
	howpublished = {\url{https://github.com/NVIDIA/cutlass}}
}

@inproceedings{huggingface,
	title        = {{Transformers: State-of-the-Art Natural Language Processing}},
	author       = {Thomas Wolf and Lysandre Debut and Victor Sanh and Julien Chaumond and others},
	year         = 2020,
	booktitle    = {EMNLP (System Demonstrations)},
	pages        = {38--45}
}

@misc{bfloat,
	title        = {{BFloat16: The Secret to High Performance on Cloud TPUs}},
	author       = {Wang, Shibo and Kanwar, Pankaj},
	year         = 2019,
	howpublished = {\url{http://bit.ly/3WEtCGm}}
}

@incollection{quantization_survey,
	title        = {{A Survey of Quantization Methods for Efficient Neural Network Inference}},
	author       = {Gholami, Amir and Kim, Sehoon and Dong, Zhen and Yao, Zhewei and others},
	year         = 2022,
	booktitle    = {{Low-Power Computer Vision}},
	publisher    = {Chapman and Hall/CRC},
	pages        = {291--326}
}

@article{quantization_survey2,
	title        = {{A Comprehensive Survey on Model Quantization for Deep Neural Networks in Image Classification}},
	author       = {Rokh, Babak and Azarpeyvand, Ali and Khanteymoori, Alireza},
	year         = 2023,
	journal      = {ACM Transactions on Intelligent Systems and Technology},
	publisher    = {ACM New York, NY},
	volume       = 14,
	number       = 6,
	pages        = {1--50}
}

@article{awq,
	title        = {{AWQ: Activation-Aware Weight Quantization for On-Device LLM Compression and Acceleration}},
	author       = {Lin, Ji and Tang, Jiaming and Tang, Haotian and Yang, Shang and others},
	year         = 2024,
	journal      = {MLSys}
}

@inproceedings{omniquant,
	title        = {{OmniQuant: Omnidirectionally Calibrated Quantization for Large Language Models}},
	author       = {Shao, Wenqi and Chen, Mengzhao and Zhang, Zhaoyang and Xu, Peng and others},
	year         = 2024,
	booktitle    = {ICLR}
}

@article{affinequant,
	title        = {{AffineQuant: Affine Transformation Quantization for Large Language Models}},
	author       = {Ma, Yuexiao and Li, Huixia and Zheng, Xiawu and Ling, Feng and others},
	year         = 2024,
	journal      = {arXiv preprint arXiv:2403.12544}
}

@inproceedings{jsq,
	title        = {{Compressing Large Language Models by Joint Sparsification and Quantization}},
	author       = {Guo, Jinyang and Wu, Jianyu and Wang, Zining and Liu, Jiaheng and others},
	year         = 2024,
	booktitle    = {ICML}
}

@article{lqer,
	title        = {{LQER: Low-Rank Quantization Error Reconstruction for LLMs}},
	author       = {Zhang, Cheng and Cheng, Jianyi and Constantinides, George A and Zhao, Yiren},
	year         = 2024,
	journal      = {arXiv preprint arXiv:2402.02446}
}

@inproceedings{quarot,
	title        = {{QuaRot: Outlier-Free 4-Bit Inference in Rotated LLMs}},
	author       = {Ashkboos, Saleh and Mohtashami, Amirkeivan and Croci, Maximilian L and Li, Bo and others},
	year         = 2024,
	booktitle    = {NeurIPS}
}

@article{marlin,
	title        = {{MARLIN: Mixed-Precision Auto-Regressive Parallel Inference on Large Language Models}},
	author       = {Frantar, Elias and Castro, Roberto L and Chen, Jiale and Hoefler, Torsten and others},
	year         = 2024,
	journal      = {arXiv preprint arXiv:2408.11743}
}

@inproceedings{vllm,
	title        = {{Efficient Memory Management for Large Language Model Serving with PagedAttention}},
	author       = {Woosuk Kwon and Zhuohan Li and Siyuan Zhuang and Ying Sheng and others},
	year         = 2023,
	booktitle    = {SOSP}
}

@inproceedings{sparta,
	title        = {{{SparTA}: {Deep-Learning} Model Sparsity via {Tensor-with-Sparsity-Attribute}}},
	author       = {Zheng, Ningxin and Lin, Bin and Zhang, Quanlu and Ma, Lingxiao and others},
	year         = 2022,
	booktitle    = {OSDI}
}

@article{flash_llm,
	title        = {{Flash-LLM: Enabling Cost-Effective and Highly-Efficient Large Generative Model Inference with Unstructured Sparsity}},
	author       = {Xia, Haojun and Zheng, Zhen and Li, Yuchao and Zhuang, Donglin and others},
	journal      = {arXiv preprint arXiv:2309.10285},
	year         = 2023
}

@inproceedings{triton,
	title        = {{Triton: An Intermediate Language and Compiler for Tiled Neural Network Computations}},
	author       = {Tillet, Philippe and Kung, H. T. and Cox, David},
	booktitle    = {MAPL 2019: Proceedings of the 3rd ACM SIGPLAN International Workshop on Machine Learning and Programming Languages},
	year         = 2019
}

@inproceedings{smoothquant,
	title        = {{SmoothQuant: Accurate and Efficient Post-Training Quantization for Large Language Models}},
	author       = {Xiao, Guangxuan and Lin, Ji and Seznec, Mickael and Wu, Hao and others},
	year         = 2023,
	booktitle    = {ICML}
}

@inproceedings{adafactor,
	title        = {{Adafactor: Adaptive Learning Rates with Sublinear Memory Cost}},
	author       = {Shazeer, Noam and Stern, Mitchell},
	booktitle    = {ICML},
	year         = 2018
}

@misc{slimpajama,
	title        = {{SlimPajama: A 627B Token Cleaned and Deduplicated Version of RedPajama}},
	author       = {Soboleva, Daria and Al-Khateeb, Faisal and Myers, Robert and Steeves, Jacob R and others},
	year         = 2023,
	url          = {https://huggingface.co/datasets/cerebras/SlimPajama-627B},
	howpublished = {\url{https://bit.ly/slimpajamas}}
}

@article{llama3,
	title        = {{The LLaMA 3 Herd of Models}},
	author       = {Dubey, Abhimanyu and Jauhri, Abhinav and Pandey, Abhinav and Kadian, Abhishek and others},
	year         = 2024,
	journal      = {arXiv preprint arXiv:2407.21783}
}

@article{fp8,
	title        = {{FP8 Formats for Deep Learning}},
	author       = {Micikevicius, Paulius and Stosic, Dusan and Burgess, Neil and Cornea, Marius and others},
	year         = 2022,
	journal      = {arXiv preprint arXiv:2209.05433}
}

@inproceedings{maskllm,
	title        = {{MaskLLM: Learnable Semi-Structured Sparsity for Large Language Models}},
	author       = {Fang, Gongfan and Yin, Hongxu and Muralidharan, Saurav and Heinrich, Greg and others},
	year         = 2024,
	booktitle    = {NeurIPS}
}

@article{gradient_flow,
	title        = {{Progressive Gradient Flow for Robust N:M Sparsity Training in Transformers}},
	author       = {Bambhaniya, Abhimanyu Rajeshkumar and Yazdanbakhsh, Amir and Subramanian, Suvinay and Kao, Sheng-Chun and others},
	year         = 2024,
	journal      = {arXiv preprint arXiv:2402.04744}
}

@article{thanos,
	title        = {{Thanos: A Block-Wise Pruning Algorithm for Efficient Large Language Model Compression}},
	author       = {Ilin, Ivan and Richtarik, Peter},
	year         = 2025,
	journal      = {arXiv preprint arXiv:2504.05346}
}

@inproceedings{slim,
	title        = {{SLiM: One-Shot Quantized Sparse Plus Low-Rank Approximation of LLMs}},
	author       = {Mozaffari, Mohammad and Yazdanbakhsh, Amir and Mehri Dehnavi, Maryam},
	year         = 2025,
	booktitle    = {ICML},
	url          = {https://openreview.net/forum?id=4UfRP8MopP}
}

@inproceedings{slope,
	title        = {{SLoPe: Double-Pruned Sparse Plus Lazy Low-Rank Adapter Pretraining of LLMs}},
	author       = {Mozaffari, Mohammad and Yazdanbakhsh, Amir and Zhang, Zhao and Dehnavi, Maryam Mehri},
	year         = 2025,
	booktitle    = {ICLR}
}

@article{gemma3,
	title        = {{Gemma 3 Technical Report}},
	author       = {Team, Gemma and Kamath, Aishwarya and Ferret, Johan and Pathak, Shreya and others},
	year         = 2025,
	journal      = {arXiv preprint arXiv:2503.19786}
}

@article{cerebras_sparsity,
	title        = {{Enabling High-Sparsity Foundational LLaMA Models with Efficient Pretraining and Deployment}},
	author       = {Agarwalla, Abhinav and Gupta, Abhay and Marques, Alexandre and Pandit, Shubhra and others},
	year         = 2024,
	journal      = {arXiv preprint arXiv:2405.03594}
}

@article{pdqp,
	title        = {{A Practical and Optimal First-Order Method for Large-Scale Convex Quadratic Programming}},
	author       = {Lu, Haihao and Yang, Jinwen},
	year         = 2023,
	journal      = {arXiv preprint arXiv:2311.07710}
}

@article{mpax,
	title        = {{MPAX: Mathematical Programming in JAX}},
	author       = {Lu, Haihao and Peng, Zedong and Yang, Jinwen},
	year         = 2024,
	journal      = {arXiv preprint arXiv:2412.09734}
}

@article{proxsparse,
	title        = {{ProxSparse: Regularized Learning of Semi-Structured Sparsity Masks for Pretrained LLMs}},
	author       = {Liu, Hongyi and Saha, Rajarshi and Jia, Zhen and Park, Youngsuk and others},
	year         = 2025,
	journal      = {arXiv preprint arXiv:2502.00258}
}

@inproceedings{gumbelsoftmax,
	title        = {{Categorical Reparameterization with Gumbel-Softmax}},
	author       = {Jang, Eric and Gu, Shixiang and Poole, Ben},
	year         = 2017,
	booktitle    = {ICLR}
}

@inproceedings{owl,
	title        = {{Outlier Weighed Layerwise Sparsity (OWL): A Missing Secret Sauce for Pruning LLMs to High Sparsity}},
	author       = {Lu Yin and You Wu and Zhenyu Zhang and Cheng-Yu Hsieh and others},
	year         = 2024,
	booktitle    = {ICML}
}

@misc{rethinkingvaluetransformercomponents,
	title        = {{Rethinking the Value of Transformer Components}},
	author       = {Wenxuan Wang and Zhaopeng Tu},
	year         = 2020,
	url          = {https://arxiv.org/abs/2011.03803},
	eprint       = {2011.03803},
	archiveprefix = {arXiv},
	primaryclass = {cs.CL}
}

@misc{layeradaptivesparsitymagnitudebasedpruning,
	title        = {{Layer-Adaptive Sparsity for the Magnitude-Based Pruning}},
	author       = {Jaeho Lee and Sejun Park and Sangwoo Mo and Sungsoo Ahn and others},
	year         = 2021,
	url          = {https://arxiv.org/abs/2010.07611},
	eprint       = {2010.07611},
	archiveprefix = {arXiv},
	primaryclass = {cs.LG}
}

@misc{qwen,
	title        = {{Qwen2.5 Technical Report}},
	author       = {Qwen and An Yang and Baosong Yang and others},
	year         = 2025,
	url          = {https://arxiv.org/abs/2412.15115},
	eprints      = {2412.15115},
	archiveprefix = {arXiv},
	primaryclass = {cs.CL}
}

@article{wiki,
	title        = {{Pointer Sentinel Mixture Models}},
	author       = {Merity, Stephen and Xiong, Caiming and Bradbury, James and Socher, Richard},
	year         = 2016,
	journal      = {arXiv preprint arXiv:1609.07843}
}

@misc{eval-harness,
	title        = {{The Language Model Evaluation Harness}},
	author       = {Gao, Leo and Tow, Jonathan and Abbasi, Baber and Biderman, Stella and others},
	year         = 2024,
	month        = {07},
	publisher    = {Zenodo},
	doi          = {10.5281/zenodo.12608602},
	url          = {https://zenodo.org/records/12608602},
	version      = {v0.4.3}
}

@inproceedings{race,
	title        = {{{RACE}: Large-Scale {Re}{A}ding Comprehension Dataset from Examinations}},
	author       = {Lai, Guokun and Xie, Qizhe and Liu, Hanxiao and Yang, Yiming and others},
	year         = 2017,
	month        = sep,
	booktitle    = {EMNLP},
	publisher    = {Association for Computational Linguistics},
	address      = {Copenhagen, Denmark},
	pages        = {785--794},
	doi          = {10.18653/v1/D17-1082},
	url          = {https://aclanthology.org/D17-1082},
	editor       = {Palmer, Martha and Hwa, Rebecca and Riedel, Sebastian}
}

@inproceedings{hellaswag,
	title        = {{HellaSwag: Can a Machine Really Finish Your Sentence?}},
	author       = {Zellers, Rowan and Holtzman, Ari and Bisk, Yonatan and Farhadi, Ali and others},
	year         = 2019,
	booktitle    = {Proceedings of the 57th Annual Meeting of the Association for Computational Linguistics}
}

@article{magnitudepruning,
	title        = {{Learning Both Weights and Connections for Efficient Neural Network}},
	author       = {Han, Song and Pool, Jeff and Tran, John and Dally, William},
	journal      = {Advances in Neural Information Processing Systems},
	volume       = 28,
	year         = 2015
}

@misc{stoicc,
	title        = {{Stoicc}},
	author       = {Rafii, Arya and Kamel, Victor and Mehri Dehnavi, Maryam},
	year         = 2025,
	howpublished = {\url{https://paramathic.github.io/stoicc-docs/}}
}

@article{adaptive1,
	title        = {{Adaptive Layer Sparsity for Large Language Models via Activation Correlation Assessment}},
	author       = {Li, Wei and Li, Lujun and Lee, Mark and Sun, Shengjie},
	year         = 2024,
	journal      = {Advances in Neural Information Processing Systems},
	volume       = 37,
	pages        = {109350--109380}
}

@article{adaptive2,
	title        = {{BESA: Pruning Large Language Models with Blockwise Parameter-Efficient Sparsity Allocation}},
	author       = {Xu, Peng and Shao, Wenqi and Chen, Mengzhao and Tang, Shitao and Zhang, Kaipeng and Gao, Peng and An, Fengwei and Qiao, Yu and Luo, Ping},
	year         = 2024,
	journal      = {arXiv preprint arXiv:2402.16880}
}

@article{adaptive3,
	title        = {{Discovering Sparsity Allocation for Layer-Wise Pruning of Large Language Models}},
	author       = {Li, Lujun and Dong, Peijie and Tang, Zhenheng and Liu, Xiang and Wang, Qiang and Luo, Wenhan and Xue, Wei and Liu, Qifeng and Chu, Xiaowen and Guo, Yike},
	year         = 2024,
	journal      = {Advances in Neural Information Processing Systems},
	volume       = 37,
	pages        = {141292--141317}
}

@misc{cublas,
	title        = {{NVIDIA cuBLAS}},
	author       = {{NVIDIA Corporation}},
	howpublished = {\url{https://docs.nvidia.com/cuda/cublas/}}
}

@inproceedings{spinfer,
	title        = {{SpInfer: Leveraging Low-Level Sparsity for Efficient Large Language Model Inference on GPUs}},
	author       = {Fan, Ruibo and Yu, Xiangrui and Dong, Peijie and Li, Zeyu and others},
	year         = 2025,
	booktitle    = {Proceedings of the Twentieth European Conference on Computer Systems},
	pages        = {243--260}
}

@article{2:4,
	title        = {{Accelerating Sparse Deep Neural Networks}},
	author       = {Mishra, Asit and Latorre, Jorge Albericio and Pool, Jeff and Stosic, Darko and others},
	year         = 2021,
	journal      = {arXiv preprint arXiv:2104.08378}
}

@misc{patterson2021carbon,
	title        = {{Carbon Emissions and Large Neural Network Training}},
	author       = {David Patterson and Joseph Gonzalez and Quoc V. Le and Chen Liang and others},
	year         = 2021,
	eprint       = {2104.10350},
	archiveprefix = {arXiv},
	primaryclass = {cs.LG}
}

@article{swiglu,
	title        = {{GLU Variants Improve Transformer}},
	author       = {Shazeer, Noam},
	year         = 2020,
	journal      = {arXiv preprint arXiv:2002.05202}
}

@inproceedings{Aghajanyan2021,
	title        = {{Intrinsic Dimensionality Explains the Effectiveness of Language Model Fine-Tuning}},
	author       = {Aghajanyan, Armen and Gupta, Sonal and Zettlemoyer, Luke},
	year         = 2021,
	month        = {August},
	booktitle    = {Proceedings of the 59th Annual Meeting of the Association for Computational Linguistics and the 11th International Joint Conference on Natural Language Processing (Volume 1: Long Papers)},
	publisher    = {Association for Computational Linguistics},
	address      = {Online},
	pages        = {7319--7328},
	doi          = {10.18653/v1/2021.acl-long.568},
	url          = {https://aclanthology.org/2021.acl-long.568}
}

@article{winogrande,
	title        = {{WinoGrande: An Adversarial Winograd Schema Challenge at Scale}},
	author       = {Sakaguchi, Keisuke and Bras, Ronan Le and Bhagavatula, Chandra and Choi, Yejin},
	year         = 2021,
	journal      = {Communications of the ACM},
	publisher    = {ACM New York, NY, USA},
	volume       = 64,
	number       = 9,
	pages        = {99--106}
}

@article{patch,
	title        = {{PATCH: Learnable Tile-Level Hybrid Sparsity for LLMs}},
	author       = {Hourri, Younes and Mozaffari, Mohammad and Dehnavi, Maryam Mehri},
	journal      = {arXiv preprint arXiv:2509.23410},
	year         = {2025}
}

@misc{nvidia_a100_2020,
	title        = {{NVIDIA A100 Tensor Core GPU Architecture}},
	author       = {NVIDIA},
	year         = 2020,
	url          = {https://images.nvidia.com/aem-dam/en-zz/Solutions/data-center/nvidia-ampere-architecture-whitepaper.pdf},
	note         = {Version 1.0}
}

@misc{nvidia_h100_2022,
	title        = {{NVIDIA Hopper Architecture Whitepaper}},
	author       = {NVIDIA},
	year         = 2022,
	url          = {https://www.nvidia.com/content/dam/en-zz/Solutions/design-visualization/perf-analyzer/nvidia-hopper-architecture-whitepaper.pdf},
	note         = {Accessed via GTC 2022 presentation materials}
}

@misc{muon,
	title        = {{Muon: An Optimizer for Hidden Layers in Neural Networks}},
	author       = {Keller Jordan and Yuchen Jin and Vlado Boza and You Jiacheng and Franz Cesista and Laker Newhouse and Jeremy Bernstein},
	year         = 2024,
	url          = {https://kellerjordan.github.io/posts/muon/}
}

@inproceedings{intrinsic_dim,
	title        = {{Measuring the Intrinsic Dimension of Objective Landscapes}},
	author       = {Li, Chunyuan and Farkhoor, Heerad and Liu, Rosanne and Yosinski, Jason},
	booktitle    = {ICLR},
	year         = 2018,
	url          = {https://openreview.net/forum?id=ryup8-WCW}
}

@article{optima,
	title        = {{OPTIMA: Optimal One-Shot Pruning for LLMs via Quadratic Programming Reconstruction}},
	author       = {Mozaffari, Mohammad and Kushnir, Samuel and Dehnavi, Maryam Mehri and Yazdanbakhsh, Amir},
	journal      = {arXiv preprint arXiv:2512.13886},
	year         = 2025
}

@techreport{chen1998evaluation,
	title        = {{Evaluation Metrics for Language Models}},
	author       = {Chen, Stanley F. and Beeferman, Douglas and Rosenfeld, Roni},
	institution  = {Carnegie Mellon University},
	year         = 1998
}

@article{candes2011robust,
	title        = {{Robust Principal Component Analysis?}},
	author       = {Cand{\`e}s, Emmanuel J. and Li, Xiaodong and Ma, Yi and Wright, John},
	journal      = {Journal of the ACM},
	volume       = 58,
	number       = 3,
	pages        = {1--37},
	year         = 2011,
	publisher    = {ACM}
}

@article{chandrasekaran2011rank,
	title        = {{Rank-Sparsity Incoherence for Matrix Decomposition}},
	author       = {Chandrasekaran, Venkat and Sanghavi, Sujay and Parrilo, Pablo A. and Willsky, Alan S.},
	journal      = {SIAM Journal on Optimization},
	volume       = 21,
	number       = 2,
	pages        = {572--596},
	year         = 2011,
	publisher    = {SIAM}
}

@inproceedings{zhou2010stable,
	title        = {{Stable Principal Component Pursuit}},
	author       = {Zhou, Zihan and Li, Xiaodong and Wright, John and Cand{\`e}s, Emmanuel J. and Ma, Yi},
	booktitle    = {IEEE International Symposium on Information Theory (ISIT)},
	pages        = {1518--1522},
	year         = 2010,
	organization = {IEEE}
}

@inproceedings{yu2017compressing,
	title        = {{On Compressing Deep Models by Low Rank and Sparse Decomposition}},
	author       = {Yu, Xiyu and Liu, Tongliang and Wang, Xinchao and Tao, Dacheng},
	booktitle    = {IEEE Conference on Computer Vision and Pattern Recognition (CVPR)},
	pages        = {7370--7379},
	year         = 2017
}

@inproceedings{zhang2025oats,
	title        = {{OATS}: Outlier-Aware Pruning Through Sparse and Low Rank Decomposition},
	author       = {Zhang, Stephen and Papyan, Vardan},
	booktitle    = {The Thirteenth International Conference on Learning Representations (ICLR)},
	year         = 2025
}

@inproceedings{makni2025hassle,
	title        = {{A Unified Framework for Sparse Plus Low-Rank Matrix Decomposition for LLMs}},
	author       = {Makni, Mehdi and Behdin, Kayhan and Xu, Zheng and Ponomareva, Natalia and Mazumder, Rahul},
	booktitle    = {The Second Conference on Parsimony and Learning (Proceedings Track)},
	year         = 2025
}

@article{behdin2025basil,
	title        = {{3BASiL}: An Algorithmic Framework for Sparse Plus Low-Rank Compression of {LLM}s},
	author       = {Behdin, Kayhan and Makni, Mehdi and Mazumder, Rahul},
	journal      = {arXiv preprint arXiv:2603.01376},
	year         = 2025
}

  \appendix
\renewcommand{\sectionautorefname}{Appendix}
\renewcommand{\subsectionautorefname}{Appendix}

\chapter{Supplementary Material for MKOR}
\label{app:mkor}
In this chapter, we report the GLUE results achieved on each task in \autoref{subsection:glue}. Next, we discuss the derivation of KFAC and SNGD approximation methods in \autoref{subsection:derivation}. We then discuss some of the features of MKOR and other optimizers and back them up with quantitative data in \autoref{subsection:numerical_instability} and \autoref{subsection:sensitivity_to_lr}. We provide scalability results of MKOR in \autoref{subsection:scalability} and provide more data to back up the low-rank features of the covariance matrices in \autoref{subsection:decaying_eigenvalues}. In \autoref{subsection:training_accuracy_experiment}, we analyze MKOR and other optimizers on the training tasks as non-convex optimizers, only concerning their performance on training tasks. We also describe the knee-point learning rate scheduler in \autoref{subsection:knee_point_scheduler}. We conclude this chapter by proving the lemmas used in the preceding sections.

\section{GLUE Results}
\label{subsection:glue}
We discussed the speedup achieved using MKOR on the GLUE dataset in \autoref{sec:experiments}. For completeness, \autoref{table:glue_accuracy} shows the metrics achieved in each of the different GLUE tasks on BERT-Large-Uncased trained on different optimizers.
\begin{table}[hbpt]
    \centering
    \small
    \caption{BERT-Large-Uncased Results on the GLUE classification tasks.}
    \begin{tabularx}{1\textwidth} {
        | >{\centering\arraybackslash \hsize=1.6\hsize}X
        | >{\centering\arraybackslash \hsize=0.94\hsize}X
        | >{\centering\arraybackslash \hsize=0.94\hsize}X
        | >{\centering\arraybackslash \hsize=0.94\hsize}X
        | >{\centering\arraybackslash \hsize=0.94\hsize}X
        | >{\centering\arraybackslash \hsize=0.94\hsize}X
        | >{\centering\arraybackslash \hsize=0.94\hsize}X
        | >{\centering\arraybackslash \hsize=0.94\hsize}X
        | >{\centering\arraybackslash \hsize=0.94\hsize}X
        | >{\centering\arraybackslash \hsize=0.94\hsize}X
        | >{\centering\arraybackslash \hsize=0.94\hsize}X | }
        \hline
        \vspace{0.04in} Optimizer    & \vspace{0.01in}Iterati-ons & \vspace{0.01in}MNLI (acc) & \vspace{0.01in}QQP (F1) & \vspace{0.01in}QNLI (acc) & SST-2 (acc) & \vspace{0.01in}COLA (mcc)  & STS-B (corr) & \vspace{0.01in}MRPC (F1)  & \vspace{0.01in}RTE (acc)   & \vspace{0.01in}Avera-ge \\
        \hline
        LAMB                        & 1,563                      & 0.841 & 0.878 & 0.913 & 0.919 & 0.516 & 0.875 & 0.812 & 0.664 & 0.8023                 \\
        \hline
        KAISA                       & 1,563                      & 0.821 & 0.854 & 0.900 & 0.921 & 0.489 & 0.878 & 0.888 & 0.617 & 0.796                  \\
        \hline
        MKOR                        & 1,500                      & 0.844 & 0.879 & 0.916 & 0.923 & 0.523 & 0.892 & 0.905 & 0.690 & 0.8214                 \\
        \hline
        MKOR                        & 600                        & 0.833 & 0.878 & 0.904 & 0.921 & 0.494 & 0.886 & 0.893 & 0.653 & 0.8078                 \\
        \hline
        MKOR-H                        & 600                         & 0.838     & 0.877     & 0.911     & 0.921     & 0.502     & 0.886     & 0.898     & 0.657     & 0.811                 \\
        \hline
        Eva & 1000 & 0.839 & 0.877 & 0.907 & 0.914 & 0.499 & 0.890 & 0.904 & 0.650 & 0.809 \\
        \hline
        
    \end{tabularx}
    \label{table:glue_accuracy}
\end{table}

\section{Derivation of NGD Approximations}
\label{subsection:derivation}

\begin{figure}
    \centering
    \includegraphics[width=\textwidth]{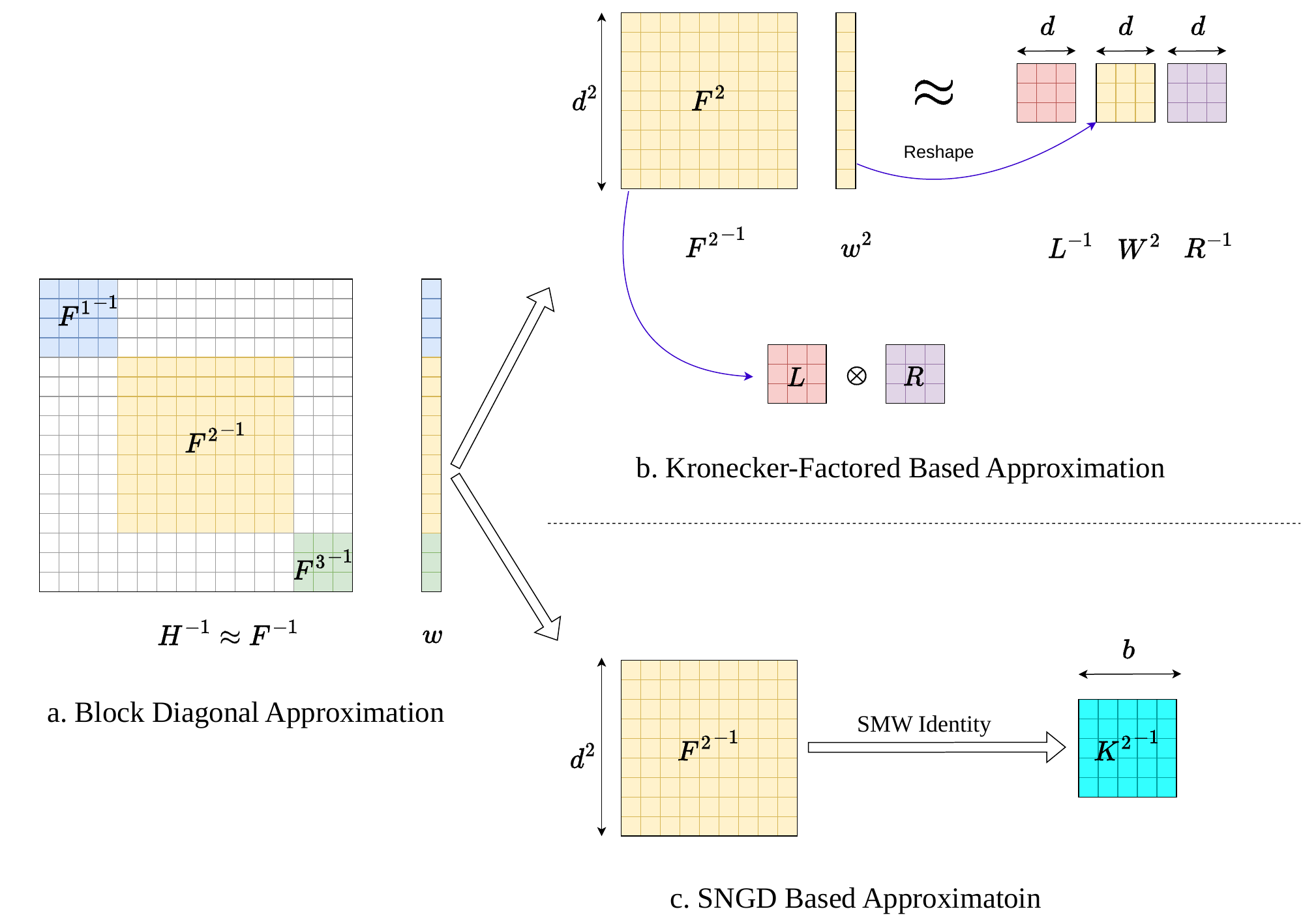}
    \caption{Approximations in second-order methods.}
    \label{fig:second_order_methods}
\end{figure}

\textbf{\textit{Natural Gradient Descent (NGD)}}
\label{subsection:ngd}
In NGD, which is a second-order method, we use the inverse of the Fisher Information Matrix (FIM) as a preconditioner to the gradients as shown in \autoref{fig:second_order_methods}-a. \autoref{eq:ngd} shows the update rule of NGD, where $F^m$ is the FIM block corresponding to block $m$. \autoref{eq:fim} shows the definition of FIM for an arbitrary layer in our model, where $x_i$ is the $i^{th}$ sample in the batch.

\begin{equation}
    \label{eq:fim}
    F^m = \frac{1}{b} \sum_{i = 1}^{b} \nabla_{x_i} \ell(\mathcal{W}, x_i) \nabla_{x_i} \ell(\mathcal{W}, x_i)^T
\end{equation}

\textbf{\textit{Kronecker Factorization (KFAC).}}
KFAC methods reformulate the FIM block as the Kronecker product of two matrices as shown in \autoref{eq:kfac_kronecker} where $g^m = \nabla_{x^m} \mathcal{L}$ and $a^m$ is the vector form of the activation output of layer $m$ and $\mathbb{E}$ is the expectation operator and $x^m$ is the input matrix of layer $m$. Please note that we have used the mixed-product property of Kronecker multiplication for getting the right hand value. 

\begin{equation}
    \label{eq:kfac_kronecker}
    F^m = \mathbb{E} [ (g^m \otimes a^{m - 1}) (g^m \otimes a^{m - 1})^T ] = \mathbb{E} [ (g^m {g^m}^T) \otimes (a^{m - 1} {a^{m - 1}}^T) ]
\end{equation}

Furthermore, we assume that $\mathbb{E} [ (g^m {g^m}^T) \otimes (a^{m - 1} {a^{m - 1}}^T) ] \approx \mathbb{E} [ g^m {g^m}^T ] \otimes \mathbb{E} [a^{m - 1} {a^{m - 1}}^T ]$, which is a strong assumption, but helps us simplify the computation further. Using the inversion property of Kronecker multiplication, we can compute the inverse of FIM using \autoref{eq:kfac_inv}.

\begin{equation}
    \label{eq:kfac_inv}
    {F^m}^{-1} w^m = \mathbb{E} [ g^m {g^m}^T ]^{-1} \otimes \mathbb{E} [a^{m - 1} {a^{m - 1}}^T ]^{-1} w^m
\end{equation}

By using the mixed Kronecker matrix-vector product property, we can get the update value in \autoref{eq:kfac_final}, which is illustrated in \autoref{fig:second_order_methods}-b. We refer to $L^m$ and $R^m$ as the left and right factors respectively. Adding momentum to the left and right factors and denoting the iteration number with a subscript to the factors, we will get \autoref{eq:left_factor_momentum} and \autoref{eq:right_factor_momentum}.

\textit{Sherman-Morrison-Woodbury-Based Natural Gradient Descent (SNGD).}
In this method, the SMW identity is used for approximating the inverse of $(F^m + \mu I) \in \mathbb{R}^{d^2 \times d^2}$, where $\mu$ is a damping factor used in the preconditioning. \autoref{eq:sngd} shows the process of computing the inverse of the FIM for a single layer in the network, where $A^m\in \mathbb{R}^{d\times b}$ is the batch of activations of layer $l$ and $G^m \in \mathbb{R}^{d\times b}$ is the batch of gradients of the loss function with respect to the inputs of that layer and $U = [\nabla_{W^m} \mathcal{L(W}, x_1), ...,\nabla_{W^m} \mathcal{L(W}, x_b)]^T \in \mathbb{R}^{d^2 \times b}$ is the concatenation of the gradients of the loss function with respect to the parameters of that layer and $\odot$ shows the Hadamard element-wise product. In this method, a kernel matrix in $\mathbb{R}^{b \times b}$ is inverted, as shown in \autoref{fig:second_order_methods}-c.

\begin{equation}
    \label{eq:sngd}
    (F^m + \mu I)^{-1} = \frac{1}{\mu} ( I - U^m({A^{m- 1}}^TA^{m-1} \odot {G^m}^TG^m + \mu I)^{-1} {U^m}^T)
\end{equation}

\section{Numerical Instability of Second-order Methods}
\label{subsection:numerical_instability}
\begin{figure}[ht]
    \centering
    \includegraphics[scale=1.5]{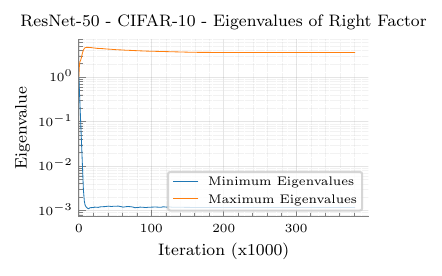}\\
    
    (a)\\
    \includegraphics[scale=1.5]{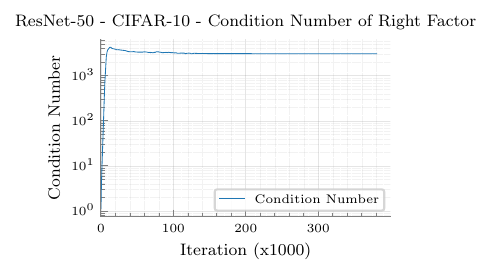}\\
    (b)\\   
    \caption{Maximum and minimum eigenvalues (\textbf{a}) and the condition number (\textbf{b}) of the right factors in KFAC when training ResNet-50 on CIFAR-10. As illustrated, the minimum eigenvalues of the factors in KFAC approach zero, meaning that the factors are singular, and hence have large condition numbers, making numerical inversion of them complex and numerically unstable.}
    \label{fig:eigenvalues}
\end{figure}

In \autoref{sec:numerical_stability}, we discussed that in second-order methods, multiple matrix inversion or root-finding algorithms need to be executed, which make the second-order methods prone to numerical instabilities. Furthermore, we discussed that left and right factors in second-order methods have large condition numbers, resulting in further issues in inversion. \autoref{fig:eigenvalues} shows the eigenvalues of the right factor and its condition number for ResNet-50 model on CIFAR-10 dataset on KFAC algorithm. Even when using damping factors and filtering out extremely small eigenvalues, the condition number of these matrices is large, motivating the use of double precision computations for avoiding numerical instabilities.

MKOR, on the other hand, doesn't suffer numerical instabilities when inverting such matrices, and its computational complexity isn't dependent on the condition number either.

\section{Sensitivity to Learning Rate}
\label{subsection:sensitivity_to_lr}
Learning rate is one of the main hyperparameters in machine learning (ML) that can directly affect the convergence time of optimization, and ML practitioners have to spend a lot of time tuning this hyperparameter. More specifically, in first-order methods, a large learning rate can easily lead to divergence and numerical instability, and in second-order methods, large learning rates can lead to exploding gradients as discussed in \autoref{section:exploding_gradient}. Using small learning rates can lead to slow convergence in both first- and second-order methods and can even jeopardize the main advantage of second-order methods, which is their faster convergence rate.

\begin{table}
    \centering
    \caption{Number of epochs necessary for convergence in different optimizers for ResNet-50 on CIFAR10. MKOR is the least sensitive optimizer to learning rate, converging in almost the same number of iterations for a wide range of learning rate, while other optimizers either diverge (\textbf{D}) or converge to a local-minimum (\textbf{$*$ superscript}).}
    \begin{tabularx}{1\textwidth} { 
    | >{\centering\arraybackslash}X 
    | >{\centering\arraybackslash}X 
    | >{\centering\arraybackslash}X 
    | >{\centering\arraybackslash}X 
    | >{\centering\arraybackslash}X | }
    \hline
    \backslashbox[\dimexpr\linewidth+2\tabcolsep]{Optimizer}{\scriptsize{Learning Rate}} & $10$ & $1$    & $0.1$ & $0.01$ \\
    \hline
    MKOR                & $94$      & $79$      & $78$  & $76$  \\
    \hline
    KAISA               & $112$     & $100$     & $90$  & $89^*$  \\
    \hline
    HyLo                & $D$       & $123^*$   & $98$  & $150^*$   \\
    \hline
    SGD                 & $D$       & $D$       & $108$ & $145^*$    \\
    \hline        
    \end{tabularx}
    \label{table:lr_sensitivity}
\end{table}

One of the main advantages of our method is its robustness against a wide range of learning rates. As \autoref{table:lr_sensitivity} shows, first-order methods are extremely sensitive to the learning rate, and the second-order methods are prone to ripples and divergent for a larger range of learning rates, and lose their performance for small learning rates. Our method, on the other hand, will converge with a high convergence rate for a wide range of learning rates, and by directly modifying the inverse of factors as discussed in \autoref{section:exploding_gradient} can find a proper equilibrium between first- and second-order methods. This table shows that our method is the least sensitive to the learning rate values and can make the job of ML practitioners for tuning this hyperparameter extremely easy.

\section{Scalability}

\label{subsection:scalability}
\begin{figure}
\centering
    \includegraphics[scale=1.2]{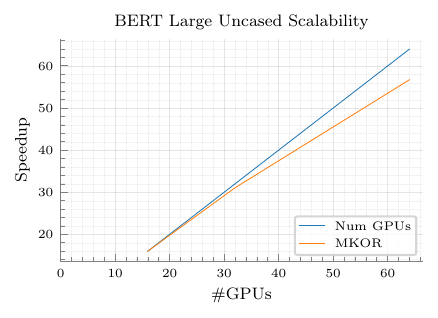}
    \caption{Scalability of MKOR.}
    \label{fig:scalability}
\end{figure}

\autoref{fig:scalability} shows the strong scalability of MKOR on BERT-Large-Uncased on up to 64 GPUs.

\section{Decaying Eigenvalues and Rank-1 Approximations} 

\label{subsection:decaying_eigenvalues}
\begin{figure}
    \centering
    \centerline{
        \includegraphics[scale=1.2]{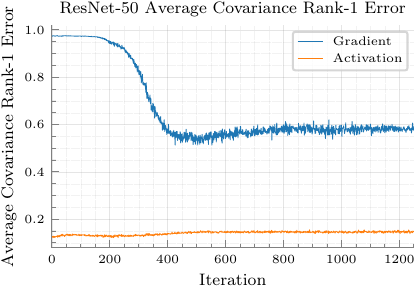}
    }
    \caption{Average covariance rank-1 approximation error for ResNet-50 in different iterations}
    \vspace{-0.03in}
    \label{fig:average_covariance_error}
\end{figure}

\autoref{fig:average_covariance_error} shows that the eigenvalues of the factors will decay as the model converges, making rank-1 approximations more effective. The reason behind the decay in the eigenvalues is that the weights are initialized randomly and the neurons work independently in the beginning of the training, but as the model converges, the neurons become more dependent on each other and thus the activation and input gradients will become linearly dependent. This is also reflected by some large error values in the distributions in \autoref{fig:rank_1_error}[a, b, c, d]. The factors in MKOR are initialized with identity, starting MKOR from a first-order method. As a result, MKOR is more robust against noise in the approximations in the first iterations (the approximation error does  not noticeably affect the factors when replacing ${L_{t-1}^{m-1}}^{-1}$ and ${R_{t-1}^{m}}^{-1}$ in \autoref{eq:rank1_left_factor_inversion} and \autoref{eq:rank1_right_factor_inversion} with identity). But as the model converges, the factors in MKOR will be mostly shaped by the training samples, making MKOR more reliable on less erroneous approximations, and the decaying eigenvalues of the factors help MKOR with that.

\section{Training Accuracy Experiments}
\label{subsection:training_accuracy_experiment}
To evaluate MKOR as an optimizer that tries to minimize a specific objective function, we have considered the case of only minimizing the loss function of models on different tasks and set all the other optimizer parameters such as weight decay to zero, since using a non-zero weight decay adds a quadratic term to the loss function and using different weight decays for different optimizers leads to optimizing different objective functions, which might be considered unfair.

\begin{figure}
    \centering
    \centerline{
        \includegraphics[scale=1.2]{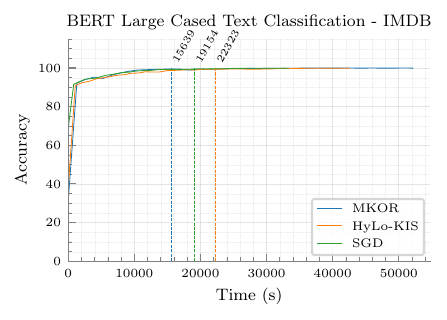}
        \includegraphics[scale=1.2]{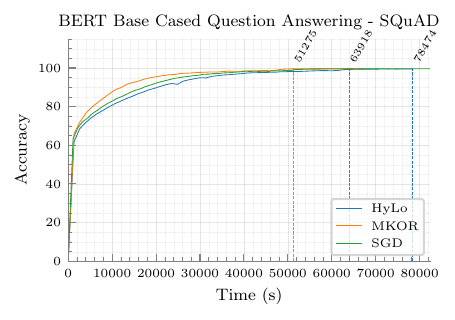}
    }
    \hspace{0.04\linewidth}
    (a) \hspace{.52\linewidth}
    (b)
    \centerline{
        \includegraphics[scale=1.2]{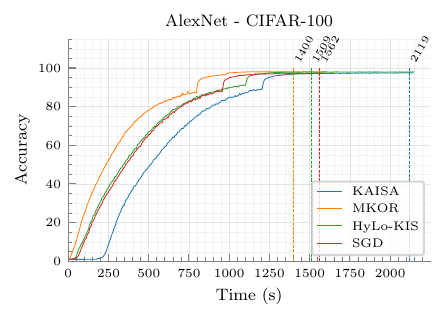}
    }
    \hspace*{0.05\linewidth}
    (c)
    \caption{Training time for distributed first- and second-order optimizers SGD, MKOR, KAISA, and HyLo on \textit{BERT-Large-Cased} on \textit{IMDB} (\textbf{a}), \textit{BERT-Base-Cased} on \textit{SQuAD} (\textbf{b}), and \textit{AlexNet} on \textit{CIFAR-100} (\textbf{c}). In all the experiments, MKOR outperforms other optimizers in convergence speed.}
    \label{fig:training_time}
\end{figure}

\begin{figure}
    \centering
    \centerline{
        \includegraphics[scale=1.2]{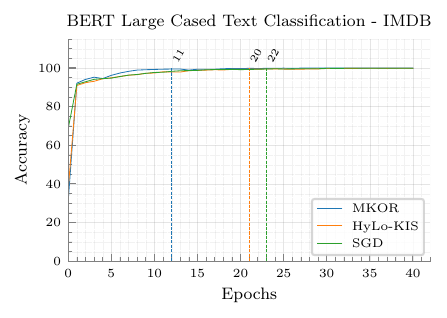}
        \includegraphics[scale=1.2]{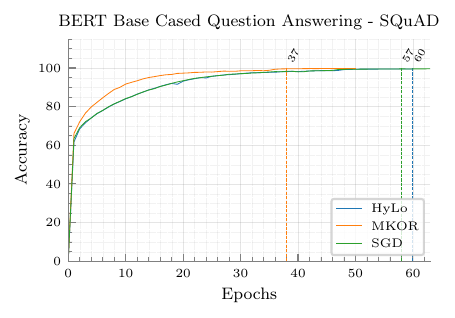}
    }
    \hspace{0.04\linewidth}
    (a) \hspace{.52\linewidth}
    (b)
    \centerline{
        \includegraphics[scale=1.2]{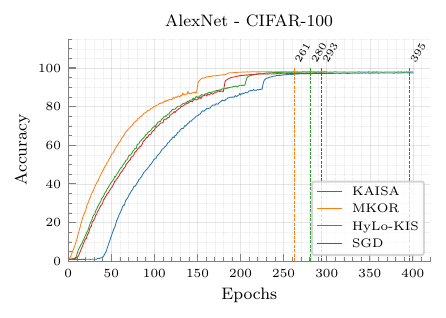}
    }
    \hspace*{0.05\linewidth}
    (c)
    \caption{Training accuracy vs. the number of epochs for distributed first- and second-order optimizers SGD, MKOR, KAISA, and HyLo on \textit{BERT-Large-Cased} on \textit{IMDB} (\textbf{a}), \textit{BERT-Base-Cased} on \textit{SQuAD} (\textbf{b}), and \textit{AlexNet} on \textit{CIFAR-100} (\textbf{c}). In all the experiments, MKOR outperforms other optimizers in convergence rate.}
    \label{fig:training_epoch}
\end{figure}

Recent work has shown the advantage of second-order methods over their first-order counterparts on multiple CNN tasks \cite{hylo, kaisa}, such as residual networks \cite{resnet}. In our training accuracy experiment, we use another CNN benchmark, \textit{AlexNet}~\cite{alexnet} with more than 20M parameters on \textit{CIFAR-100}~\cite{cifar} consisting of 50K training and 10K validation images of 100 classes. \autoref{fig:training_time}-c and \autoref{fig:training_epoch}-c show the convergence properties of different optimizers. MKOR is $1.26\times$, $1.31\times$, and $1.58\times$ faster than HyLo-KIS, SGD, and KAISA respectively. The reason for the low convergence speed of KAISA is that we needed to use small learning rates for avoiding exploding gradients in it, which has damaged its convergence rate.

\textit{BERT} is a large language model with two variants, \textit{BERT-Base} with more than 108M parameters and \textit{BERT-Large} with more than 335M parameters. As shown in \autoref{fig:training_time}-a and \autoref{fig:training_epoch}-a, we have fine-tuned \textit{BERT-Large} on the \textit{IMDB} dataset which is a text classification task with 25K training and 25K test samples. MKOR outperforms SGD and HyLo-KIS by a speedup factor of $1.22\times$ and $1.43\times$ respectively. We have also fine-tuned \textit{BERT-Base} on \textit{SQuAD} dataset, which is a question answering task with 87.6K training and 10.6K test samples. MKOR achieves $1.26\times$ and $1.56\times$ speedup over SGD and HyLo respectively. Using a wide range of learning rates, KAISA could not converge on any of our \textit{BERT} experiments which are based on the HuggingFace implementation of BERT, and the reason for lack of convergence of KAISA is exploding gradients.

\section{Knee-Point Learning Rate Scheduler}\footnote{The knee-point learning rate is not used in any of the experiments in the main chapters.}
\label{subsection:knee_point_scheduler}
While using large learning rates is crucial for utilizing the higher convergence rate of second-order methods, it is necessary to reduce the learning rate after some iterations so that the model converges, and the  number of iterations for changing the learning rate is not known in advance. In practice, machine learning practitioners will manually find the  number of iterations by trial and error or use predefined functions \cite{triangular_lr, cosine_annealing_lr} that don't necessarily work ideally in all optimization problems. We used knee-point learning rate scheduler in \autoref{subsection:training_accuracy_experiment}.

To fully utilize the potentials of the optimizers, we have designed a learning rate scheduler that monitors the rate of improvement in accuracy or decrease in the loss function value, and based on that decides when to decrease the learning rate. The scheduler detects knee-points in the accuracy/loss and decreases the learning rate when a knee-point is observed.

By definition, knee-points are defined as the points where the average accuracy/loss rate is less than $\beta$ times the increment/decrement in the accuracy/loss since using the current learning-rate. For averaging the accuracy/loss rate, we use an exponential moving average, and $\beta$ is a hyperparameter that we can choose to show how much the scheduler can tolerate lack of improvement to detect the accuracy/loss.

\section{Proofs}

\textbf{\autoref{lemma:positive_definite}}:

\begin{proof}
    Given a positive definite matrix $J_{t-1}$ and a vector $j$ and scalar $0 < \gamma < 1$, we show that \autoref{eq:rank1_factor_inversion} results in a positive-definite matrix.
    \begin{equation}
        \label{eq:rank1_factor_inversion}
        {J_{t}}^{-1} = \gamma {J_{t - 1}}^{-1} + \frac{(1 - \gamma)}{\gamma^2(1 + \gamma (1 - \gamma) {j}^T{J_{t - 1}}^{-1} {j})} {J_{t - 1}}^{-1} {j} {j}^T {J_{t - 1}}^{-1}
    \end{equation}
    Since $\gamma > 0$ and $J_{t-1}$ is a positive-definite matrix, $\gamma J_{t-1}$ is also positive-definite.

    Also, since $J_{t-1}$ is a positive-definite matrix, $\forall x \neq 0: x^T J_{t-1} x > 0$. 
    \begin{equation}
        j^T L_{t-1}^{-1} j > 0 \xrightarrow[]{0 < \gamma < 1} 1 + \gamma (1 - \gamma) {j}^T{J_{t - 1}}^{-1} {j} > 0
    \end{equation}
    Now we show  that ${J_{t - 1}}^{-1} {j} {j}^T {J_{t - 1}}^{-1}$ is also positive-definite.
    \begin{equation}
        \forall x \neq 0: x^T {J_{t - 1}}^{-1} {j} {j}^T {J_{t - 1}}^{-1}x = (j^T J_{t-1}^{-1} x)^T (j^T J_{t-1}^{-1} x) = \|(j^T J_{t-1}^{-1} x)\|^2 > 0
    \end{equation}
    Since both matrices on the right-hand-side of \autoref{eq:rank1_factor_inversion} are positive-definite and the sum of two positive-definite matrices is a positive-definite matrix, the left-hand-side of it will be also positive definite.
\end{proof}

\textbf{\autoref{lemma:loss_reduction}:}
\begin{proof}
    By defining $P = ( \zeta L^{-1} + (1 - \zeta) I) \otimes  (\zeta R^{-1} + (1 - \zeta) I)$, \autoref{eq:taylor_expansion} will hold.
    
    \begin{equation}
        \label{eq:taylor_expansion}
        \mathcal{\hat{L}}(w_0 - \Delta w) - \mathcal{L}(w_0) = -\nabla \mathcal{L}(w_0)^T P \nabla \mathcal{L}(w_0)
    \end{equation}

    Now, we will show that $P$ is a positive-semi-definite matrix. By using the associativity property of Kronecker multiplication, we can represent $P$ as in \autoref{eq:P}. Please note that different identity matrices in \autoref{eq:P} have different shapes.
    
    \begin{equation}
        \label{eq:P}
        P = \zeta^2 L^{-1} \otimes R^{-1} + \zeta (1 - \zeta) L^{-1} \otimes I + \zeta (1 - \zeta) I \otimes R^{-1} + I
    \end{equation}
    
    Since matrices $L$ and $R$ are positive-semi-definite, the Kronecker products $L^{-1} \otimes R^{-1}$, $L^{-1} \otimes I$, and $I \otimes R^{-1}$ are also positive-semi-definite. As a result, for any non-zero vector $x$, $x^T P x > 0$. So, based on \autoref{eq:taylor_expansion}, $-\nabla \mathcal{L}(w_0)^T P \nabla \mathcal{L}(w_0)$. As a result, \autoref{eq:inequality} holds.
    \begin{equation}
        \label{eq:inequality}
        \mathcal{\hat{L}}(w_0 - \Delta w) < \mathcal{L}(w_0)
    \end{equation}
\end{proof}

\textbf{\autoref{lemma:quantization_error}}
\begin{proof}
    Considering the quantization of matrix $J$ and vector $j$ in \autoref{eq:quantization_formula} and assuming the maximum quantization error is $\epsilon$ and the maximum values in vector $j$ and matrix $J$ is $m$, we can consider one of the three possible cases:
    \begin{enumerate}
        \item \textit{Vector-Vector Dot Product}: The resulting error is $O(2\epsilon m^2 d)$ because the error of each multiplication can at most be $2m \epsilon + \epsilon^2$ and by adding the $d$ multiplications, the maximum error can be $(2m + 1)d \epsilon$.
        \item \textit{Vector-Matrix Product}: The resulting error is $O(2\epsilon m^2 d)$ because a vector-matrix product can be considered as multiple independent vector-vector dot products.
        \item \textit{Vector-Vector Outer Product}: The resulting error is $O(2\epsilon m)$ because each element in the resulting matrix is computed using a multiplication with the maximum error of $2\epsilon m^2 + \epsilon ^ 2$.
    \end{enumerate}
    
    \begin{equation}
        \label{eq:quantization_formula}
        \gamma {J} + \frac{(1 - \gamma)}{\gamma^2(1 + \gamma (1 - \gamma) {j}^T{J} {j})} {J} {j} {j}^T {J^T}
    \end{equation}

    The error imposed by quantizing $\gamma J$ is at most $\gamma \epsilon$. The quantization error in the denominator of the fraction in \autoref{eq:quantization_formula} is negligible in comparison to $1$ and won't change the growth order of the final error term. The quantization error of $Jj$ is $O(2m^2d\epsilon)$ as discussed earlier in the case of vector-matrix product. $Jjj^TJ^T = (Jj) (Jj)^T$ is a vector-vector outer product, resulting in $O(4m^3d^2)$ error. 

    So the final error quantization error is $O((\gamma + 4\frac{(1 - \gamma)}{\gamma^2} m^3 d^2) \epsilon)$
    
\end{proof}

\chapter{Supplementary Material for \slope}
This chapter provides supplementary material for the \slope chapter. We begin with a comparison against dynamic sparsity using SR-STE in \autoref{section:rs_ste}, followed by an analysis of the cuSPARSELt initialization overhead in \autoref{section:setup_overhead}. We present BERT-Large-Uncased pretraining and downstream evaluation results in \autoref{section:bert_loss}, and discuss the performance overhead of the bidirectional mask in \autoref{section:transposable_slow_down}. \autoref{section:lossy_backward_effect} analyzes the sparsity ratio in the double-pruned backward pass, and \autoref{section:pruning_matrices} examines sensitivity to the choice of pruning matrix. Implementation details are provided in \autoref{section:implementation_details}, task-specific GLUE results in \autoref{section:glue_results}, and the integration with Flash Attention in \autoref{sec:flash_attention}. We compare with dense models in \autoref{sec:gpt2_half} and conclude with proofs of the lemmas used in the chapter.

\section{Comparison with Dynamic Sparsity: SR-STE}
\label{section:rs_ste}
We pretrained GPT2-Small (\autoref{section:accuracy_results}) using the SR-STE method~\cite{nm_scratch} and reported the perplexity results in \autoref{fig:gpt_perplexity}.
SR-STE aims to mitigate the Sparse Architecture Divergence (SAD) by dynamically adjusting the sparsity mask throughout training. We have tested various decay factor hyperparameters to find the optimal optimization strategy for SR-STE.

To understand the performance gap between SR-STE and \slopespace (our method) for the same training budget, we analyzed the mask dynamics in SR-STE. We plotted the \textit{average number of mask elements} changes during training compared to the final converged mask sparsity pattern.
High mask change values indicate that training resources are spent on updating weights that ultimately get pruned and do not necessarily contribute to the final model accuracy.

\begin{figure}[ht]
\begin{center}
{\includegraphics[scale=1]{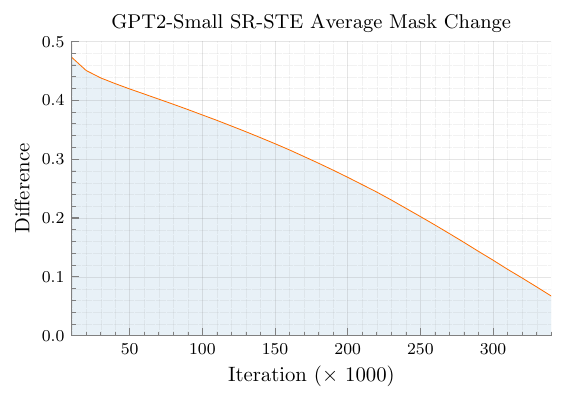}}
\caption{Average mask difference between each iteration and the converged sparsity pattern in GPT2-Small pretraining using SR-STE. The highlighted area shows the ratio of the resources used for updating weights that are pruned and not used in the inference of the model.
}
\label{fig:rs_ste_mask_change}
\end{center}
\end{figure}

\autoref{fig:rs_ste_mask_change} shows this average mask difference per iteration relative to the converged model.
As training progresses, the mask difference decreases, demonstrating SR-STE's convergence to a specific sparsity pattern. However, in \slopespace, where all resources are dedicated to optimizing weights under a \textit{static} mask\footnote{We determine the pruning mask at the very first iteration and maintain it for the rest of training.}, SR-STE's dynamic approach leads to wasted computation (represented by the area under the curve in \autoref{fig:rs_ste_mask_change}). Consequently, for the same training budget, \slopespace achieves a lower perplexity in comparison to SR-STE due to its static mask approach.

\section{cuSPARSELt Initialization Overhead: Static vs. Dynamic Sparsity}
\label{section:setup_overhead}
This section analyzes the time breakdown of the cuSPARSELt SpMM pipeline, highlighting the significant overheads associated with dynamically changing sparsity masks.
The cuSPARSELt SpMM operation consists of two main phases: \textit{(1) Setup} and \textit{(2) Matrix Multiplication}. The setup phase involves initializing matrix handles and compressing the 2:4 sparse matrix. This compression copies non-zero values into a contiguous memory layout and generates indices for those values. The matrix multiplication phase leverages this metadata to perform the sparse matrix-matrix multiplication.

\begin{figure}[ht]
\begin{center}
{\includegraphics{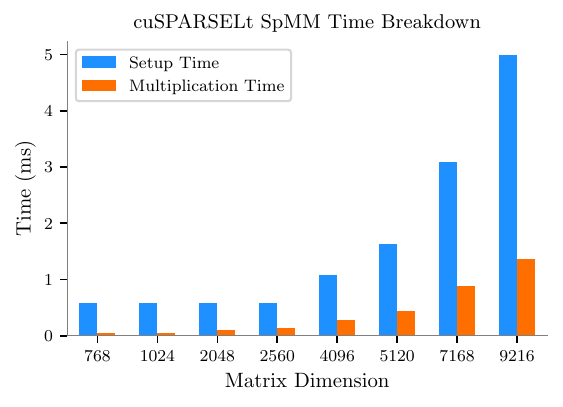}}
\caption{The setup and multiplication time for square matrices using the cuSPARSELt SpMM backend.}
\label{fig:setup_overhead}
\end{center}
\end{figure}

\autoref{fig:setup_overhead} shows the setup and multiplication time for square matrices using the cuSPARSELt SpMM backend.  As evident from the figure, the setup overhead is significantly larger than the actual matrix multiplication time.
For \slopespace, which employs static sparsity masks, the setup cost is incurred only once and becomes negligible compared to the numerous matrix multiplications performed during training and inference.  However, for dynamic sparsity patterns, such as Fully Sparse Training~\cite{2_4_pretraining}, Bidirectional Masks~\cite{n_m2}, and other similar methods\cite{n_m1, n_m3_dominosearch, n_m5_step, nm_scratch}, this setup overhead can be substantial, leading to reduced speedup (as observed in \autoref{section:speedup_results} for Fully Sparse Training) or  slowdowns in some configurations (as discussed in ~\autoref{section:transposable_slow_down}).\footnote{A recent work observed a similar overhead using dynamic sparsity in cuSPARSELt SpMM pipeline~\cite{nm_attention}.}

\section{BERT-Large-Uncased: Pretraining and Downstream Evaluation}
\label{section:bert_loss}
\begin{figure}[ht]
\begin{center}
{\includegraphics[width=0.6\columnwidth]{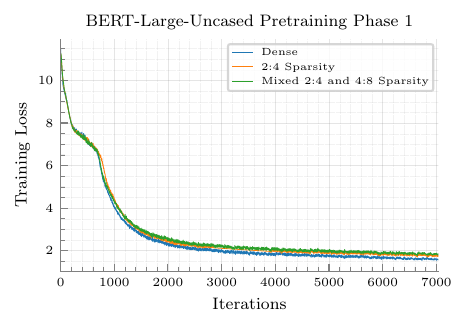} \includegraphics[width=0.6\columnwidth]{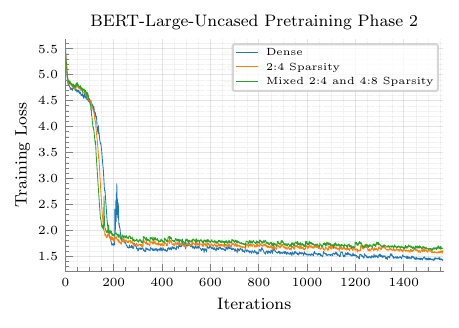}}
\caption{Training loss of BERT-Large-Uncased on WikiCorpus dataset for phase 1 and 2.}
\label{fig:slope_bert_loss}
\end{center}
\end{figure}

BERT-Large-Uncased pretraining consists of two phases, as illustrated in \autoref{fig:slope_bert_loss}. Phase 1 comprises 7,038 iterations with a global batch size of 65,536 and a sequence length of 128. Phase 2 includes 1,563 iterations with a global batch size of 32,768 and a sequence length of 512.

\autoref{fig:slope_bert_loss} shows the training loss for both phases under different sparsity settings.  We observe that higher sparsity ratios generally lead to higher training loss in both phases. Interestingly, the loss/perplexity gap does not directly correlate with the observed accuracy drops in downstream tasks~\cite{perplexity, perplexity2, perplexity3}.

We evaluated the pretrained BERT-Large-Uncased models on the SQuAD v1.1~\cite{squad} and GLUE~\cite{glue} benchmarks.
SQuAD v1.1, a comprehensive question-answering dataset based on Wikipedia, is widely used for LLM training. We report the F1 score for SQuAD throughout the chapter.
GLUE, a diverse benchmark for natural language understanding tasks, provides a single aggregated score across various challenges, facilitating model comparisons. The chapter presents the average metric score for GLUE, while task-specific metrics are detailed in ~\autoref{section:glue_results}.

\section{Performance overhead of bidirectional mask}
\label{section:transposable_slow_down}
\autoref{table:bi_directional_slowdown} presents the runtime results of Bidirectional Masks~\cite{n_m2}, a state-of-the-art N:M sparsity method.
Our analysis demonstrates that the mask search and associated overheads of this approach result in significant slowdowns compared to dense baselines. For these experiments, we utilized the repository provided in~\cite{n_m2} and employed the same models used in their evaluation.
\begin{table}[t]
\caption{End-to-end slow-down of Bi-directional Mask \cite{n_m2} in comparison to the dense baseline.}
\label{table:bi_directional_slowdown}
\vskip 0.15in
\begin{center}
\begin{small}
\begin{sc}
\begin{tabular}{ccc}
\toprule
Model   & Dataset & Slow-down ($\times$) \\
\midrule
MobileNet v2 & CIFAR10 & 5.08 \\
\hline
ResNet-32 & CIFAR10 & 5.07 \\
\hline
VGG19 & CIFAR10 & 8.41 \\
\hline
ResNet-18 & ImageNet & 3.66 \\
\hline
ResNet-50 & ImageNet & 3.01 \\
\bottomrule
\end{tabular}
\end{sc}
\end{small}
\end{center}
\vskip -0.1in
\end{table}

\section{Sparsity ratio analysis of double-pruned backward pass}
\label{section:lossy_backward_effect}
\begin{figure}[ht]
\begin{center}
\centerline{\includegraphics[scale=1.2]{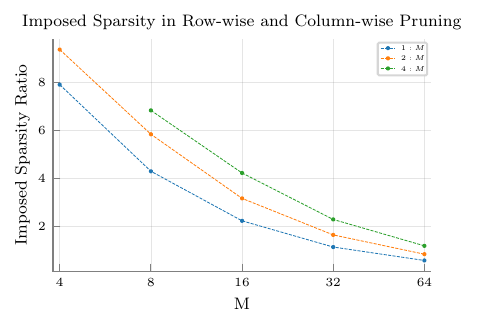}}
\caption{The imposed sparsity ratio when pruning the weight matrices in the backward pass.}
\label{fig:sparsity_ratios}
\end{center}
\end{figure}

As described in ~\autoref{section:sparse_pretraining}, our proposed sparse pretraining approach involves pruning weights in both the forward and backward passes.
During the backward pass, we apply both row-wise and column-wise pruning, which introduces additional zero values to the column-wise pruned weight matrices used in the forward pass.
\autoref{lemma:sparsity_ratio} demonstrates that the resulting sparsity ratio can be calculated using ~\autoref{eq:sparsity_ratio}.
\autoref{fig:sparsity_ratios} visualizes the imposed sparsity ratios for various N:M sparsity patterns.  As expected, smaller N/M ratios lead to lower imposed sparsity ratios.  Moreover, in most cases, the imposed sparsity ratio is significantly smaller than the original matrix's density ratio.

\section{Sensitivity to the choice of pruning matrix}
\label{section:pruning_matrices}
In linear layers, three matrices are involved in the forward and backward passes: the input, the output gradient, and the weights. Pruning each of these matrices can have distinct effects on model performance.

To identify the optimal pruning strategy, we conducted an experiment where we pretrained GPT2-Small for 100,000 iterations (a quarter of the full pretraining) while systematically applying both static and dynamic pruning to each of the three matrices.
Static pruning involves generating a random mask at initialization and applying it throughout training. Dynamic pruning, on the other hand, prunes matrices based on their magnitude at each iteration. For dynamic pruning, the dense matrix values are computed and stored, and then pruned at every step.

\begin{figure}[ht]
\begin{center}
\centerline{\includegraphics[scale=2]{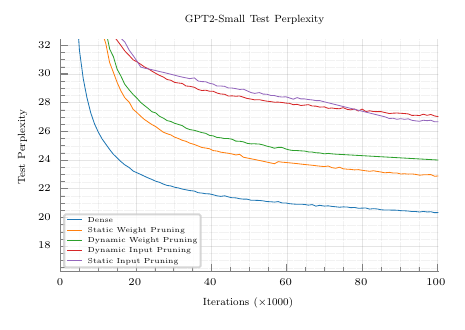}}
\caption{Validation perplexity on GPT2-Small pretraining for 100,000 iterations for different matrix pruning settings. Pruning the output gradients leads to divergence within a few iterations and hence is not reported.}
\label{fig:pruning_matrices}
\end{center}
\end{figure}

~\autoref{fig:pruning_matrices} presents the validation perplexity for these experiments. Notably, pruning the output gradient led to model divergence after a few iterations and is not shown in the figure.

\niparagraph{Analysis.}
As shown in ~\autoref{fig:pruning_matrices}, static pruning consistently achieved lower perplexities. This behavior suggests that focusing computational resources on elements that remain active throughout training can lead to improved performance. Furthermore, pruning weights resulted in lower perplexities compared to pruning inputs, indicating that weights are generally a better target for pruning.

\niparagraph{Intuition.}
Pruning weights is analogous to removing connections between neurons. Pruning activation tensors is similar to introducing a non-linear function (akin to max-pooling) before each linear layer.  Pruning output gradients, however, lacks practical justification and introduces errors into the backward pass, leading to model divergence.

\section{Implementation details}
\label{section:implementation_details}

This section details the implementation of the custom functions and CUDA kernels used in \autoref{alg:training} to facilitate efficient sparse training.

\noindent{}\textbf{Initialization, sparse matrix setup, and SpMM kernels.}
Before utilizing the cuSPARSELt APIs, a crucial initialization phase ensures proper configuration of essential variables for our computational task.
Following initialization, we configure the sparse data formats tailored for sparse matrices. This involves initializing matrix descriptors, pruning the matrices, and compressing them into a more compact representation.
cuSPARSELt employs an automated search to determine the optimal kernel for executing SpMM.
While setting up these sparse data formats incurs a non-negligible computational cost, this overhead is mitigated by the repetitive nature of matrix multiplications during the training process.

\noindent{}\textbf{Prune and compress.}
The gradient of the loss function with respect to the weights requires pruning using the same mask as the weight matrix.
Consequently, it contains 50\% extra zero values in the dense format.
To address this redundancy, we developed an optimized CUDA kernel, integrated into PyTorch, that masks the gradients accordingly, eliminating the storage of unnecessary data and reducing memory usage.
The output of this operation is a new matrix in $\mathbb{R}^{d_{out} \times \frac{d_{in}}{2}}$.

\noindent{}\textbf{Sparse matrix addition.}
The cuSPARSELt sparse data format does not natively support addition operations. However, for matrices $A$ and $B$ sharing the same sparsity patterns, we developed an optimized CUDA kernel seamlessly integrated into the PyTorch training workflow. This kernel efficiently computes linear combinations of the form $\beta A + \gamma B$, where $\beta$ and $\gamma$ are arbitrary user-defined constants. This functionality is particularly useful for adding sparse weights to gradients in optimizers that utilize weight decay.

\noindent{}\textbf{Update Sparse Matrix.}
After the optimizer updates the weight tensor values based on its rules, we need to update the sparse matrix format to reflect these changes.  We implemented an optimized CUDA kernel that copies the weight tensors from the PyTorch format into the cuSPARSELt data type, enabling efficient storage and manipulation of sparse weights.

\section{Task-specific GLUE results}
\label{section:glue_results}
The GLUE benchmark~\cite{glue} comprises eight distinct natural language understanding classification tasks. While \autoref{section:results} presented the average GLUE score as a measure of overall model performance, this section provides a more detailed analysis by presenting the complete task-specific results for each training setting in ~\autoref{tab:glue_complete}.

\begin{table}
\begin{scriptsize}
    \centering
    \caption{GLUE results for each task in the experiments discussed in \autoref{section:results}.}
    \begin{tabular}{cccccccccccccc}
         & & & First & Last \\
         Method &Phase &Rank &12 &12 &CoLA &SST-2 &MRPC  &STS-B  &QQP &RTE &MNLI &QNLI \\
          & & & \scriptsize{Blocks} & \scriptsize{Blocks} & (mcc) & (acc) & (f1) & (corr) & (f1) & (acc) & (acc) & (acc) \\
          \midrule
        Dense & 1,2 & 0 & 2:4 & 2:4 & 51.6 & 91.9 & 81.2 & 87.5 & 87.8 & 66.4 & 84.1 & 91.3 \\
        \midrule
        \scriptsize{\slopespace} \\
        \scriptsize{MLP Mixer} & 2 & 0 & 2:4 & 2:4 & 41.8 & 91.4 & 88.7 & 87.2 & 85.9 & 65 & 82.1 & 90.1 \\
        \scriptsize{Only} \\
        \midrule
        \scriptsize{\slopespace} \\
        \scriptsize{MLP Mixer +}  & 2 & 0 & 2:4 & 2:4 & 38.8 & 90.4 & 85.9 & 86.4 & 85.9 & 63.5 & 81.5 & 89.3 \\
        \scriptsize{Self-Attention} \\
        \midrule
        \scriptsize{\slopespace with} \\
        \scriptsize{Non-Lazy} & 2 & 40 & 2:4 & 2:4 & 43.3 & 90.8 & 89 & 87 & 86 & 64.6 & 82.3 & 89.6 \\
        Adapters \\
        \midrule
        \scriptsize{\slopespace with} \\
        \scriptsize{Non-Lazy}  & 2 & 40 & 2:8 & 2:4 & 29 & 89.7 & 83.7 & 85.6 & 85.2 & 66.8 & 79.9 & 87.4 \\
        Adapters \\
        \midrule
        \scriptsize{\slopespace with} \\
        \scriptsize{Non-Lazy}  & 2 & 40 & 2:4 & 2:8 & 44.1 & 91.1 & 89.8 & 86.6 & 86.3 & 62.5 & 82.3 & 89.6 \\
        Adapters \\
        \midrule
        \slopespace & 1,2 & 0 & 2:4 & 2:4 & 37.9 & 91.4 & 85.4 & 86.6 & 85.8 & 62.5 & 80.7 & 88.6 \\
        \midrule
        \slopespace & 1,2 & 4 & 2:4 & 2:4 & 38.5 & 91.4 & 85.8 & 86.8 & 85.8 & 63.9 & 80.8 & 88.4 \\
        \midrule
        \slopespace & 1,2 & 16 & 2:4 & 2:4 & 39.2 & 91.3 & 86.4 & 86.6 & 86 & 63.5 & 80.8 & 88.2 \\
        \midrule
        \slopespace & 1,2 & 64 & 2:4 & 2:4 & 42.7 & 90.3 & 85.1 & 86.8 & 85.7 & 66.4 & 80.3 & 88.5 \\
        \midrule
        WANDA & N/A & 0 & 2:4 & 2:4 & 43.0 & 91.4 & 88.3 & 86.9 & 86.1 & 63.5 & 81.9 & 89.6 \\
        \midrule
        WANDA & N/A & 0 & 2:8 & 2:4 & 4.6 & 0.88 & 81.3 & 81 & 83.3 & 53.8 & 76.7 & 83.9  \\
        \midrule
        WANDA & N/A & 0 & 2:4 & 2:8 & 42.1 & 91.7 & 84.4 & 87.2 & 85.6 & 63.5 & 81.5 & 81.9 \\
        \bottomrule
    \end{tabular}
    \label{tab:glue_complete}
    \end{scriptsize}
\end{table}

\section{Integration with Flash Attention}
\label{sec:flash_attention}

To show the compatibility of \slopespace with other optimization methods, we integrate \slopespace with FlashAttention-2 \cite{flashattention2} and show that these approaches are orthogonal in practice and can boost the performance of the model separately.
\autoref{tab:flash_attention} summarizes the speedup achieved with and without \slopespace or FlashAttention-2. As it can be observed, each of these methods can improve the speed of the model both in training and inference, and adding them together will increase the speedup even further.

\begin{table}[t]
\caption{Speedup of \slopespace and FlashAttention-2 (FA2) on OPT models.}
\label{tab:flash_attention}
\centering
\begin{scriptsize}
\begin{sc}
\begin{tabular}{c|ccc|ccc|cc|cc}
\toprule
\textbf{Model} & \multicolumn{3}{c}{\textbf{Training}} & \multicolumn{3}{c}{\textbf{Inference}} & \multicolumn{2}{c}{\textbf{Inference}} & \multicolumn{2}{c}{\textbf{Inference}} 
\\ 
\textbf{Size}& FA2 & \slopespace & \slopespace + FA2 & FA2 & \slopespace & \slopespace + FA2 & FA2 & \slopespace + FA2 & FA2 & \slopespace + FA2
\\
\midrule
66B & 1.28 & 1.13 & 1.53 & 1.36 & 1.34 & 1.99 & 1.31 & 1.95 & 1.30 & 1.91
\\
\midrule
30B & 1.36 & 1.14 & 1.66 & 1.46 & 1.32 & 2.24 & 1.28 & 2.24 & 1.27 & 2.20
\\
\midrule
13B & 1.47 & 1.12 & 1.84 & 1.61 & 1.30 & 2.48 & 1.30 & 2.24 & 1.12 & 2.19 
\\
\midrule
6.7B & 1.60 & 1.08 & 1.94 & 1.71 & 1.21 & 2.50 & 1.13 & 2.50 & 1.12 & 2.45
\\
\midrule
2.6B & 2.26 & 1.05 & 2.56 & 2.47 & 1.07 & 3.23 & 1.05 & 3.09 & 1.00 & 2.92
\\
\bottomrule
\end{tabular}
\end{sc}
\end{scriptsize}
\end{table}

% intentionally empty (placeholder kept so the file survives archiving)

\section{Comparison with dense models}

\label{sec:gpt2_half}

\newtext{To compare the performance of sparse models with dense models of the same size, we have conducted an experiment with GPT2-Small, in which we have reduced the number of transformer blocks in the model to half of GPT2-Small. We call this new configuration GPT2-Half. \autoref{tab:gpt_half} and \autoref{tab:gpt_glue} summarize the accuracy results for GPT2-Half on different zero-shot downstream tasks.

It can be observed that \slopespace outperforms GPT2-Half on average, while dynamic sparse training methods, such as SR-STE perform worse than it. Additionally, it is clear that adding low-rank adapters to the model improves the accuracy of all sparse pretraining methods.
}

\begin{table}[ht]
    \centering
    \caption{Performance comparison across different GPT models, sparsity methods, and LoRA ranks on various tasks. E-SR-STE stands for Extended SR-STE.}
    \vspace{3pt}
    \resizebox{\textwidth}{!}{%
    \begin{tabular}{lccccccc}
        \toprule
        Model        & Method & LoRA (r) & MMLU & Arc Challenge & Open Book QA & Average \\
        \midrule
        GPT2-Small   & Dense    & r = 0    & 22.9 & 20.7           & 16.2         & 19.94   \\
        GPT2-Small       & \slopespace  & r = 0    & 23.0 & 19.3           & 16.0         & 19.43   \\
        GPT2-Small     & \slopespace  & r = 0.05\% & 23.0 & 19.4          & 16.2         & 19.53   \\
        GPT2-Small      & \slopespace  & r = 2.1\%  & 23.0 & 19.3          & 16.4         & 19.57   \\
        GPT2-Small      & E-SR-STE & r = 0    & 24.1 & 18.3           & 12.6         & 18.33   \\
        GPT2-Small      & E-SR-STE & r = 0.05\% & 24.1 & 18.4          & 14.2         & 18.90   \\
        GPT2-Small      & E-SR-STE & r = 2.1\%  & 24.2 & 18.3          & 14.2         & 18.90   \\
        GPT2-Half    & Dense       & r = 0    & 22.9 & 19.5           & 16.0         & 19.47   \\
        \bottomrule
    \end{tabular}
    }
    \label{tab:gpt_half}
\end{table}

\section{Zero-shot GLUE results for GPT}

\label{sec:gpt_glue}

\newtext{We have tested the accuracy of the models on the zero-shot GLUE tasks in Language Model Evaluation Harness \citep{lmharness}. \autoref{tab:gpt_glue} summarizes the achieved GLUE results by different models. It can be seen that \slopespace outperforms SR-STE and GPT2-Half on average. Additionally, SR-STE performs better than GPT2-Half in GLUE task. 
}

\begin{table}[th]
    \centering
    \caption{Performance comparison of GPT models using different sparsity methods and LoRA ranks on GLUE tasks. E-SR-STE stands for Extended SR-STE.}
    \vspace{3pt}
    \resizebox{\textwidth}{!}{%
    \begin{tabular}{lccccccccccc}
        \toprule
        Model          & Method & LoRA (r) & CoLA & MNLI & MNLI & MRPC  & QNLI & QQP  & RTE  & SST2 & Avg \\
                      &        &          &      & (m)  & (mm) &       &      &      &      &      & \\
        \midrule
        GPT2-Small     & Dense  & r = 0    & 0    & 32.4 & 33.2 & 66.9  & 50.3 & 51.8 & 49.8 & 59.3 & 43.2  \\
        GPT2-Small     & \slopespace  & r = 0    & 0    & 34.3 & 34.0 & 72.5  & 50.0 & 48.5 & 50.0 & 52.3 & 42.8  \\
        GPT2-Small     & \slopespace  & r = 0.05\% & 0  & 34.3 & 34.1 & 72.6  & 49.8 & 48.8 & 50.9 & 52.3 & 42.9  \\
        GPT2-Small     & \slopespace  & r = 2.1\%  & 0  & 34.3 & 34.0 & 71.6  & 50.0 & 49.0 & 52.0 & 52.6 & 43.1  \\
        GPT2-Small     & E-SR-STE & r = 0    & 0    & 33.6 & 33.9 & 57.1  & 50.7 & 50.4 & 55.2 & 54.7 & 42.5  \\
        GPT2-Small     & E-SR-STE & r = 0.05\% & 0  & 33.1 & 33.6 & 57.9  & 51.0 & 50.5 & 55.4 & 55.0 & 42.6  \\
        GPT2-Small     & E-SR-STE & r = 2.1\%  & 0  & 33.3 & 33.5 & 58.2  & 51.0 & 50.5 & 55.2 & 55.2 & 42.6  \\
        GPT2-Half      & Dense  & r = 0    & 0.0  & 33.9 & 33.8 & 53.6  & 51.1 & 47.7 & 56.7 & 50.6 & 41.1  \\
        \bottomrule
    \end{tabular}
    }
    \label{tab:gpt_glue}
\end{table}

\section{Extended SR-STE and \transposable~ implementation details}

\newtext{
Before we proceed with the details of Extended SR-STE and \transposable, we clarify the notations used in this chapter and the \transposable~ paper \citep{2_4_pretraining} in \autoref{tab:slope_notation}
}

\begin{table}[ht]
\caption{Description of Key Terms}
\label{tab:slope_notation}

\centering
\begin{tabular}{|c|p{0.6\linewidth}|}
\hline
\textbf{Term} & \textbf{Description} \\ \hline
Sparse Pretraining & Common notation used in SLoPe and FST, indicating the use of  
sparse weights during pretraining. \\ \hline
Dense Finetuning & Notation used in the FST paper, indicating an extended  
pretraining phase. \\ \hline
Downstream Finetuning & Performance after pretraining concludes, used to finetune the  
model for specific downstream tasks. \\ \hline
FST & Extended pretraining technique focused on dense finetuning. \\ \hline
Extended SR-STE & Variation of sparse pretraining extended with additional
fine-tuning. \\ \hline
\end{tabular}
\end{table}

\newtext{
 We compare \slopespace with \transposable~ exclusively for \textit{training speedups and memory savings}. Comparing pretraining quality between \slopespace and \transposable~ is less meaningful because the final models produced by these methods differ significantly in the number of parameters. Specifically, \transposable~ produces a dense model after sparse pretraining and dense finetuning (99\% sparse pretraining + 1\% sparse + low-rank adaptation), while \slopespace produces a sparse model augmented with lightweight low-rank adapters. The number of parameters in the \transposable~ model is approximately 2$\times$ larger than in \slopespace, which makes a direct quality comparison imbalanced. 

 We compare SLoPe with Extended SR-STE in terms of model quality, focusing on understanding the dynamics between static and dynamic masking under an equal number of parameters. This allows for a fair, "apple-to-apple" comparison between the methods (iso-params). We refer to this method as "Extended SR-STE" because, while the original SR-STE approach was designed for use with SGD, the FST paper extended it to support other optimizers.

 The \transposable and Extended SR-STE code are available in \autoref{code:fst} and  \autoref{code:sr_ste} respectively.

}

\begin{lstlisting}[style=PythonStyle, caption={\transposable~ Algorithm}, label={code:fst}]
def forward(ctx, x, weight, weight_sparse, weight_sparse_T, bias):
    ctx.save_for_backward(input, weight_sparse_T, bias)
    ctx.shape = x.shape
    x = x.view(-1, x.shape[-1])
    output = torch.mm(x, weight_sparse.t())
    if bias is None:
        return output.view(*ctx.shape[:-1], -1)
    else:
        return output.view(*ctx.shape[:-1], -1) + bias

def backward(ctx, grad_output):
    grad_output = grad_output.half()
    x, weight_T, bias = ctx.saved_tensors
    grad_input = grad_weight = grad_bias = None
    if ctx.needs_input_grad[0]:
        if grad_output.stride() == (0, 0, 0):
            grad_output = torch.ones_like(grad_output, device=grad_output.device, dtype=grad_output.dtype)
        grad_output = grad_output.view(-1, grad_output.shape[-1])
        grad_input = torch.mm(grad_output, weight_T.t()).view(
            ctx.shape)
    if ctx.needs_input_grad[1]:
        x = x.view(-1, input.shape[-1])
        grad_output = grad_output.view(-1, grad_output.shape[-1])
        grad_weight = torch.mm(to_sparse_semi_structured(grad_output.t(), MVUE24=True), x)
    if ctx.needs_input_grad[2]:
        grad_bias = grad_output.sum(0)
    return grad_input, grad_weight, None, None, grad_bias
\end{lstlisting}

\begin{lstlisting}[style=PythonStyle, caption={Extended SR-STE Algorithm. The weights are stored as dense and are pruned on-the-fly.}, label={code:sr_ste}]
def forward(ctx, input, weight, mask, weight_factor):
    sparse_weight = weight.clone().detach()
    sparse_weight[mask] = 0.
    ctx.save_for_backward(input, sparse_weight, weight_factor * mask * weight)
    output = torch.matmul(input, sparse_weight.t())

    output = output.clone()
    return output

@staticmethod
def backward(ctx, grad_output):
    input, weight, weight_addition_term = ctx.saved_tensors
    input_shape = input.shape
    if input.dim() == 3:
        new_batch_size = input_shape[0] * input_shape[1]
        input = input.reshape(new_batch_size, -1)
        grad_output = grad_output.reshape(new_batch_size, -1)
    grad_output, grad_output_mask = prune_column_wise(grad_output)
    grad_weight = torch.matmul(grad_output.t(), input)
    grad_weight += weight_addition_term

    grad_input = torch.matmul(grad_output, weight)
    grad_input = grad_input.reshape(input_shape)
    return grad_input, grad_weight, None
\end{lstlisting}

\section{Comparison of Depth and Width Pruning}

\newtext{
Depth pruning refers to reducing the number of layers in a model, while width pruning means reducing the size of the weights inside each layer in the model. We have conducted an experiment with depth and width pruning on LLaMA-2-7B \citep{llama2} and Gemma-2-2B and Gemma-2-9B \citep{gemma2} to compare the effects of depth and width pruning on the performance of the models. The configurations used for the models are summarized in \autoref{tab:llama_2_7b}, \autoref{tab:gemma_2b}, \autoref{tab:gemma_9b}. Similar to \citep{2_4_pretraining}, we reduced the aspect ratio of the Up-Sample and Down-Sample modules to half. Please note that this mechanism gives an advantage to width pruning methods, as the number of parameters in the Self-Attention modules remain intact.
}

\begin{table}[ht]
\caption{Model Configurations for LLaMA-2 7B}
\label{tab:llama_2_7b}
\centering
\begin{tabular}{|l|l|}
\hline
\textbf{Pruning Method} & \textbf{Attributes} \\ \hline
Baseline &
\begin{tabular}[c]{@{}l@{}}
base\_emb\_dim: 4096 \\
base\_num\_query\_heads: 32 \\
base\_num\_kv\_heads: 32 \\
base\_mlp\_dim: 11008 \\
base\_num\_decoder\_layers: 32 \\
head\_dim: 128
\end{tabular} \\ \hline
Depth Pruning &
\begin{tabular}[c]{@{}l@{}}
base\_emb\_dim: 4096 \\
base\_num\_query\_heads: 32 \\
base\_num\_kv\_heads: 32 \\
base\_mlp\_dim: 11008 \\
base\_num\_decoder\_layers: 16 \# half the number of layers \\
head\_dim: 128
\end{tabular} \\ \hline
Width Pruning &
\begin{tabular}[c]{@{}l@{}}
base\_emb\_dim: 4096 \\
base\_num\_query\_heads: 32 \\
base\_num\_kv\_heads: 32 \\
base\_mlp\_dim: 5504 \# half the number of dimensions \\
base\_num\_decoder\_layers: 32 \\
head\_dim: 128
\end{tabular} \\ \hline
\end{tabular}
\end{table}

\begin{table}[ht]
\caption{Model Configurations for Gemma-9B}
\label{tab:gemma_9b}
\centering
\begin{tabular}{|l|l|}
\hline
\textbf{Pruning Method} & \textbf{Attributes} \\ \hline
Baseline & 
\begin{tabular}[c]{@{}l@{}}
base\_emb\_dim: 3584 \\
base\_num\_query\_heads: 16 \\
base\_num\_kv\_heads: 8 \\
base\_mlp\_dim: 14336 \\
base\_num\_decoder\_layers: 20 \# merged local and global attention \\
head\_dim: 256
\end{tabular} \\ \hline
Depth Pruning & 
\begin{tabular}[c]{@{}l@{}}
base\_emb\_dim: 3584 \\
base\_num\_query\_heads: 16 \\
base\_num\_kv\_heads: 8 \\
base\_mlp\_dim: 14336 \\
base\_num\_decoder\_layers: 10 \# half the merged layers \\
head\_dim: 256
\end{tabular} \\ \hline
Width Pruning & 
\begin{tabular}[c]{@{}l@{}}
base\_emb\_dim: 3584 \\
base\_num\_query\_heads: 16 \\
base\_num\_kv\_heads: 8 \\
base\_mlp\_dim: 7168 \# half the number of dimensions \\
base\_num\_decoder\_layers: 20 \# merged local and global attention \\
head\_dim: 256
\end{tabular} \\ \hline
\end{tabular}
\end{table}

\begin{table}[ht]
\caption{Model Configurations for Gemma-2B}
\label{tab:gemma_2b}
\centering
\begin{tabular}{|l|l|}
\hline
\textbf{Pruning Method} & \textbf{Attributes} \\ \hline
Baseline &
\begin{tabular}[c]{@{}l@{}}
base\_emb\_dim: 2304 \\
base\_num\_query\_heads: 8 \\
base\_num\_kv\_heads: 4 \\
base\_mlp\_dim: 9216 \\
base\_num\_decoder\_layers: 12 \# merged local and global attention \\
head\_dim: 256
\end{tabular} \\ \hline
Depth Pruning & 
\begin{tabular}[c]{@{}l@{}}
base\_emb\_dim: 2304 \\
base\_num\_query\_heads: 8 \\
base\_num\_kv\_heads: 4 \\
base\_mlp\_dim: 9216 \\
base\_num\_decoder\_layers: 6 \# half the merged layers \\
head\_dim: 256
\end{tabular} \\ \hline
Width Pruning & 
\begin{tabular}[c]{@{}l@{}}
base\_emb\_dim: 2304 \\
base\_num\_query\_heads: 8 \\
base\_num\_kv\_heads: 4 \\
base\_mlp\_dim: 4608 \# half the number of dimensions \\
base\_num\_decoder\_layers: 12 \# merged local and global attention \\
head\_dim: 256
\end{tabular} \\ \hline
\end{tabular}
\end{table}

\begin{figure}[ht]
\begin{center}
{\includegraphics[width=0.49\columnwidth]{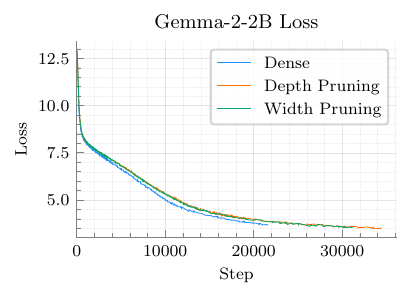} \includegraphics[width=0.49\columnwidth]{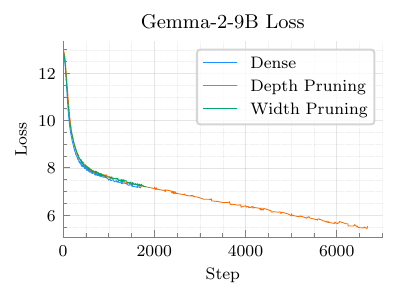}
\includegraphics[width=0.49\columnwidth]{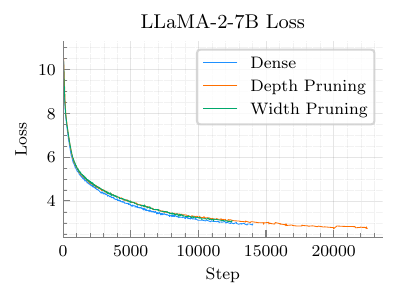}}
\caption{Comparison of the loss of depth and width pruning methods.}
\label{fig:gemma_loss}
\end{center}
\end{figure}

\newtext{
Preliminary retraining loss curves, as shown in \autoref{fig:gemma_loss} suggest no significant difference between depth-pruning and width-pruning during pretraining. Interestingly, in some cases, depth-pruning appears to outperform width-pruning.
}

\section{Proofs}
\subsubsection{\autoref{lemma:sparsity_ratio}}
\label{lemma_proof}

\textit{Proof.} Considering a matrix with $N:M$ column-wise pruned sparsity pattern, we want to prune the matrix using $N:M$ sparsity pattern row-wise as well. Let's define random variable $X$ as the number of added non-zeros to $M$ row-wise consecutive elements and $Y$ as the number of non-zeros in $M$ row-wise consecutive elements. 

\begin{equation}
    \label{eq:average_added_nonzero_1}
    E [ X ] = \sum_{i=1}^{M - N} Pr [ X = i ]i
\end{equation}

Replacing $Pr[X = i] = Pr[ Y = N + i]$ in \autoref{eq:average_added_nonzero_1}, we will get \autoref{eq:average_added_nonzero_2}, where we used a change in dummy variable $j = N + i$.

\begin{equation}
    \label{eq:average_added_nonzero_2}
    E [ X ] = \sum_{i=1}^{M - N} Pr[Y = N + i] i = \sum_{j = N + 1}^M Pr[Y = j] (j - N) 
\end{equation}

Considering the definition of $Y$, it can be inferred that random variable $Y$ has binomial distribution with a success probability of $\frac{N}{M}$. As a result \autoref{eq:single_nonzero_probability} shows the probability mass distribution of $Y$.

\begin{equation}
    \label{eq:single_nonzero_probability}
    Pr[Y = j] = {\binom{M}{j}} s^j (1 - s)^{M - j}; s \triangleq \frac{N}{M}
\end{equation}

By replacing \autoref{eq:single_nonzero_probability} in \autoref{eq:average_added_nonzero_2}, we will get \autoref{eq:average_added_nonzero_3}.

\begin{equation}
    \label{eq:average_added_nonzero_3}
    E[X] = \sum_{j = N + 1}^M {\binom{M}{j}} s^j (1-s)^{M - j} (j - N)
\end{equation}

Let's define random variable $Z$ as the added sparsity ratio to the matrix by the extra pruning. Since $X$ was the number of added non-zeros in $M$ consecutive elements, $E[Z] = \frac{1}{M} E[X]$, and hence:

\begin{equation}
    E[Z] = D(A^R) - D(A^{R,C}) = \sum_{j = N + 1}^M {\binom{M}{j}} s^j (1-s)^{M - j} \frac{j - N}{M}
\end{equation}

\subsubsection{\autoref{theorem:convergence}}

\textit{Proof.} In an optimization problem, we are aiming to find the optimal solution to \autoref{eq:optimization}.

\begin{equation}
    \label{eq:optimization}
    \min_{W_i} E_X[\mathcal{L}(X, W_i)]
\end{equation}

When using backpropagation, which is based on the chain rule in derivation, we compute the gradient in \autoref{eq:gradient_1}.

\begin{equation}
    \label{eq:gradient_1}
    E_X[\nabla_{X_i} \mathcal{L}(X, W_i)] = E_X[\nabla_{Y_i} \mathcal{L} W] 
\end{equation}

Let's define random variable $M$ as a uniformly random mask of 0's and 1's. The mask will be $1$ at each point with a probability of $\frac{N}{M}$. Let's define $O \triangleq E[M]$. $O$ is a matrix of all $\frac{N}{M}$'s. As a result $O \odot W = \frac{N}{M} W$.

\begin{equation}
    E_X[\nabla_{Y_i} \mathcal{L} W] 
 = E_X[\nabla_{Y_i}\mathcal{L} (\frac{M}{N} O \odot W)] = E_X[\nabla_{Y_i}\mathcal{L} (\frac{M}{N} E_M{M} \odot W)]
\end{equation}

By using the linearity of derivation and expectation operators, we can get the result in \autoref{eq:gradient_final}, which proves the theorem.

\begin{equation}
    \label{eq:gradient_final}
    E_X[\nabla_{X_i} \mathcal{L}(X, W_i)] = \frac{M}{N} E_M [E_X [\nabla_{Y_i} \mathcal{L} (M \odot W)]]
\end{equation}

\chapter{Supplementary Material for \optima}
This chapter provides supplementary material for the \optima chapter. \autoref{app:calibration_size} evaluates the sensitivity of \optima to the size of the calibration dataset.

\section{Calibration dataset size sensitivity}
\label{app:calibration_size}

\begin{figure}
    \centering
    \includegraphics{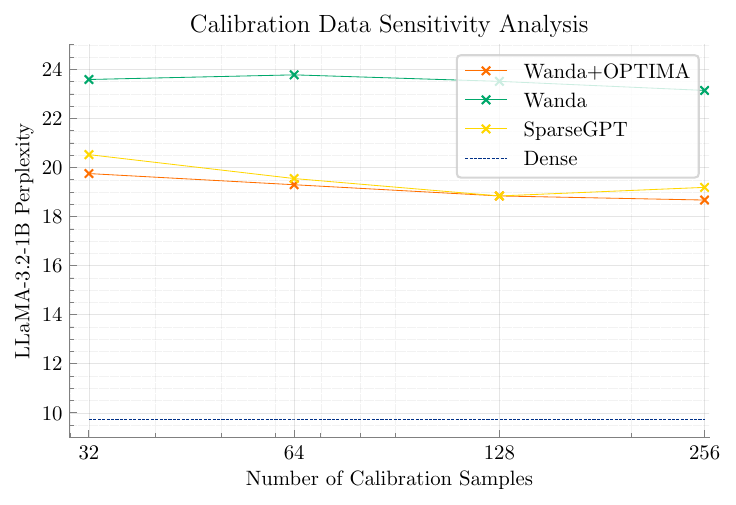}
    \caption{Sensitivity analysis for the number of calibration samples for different pruning methods.}
    \label{fig:calibration_size}
\end{figure}

Similar to previous work (SparseGPT, Wanda, Thanos), \optima leverages a set of calibration data from the C4 dataset to prune the models. \autoref{fig:calibration_size} shows the perplexity of LLaMA-3.2-1B on WikiText2 dataset when pruning the models with various numbers of calibration samples. Our results indicate that unlike the other methods (Wanda and SparseGPT) that have stochastic behavior as the number of samples increases, \optima shows consistent improvement in model quality. However, the improvements are not significant, suggesting robustness to dataset size.

\chapter{Supplementary Material for \patch}
This chapter provides supplementary material for the \patch chapter. \autoref{app::stoicc} describes the integration with STOICC. \autoref{app::per_task} reports per-task accuracy results, and \autoref{app::tile_transfer_learning} explores tile transfer learning across different models.

\section{STOICC Integration}
\label{app::stoicc}

Triton \citep{triton} enables developers to write efficient GPU kernels with a Python-like syntax, but it natively supports only dense matrix operations and cannot handle sparsity. To accelerate the mixed-tile format produced by \patch, we employ the STOICC compiler \citep{stoicc}. STOICC extends Triton with a sparse code-generation backend that allows tiles within a matrix to be either dense or sparse, enabling mixed execution within a single matrix multiplication.

We rely on STOICC’s inspector to autotune both tile sizes and execution schedules (i.e., alternative kernel execution schemes such as split-$K$ parallelism) for the prefill and decoding stages of LLM inference. Matrix compression and metadata generation are determined by the chosen tile size, which must remain consistent across both stages. To address this, we first autotune the decoding stage, which is the primary bottleneck of autoregressive generation, since it is executed once per generated token (e.g., 128 times for 128 new tokens), unlike the single pass of prefill. The optimal tile size identified for decoding is then fixed and reused for prefill, where we perform a second round of autotuning over the remaining independent parameters. 

In contrast, for fully 2:4 sparse matrices, compression is independent of the block size, so they can be autotuned in the same way as dense kernels in Triton without this coupling constraint.

The pseudocode outlining this process, including the handling of dense, fully 2:4 sparse, and mixed-sparsity modules, is provided in \autoref{code:stoicc_pseudo}.

\definecolor{keywordcolor}{rgb}{0.6, 0.1, 0.6}
\definecolor{commentcolor}{rgb}{0.4, 0.4, 0.4}
\definecolor{stringcolor}{rgb}{0.1, 0.5, 0.1}
\definecolor{backgroundcolor}{rgb}{0.95, 0.95, 0.95}

\definecolor{classcolor}{rgb}{0.45, 0.15, 0.65}
\definecolor{funccolor}{rgb}{0.10, 0.35, 0.80}
\definecolor{varcolor}{rgb}{0.00, 0.00, 0.00} 
\definecolor{rulecolor}{gray}{0.25}

\lstdefinelanguage{Pseudo}{
  morekeywords={for,each,in,if,elif,else,continue,return,where},
  sensitive=true,
  morecomment=[l]{//},
  morestring=[b]"
}

\lstdefinestyle{PseudoStyle}{
    backgroundcolor=\color{backgroundcolor},
    basicstyle=\ttfamily\small,
    language=Pseudo,
    keywordstyle=\color{keywordcolor}\bfseries,
    commentstyle=\color{commentcolor}\itshape,
    stringstyle=\color{stringcolor},
    showstringspaces=false,
    frame=single,
    rulecolor=\color{rulecolor},
    numbers=left,
    numberstyle=\tiny\color{gray},
    numbersep=6pt,
    breaklines=true,
    tabsize=2,
    keepspaces=true,
    captionpos=b,
    emph={Inspector,MixedModule,Tensor, STOICC},
    emphstyle=\color{classcolor}\bfseries,
    emph={[2]create_configs,select_2_4_backend,inspect,compress,
          create_module,replace,set_configs, get_sparsity_ratio},
    emphstyle={[2]\color{funccolor}},
    moredelim=**[is][\color{varcolor}]{|}{|},
     mathescape=true,
    escapeinside={(*@}{@*)},
    alsoletter={\in}, 
}

\begingroup
\renewcommand{\lstlistingname}{PseudoCode}

\begin{lstlisting}[style=PseudoStyle,
  caption={Tuning and Converting Model Weights to Mixed Format.},
  label={code:stoicc_pseudo}]
def tune_and_convert_model(|M|, |backend_name|):   
    // backend_name (*@ {\color{commentcolor}$\in$} @*) {"STOICC", "cuSPARSELt"}
    |2_4_backend| = select_2_4_backend(|backend_name|)

    // create all configs & schedules to tune over
    |base_configs| = STOICC.create_configs()
    |inspector|    = Inspector()

    for each |module| in |M|:
        |s| = get_sparsity_ratio(|module.weight|)

        // Keep dense Torch (cuBLAS) module
        if |s| == 0:
            continue

        // Use STOICC or cuSPARSELt for fully 2:4
        elif |s| == 0.5:
            |c|          = |2_4_backend|.compress(|module.weight|)
            |new_module| = |2_4_backend|.create_module(|c|)
            replace(|module|, |new_module|)
            continue

        else:
            |decoding_input| = Tensor(|BS|, |module.weight.shape[1]|)
            |prefill_input|  = Tensor(|BS| * |SL|, |module.weight.shape[1]|)

            // Tune on decoding input first
            |inspector|.set_configs(|base_configs|)
            |best_cfg_dec| = |inspector|.inspect(
                |decoding_input|, 
                |module.weight|, 
                isASparse=False)
            |BN| = |best_cfg_dec|["BLOCK_N"]
            |BK| = |best_cfg_dec|["BLOCK_K"]

            // Tune on prefill using decoding tile sizes
            |prefill_cfg| = STOICC.create_configs(BLOCK_N=|BN|, BLOCK_K=|BK|)
            |inspector|.set_configs(|prefill_cfg|)
            |best_cfg_pre| = |inspector|.inspect(
                |prefill_input|, 
                |module.weight|, 
                isASparse=False)

            |c| = |inspector|.compress(|module.weight|, |BN|, |BK|)
            |mixed_module| = MixedModule(|c|, |best_cfg_dec|, |best_cfg_pre|)
            replace(|module|, |mixed_module|)

    return |M|
\end{lstlisting}

\endgroup

\autoref{table::speedup} reports the measured throughput (tokens processed per second) of LLaMA-2 7B at sparsity levels of 45\%, 35\%, and 25\% with a batch size of 16 on an A6000 GPU. To reduce CPU overhead from launching Triton kernels in PyTorch, we executed generation through CUDA graphs, capturing both the prefill and decoding stages. With sparsity ratios between 25\% and 45\%, our heterogeneous approach achieves 1.18$\times$–1.38$\times$ end-to-end acceleration over the dense baseline. We also report timings on A100 in \autoref{table::speedup_a100}.

\begin{table}[ht]
\centering
\caption{Throughput of LLaMA-2 7B with mixed sparsity compared to the dense model. Measurements taken on an A6000 GPU with batch size 16. Throughput is reported in tokens processed/sec.}
\label{table::speedup}
\begin{tabular}{ccccc}
\toprule
\textbf{Sparsity} & \textbf{Prefill length} & \textbf{Tokens generated} & \textbf{Throughput (tok/s)} & \textbf{Speedup vs. dense} \\
\midrule
0\%  & 128 & 128  & 1023.80 & 1.00$\times$ \\
25\% & 128 & 128  & 1212.79 & 1.18$\times$ \\
35\% & 128 & 128  & 1304.46 & 1.27$\times$ \\
45\% & 128 & 128  & 1410.20 & 1.38$\times$ \\
\midrule
0\%  & 128 & 1024 & 435.42  & 1.00$\times$ \\
25\% & 128 & 1024 & 493.33 & 1.13$\times$ \\
35\% & 128 & 1024 & 515.39 & 1.18$\times$ \\
45\% & 128 & 1024 & 542.87 & 1.25$\times$ \\
\bottomrule
\end{tabular}
\end{table}

\begin{table}[tb]
\centering
\caption{Throughput of LLaMA-2 7B with mixed sparsity compared to the dense model. Measurements taken on an A100 GPU with batch size 16. Throughput is reported in tokens processed/sec.}
\label{table::speedup_a100}
\begin{tabular}{ccccc}
\toprule
\textbf{Sparsity} & \textbf{Prefill length} & \textbf{Tokens generated} & \textbf{Throughput (tok/s)} & \textbf{Speedup vs. dense} \\
\midrule
0\%  & 128 & 128  & 1876.24& 1.00$\times$ \\
25\% & 128 & 128  & 2002.02& 1.07$\times$\\
35\% & 128 & 128  & 2088.98& 1.11$\times$\\
45\% & 128 & 128  & 2180.88& 1.16$\times$\\
\midrule
0\%  & 128 & 1024 & 812.55& 1.00$\times$ \\
25\% & 128 & 1024 & 864.66& 1.06$\times$\\
35\% & 128 & 1024 & 885.90& 1.09$\times$\\
45\% & 128 & 1024 & 907.12& 1.12$\times$\\
\bottomrule
\end{tabular}
\end{table}

\section{Per Task Results}
\label{app::per_task}

This appendix provides detailed per-task accuracy results for the models evaluated in \autoref{sec:patch_results}, covering eight zero-shot downstream tasks: MMLU, PIQA, ARC-Easy, ARC-Challenge, Winogrande, OpenBookQA, RACE, and HellaSwag. The results are presented for each model at various sparsity levels and pruning methods, including our proposed \patchjoint and \patchtile variants, alongside baseline methods such as Magnitude, Wanda, SparseGPT, Thanos, ProxSparse, and MaskLLM. These tables complement the average accuracy and perplexity results reported in \autoref{table::joint_patch} and \autoref{table::tiled_patch} of this chapter, offering a granular view of model performance across individual tasks.

For smaller models (Qwen-2.5 0.5B, LLaMA-3.2 1B, and Gemma-3 1B), we report results using the \patchjoint variant, which jointly optimizes dense tile locations and sparsity patterns within sparse tiles. For larger models (LLaMA-2 7B and LLaMA-3.1 8B), we report results using the memory-efficient \patchtile variant, which optimizes dense tile selections with a fixed 2:4 sparsity mask. The per-task accuracies highlight the effectiveness of our approaches in maintaining robust performance across diverse tasks, even at high sparsity levels, compared to baseline methods.

The following tables detail the per-task accuracies for each model:

\begin{itemize}
    \item \textbf{Qwen-2.5 0.5B}: \autoref{table:qwen_joint_patch} presents the per-task accuracies for the \patchjoint variant and baselines at 0\% and 50\% sparsity, with \patchjoint evaluated at 25\%, 35\%, and 45\% sparsity.
    \item \textbf{LLaMA-2 7B}: \autoref{table:llama2_tile_patch} shows the per-task accuracies for the \patchtile variant and baselines, with \patchtile evaluated at 25\%, 35\%, and 45\% sparsity.
    \item \textbf{LLaMA-3.1 8B}: \autoref{table:llama31_tile_patch} provides the per-task accuracies for the \patchtile variant and baselines, with \patchtile at 25\%, 35\%, and 45\% sparsity.
    \item \textbf{LLaMA-3.2 1B}: \autoref{table:llama32_joint_patch} reports the per-task accuracies for the \patchjoint variant and baselines, with \patchjoint at 25\%, 35\%, and 45\% sparsity.
    \item \textbf{Gemma-3 1B}: \autoref{table:gemma_joint_patch} details the per-task accuracies for the \patchjoint variant and baselines, with \patchjoint at 25\%, 35\%, and 45\% sparsity.
\end{itemize}

These results enable a deeper analysis of the task-specific performance trends, demonstrating the flexibility and robustness of \patchjoint and \patchtile in achieving high accuracy across diverse tasks while maintaining hardware-friendly sparsity patterns.

\begin{table}[H]
\centering
\small
\caption{Model quality (task accuracy across eight zero-shot tasks, reported in \%) for Qwen-2.5 0.5B with different pruning methods. \patchjoint optimizes dense tile locations and sparsity patterns, enabling a flexible sparsity-quality tradeoff.}
\label{table:qwen_joint_patch}
\setlength{\tabcolsep}{3pt}
\resizebox{\textwidth}{!}{
\begin{tabular}{l l l c c c c c c c c c}
\toprule
\textbf{Sparsity} & \textbf{Method} & \textbf{Pattern} & \textbf{MMLU} & \textbf{PIQA} & \textbf{ARC-E} & \textbf{ARC-C} & \textbf{WinoG.} & \textbf{OBQA} & \textbf{RACE} & \textbf{HellaS.} & \textbf{Avg} \\
\midrule
0\% & Dense & - & 47.71 & 70.24 & 64.48 & 29.52 & 56.20 & 24.20 & 35.02 & 40.63 & 46.00 \\
\midrule
50\% & Magnitude & 2:4 & 23.00 & 54.24 & 31.23 & 19.20 & 49.96 & 13.60 & 23.44 & 26.59 & 30.16 \\
     & Wanda & 2:4 & 24.43 & 58.71 & 43.18 & 17.75 & 51.62 & 12.20 & 26.32 & 29.58 & 32.97 \\
     & SparseGPT & 2:4 & 22.93 & 60.77 & 46.60 & 20.82 & 52.88 & 14.00 & 29.57 & 30.93 & 34.81 \\
     & Thanos & 2:4 & 22.97 & 60.17 & 45.37 & 19.20 & 53.59 & 15.20 & 31.00 & 31.31 & 34.85 \\
     & ProxSparse & 2:4 & 23.00 & 57.34 & 40.53 & 18.26 & 48.62 & 14.00 & 25.65 & 29.02 & 32.05 \\
     & MaskLLM & 2:4 & 25.11 & 67.03 & 56.57 & 23.98 & 52.57 & 20.20 & 33.30 & 35.90 & 39.33 \\
\midrule
\rowcolor{PatchShade} 45\% & \patchjoint & Dense/2:4 Tiles & 27.39 & 68.44 & 59.13 & 25.77 & 53.67 & 19.80 & 32.15 & 35.99 & 40.29 \\
\rowcolor{PatchShade} 35\% & \patchjoint & Dense/2:4 Tiles & 29.04 & 68.88 & 60.40 & 26.37 & 55.09 & 20.40 & 32.44 & 36.58 & 41.15 \\
\rowcolor{PatchShade} 25\% & \patchjoint & Dense/2:4 Tiles & 30.89 & 69.15 & 62.79 & 29.10 & 55.33 & 20.00 & 34.16 & 37.71 & 42.39 \\
\bottomrule
\end{tabular}
}

\end{table}

\begin{table}[H]
\centering
\small
\caption{Model quality (task accuracy across eight zero-shot tasks, reported in \%) for LLaMA-2 7B with different pruning methods. \patchtile optimizes tile-based sparsity, enabling a flexible sparsity-quality tradeoff.}
\label{table:llama2_tile_patch}
\setlength{\tabcolsep}{3pt}
\resizebox{\textwidth}{!}{
\begin{tabular}{l l l c c c c c c c c c}
\toprule
\textbf{Sparsity} & \textbf{Method} & \textbf{Pattern} & \textbf{MMLU} & \textbf{PIQA} & \textbf{ARC-E} & \textbf{ARC-C} & \textbf{WinoG.} & \textbf{OBQA} & \textbf{RACE} & \textbf{HellaS.} & \textbf{Avg} \\
\midrule
0\% & Dense & - & 41.82 & 78.07 & 76.35 & 43.52 & 69.06 & 31.40 & 39.52 & 57.13 & 54.61 \\
\midrule
50\% & Magnitude & 2:4 & 25.82 & 70.02 & 61.78 & 30.12 & 61.01 & 21.80 & 31.48 & 45.45 & 43.44 \\
     & Wanda & 2:4 & 25.80 & 71.00 & 63.80 & 30.29 & 61.09 & 25.20 & 35.50 & 41.75 & 44.30 \\
     & SparseGPT & 2:4 & 26.17 & 70.73 & 63.80 & 30.63 & 65.04 & 24.00 & 37.13 & 43.18 & 45.09 \\
     & Thanos & 2:4 & 25.27 & 70.78 & 63.43 & 30.97 & 64.56 & 23.80 & 36.46 & 43.11 & 44.80 \\
     & ProxSparse & 2:4 & 26.77 & 71.60 & 65.70 & 33.02 & 62.90 & 24.20 & 35.31 & 47.84 & 45.92 \\
     & MaskLLM & 2:4 & 27.65 & 74.76 & 69.44 & 35.58 & 65.04 & 26.80 & 38.56 & 51.15 & 48.62 \\
\midrule
\rowcolor{PatchShade} 45\% & \patchtile & Dense/2:4 Tiles & 27.28 & 75.41 & 70.16 & 35.84 & 65.27 & 27.60 & 38.76 & 51.61 & 48.99 \\
\rowcolor{PatchShade} 35\% & \patchtile & Dense/2:4 Tiles & 29.93 & 76.71 & 70.88 & 36.95 & 65.67 & 28.20 & 39.33 & 52.96 & 50.08 \\
\rowcolor{PatchShade} 25\% & \patchtile & Dense/2:4 Tiles & 32.33 & 76.99 & 72.81 & 38.57 & 68.27 & 29.80 & 39.52 & 54.34 & 51.58 \\
\bottomrule
\end{tabular}
}

\end{table}

\begin{table}[H]
\centering
\small
\caption{Model quality (task accuracy across eight zero-shot tasks, reported in \%) for LLaMA-3.1 8B with different pruning methods. \patchtile optimizes tile-based sparsity, enabling a flexible sparsity-quality tradeoff.}
\label{table:llama31_tile_patch}
\setlength{\tabcolsep}{3pt}
\resizebox{\textwidth}{!}{
\begin{tabular}{l l l c c c c c c c c c}
\toprule
\textbf{Sparsity} & \textbf{Method} & \textbf{Pattern} & \textbf{MMLU} & \textbf{PIQA} & \textbf{ARC-E} & \textbf{ARC-C} & \textbf{WinoG.} & \textbf{OBQA} & \textbf{RACE} & \textbf{HellaS.} & \textbf{Avg} \\
\midrule
0\% & Dense & - & 63.57 & 80.09 & 81.44 & 51.37 & 73.48 & 33.40 & 39.14 & 60.02 & 60.31 \\
\midrule
50\% & Magnitude & 2:4 & 23.06 & 63.82 & 45.33 & 25.94 & 53.91 & 15.20 & 26.70 & 33.49 & 35.93 \\
     & Wanda & 2:4 & 27.85 & 68.88 & 58.33 & 26.71 & 60.93 & 19.00 & 33.78 & 38.70 & 41.77 \\
     & SparseGPT & 2:4 & 31.82 & 70.46 & 63.85 & 31.74 & 64.56 & 21.60 & 37.22 & 42.99 & 45.53 \\
     & Thanos & 2:4 & 34.23 & 70.40 & 63.13 & 31.40 & 63.61 & 23.20 & 37.03 & 42.75 & 45.72 \\
     & ProxSparse & 2:4 & 29.89 & 71.71 & 62.63 & 33.28 & 58.56 & 23.80 & 35.22 & 46.03 & 45.14 \\
     & MaskLLM & 2:4 & 42.47 & 77.04 & 73.15 & 40.19 & 68.43 & 28.80 & 38.28 & 54.04 & 52.80 \\
\midrule
\rowcolor{PatchShade} 45\% & \patchtile & Dense/2:4 Tiles & 47.32 & 77.96 & 73.61 & 41.89 & 68.03 & 29.00 & 36.56 & 54.44 & 53.60 \\
\rowcolor{PatchShade} 35\% & \patchtile & Dense/2:4 Tiles & 51.15 & 77.97 & 76.14 & 42.41 & 69.46 & 31.40 & 38.18 & 55.54 & 55.28 \\
\rowcolor{PatchShade} 25\% & \patchtile & Dense/2:4 Tiles & 52.95 & 77.75 & 77.57 & 44.62 & 70.56 & 31.80 & 39.90 & 56.69 & 56.48 \\
\bottomrule
\end{tabular}
}

\end{table}

\begin{table}[H]
\centering
\small
\caption{Model quality (task accuracy across eight zero-shot tasks, reported in \%) for LLaMA-3.2 1B with different pruning methods. \patchjoint optimizes dense tile locations and sparsity patterns, enabling a flexible sparsity-quality tradeoff.}
\label{table:llama32_joint_patch}
\setlength{\tabcolsep}{3pt}
\resizebox{\textwidth}{!}{
\begin{tabular}{l l l c c c c c c c c c}
\toprule
\textbf{Sparsity} & \textbf{Method} & \textbf{Pattern} & \textbf{MMLU} & \textbf{PIQA} & \textbf{ARC-E} & \textbf{ARC-C} & \textbf{WinoG.} & \textbf{OBQA} & \textbf{RACE} & \textbf{HellaS.} & \textbf{Avg} \\
\midrule
0\% & Dense & - & 37.57 & 74.54 & 65.53 & 31.32 & 60.62 & 26.40 & 37.89 & 47.76 & 47.70 \\
\midrule
50\% & Magnitude & 2:4 & 23.31 & 53.81 & 27.74 & 18.94 & 51.38 & 11.80 & 24.02 & 26.26 & 29.66 \\
     & Wanda & 2:4 & 22.90 & 58.11 & 37.08 & 19.20 & 49.09 & 13.20 & 25.17 & 28.11 & 31.61 \\
     & SparseGPT & 2:4 & 22.93 & 61.43 & 45.03 & 22.35 & 54.93 & 15.80 & 29.86 & 32.08 & 35.55 \\
     & Thanos & 2:4 & 23.12 & 62.40 & 44.91 & 21.76 & 54.30 & 16.00 & 31.10 & 32.09 & 35.71 \\
     & ProxSparse & 2:4 & 22.96 & 60.83 & 39.44 & 20.31 & 51.54 & 16.80 & 25.17 & 31.37 & 33.55 \\
     & MaskLLM & 2:4 & 26.28 & 69.10 & 57.41 & 25.85 & 55.48 & 21.40 & 32.82 & 39.94 & 41.04 \\
\midrule
\rowcolor{PatchShade} 45\% & \patchjoint & Dense/2:4 Tiles & 23.81 & 70.89 & 60.77 & 27.22 & 56.27 & 22.80 & 34.07 & 40.78 & 42.08 \\
\rowcolor{PatchShade} 35\% & \patchjoint & Dense/2:4 Tiles & 25.13 & 71.32 & 60.27 & 29.18 & 57.06 & 22.00 & 34.64 & 42.17 & 42.72 \\
\rowcolor{PatchShade} 25\% & \patchjoint & Dense/2:4 Tiles & 28.59 & 71.44 & 61.57 & 28.67 & 58.25 & 23.20 & 35.22 & 43.52 & 43.81 \\
\bottomrule
\end{tabular}
}

\end{table}

\begin{table}[H]
\centering
\small
\caption{Model quality (accuracy across eight zero-shot tasks) for Gemma-3 1B with different pruning methods. \patchjoint optimizes dense tile locations and sparsity patterns, enabling a flexible sparsity-quality tradeoff.}
\label{table:gemma_joint_patch}
\setlength{\tabcolsep}{3pt}
\resizebox{\textwidth}{!}{
\begin{tabular}{l l l c c c c c c c c c}
\toprule
\textbf{Sparsity} & \textbf{Method} & \textbf{Pattern} & \textbf{MMLU} & \textbf{PIQA} & \textbf{ARC-E} & \textbf{ARC-C} & \textbf{WinoG.} & \textbf{OBQA} & \textbf{RACE} & \textbf{HellaS.} & \textbf{Avg} \\
\midrule
0\% & Dense & - & 24.95 & 75.03 & 71.84 & 34.90 & 58.64 & 28.60 & 34.83 & 47.26 & 47.01 \\
\midrule
50\% & Magnitude & 2:4 & 23.08 & 59.79 & 37.29 & 17.66 & 50.59 & 14.00 & 22.87 & 27.97 & 31.66 \\
     & Wanda & 2:4 & 23.96 & 59.52 & 48.02 & 18.34 & 51.22 & 14.20 & 27.85 & 30.18 & 34.16 \\
     & SparseGPT & 2:4 & 23.62 & 62.79 & 49.83 & 19.03 & 51.54 & 15.20 & 30.62 & 31.99 & 35.58 \\
     & Thanos & 2:4 & 23.44 & 62.24 & 48.86 & 18.34 & 50.12 & 15.60 & 30.81 & 31.28 & 35.09 \\
     & ProxSparse & 2:4 & 23.10 & 64.25 & 50.72 & 21.59 & 53.43 & 18.00 & 29.09 & 32.86 & 36.63 \\
     & MaskLLM & 2:4 & 25.03 & 69.91 & 60.27 & 27.65 & 56.27 & 21.20 & 34.55 & 39.84 & 41.84 \\
\midrule
\rowcolor{PatchShade} 45\% & \patchjoint & Dense/2:4 Tiles & 23.54 & 71.65 & 63.97 & 27.47 & 57.30 & 23.60 & 33.49 & 41.39 & 42.80 \\
\rowcolor{PatchShade} 35\% & \patchjoint & Dense/2:4 Tiles & 25.38 & 72.31 & 63.80 & 27.39 & 56.67 & 24.00 & 34.74 & 42.07 & 43.30 \\
\rowcolor{PatchShade} 25\% & \patchjoint & Dense/2:4 Tiles & 25.45 & 71.87 & 66.16 & 30.55 & 57.85 & 22.80 & 34.55 & 43.33 & 44.07 \\
\bottomrule
\end{tabular}
}

\end{table}

\section{Tile Transfer Learning}
\label{app::tile_transfer_learning}

We also test whether initializing tile logits with priors from one-shot pruning methods improves performance, as done in MaskLLM \citep{maskllm}. In our case, the initialization is derived from one-shot pruning with unstructured sparsity. We initialize tiles that retain more nonzeros after unstructured pruning with positive logits (favoring dense assignment), while the remaining tiles receive negative logits, controlled by a strength parameter. The number of tiles initialized as dense is selected such that the overall layer-wise sparsity target is satisfied. As shown in \autoref{ablation:tile_prior}, the choice of prior has little impact on final performance: all priors yield nearly identical perplexity, with random initialization often performing best. This is likely because the global sparsity target enables dynamic reallocation of sparsity across layers during training, overriding the effect of any fixed initialization. For consistency with prior work, we adopt SparseGPT initialization in all experiments.

\begin{table}[H]
\centering
\small
\setlength{\tabcolsep}{6pt}
\caption{Perplexity ($\downarrow$) under different tile prior initializations. All priors yield nearly identical performance, suggesting that the global sparsity target allows dynamic reallocation of sparsity during training, overriding the influence of fixed initialization.}

\label{ablation:tile_prior}
\begin{tabular}{l c c c c c}
\toprule
\textbf{Sparsity (0.5B)} & \textbf{Nothing} & \textbf{SparseGPT} & \textbf{Wanda} & \textbf{Magnitude} & \textbf{Random} \\
\midrule
45\% & 14.80 & 14.57 & 14.50 & \textbf{14.48} & 14.51 \\
35\% & 13.97 & 13.84 & 13.87 & 13.85 & \textbf{13.79} \\
25\% & 13.47 & 13.47 & 13.37 & 13.44 & \textbf{13.33} \\
\bottomrule
\end{tabular}
\end{table}

\chapter{Supplementary Material for \slim}
This chapter provides supplementary material for the \slim chapter. We begin by defining the notations used throughout this work in \autoref{appendix:notation}. \autoref{appendix:input_quantization} evaluates input quantization, and \autoref{appendix:additional_finetuning_results} presents additional fine-tuning results. Language modeling experiments are reported in \autoref{appendix:language_modeling}, with additional sparse and quantized accuracy results in \autoref{appendix:accuracy_results}. \autoref{appendix:sparsity_vs_quantization} compares sparsity and quantization, while \autoref{appendix:speedup} provides additional speedup results. \autoref{appendix:fine_tuning_overhead} details the fine-tuning costs, and theoretical analyses of memory and computation reductions are provided in \autoref{appendix:memory_reduction} and \autoref{appendix:compute_reduction}, respectively. \autoref{sec:compression_cost} reports compression costs, \autoref{sec:rank_analysis} explores the impact of rank choices on low-rank adapters, and \autoref{appendix:dataset_sensitivity} examines sensitivity to the calibration dataset. \autoref{appendix:sparsity_analysis} analyzes the effects of different sparsity ratios, and \autoref{appendix:group_quantization_challenges} discusses the challenges of group quantization.

\section{Notations}
\label{appendix:notation}

\autoref{tab:slim_notation} details the key notations, particularly for \autoref{sec:results}.

\begin{table}[htbp]
\caption{Key notation definitions used in the experimental results (\autoref{sec:results}).}
\label{tab:slim_notation}
\begin{center}
\begin{tabular}{|c|p{0.7\linewidth}|}
\hline
Term & Description
\\
\hline
Naive-LoRA & A one-shot low-rank adapter that minimizes the norm of the difference between the original and the compressed weights.
\\
\hline
\slim-LoRA & A saliency-based one-shot low-rank adapter that minimizes the saliency of the difference between the original and the compressed weights.
\\
\hline
$Q$ (Superscript) & $^Q$ indicates that the compression method quantizes the low-rank adapters as well.
\\
\hline
+ FT & + FT shows a short fine-tuning phase on 300,000 tokens from the C4 dataset. 
\\
\hline
\end{tabular}
\end{center}
\end{table}

\section{Input Quantization}
\label{appendix:input_quantization}

We evaluate \slimspace with 8-bit input quantization to assess its impact on accuracy. We use AbsMax uniform quantization with a single parameter per input tensor and apply FP8 format \citep{fp8} for weight quantization. The choice between E4M3 and E5M2 depends on the tensor’s maximum value; if it exceeds E4M3’s range, we switch to E5M2 for greater expressivity. Next, we examine how input quantization affects model accuracy.

\autoref{tab:input_quantization_accuracy} presents accuracy results for different \slimspace variants with input quantization. A comparison with \autoref{tab:fine-tuning-accuracy-full}, which reports accuracy without input quantization, reveals minimal accuracy loss, demonstrating \slim's robustness. For further validation, we extend these experiments to language modeling tasks (\autoref{appendix:language_modeling}).

\begin{table}[htbp]
\caption{Average zero-shot accuracy of LLaMA-2 and OPT models with \textbf{4-bit weight and 8-bit input quantization}. $\uparrow$ indicates better performance.}
\label{tab:input_quantization_accuracy}
\begin{center}
\resizebox{\textwidth}{!}{%
\begin{tabular}{llllllllll}
\multicolumn{1}{c}{Pruning/LoRA}  &\multicolumn{1}{c}{Weight} & \multicolumn{6}{c}{OPT} & \multicolumn{2}{c}{LLaMA-2}
\\
\multicolumn{1}{c}{Method}  &\multicolumn{1}{c}{Quantization} & 125M & 350M & 1.3B & 2.7B & 6.7B & 13B & 7B & 13B
\\ \hline \\[-8pt]
Dense & - & 35.9 & 37.1 & 43.4 & 45.5 & 48.3 & 48.7 & 56.6 & 60.8 
\\ \hline \\[-8pt]
\textbf{50\% 2:4} \\
\slim-LoRA & \quantization & 34.85 & 34.27 & 40.29 & 42.58 & 45.78 & 46.21 & 50.99 & 54.66
\\
\slim-LoRA + FT & \quantization & 35.28 & 34.33 & 41.14 & 43.29 & 46.44 & 47.33 & 51.77 & 56.28
\\
\slim-LoRA$^Q$ & \quantization & 34.30 & 33.85 & 39.92 & 41.99 & 46.08 & 45.94 & 50.70 & 53.56
\\
\slim-LoRA$^Q$ + FT & \quantization & 34.92 & 34.80 & 41.66 & 43.69 & 46.03 & 46.87 & 50.26 & 56.28
\\ \hline \\[-8pt]
\textbf{50\% Unstructured}  \\
\slim-LoRA & \quantization & 35.12 & 34.86 & 41.94 & 43.53 & 47.27 & 47.70 & 54.28 & 57.82
\\
\slim-LoRA + FT & \quantization & 35.18 & 35.30 & 42.37 & 44.02 & 47.01 & 48.52 & 54.43 & 57.70
\\
\slim-LoRA$^Q$ & \quantization & 35.26 & 34.67 & 41.48 & 43.46 & 47.25 & 47.76 & 53.91 & 57.16
\\
\slim-LoRA$^Q$ + FT & \quantization & 35.52 & 35.31 & 42.66 & 44.50 & 47.08 & 48.53 & 53.23 & 57.55
\\ \hline 
\end{tabular}
}
\end{center}
\end{table}

\section{Additional fine-tuning results}
\label{appendix:additional_finetuning_results}

To complement the results in \autoref{sec:results}, we provide accuracy measurements for PEFT-based fine-tuning of low-rank adapters on the OPT and LLaMA-2 model families in \autoref{tab:fine-tuning-accuracy-full} while showing the accuracy results without fine-tuning for comparison. The results confirm the previously observed trend: lightweight fine-tuning enhances the accuracy of all baselines, with \slim-LoRA achieving the most significant improvements due to its saliency-based design.

\begin{table}[htbp]
\caption{Effects of fine-tuning on the average zero-shot accuracy of LLaMA-2 and OPT models with 50\% sparsity and 4-bit weight quantization. $\uparrow$ indicates better performance.}
\label{tab:fine-tuning-accuracy-full}
\begin{center}
\resizebox{\textwidth}{!}{%
\begin{tabular}{llllllllll}
\multicolumn{1}{c}{Pruning/LoRA}  &\multicolumn{1}{c}{Weight} & \multicolumn{6}{c}{OPT} & \multicolumn{2}{c}{LLaMA-2}
\\
\multicolumn{1}{c}{Method}  &\multicolumn{1}{c}{Quantization} & 125M & 350M & 1.3B & 2.7B & 6.7B & 13B & 7B & 13B
\\ \hline \\[-8pt]
Dense & - & 35.9 & 37.1 & 43.4 & 45.5 & 48.3 & 48.7 & 56.6 & 60.8 
\\ \hline \\[-8pt]
\textbf{50\% 2:4} \\
Naive-LoRA & \quantization & 34.28 & 33.38 & 38.36 & 41.21 & 44.91 & 45.25 & 48.45 & 51.94
\\
Naive-LoRA + FT & \quantization & 34.41 & 34.70 & 39.72 & 42.88 & 46.16 & 46.76 & 50.89 & 55.70
\\
\slim-LoRA & \quantization & 34.62 & 34.36 & 40.61 & 42.73 & 45.99 & 46.09 & 51.15 & 54.94
\\
\slim-LoRA + FT & \quantization & \textbf{35.03} & 34.58 & 41.11 & 43.35 & \textbf{46.71} & \textbf{47.25} & \textbf{52.12} & \textbf{56.60}
\\
\slim-LoRA$^Q$ & \quantization & 34.43 & 34.30 & 40.11 & 42.37 & 46.33 & 46.24 & 51.02 & 53.55
\\
\slim-LoRA$^Q$ + FT & \quantization & 34.92 & \textbf{34.85} & \textbf{41.84} & \textbf{43.87} & 46.31 & 46.91 & 48.31 & 56.50
\\ \hline \\[-8pt]
\textbf{50\% Unstructured}  \\
Naive-LoRA & \quantization & 34.77 & 34.23 & 40.40 & 43.37 & 46.64 & 47.30 & 51.52 & 55.33
\\
Naive-LoRA + FT & \quantization & 35.70 & 35.47 & 41.89 & 44.16 & 47.08 & 47.78 & 52.90 & 57.08
\\
\slim-LoRA & \quantization & 35.20 & 35.32 & 41.85 & 43.48 & 47.08 & 47.96 & 54.26 & 57.85
\\
\slim-LoRA + FT & \quantization & 35.59 & \textbf{35.71} & 42.37 & \textbf{44.58} & \textbf{47.69} & 48.26 & \textbf{54.69} & \textbf{57.96}
\\
\slim-LoRA$^Q$ & \quantization & \textbf{35.35} & 35.13 & 41.74 & 43.63 & 47.16 & 47.86 & 54.18 & 57.33
\\
\slim-LoRA$^Q$ + FT & \quantization & 35.65 & 35.67 & \textbf{42.74} & 44.54 & 47.48 & \textbf{48.40} & 53.57 & 57.78
\\ \hline 
\end{tabular}
}
\end{center}
\end{table}

\section{Language modeling experiments}
\label{appendix:language_modeling}

We evaluate all benchmarks from \autoref{sec:results} and \autoref{appendix:input_quantization} on the WikiText2 language modeling task. \autoref{tab:sparse_quantized_24_perplexity} and \autoref{tab:sparse_quantized_unstructured_perplexity} show perplexity results for 4-bit quantized models with 2:4 and unstructured sparsity, respectively. \autoref{tab:input_quantization_perplexity} summarizes the results for 8-bit input quantization. To examine sparsity and quantization independently, \autoref{tab:sparse_perplexity} and \autoref{tab:quantized_perplexity} report results for pruning-only and quantization-only models. Consistent with \autoref{sec:results}, \slimspace achieves superior performance across all settings.

\begin{table}[H]
\caption{Perplexity of LLaMA-2 and OPT models with \textbf{2:4 sparsity and 4-bit weight quantization} on WikiText-2 dataset language modeling task. $\downarrow$ indicates better performance.}
\label{tab:sparse_quantized_24_perplexity}
\begin{center}
\resizebox{\textwidth}{!}{%
\begin{tabular}{llllllllll}
\multicolumn{1}{c}{Pruning/LoRA}  &\multicolumn{1}{c}{Weight} & \multicolumn{6}{c}{OPT} & \multicolumn{2}{c}{LLaMA-2}
\\
\multicolumn{1}{c}{Method}  &\multicolumn{1}{c}{Quantization} & 125M & 350M & 1.3B & 2.7B & 6.7B & 13B & 7B & 13B
\\ \hline \\
Dense & - & 27.66 & 22.00 & 14.62 & 12.47 & 10.86 & 10.13 & 5.12 & 4.57
\\ \hline \\
Magnitude & Group AbsMax & 5.1E2 & 4.4E2 & 1.2E3 & 1.3E3 & 3.6E2 & 4.9E2 & 86.34 & 8.98
\\
SparseGPT & Group OPTQ & 78.18 & 59.86 & 27.36 & 18.62 & 15.31 & 13.25 & 15.01 & 8.97
\\
Wanda &  Group AbsMax & 1.8E2 & 1.3E2 & 32.76 & 24.48 & 17.29 & 16.86 & 13.46 & 8.70
\\
Wanda & AWQ & 9.3E1 & 8.1E5 & 29.56 & 22.91 & 16.28 & 16.72 & 12.79 & OOM
\\
Wanda & OmniQuant & 9.7E1 & NaN & 33.61 & 25.89 & 19.09 & OOM & 12.77 & OOM
\\
Wanda & AffineQuant & 9.7E1 & NaN & 30.32 & 1.6E3 & 16.85 & OOM & 12.21 & OOM
\\
JSQ & JSQ & 3.5E3 & 2.5E4 & 67.36 & 3.2E3 & 22.50 & 5.5E2 & 11.69 & 8.05
\\ \hline \\
Naive-LoRA & Group AbsMax & 69.23 & 50.02 & 20.52 & 16.05 & 12.83 & 13.12 & 8.04 & 6.38
\\
Naive-LoRA & \quantization & 83.08 & 58.69 & 27.06 & 20.92 & 14.29 & 13.20 & 8.19 & 7.09
\\
Naive-LoRA + FT & \quantization & 51.82 & 38.84 & 20.59 & 16.19 & 13.13 & 12.55 & 6.96 & 6.01
\\ \hline \\
\slim-LoRA & \quantization & 57.91 & 50.09 & 19.64 & 15.65 & 12.71 & 12.13 & 7.56 & 6.50
\\
\slim-LoRA + FT & \quantization & 44.03 & 37.32 & 18.25 & 14.89 & 12.68 & 12.06 & \textbf{6.70} & 6.60
\\
\slim-LoRA$^Q$ & \quantization & 53.09 & 46.96 & 19.62 & 16.01 & 12.48 & 12.15 & 7.75 & 6.96
\\
\slim-LoRA$^Q$ + FT & \quantization & \textbf{42.80} & \textbf{37.39} & \textbf{18.38} & \textbf{15.40} & \textbf{12.65} & \textbf{12.35} & 7.08 & \textbf{6.36}
\\ \hline 
\end{tabular}
}
\end{center}
\end{table}

\begin{table}[htbp]
\caption{Perplexity of LLaMA-2 and OPT models with \textbf{4-bit weight and 8-bit input quantization}. $\downarrow$ indicates better performance.}
\label{tab:input_quantization_perplexity}
\begin{center}
\resizebox{\textwidth}{!}{%
\begin{tabular}{llllllllll}
\multicolumn{1}{c}{Pruning/LoRA}  &\multicolumn{1}{c}{Weight} & \multicolumn{6}{c}{OPT} & \multicolumn{2}{c}{LLaMA-2}
\\
\multicolumn{1}{c}{Method}  &\multicolumn{1}{c}{Quantization} & 125M & 350M & 1.3B & 2.7B & 6.7B & 13B & 7B & 13B
\\ \hline \\[-8pt]
Dense & - & 27.66 & 22.00 & 14.62 & 12.47 & 10.86 & 10.13 & 5.12 & 4.57
\\ \hline \\[-8pt]
\textbf{50\% 2:4} \\
\slim-LoRA & \quantization & 48.4 & 49.6 & 16.6 & 16.2 & 12.9 & 12.3 & 7.2 & 6.5
\\
\slim-LoRA + FT & \quantization & 39.8 & 37.5 & 18.3 & 15.5 & 12.8 & 12.1 & 6.6 & 5.8
\\
\slim-LoRA$^Q$ & \quantization & 54.2 & 50.8 & 20.8 & 16.8 & 13.0 & 12.4 & 7.8 & 7.0
\\
\slim-LoRA$^Q$ + FT & \quantization & 43.4 & 39.1 & 19.3 & 16.0 & 13.1 & 12.6 & 7.1 & 5.8
\\ \hline \\[-8pt]
\textbf{50\% Unstructured}  \\
\slim-LoRA & \quantization & 36.8 & 31.1 & 16.8 & 14.0 & 11.7 & 10.9 & 6.1 & 5.4
\\
\slim-LoRA + FT & \quantization & 33.8 & 28.6 & 16.5 & 14.0 & 12.0 & 11.5 & 5.9 & 5.2
\\
\slim-LoRA$^Q$ & \quantization & 39.5 & 31.3 & 17.3 & 14.2 & 11.8 & 10.9 & 6.3 & 5.6
\\
\slim-LoRA$^Q$ + FT & \quantization & 35.6 & 29.1 & 17.0 & 14.3 & 12.2 & 11.7 & 6.2 & 5.5
\\ \hline 
\end{tabular}
}
\end{center}
\end{table}

\begin{table}[H]
\caption{Perplexity of LLaMA-2 and OPT models with \textbf{unstructured sparsity and 4-bit weight quantization} on WikiText-2 dataset language modeling task. $\downarrow$ indicates better performance.}
\label{tab:sparse_quantized_unstructured_perplexity}
\begin{center}
\resizebox{\textwidth}{!}{%
\begin{tabular}{llllllllll}
\multicolumn{1}{c}{Pruning/LoRA}  &\multicolumn{1}{c}{Weight} & \multicolumn{6}{c}{OPT} & \multicolumn{2}{c}{LLaMA-2}
\\
\multicolumn{1}{c}{Method}  &\multicolumn{1}{c}{Quantization} & 125M & 350M & 1.3B & 2.7B & 6.7B & 13B & 7B & 13B
\\ \hline \\
Dense & - & 27.66 & 22.00 & 14.62 & 12.47 & 10.86 & 10.13 & 5.12 & 4.57
\\ \hline \\
Magnitude & Group AbsMax & 3.2E2 & 1.1E2 & 3.2E3 & 3.6E2 & 7.2E2 & 5.4E3 & 17.18 & 6.77
\\
SparseGPT & Group OPTQ & 42.60 & 34.19 & 21.41 & 14.30 & 12.15 & 11.26 & 8.28 & 5.92
\\
Wanda &  Group AbsMax & 62.64 & 39.60 & 19.93 & 15.01 & 12.31 & 12.46 & 6.80 & 5.75
\\
Wanda & AWQ & 42.49 & 3.8E5 & 18.80 & 14.67 & 12.17 & 12.34 & 7.28 & OOM
\\
Wanda & OmniQuant & 43.55 & NaN & 20.58 & 15.82 & 13.29 & OOM & 7.40 & OOM
\\
Wanda & AffineQuant & 43.66 & NaN & 19.40 & 14.94 & 12.39 & OOM & 7.21 & OOM
\\
JSQ & JSQ & 3.2E3 & 1.9E4 & 23.88 & 2.3E2 & 15.13 & 1.2E5 & 6.63 & 5.73
\\ \hline \\
Naive-LoRA & Group AbsMax & 40.37 & 30.99 & 17.02 & 13.91 & 11.68 & 11.38 & 6.12 & 5.28
\\
Naive-LoRA & \quantization & 46.66 & 33.90 & 19.46 & 15.36 & 12.16 & 11.41 & 6.56 & 5.58
\\
Naive-LoRA + FT & \quantization & 38.05 & 29.27 & 17.52 & 14.39 & 12.28 & 11.84 & 6.10 & 5.28
\\ \hline \\
\slim-LoRA & \quantization & 39.62 & 31.51 & 16.52 & 13.65 & 11.42 & 10.82 & 6.16 & 5.36
\\
\slim-LoRA + FT & \quantization & \textbf{34.92} & \textbf{28.67} & \textbf{16.16} & \textbf{13.66} & 11.83 & 11.47 & \textbf{5.36} & \textbf{5.19}
\\
\slim-LoRA$^Q$ & \quantization & 38.79 & 30.16 & 16.64 & 13.82 & 11.43 & 10.80 & 6.26 & 5.58
\\
\slim-LoRA$^Q$ + FT & \quantization & 35.17 & 28.31 & 16.46 & 13.96 & \textbf{11.42} & \textbf{10.80} & 5.94 & 5.46
\\ \hline 
\end{tabular}
}
\end{center}
\end{table}

\begin{table}[H]
\caption{Perplexity of LLaMA-2 and OPT models with pruning on WikiText-2 dataset language modeling task. The quantization is disabled in this experiment. $\downarrow$ indicates better performance.}
\label{tab:sparse_perplexity}
\begin{center}
\begin{tabular}{lllllllll}
\multicolumn{1}{c}{Pruning/LoRA}  & \multicolumn{6}{c}{OPT} & \multicolumn{2}{c}{LLaMA-2}
\\
\multicolumn{1}{c}{Method}  & 125M & 350M & 1.3B & 2.7B & 6.7B & 13B & 7B & 13B
\\ \hline \\[-8pt]
Dense & 27.66 & 22.00 & 14.62 & 12.47 & 10.86 & 10.13 & 5.12 & 4.57
\\ \hline \\[-8pt]
\textbf{2:4 Sparsity} \\
Magnitude & 341.5 & 417.1 & 427.2 & 1.2E3 & 264.1 & 4.0E4 & 9.1E4 & 2.0E5
\\
SparseGPT & 60.7 & 50.7 & 23.8 & 17.2 & 14.1 & 12.9 & 10.2 & 8.3
\\
Wanda & 81.6 & 116.0 & 27.8 & 21.4 & 16.0 & 16.4 & 12.0 & 8.5
\\
Naive-LoRA & 46.9 & 45.0 & 18.8 & 15.2 & 12.5 & 12.9 & 8.1 & 6.5
\\ 
Naive-LoRA + FT & 39.6 & 35.1 & \textbf{15.0} & 16.3 & 12.7 & 12.3 & 6.5 & \textbf{5.7}
\\
\slim-LoRA & 45.2 & 43.6 & 18.6 & 15.0 & \textbf{12.4} & 12.6 & 7.3 & 6.2
\\
\slim-LoRA  + FT & \textbf{37.1} & \textbf{33.7} & 17.0 & \textbf{14.2} & 12.4 & \textbf{12.1} & \textbf{6.4} & 5.8
\\ \hline \\[-8pt]
\textbf{50\% Unstructured} \\
Magnitude & 193.4 & 97.8 & 1.7E3 & 265.2 & 968.7 & 2.4E4 & 9.9E4 & 1.1E5
\\
SparseGPT & 36.7 & 31.8 & 17.6 & 13.4 & 11.5 & 11.1 & 6.5 & 5.6
\\
Wanda & 39.3 & 36.4 & 18.3 & 14.3 & 12.0 & 12.3 & 6.4 & 5.4
\\
Naive-LoRA & 33.3 & 29.1 & 16.3 & 13.5 & 11.5 & 11.2 & 6.2 & 5.4
\\ 
Naive-LoRA + FT & 31.9 & 27.5 & 16.3 & 13.8 & 12.0 & 11.6 & 5.8 & 5.1
\\
\slim-LoRA & 32.7 & 29.0 & 15.9 & 13.2 & \textbf{11.2} & \textbf{10.8} & 5.9 & 5.2
\\
\slim-LoRA  + FT & \textbf{31.0} & \textbf{26.8} & \textbf{15.5} & \textbf{13.1} & 11.6 & 11.0 & \textbf{5.8} & \textbf{4.7}
\\ \hline 
\end{tabular}

\end{center}
\end{table}

\begin{table}[H]
\caption{Perplexity of LLaMA-2 and OPT models with quantization on WikiText-2 dataset language modeling task. The sparsity is disabled in this experiment. $\downarrow$ indicates better performance.}
\label{tab:quantized_perplexity}
\begin{center}
\resizebox{\textwidth}{!}{%
\begin{tabular}{llllllllll}
\multicolumn{1}{c}{Quantization} & \multicolumn{1}{c}{Low-rank}  & \multicolumn{6}{c}{OPT} & \multicolumn{2}{c}{LLaMA-2}
\\
\multicolumn{1}{c}{Method} & \multicolumn{1}{c}{Adapter} & 125M & 350M & 1.3B & 2.7B & 6.7B & 13B & 7B & 13B
\\ \hline \\
Dense & - & 27.66 & 22.00 & 14.62 & 12.47 & 10.86 & 10.13 & 5.12 & 4.57
\\ \hline \\
OPTQ & - & 33.0 & 24.4 & 16.0 & 13.0 & 11.3 & 10.3 & 6.1 & 4.9
\\
AWQ & - & 29.1 & 2.7E5 & 14.9 & 12.7 & 11.0 & 10.2 & 6.0 & OOM
\\
OmniQuant & - & 30.2 & NaN & 15.8 & 13.3 & 11.6 & OOM & 5.7 & OOM
\\
AffineQuant & - & 28.7 & NaN & 14.9 & 12.6 & 11.0 & OOM & 5.7 & OOM
\\
\hline
\\
Group AbsMax & - & 35.1 & 23.3 & 15.5 & 12.9 & 11.1 & 10.3 & 5.4 & 4.7
\\
Group AbsMax & Naive-LoRA & 30.4 & 22.9 & 15.1 & 12.7 & 11.0 & 10.2 & 5.3 & 4.7
\\
Group AbsMax & \slim-LoRA & \textbf{29.3} & \textbf{22.8} & 15.0 & \textbf{12.7} & \textbf{10.9} & 10.2 & \textbf{5.2} & \textbf{4.7}
\\
\hline
\\
\quantization & - & 1.4E3 & 26.0 & 1.7E3 & 33.1 & 31.0 & 6.7E2 & 1.3E5 & 7.8E4
\\
\quantization & Naive-LoRA & 32.1 & 24.1 & 15.6 & 13.4 & 11.2 & 10.5 & 5.4 & 4.8
\\
\quantization & \slim-LoRA & 30.8 & 23.1 & \textbf{15.2} & 12.9 & 11.1 & 10.3 & 5.4 & 4.8
\\
\quantization & \slim-LoRA + FT & 30.7 & 23.5 & 15.3 & 13.3 & 11.6 & \textbf{10.0} & 5.3 & 4.7
\\ \hline
\end{tabular}}%
\end{center}
\end{table}

\section{Additional Sparse and Quantized Results}
\label{appendix:accuracy_results}

In \autoref{sec:results}, we provided the accuracy results for different pruning and quantization methods. When using Wanda for pruning, we only reported the best quantization method out of Group AbsMax, AWQ, OmniQuant, and AffineQuant. For completeness, we have provided the accuracy achieved by each of these quantization methods separately in \autoref{tab:additional_sparse_quantized_accuracy}.

Methods like OmniQuant and AffineQuant encounter difficulties in quantizing OPT-350M, resulting in NaN values. Additionally, approaches such as AWQ, OmniQuant, and AffineQuant cause memory issues (OOM) when attempting to compress the models on a single A100-40GB GPU.

\begin{table}[H]
\caption{Average zero-shot accuracy of LLaMA-2 and OPT models with \textbf{2:4 sparsity and 4-bit weight quantization}. $\uparrow$ indicates better performance.}
\label{tab:additional_sparse_quantized_accuracy}
\begin{center}
\resizebox{\textwidth}{!}{%
\begin{tabular}{llllllllll}
\multicolumn{1}{c}{Pruning/LoRA}  &\multicolumn{1}{c}{Weight} & \multicolumn{6}{c}{OPT} & \multicolumn{2}{c}{LLaMA-2}
\\
\multicolumn{1}{c}{Method}  &\multicolumn{1}{c}{Quantization} & 125M & 350M & 1.3B & 2.7B & 6.7B & 13B & 7B & 13B
\\ \hline \\[-8pt]
Dense & - & 35.9 & 37.1 & 43.4 & 45.5 & 48.3 & 48.7 & 56.6 & 60.8 
\\ \hline \\[-8pt]
\textbf{2:4 Sparsity} \\
Wanda &  Group AbsMax & 33.27 & 32.79 & 37.47 & 39.45 & 42.95 & 43.64 & 43.89 & 48.94
\\
Wanda & AWQ & 33.33 & 31.50 & 38.43 & 40.00 & 43.41 & 44.07 & 44.86 & OOM
\\
Wanda & OmniQuant & 33.37 & NaN & 37.35 & 39.39 & 41.50 & OOM & 43.95 & OOM
\\
Wanda & AffineQuant & 33.39 & NaN & 37.48 & 33.51 & 42.88 & OOM & 44.62 & OOM
\\ \hline \\[-8pt]
\textbf{50\% Unstructured} \\
Wanda &  Group AbsMax & 34.67 & 33.89 & 40.38 & 42.77 & 45.88 & 46.60 & 51.76 & 56.76
\\
Wanda & AWQ & 35.11 & 31.57 & 41.02 & 42.89 & 46.52 & 46.84 & 50.68 & OOM
\\
Wanda & OmniQuant & 34.85 & NaN & 39.84 & 42.16 & 44.67 & OOM & 50.51 & OOM
\\
Wanda & AffineQuant & 34.64 & NaN & 41.23 & 42.68 & 46.05 & OOM & 53.62 & OOM
\\
\hline
\end{tabular}}%
\end{center}
\end{table}

\section{Sparsity vs. quantization}
\label{appendix:sparsity_vs_quantization}

A natural question that arises when compressing models is whether it is more efficient to reduce the model size through pruning or quantization. To answer this question, we conduct a set of experiments, which evaluate the perplexity of different models under three different conditions, all with around $8\times$ model size reduction factor: (1) 2-bit weight quantization with no sparsity, (2) 4-bit weight quantization with 50\% unstructured sparsity, and (3) 4-bit weight quantization with 50\% 2:4 sparsity. We have used \slim-LoRA with \quantization in all the experiments. The accuracy and perplexity results of these experiments are summarized in \autoref{tab:sparsity_vs_quantization_accuracy} and \autoref{tab:sparsity_vs_quantization_perplexity}, showing that combining sparsity and quantization yields better results in comparison to quantization-only settings with lower bitwidth.

\begin{table}[H]
    \centering
    \caption{Average zero-shot accuracy of different models using different pruning and quantization schemes. $\uparrow$ indicates better performance. Combining sparsity and quantization provides better accuracy results in comparison to solely using quantization.}
    \begin{tabular}{lccccccccc}\\
        & & \multicolumn{6}{c}{OPT} & \multicolumn{2}{c}{LLaMA-2}
        \\
        \multicolumn{1}{c}{Quantization} & \multicolumn{1}{c}{Sparsity}   & 125M & 350M & 1.3B & 2.7B & 6.7B & 13B & 7B & 13B
        \\ \hline \\[-8pt]
        2-bit & - & 33.5 & 32.5 & 38.5 & 39.2 & 43.8 & 44.4 & 42.4 & 44.9
        \\
        4-bit & 2:4 & 34.6 & 34.4 & 40.6 & 42.7 & 46.0 & 46.1 & 51.2 & 54.9
        \\
        4-bit & 50\% Unstructured & 35.2 & 35.3 & 41.9 & 43.5 & 47.1 & 48.0 & 54.3 & 57.9
        \\
        \hline
    \end{tabular}
    \label{tab:sparsity_vs_quantization_accuracy}
\end{table}

\begin{table}[H]
    \centering
    \caption{Perplexity of different models on WikiText-2 dataset using different pruning and quantization schemes. $\downarrow$ indicates better performance. Combining sparsity and quantization provides better accuracy results in comparison to solely using quantization.}
    \begin{tabular}{lccccccccc}\\
        & & \multicolumn{6}{c}{OPT} & \multicolumn{2}{c}{LLaMA-2}
        \\
        \multicolumn{1}{c}{Quantization} & \multicolumn{1}{c}{Sparsity}   & 125M & 350M & 1.3B & 2.7B & 6.7B & 13B & 7B & 13B
        \\ \hline \\[-8pt]
        2-bit & - & 116.2 & 169.7 & 35.1 & 27.1 & 16.2 & 15.0 & 12.5 & 11.7
        \\
        4-bit & 2:4 & 47.5 & 45.6 & 18.8 & 15.7 & 12.4 & 12.1 & 7.2 & 6.5
        \\
        4-bit & 50\% Unstructured & 36.3 & 29.9 & 16.3 & 13.7 & 11.4 & 10.8 & 6.0 & 5.4
        \\
        \hline
    \end{tabular}
    \label{tab:sparsity_vs_quantization_perplexity}
\end{table}

\section{Additional speedup results}
\label{appendix:speedup}

\begin{figure}[H]
    \centering
    \includegraphics[width=\textwidth]{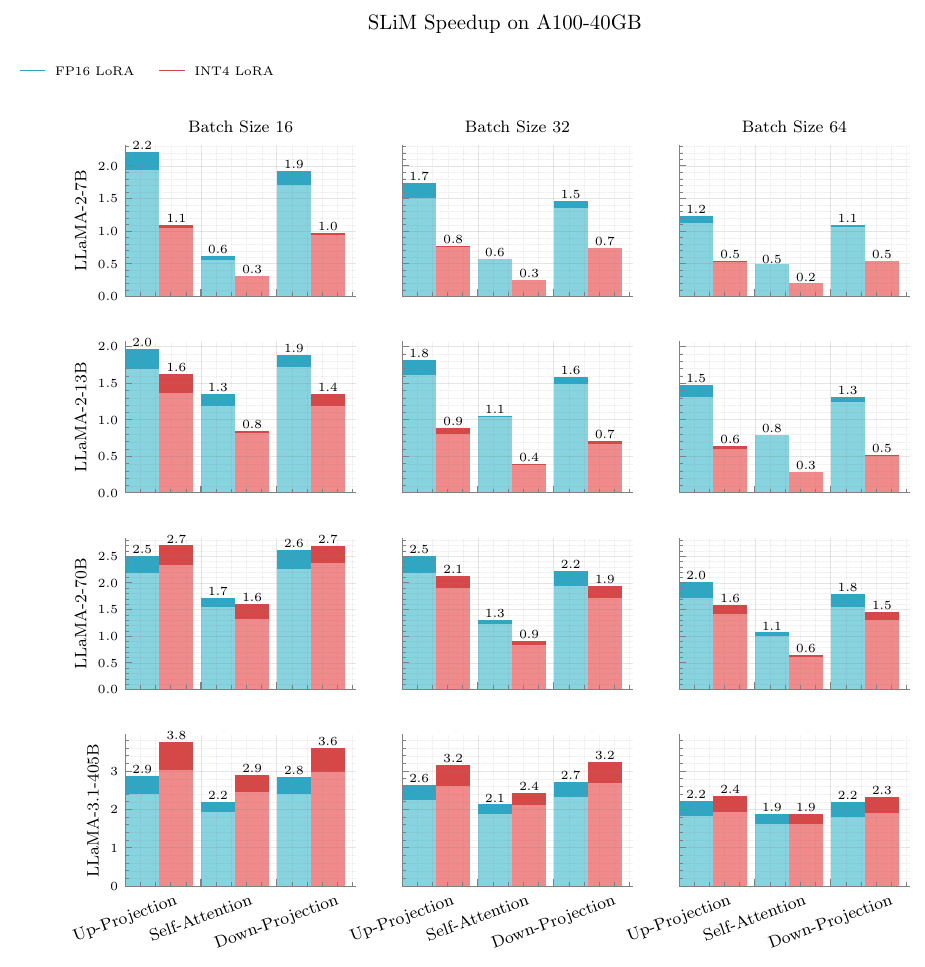}
    \caption{\slimspace speedup for LLaMA-2 family of models on NVIDIA A100-40GB GPUs.}
    \label{fig:a100_speedup}
\end{figure}

\autoref{sec:results} presents the speedup of \slimspace on consumer-grade GPUs, while this section provides results on NVIDIA A100-40GB GPUs. \autoref{fig:a100_speedup} summarizes the speedup for the LLaMA-2 and LLaMA-3.1 model families, including LLaMA-3.1-405B, highlighting \slim's scalability to large models. As with consumer-grade devices, larger models achieve higher speedups. However, for smaller models like LLaMA-2-7B, Sparse Marlin's sparse quantized matrix multiplication kernels lead to a slowdown on A100 GPUs, which does not occur on consumer-grade GPUs and is not specific to \slimspace.

\section{Fine-tuning costs} 
\label{appendix:fine_tuning_overhead}

Fine-tuning compressed models can recover lost accuracy, but the high parameter count leads to substantial time and memory costs. In our experiments, we fine-tuned models with low-rank adapters, where the quantized weights are frozen and only the adapters are fine-tuned. This results in a more parameter-efficient approach, reducing both memory and computational costs. When no low-rank adapter is used, the straight-through estimator (STE) fine-tunes the quantized weights.

\autoref{tab:fine-tuning-timing} presents the fine-tuning results for 300,000 tokens from the C4 dataset, using a batch size of 64 and sequence length of 1024 on a single H100 GPU. Fine-tuning models without low-rank adapters took 12 hours for 125M parameter models and over 36 days for 13B parameter models. Given these high costs, completing fine-tuning was challenging with our limited resources. In contrast, using low-rank adapters and freezing the sparse quantized weights made fine-tuning more efficient, enabling us to report accuracy results in \autoref{tab:sparse_quantized}.

\begin{table}[H]
\caption{The required time for fine-tuning the models with a single H100 GPU on 300,000 tokens from the C4 dataset with a batch size of 64 and a sequence length of 1024.}
\label{tab:fine-tuning-timing}
\begin{center}
\begin{tabular}{llllllllll}
\multicolumn{1}{c}{Pruning}  &\multicolumn{1}{c}{Weight} & \multicolumn{6}{c}{OPT} & \multicolumn{2}{c}{LLaMA-2}
\\
\multicolumn{1}{c}{Method}  &\multicolumn{1}{c}{Quantization} & 125M & 350M & 1.3B & 2.7B & 6.7B & 13B & 7B & 13B
\\ \hline \\
Magnitude & Group AbsMax & 
\\
SparseGPT & OPTQ & 12h & 43h & 164h & 361h & 866h & 867h & 842h & 844h
\\
Wanda & Group AbsMax & 
\\ \hline \\
\slim-Naive & \quantization &\multirow{2}{*}{1.5h} & \multirow{2}{*}{3h} & \multirow{2}{*}{6h} & \multirow{2}{*}{8h} & \multirow{2}{*}{16h} & \multirow{2}{*}{18h} & \multirow{2}{*}{14h} & \multirow{2}{*}{14h}
\\
\slim-LoRA & \quantization & 
\\ \hline 
\end{tabular}
\end{center}
\end{table}

\section{Memory reduction analysis}
\label{appendix:memory_reduction}

\slimspace prunes and quantizes the models and adds additional low-rank adapters to them. Additionally, it supports quantization methods for the low-rank adapters to reduce their overheads. In the following, we provide an analysis of the memory reduction when using \slimspace and other pruning and quantization methods.

Assuming the hidden dimension of a model is $d$ and the low-rank adapter ratio used in the model is of rank $r < 1$. Furthermore, by denoting the number of transformer blocks with $n$ and the vocabulary size of the model by $V$ and by denoting the ratio of the up-projection and down-projection layers in the model by $a$, we can get the memory reduction as the ratio of $ \frac{\text{Compressed Model Size}}{\text{Dense Model Size}}$ from \autoref{eq:memory_reduction}.

\begin{equation}
    \label{eq:memory_reduction}
    \text{Memory Reduction} =  \frac{n(4d^2/2 + 4\times2d^2r + 2d^2a/2 + 2d(dr + dra)) + dV}{n(4d^2 + 2d^2a) + dV}
\end{equation}

\autoref{tab:memory_reduction} summarizes the memory reduction of different pruning and quantization methods. Please note that when using low-rank adapters (in Naive-LoRA and \slim-LoRA), we assume a rank of $r=0.1$.

\begin{table}[H]
    \centering
    \caption{Theoretical memory reduction ($\times$) of different compression methods across various OPT and LLaMA models. In Quantized \slimspace, the low-rank adapters are also quantized.($\downarrow$ indicates better performance.)}
    \begin{tabular}{lcccccccc}\\
        \multicolumn{1}{c}{Compression} & \multicolumn{6}{c}{OPT} & \multicolumn{2}{c}{LLaMA-2}
        \\
        \multicolumn{1}{c}{Method}   & 125M & 350M & 1.3B & 2.7B & 6.7B & 13B & 7B & 13B
        \\
        \hline
        SparseGPT + OPTQ            & 0.40  & 0.30  & 0.25 & 0.17 & 0.15 & 0.14 & 0.15 & 0.14 \\
        Wanda + AbsMax              & 0.40  & 0.30  & 0.25 & 0.17 & 0.15 & 0.14 & 0.15 & 0.14 \\
        Naive-LoRA + AbsMax          & 0.50  & 0.42 & 0.38 & 0.31 & 0.30  & 0.29 & 0.31 & 0.30  \\
        \slim-LoRA + \quantization           & 0.50  & 0.42 & 0.38 & 0.31 & 0.30  & 0.29 & 0.31 & 0.30  \\
        \slim-LoRA$^Q$ + \quantization & 0.42 & 0.33 & 0.28 & 0.20  & 0.19 & 0.18 & 0.19 & 0.18 \\
        \hline
    \end{tabular}
    \label{tab:memory_reduction}
\end{table}

\section{Computation reduction analysis}
\label{appendix:compute_reduction}

\slimspace and other compression methods reduce the number of floating point operations (FLOPs) at the inference of models. Additionally, the low-rank adapters used in \slimspace and Wanda SVD can add additional computational overheads to the inference of the models. Following JSQ \citep{jsq}, in this section, we provide an analysis of the FLOP reduction in the inference of different methods. It is noteworthy that even though quantization can reduce the memory overhead of models, since all the computations are done in floating point format, it does not lead to a reduction in the computation of the inference.

Assuming the hidden dimension of a model is $d$ and the low-rank adapter ratio used in the model is of rank $r < 1$. Furthermore, by denoting the number of transformer blocks with $n$ and the vocabulary size of the model by $V$ and by denoting the ratio of the up-projection and down-projection layers in the model by $a$, we can get the FLOP reduction as the ratio of $ \frac{\text{Dense Inference FLOP Count}}{\text{Compressed Inference FLOP Count}}$ from \autoref{eq:compute_reduction}, where $b$ is the batch size, and is canceled in the numerator and the denominator of the equation.

\begin{equation}
    \label{eq:compute_reduction}
    \text{FLOP Reduction} =  \frac{n(4bd^2 + 2bd^2a) + bdV}{n(4bd^2/2 + 4\times2bd^2r + 2bd^2a/2 + 2b(d^2r + d^2ra)) + bdV}
\end{equation}

\autoref{tab:flop_reduction} summarizes the FLOP reduction of different compression methods. As can be seen, the overhead of adding the low-rank adapters ($r = 0.1$) in \slim-LoRA and Naive-LoRA is not significant.

\begin{table}[H]
    \centering
    \caption{Compute (FLOP) reduction ratios ($\times$) of different compression methods across various OPT and LLaMA models. In Quantized \slimspace, the low-rank adapters are also quantized. ($\uparrow$ indicates better performance.)}
    \vspace{5pt}
    \begin{tabular}{lcccccccc}
        \multicolumn{1}{c}{Compression} & \multicolumn{6}{c}{OPT} & \multicolumn{2}{c}{LLaMA-2} \\
        \multicolumn{1}{c}{Method}   & 125M & 350M & 1.3B & 2.7B & 6.7B & 13B & 7B & 13B \\
        \hline
        SparseGPT + OPTQ             & 1.52  & 1.66  & 1.75 & 1.91 & 1.94 & 1.96 & 1.95 & 1.97 \\
        Wanda + AbsMax               & 1.52  & 1.66  & 1.75 & 1.91 & 1.94 & 1.96 & 1.95 & 1.97 \\
        Naive-LoRA + AbsMax           & 1.32  & 1.39  & 1.43 & 1.50 & 1.51 & 1.52 & 1.49 & 1.49 \\
        \slim-LoRA + \quantization            & 1.32  & 1.39  & 1.43 & 1.50 & 1.51 & 1.52 & 1.49 & 1.49 \\
        \slim-LoRA$^Q$ + \quantization  & 1.32  & 1.39  & 1.43 & 1.50 & 1.51 & 1.52 & 1.49 & 1.49 \\
        \hline
    \end{tabular}
    \label{tab:flop_reduction}
\end{table}

\section{Compression costs} 
\label{sec:compression_cost}

The computational cost of compression methods varies depending on their complexity. While all approaches can compress a single layer at a time, the memory usage is similar across methods, as each stores only one layer in the GPU's global memory. Techniques like Wanda, which rely on matrix multiplication, are faster than more complex methods like SparseGPT, which computes the inverse Hessian matrix for each layer. Adding low-rank adapters to Wanda-SVD and \slimspace increases computational complexity due to the need for singular value decomposition (SVD), making them comparable to SparseGPT in terms of computation.

\autoref{tab:compression_time} summarizes the time required to compress various models using the discussed methods. Methods incorporating low-rank adapters (\slimspace and Wanda-SVD) generally take longer to compress due to their higher complexity. Interestingly, SparseGPT's compression time is comparable to methods with low-rank adapters, despite only performing pruning and quantization. The saliency-based approach in \slimspace does not add significant overhead compared to Wanda-SVD, maintaining efficiency despite its added complexity.

\begin{table}[H]
\caption{The required compression time for different models and compression methods using a single H100 GPU.}
\label{tab:compression_time}
\begin{center}
\begin{tabular}{llllllllll}
\multicolumn{1}{c}{Pruning}  &\multicolumn{1}{c}{Weight} & \multicolumn{6}{c}{OPT} & \multicolumn{2}{c}{LLaMA-2}
\\
\multicolumn{1}{c}{Method}  &\multicolumn{1}{c}{Quantization} & 125M & 350M & 1.3B & 2.7B & 6.7B & 13B & 7B & 13B
\\ \hline \\
Magnitude & AbsMax & 1s & 1s & 1s & 1s & 2s & 4s & 2s & 4s
\\
SparseGPT & OPTQ & 1m & 2m & 5m & 11m & 22m & 41m & 25m & 46m
\\
Wanda & \quantization & 0.5m & 1m & 3m & 5m & 8m & 13m & 8m & 14m 
\\ \hline \\
Wanda-SVD & \quantization & 1m & 2m & 7m & 13m & 33m & 60m & 38m & 67m
\\
\slimspace & \quantization & 1m & 2m & 7m & 13m & 34m & 63m & 39m & 68m
\\ \hline 
\end{tabular}
\end{center}
\end{table}

\section{Rank analysis} 
\label{sec:rank_analysis}
The key hyperparameter in low-rank approximation is the rank of the adapters. While increasing the rank reduces approximation error, it also leads to higher computational and memory overhead. Therefore, it is crucial to analyze the trade-off between the accuracy improvements and the overhead introduced by the chosen approximation rank.

Assuming the rank of the low-rank adapter is $rd$, where $r < 1$ is a fixed factor and $d$ is the dimension of the weights in a square feed-forward layer, the low-rank adapters are represented as $\mathcal{L}, \mathcal{R}^T \in \mathbb{R}^{d \times rd}$, resulting in a memory overhead of $\mathcal{O}(2rd^2)$ for storing them. To compute $\mathcal{XLR}$, where $\mathcal{X} \in \mathbb{R}^{b \times d}$ is the input with a batch size of $b$, the computational complexity is $\mathcal{O}(2brd^2)$. Given that the original memory and computational complexity of the layer are $\mathcal{O}(d^2)$ and $\mathcal{O}(bd^2)$, respectively, the overhead introduced by the low-rank adapters becomes negligible when $r \ll 1$.

\begin{figure}[htbp]
\begin{center}
\includegraphics[scale=0.99]{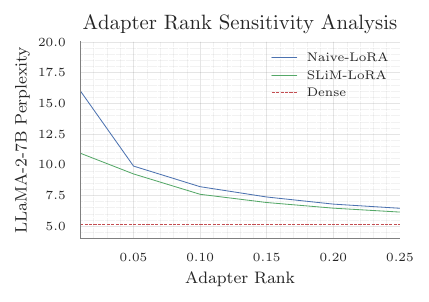}
\includegraphics[scale=0.99]{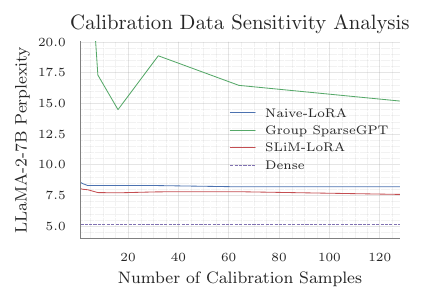}
\\
\textbf{(a)} \hspace{2.5in} \textbf{(b)}
\end{center}
\caption{Sensitivity analysis for the rank of the adapter (\textbf{a}) and the number of calibration samples (\textbf{b}) for different one-shot compression methods. For Naive-LoRA and \slim-LoRA, we have used the \quantization quantization method, and for the SparseGPT, we have used the Group quantization version of OPTQ.}
\label{fig:rank_analysis}
\end{figure}

\autoref{fig:rank_analysis}-a shows the average zero-shot accuracy of the OPT-6.7B and LLaMA-2-7B models for various ranks. As expected, increasing the rank leads to improved model accuracy. Based on these results, a rank of $r = 0.1$ provides a substantial boost in accuracy without introducing significant overhead to inference.

\section{Effects of calibration sample count}
\label{sec:calibration_sample}
Similar to previous work (SparseGPT, Wanda, AWQ, OmniQuant, and AffineQuant), \slimspace leverages a set of calibration data from the C4 dataset to assess weight saliency for pruning and low-rank approximations. \autoref{fig:rank_analysis}-b illustrates the perplexity of LLaMA-2-7B using varying numbers of calibration samples. As shown, \slimspace demonstrates low sensitivity to the number of calibration samples, making it effective even in scenarios with limited data.

\section{Sensitivity to calibration dataset}
\label{appendix:dataset_sensitivity}

Similar to other pruning and quantization methods such as Wanda, SparseGPT, OPTQ, and AWQ, \slimspace relies on a calibration dataset to evaluate weight saliency. The C4 \citep{c4} and SlimPajama \citep{slimpajama} datasets are among the most commonly used calibration sets for LLM compression. \autoref{tab:dataset_sensitivity} presents the perplexity results for \slim-LoRA and \quantization across different calibration datasets. The results indicate that \slimspace is largely insensitive to the choice of dataset, achieving comparable accuracy regardless of the calibration dataset used.

\begin{table}[H]
    \centering
    \caption{Perplexity of different models on WikiText-2 dataset using \slim-LoRA with 4-bit quantization using \quantization with different calibration datasets. $\downarrow$ indicates better performance.}
    \begin{tabular}{lcccccccc}\\
        \multicolumn{1}{c}{Calibration} & \multicolumn{6}{c}{OPT} & \multicolumn{2}{c}{LLaMA-2}
        \\
        \multicolumn{1}{c}{Dataset}   & 125M & 350M & 1.3B & 2.7B & 6.7B & 13B & 7B & 13B
        \\ \hline \\[-8pt]
        \textbf{50\% 2:4} \\
        C4  & 57.91 & 50.09 & 19.64 & 15.65 & 12.71 & 12.13 & 7.56 & 6.50
        \\
        SlimPajama & 46.27 & 44.77 & 19.35 & 16.04 & 12.56 & 12.32 & 7.15 & 6.49
        \\ \hline \\[-8pt]
        \textbf{50\% Unstructured} \\
        C4 & 39.62 & 31.51 & 16.52 & 13.65 & 11.42 & 10.82 & 6.16 & 5.36
        \\
        SlimPajama & 36.49 & 29.94 & 16.64 & 14.08 & 11.61 & 11.02 & 5.99 & 5.34
        \\
        \hline
    \end{tabular}
    \label{tab:dataset_sensitivity}
\end{table}

\section{Sparsity analysis}
\label{appendix:sparsity_analysis}

\begin{figure}[H]
    \centering
    \includegraphics[scale=1]{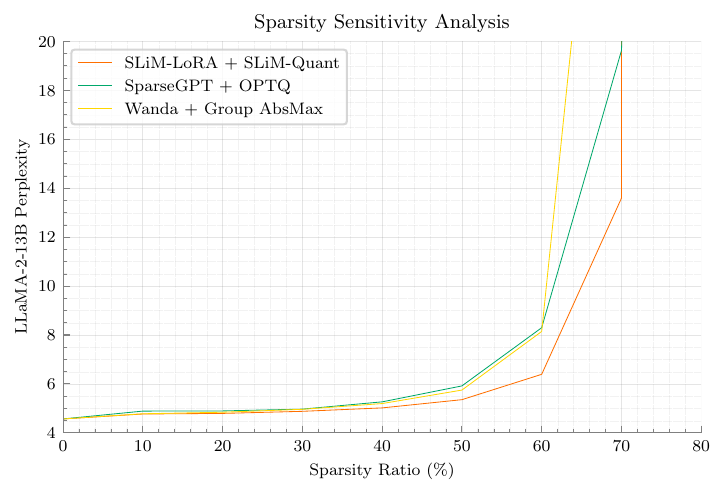}
    \caption{Sparsity analysis on LLaMA-2-13B model using perplexity on WikiText-2 dataset. $\downarrow$ indicates better performance.}
    \label{fig:sparsity_analysis}
\end{figure}

To analyze the impact of sparsity on model accuracy, we conduct experiments on LLaMA-2-13B with 4-bit quantization, pruning it to varying sparsity ratios. \autoref{fig:sparsity_analysis} presents the perplexity results for \slim-LoRA with \quantization, SparseGPT with OPTQ, and Wanda with Group AbsMax. As expected, increasing the sparsity ratio leads to higher perplexity, indicating a trade-off between compression and accuracy. Notably, \slim-LoRA combined with \quantization maintains competitive accuracy up to 60\% sparsity, whereas other methods experience noticeable degradation at lower sparsity levels.

\section{Group quantization challenges}
\label{appendix:group_quantization_challenges}

Group quantization allows sharing the same quantization parameters for a small group of the elements in the quantized matrix, leading to smaller error. But, using group quantization adds additional challenges to the training and inference of the model, e.g. \textbf{more complicated implementation} and \textbf{additional memory and compute overheads}. 

The state-of-the-art group quantization GPU kernel, dense and sparse Marlin \citep{marlin}, consists of thousands of lines of CUDA code optimized for only a limited number of GPU architectures, showcasing the amount of effort needed to implement a version of group quantization. Furthermore, other libraries and frameworks, such as Triton \citep{triton} and CUTLASS \citep{cutlass} do not provide support for 4-bit group quantization, limiting its flexibility and possibility of modification.

Furthermore, using group quantization can lead to an additional overhead during matrix multiplication, since more parameters need to be loaded for dequantizing each group. As an example, \autoref{tab:group_quantization_slow_down} shows the slow-down of using group quantization on the down-projection matrices in different LLaMA-2 and LLaMA-3.1 models on a NVIDIA A100-40GB GPU, with a batch size of 16.

\begin{table}[htbp]
\caption{Group quantization slow-down ($\times$) on different LLaMA-2 and LLaMA-3.1 models. $\downarrow$ indicates worse.}
\label{tab:group_quantization_slow_down}
\begin{center}
\begin{tabular}{l|llll}
Model & LLaMA-2-7B & LLaMA-2-13B & LLaMA-2-70B & LLaMA-3.1-405B
\\
\hline
\\
Slow-Down ($\times$) & 0.94 & 0.95 & 0.95 & 0.94
\\
\hline
\end{tabular}
\end{center}
\end{table}

\end{document}